%% file: neurips_2024.tex
\documentclass{article}

\PassOptionsToPackage{compress,sort,square}{natbib}

\usepackage[preprint]{neurips_2024}

\usepackage[utf8]{inputenc} %
\usepackage[T1]{fontenc}    %
\usepackage{hyperref}
\hypersetup{
	colorlinks=true,
	linkcolor=blue,
	filecolor=magenta,      
	urlcolor=teal,
	citecolor=blue,
}     %
\usepackage{url}            %
\usepackage{booktabs}       %
\usepackage{amsfonts}       %
\usepackage{nicefrac}       %
\usepackage{microtype}      %
\usepackage{xcolor}         %

\usepackage{graphicx}
\usepackage{subfigure}
\usepackage{mathtools} %

\usepackage{wrapfig}
\usepackage{rotating}

\usepackage{pifont}

\usepackage{amsmath}
\usepackage{amssymb}
\usepackage{mathtools}
\usepackage{amsthm}

\usepackage[capitalize,noabbrev]{cleveref}

\usepackage[textsize=tiny,textwidth=2.5cm]{todonotes}
\usepackage[vskip=1em,font=itshape,leftmargin=2em,rightmargin=2em]{quoting}
\usepackage{titlesec}

\title{Hallmarks of Optimization Trajectories in Neural Networks: Directional Exploration and Redundancy }%

\author{%
  Sidak Pal Singh\thanks{Correspondence to \href{mailto:ssidak@ethz.ch}{\texttt{ssidak@ethz.ch}}} \\
  ETH Zürich \& MPI-IS Tübingen
   \\
  \And
  Bobby He \\
  ETH Zürich \\
  \And
  Thomas Hofmann \\
  ETH Zürich \\
  \And
  Bernhard Schölkopf\\
  MPI-IS Tübingen \& ETH Zürich\\
}

\input{math_defs}

\begin{document}
\maketitle

\begin{abstract}%
	We propose a fresh take on understanding the mechanisms of neural networks by analyzing the rich directional structure of optimization trajectories, represented by their pointwise parameters. Towards this end, we introduce some natural notions of the complexity of optimization trajectories, both qualitative and quantitative, which hallmark the directional nature of optimization in neural networks: when is there redundancy, and when exploration. We use them to reveal the inherent nuance and interplay involved between various optimization choices, such as momentum and weight decay. Further, the trajectory perspective helps us see the effect of scale on regularizing the directional nature of trajectories, and as a by-product, we also observe an intriguing heterogeneity of Q,K,V dynamics in the middle attention layers in LLMs and which is homogenized by scale. Importantly, we put the significant directional redundancy observed to the test by demonstrating that training only scalar batchnorm parameters some while into training matches the performance of training the entire network, which thus exhibits the potential of hybrid optimization schemes that are geared towards efficiency.
	
\end{abstract}

\section{Introduction}
Given a network architecture and the training task, the loss landscape --- which is the high-dimensional surface whose each point characterizes the fit of the parameters to the task objective --- entails the possible trajectories (or paths) that might be followed by an optimization algorithm, such as stochastic gradient descent (SGD). The particular sets of optimization trajectories are determined, to no lesser extent, by the particular optimization choices and hyperparameters, such as the learning rate, momentum, batch size, weight decay, and more. We might even say that the regions and topographical features of the landscape that are never encountered or realized in typical optimization trajectories, might as well not be in the landscape at all. \mbox{\textit{Essentially}}, the optimization trajectories are the probes through which the loss landscape is accessed.  

Consequently, a significant body of literature builds around the principle of an inherently regular manner of traversing in the landscape, i.e., the implicit bias~\citep{gunasekar2018characterizing,li2019algorithmic,li2020explaining,moroshko2020implicit}, facilitated by optimization algorithms. This preferential landscape access is in the face of the actual surface level, possibly treacherous, non-convexity, and thus suggesting that the network stays clear of sub-optimal local minima. Therefore, despite the complexity of the neural landscapes, implicit bias lends a formal and reasonable support to the empirical success of massively over-parameterized neural networks. If that is indeed the case, we might expect to see traits and hallmarks of regularity in the sequence of steps that make up the optimization trajectories of neural networks in the loss landscape. In other words,  it provokes the following questions,

\begin{quote}
	\textit{How are these trajectories structured? Do these paths have a lot of zigzags and bends, reaching the solution winding and coiling, or are they straight and direct? And does this depend upon the phase of optimization (early vs late)? %
	}
\end{quote}

This, in essence, is the key research theme of our present study. More precisely, we explore and develop key qualitative as well as quantitative indicators (hallmarks) about the complexity/regularity of the optimization trajectory. Towards this end, we analyze and compare multiple intermediate checkpoints amongst themselves, across different scenarios and large-scale case studies.  A qualitative hallmark, which we call the \textit{`Trajectory Map'}, conveys the directional (dis)similarity of the parameters and visually depicts the nature of optimization within and across various stages of training, i.e., at a pan-trajectory level. Our quantitative hallmarks are functions of these trajectory maps, measuring various notions of lengths and angles, over and about the sequence of steps in the trajectory. \looseness=-1

The focus of our investigation is to study the properties of trajectories and their ensuing implications about the nature of found solutions since it  (a) brings in a level of architecture agnosticity and helps unlock shared insights onto features of optimization, (b) contains an intrinsic data-dependence that, for all intents and purposes, necessitates no explicit inference over additional data samples is needed (which anyway might not be possible due to resource or privacy constraints), (c) allows analyzing and prognosing the developing solution strategy on-the-fly, over the course of training (instead of waiting all the way until convergence to assess the solution quality), and (d) provides potential hints at the bottlenecks and redundancies plaguing the optimization procedure.

\textbf{Our contributions are}: 1) We propose the novel perspective of trajectory maps and showcase its use for hallmarking the directional nature of optimization. 2) We utilize it to theoretically show the intertwined effects of momentum and weight decay on obtuse angles between updates. 3) We show how scale provides a regularizing effect on the directional complexity of trajectories, and also find an interesting observation about Q,K,V heterogeneity in the middle attention layers. 4) We put the demonstrated directional redundancy to the test, by showing that only training the scalar parameters of (batch) normalization layers a short while into training suffices for matching the performance, thus laying the seeds for efficient optimization hybrids.

\section{Methodology}

\textbf{Matrix representation of Trajectory.} Let us assume the optimization trajectory consists of a set $\mathcal{T}$ of points $\lbrace\btheta_t\rbrace_{t=0}^T$, each denoting the (flattened) parameters of the network encountered at some step and which live in the parameter space $\btheta=\Reals{p}$, i.e., $\mathcal{T}\subseteq\btheta$. This set of points need not contain the entire set of points visited in the course of optimization but instead can represent a subset of points, possibly sampled at an interval of $k$ points.
It will be convenient to organize this set of points, which define the trajectory, in the form of a matrix, $\bTheta \in \Reals{ (T+1) \times p}$, whose first dimension $T+1$ makes explicit the inclusion of the initialization $\btheta_0$. \looseness=-1

Elsewhere it might be useful to analyze the trajectory relative to some point $\btheta_\tau$ as the origin, so there we will instead consider the set of points $\mathcal{T}_\tau = \lbrace\btheta_t - \btheta_\tau \rbrace_{t=0}^T$, and correspondingly organize it in the matrix $\bTheta_\tau \in \Reals{(T+1)\times p}$. When $\tau$ is itself one of the points of the trajectory, then we will omit the row of zeros and shape the matrix as $\Reals{T\times p}$. A natural point from where to contextualize the trajectory would be the initialization $\btheta_0$, and this relative trajectory will then be denoted as ${\bTheta}_0$ (where the subscript $0$ is not to be confused for the usual origin $O$, namely, $\btheta_O = \mathbf{0}$).%

\textbf{Trajectory Map.} Analyzing the matrix $\bTheta$ or $\bTheta_\tau$, on its own, might get cumbersome as the size of modern networks ranges in millions and billions of parameters. Hence, we will resort to looking at functions of the kernel matrix $\Km = \bTheta \bTheta^\top$ which would be a square matrix of shape $n=T+1$, or for $\tau\neq O$, the relative kernel matrix $\Km_\tau = \bTheta_\tau \bTheta_\tau^\top$ of shape $n=T+1$ or $n=T$ depending if the point $\btheta_\tau$ is a part of the trajectory or not. Further, it will also be helpful to isolate and analyze the directional aspect of the trajectory, for which we will normalize the set of points by their norm, and in effect, consider the set $\widehat{\mathcal{T}}_\tau = \lbrace\frac{\btheta_t - \btheta_\tau}{\|\btheta_t - \btheta_\tau\|_2} \rbrace_{t=0}^T$ with the respective matrix $\widehat{\bTheta}_\tau$. As a result, the ensuing kernel matrix $\widehat{\bTheta}_\tau\widehat{\bTheta}_\tau^\top$, which we will refer to as $\Cm_\tau$ (or $\Cm:=\Cm_O$ for the usual origin $\tau=O$), will contain the relative cosine similarities between every pair of points in the trajectory. So, $(\Cm_\tau)_{ij} $ is,%

\vspace{-4mm}
$$\text{cos-sim}(\btheta_i - \btheta_\tau, \btheta_j - \btheta_\tau) = \dfrac{\langle \btheta_i - \btheta_\tau, \,\btheta_j - \btheta_\tau\rangle}{\|\btheta_i - \btheta_\tau\|_2 \, \|\btheta_j - \btheta_\tau\|_2}\,.
$$

Hereafter, we will refer to $\Cm$ as the \textit{Trajectory Map} (TM) and $\Cm_\tau$ (for $\tau\neq O$) as the \textit{Relative Trajectory Map} (RTM). We would like to remark that, although not necessary, here we are essentially considering linear kernels, for as we will see they deliver a great mileage by themselves. The TM will be our qualitative hallmark of choice for analyzing optimization trajectories.

\textbf{Quantitative Hallmarks.} Besides visualizing the TM as the qualitative hallmark, we will consider the following set of indicators for quantitatively hallmarking the optimization trajectories. 

\textit{(i) Mean Directional Similarity (MDS):} We take the cosine similarity averaged over the entire trajectory map, i.e., over every pair of points in the trajectory. This can be written as, $\omega := \frac{1}{n^2} \,\, \mathds{1}_{n}^\top \, \cdot\, \Cm \, \cdot\, \mathds{1}_{n}$, where $\mathds{1}_n^\top = (1 \cdots 1)^\top \in \Reals{1\times n}$ denotes the vector of all ones and $n=|\mathcal{T}|$ is the cardinality of the trajectory. By using the form of the matrix $\Cm$ discussed before, we can further rewrite MDS as, $\omega = \left\|\frac{1}{n} \, \, \widehat{\bTheta}^\top \mathds{1}_{n}\right\|^2$. Now, it becomes apparent that MDS essentially projects all the trajectory points onto the unit sphere, computes their average and finally takes the squared norm. \\

To get a better sense of MDS, we can consider its two possible extremes: (a) all the parameter unit-vectors cancel out, yielding a value of $\omega=0$. For instance, this would happen in the scenario when the points in the trajectory are exactly following a circular orbit around the origin; or, (b) when each of the parameters point in the same direction, implying that the trajectory is simply a linear path, with $\omega=1$. Knowing the nature of these two extremes, we can expect neither to be desirable in an ideal trajectory which leads up to a generalizing solution.\textit{ Thus, hitting its sweet spot would be the target, or where that is unknown, at least avoiding these extremes.} \looseness=-1

Furthermore, we can also consider a variant of MDS around initialization, by switching the trajectory map $\Cm$ to its relative version $\Cm_\tau$ at $\tau=0$, resulting in $\omega_0 = \frac{1}{n^2} \,\, \mathds{1}_{n}^\top \, \cdot\, \Cm_0 \, \cdot\, \mathds{1}_{n} = \left\|\frac{1}{n}\,\widehat{\bTheta}_0^\top \mathds{1}_{n} \right\|^2$, where $n=|\mathcal{T}_0|$. This can be useful when the initialization is rather distant from the origin, and might wash off the directional nature of the trajectory. 

\begin{wrapfigure}{R}{0.5\textwidth}
	\centering    
	\includegraphics[trim=5 7 10 3, clip,width=0.5\textwidth]{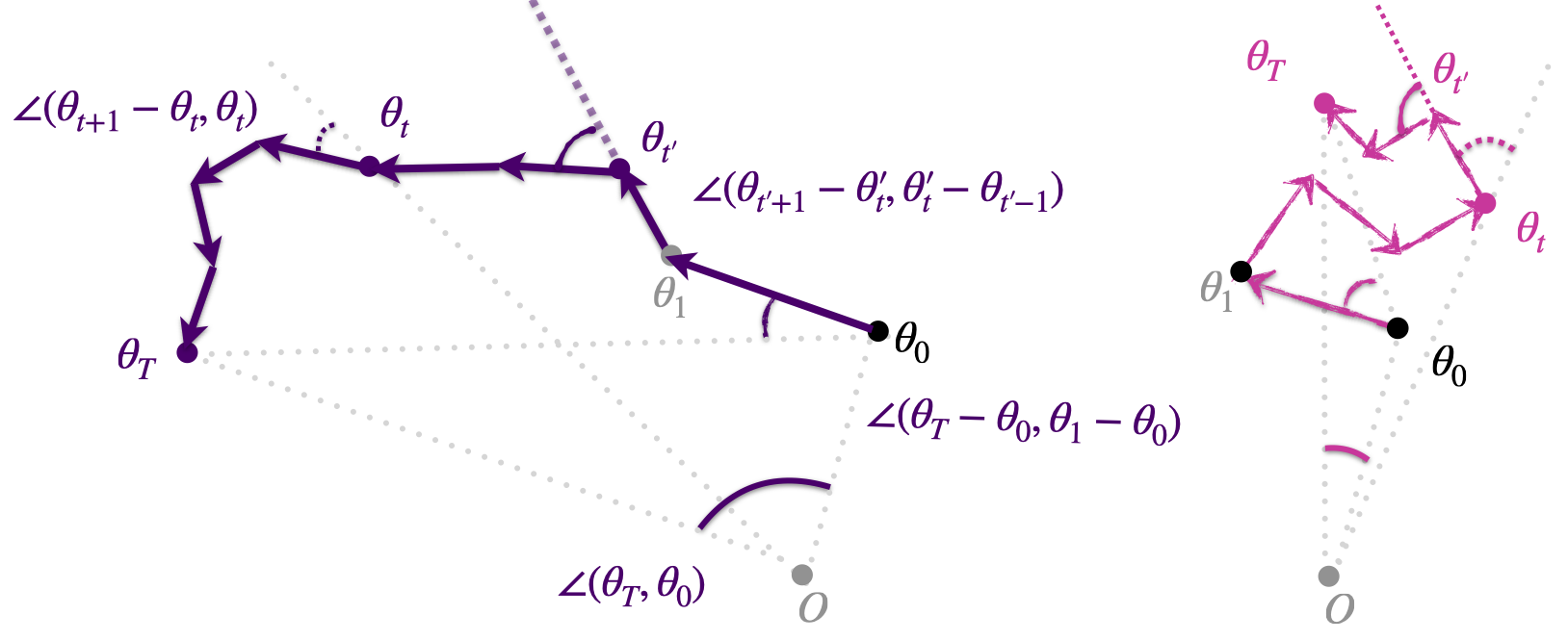}
	\caption{Illustration of two trajectories and angular measures\looseness=-1.}  
    \label{fig:angular}
\end{wrapfigure}

(ii) \textit{Angular Measures:  } One such key measure would be the angle between consecutive (net) updates, i.e.,  $\angle (\btheta_{t+1}-\btheta_t, \btheta_{t}-\btheta_{t-1})$.  Next, we can consider a cone with the vertex at initialization and track its apex angle in the form of $\angle (\btheta_{t}-\btheta_{0}, \btheta_{1}-\btheta_{0})$. Likewise, we can also consider a cone at origin and compute the apex angle there, $\angle (\btheta_{t}, \btheta_{0})$. This measure at origin will give us an idea of the amount of directional movement across a global scale, while the one centered at the initialization will indicate a more local or relative scale.\looseness=-1

(iii) \textit{Norm-based Measures:} Besides, we also include common measures such as: parameter norms $\|\btheta_t\|_2$, distance from initialization $\|\btheta_t-\btheta_0\|_2$, norm between consecutive points $\|\btheta_{t+1}-\btheta_t\|_2$.

\textbf{Remark.} For extremely large neural networks, building the underlying kernel matrices can start being a tad bit resource-expensive. In principle, there is a rich body of work in kernel methods that has focused on developing efficient approximations~\citep{davis2014asymmetric,chen2021spann}. But, as we foray into this novel trajectory perspective, we do not resort to such approximations for the sake of accuracy.\looseness=-1

\section{A Tale of Hyperparameters}
There is an emerging folk intuition that, over the past decade, networks have co-evolved hand-in-hand, amidst other things, with  a certain particular set of optimizers and hyperparameter choices, deviating from which tends to produce significantly poor results. Let us see one such case in  action, for a ResNet50 trained on ImageNet with SGD, achieving the familiar \mbox{top-1} accuracy of $\sim 76\%$. Further, as is usual convention, this network was trained via a learning rate $\eta=0.1$, momentum $\mu=0.9$, batch size $B=256$, weight decay $\lambda=0.0001$ for $90$ epochs, with a multiplicative decay by a factor of $0.1$ at epochs $30$ and $60$.

\begin{wrapfigure}{R}{0.45\textwidth}
\centering    
\includegraphics[trim=5 7 10 3, clip,width=0.4\textwidth]{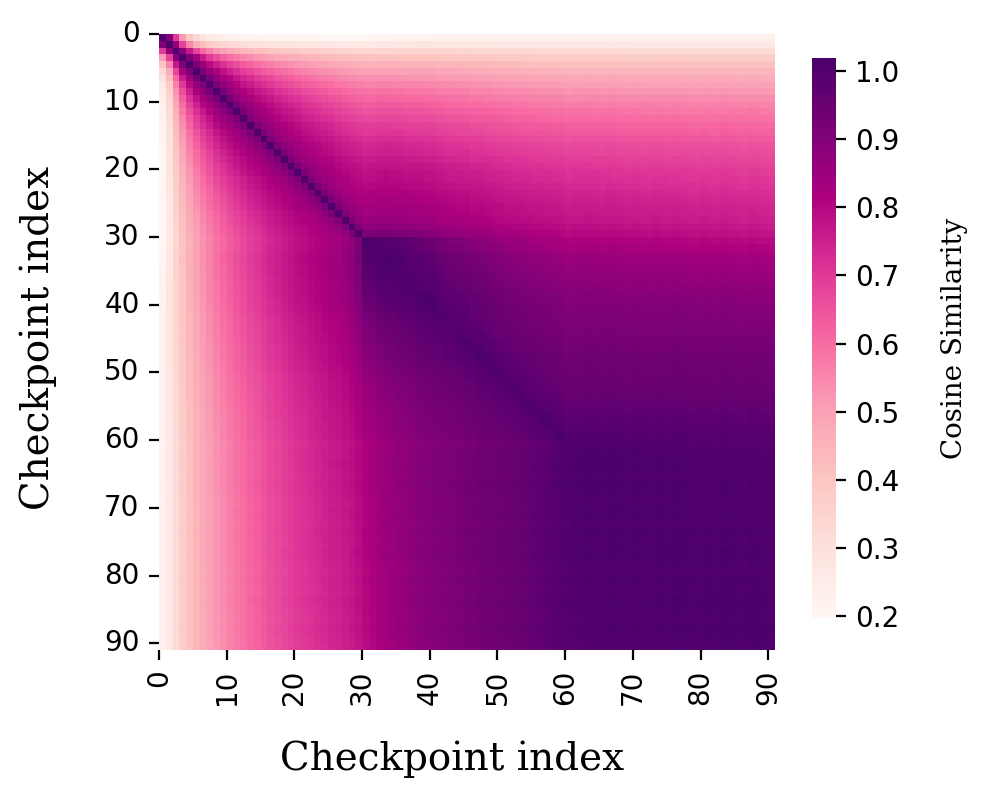}
\caption{Trajectory Map of ResNet50 on ImageNet, $\omega=0.764$\looseness=-1.}  
 \label{fig:rnteaser}
\end{wrapfigure}

\paragraph{Visual Tour of the Trajectory.} Before we start the hyperparameter excursion, let us look at the nature of the optimization trajectory, as visualized via the proposed qualitative hallmark, namely the trajectory map $\Cm$. In particular, we save epochly checkpoints from the initialization until the final epoch, and thereby giving us a total of $91$ checkpoints. We plot this $91\times91$ matrix in Figure~\ref{fig:rnteaser}. 

For starters, we can easily make out three\footnote{If we look closely, there seems to be another phase transition neighbouring the initialization and the subsequent couple of epochs, giving rise to a thin horizontal and vertical sliver of relatively lighter colour in this figure.\looseness=-1} distinct phases of optimization which are marked by an increased darkening of the pixels and their locations are precisely where learning rate decay was applied. The onset of these phases also seems to bring about an increased cosine similarity of the parameters contained within these phases, which seems to imply that following the learning rate decay the optimization is honing into a progressively confined subspace of the landscape. %

Next, the mean directionality score (MDS) for this particular figure comes out to be $\omega=0.764$. To contextualize this value, we should remind ourselves that we are working in a space of $\sim 25.6$ million dimensions. This should serve to emphasize that network optimization trajectories are highly structured, and not merely random points\footnote{In Section~\ref{app:random}, we analyze the relative trajectory map and MDS for a random walk/Brownian motion, and in comparison we find that (expectedly) the trajectory maps of neural networks are more directionally redundant.} in high-dimensions whose cosine similarity goes to zero. As a further note, the cosine similarity of two different instantiations of the ResNet50 parameters, from the same (and the usual) random initialization scheme, gives a value\footnote{A  reason for this high value of cosine similarity is the presence of BatchNorm or LayerNorm learnable scale parameters, which are typically initialized to all ones.} of $0.374$. However, besides these comparisons, it is still somewhat unclear what the value of MDS tells us about the trajectory. In particular, we need to ask ourselves (i) How does it rank in comparison to possibly other trajectories?  (ii) Is it a high enough value, or should we aim for something much larger? (iii) But since the maximum value of $\omega$ can be $1$ and since we have initialization in the mix here, obtaining such a value would not be so desirable as it would effectively mean the absence of any feature learning~\citep{chizat2020lazy}, which seems to be a critical component\footnote{Also, a certain amount of directional exploration is crucial as evident from our analysis of the grokking~\citep{power2022grokking} phenomenon discussed in Section~\ref{app:grokking}.} behind deep learning's success. \looseness=-1

\begin{figure*}[h]
	\centering    
	\includegraphics[trim=5 10 10 3, clip,width=0.85\textwidth]{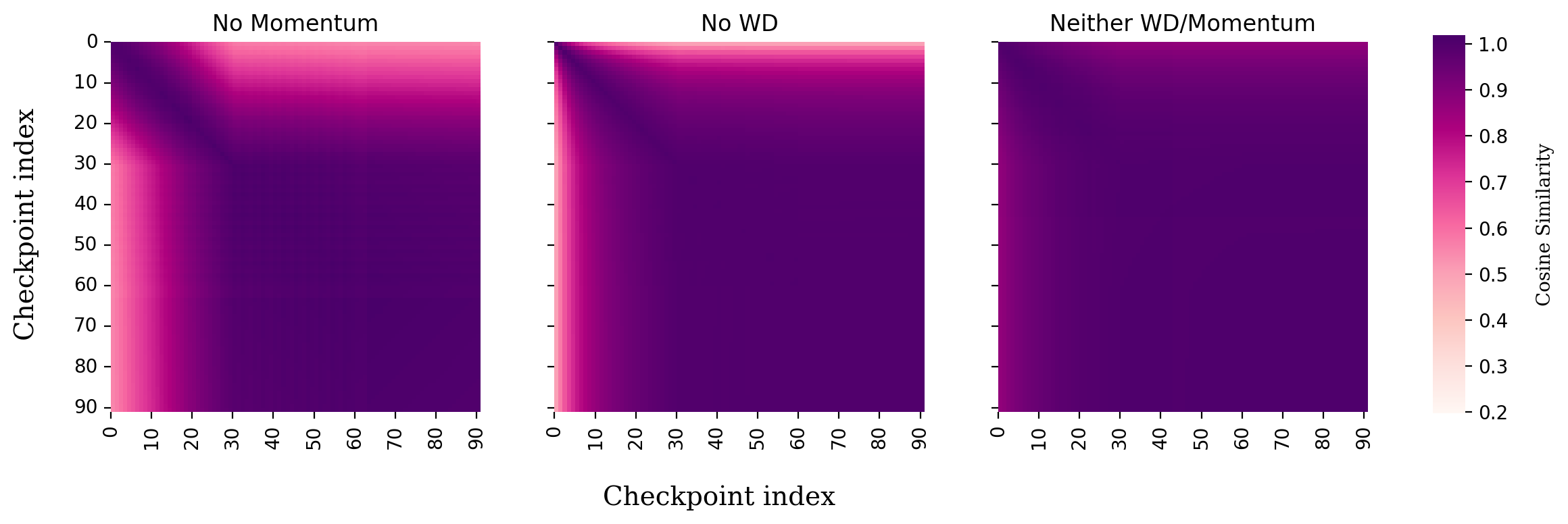}
	\caption{Trajectory Maps of ResNet50 trained on ImageNet without momentum or weight decay or both. The scale of cosine similarity is adjusted to match the values in Figure~\ref{fig:rnteaser} and allows for a direct comparison. The values of MDS are, $\omega=0.901, 0.931, 0.979$ respectively.}
 \label{fig:rn-other}
\end{figure*}

To better address these questions, let us now tinker slightly with certain hyperparameters. In particular, we will only touch just two of them, namely, momentum and weight decay, and turn them off one at a time. Importantly, everything else remains fixed. The ensuing trajectory maps can be found in Figure~\ref{fig:rn-other}, where either momentum is turned off ($\mu=0$) or weight decay ($\lambda=0$) or both ($\mu=0, \lambda=0$). The first thing that stands out is that the trajectory maps become darker in colour, suggesting an increased cosine similarity, and especially the subfigure without both momentum  and weight decay looks rather conspicuous, with essentially just one dark grid of extremely large cosine similarities. In particular, MDS here is $\omega=0.901, 0.931, 0.979$, respectively. This toggling of the hyperparameters not only has a significant effect on the trajectory maps but also their top-1 accuracies, which are $72.63\%, 70.86\%, 68.73\%$ respectively.%

These results would indicate that the optimization hyperparameters seem to be leading to qualitatively different solutions, with the presence of weight decay or momentum encouraging more directional exploration, whereas in their absence, the optimization latches on to a nearby solution (at least, in a directional sense). However, these results also seem to run against the  simplistic pictures one might conceive of momentum and weight decay. For instance, intuitively, the use of momentum should add strength to the previous gradient directions and lead to increased directional similarity, but we find the opposite to hold. Similarly, in the absence of weight decay, the network is not constrained to remain in a  ball around the origin and, in principle, there should be more license to explore in the landscape. Or, what are these intuitive notions not taking into account?

\section{A Ride with Momentum \& Decay}

In order to obtain a more refined understanding of the above observation, we would like to inspect in detail the angular and norm-based measures of the trajectory. We will largely\footnote{In the Appendix~\ref{app:direc-effects}, we also present a wider set of results, namely, the directional effects of hyper-parameters such as learning rate and batch size, recent regularizers like Sharpness-Aware Minimization~\citep{foret2021sharpnessaware} (SAM), as well as more datasets and architectures such as Vision Tranformers on ImageNet and VGG16 on CIFAR10, and for different amounts of label noise in the dataset.} focus on momentum and weight decay, given the especially intriguing directional nature of the trajectories caused by them. Now, let us start by taking a look at how aligned are the net or the aggregate updates\footnote{We qualify this by `net' or `aggregate' as we are working on somewhat coarser granularity (i.e., $1$ epoch) than every update or step. But our current granularity is still rich enough to allow for the presented trends to persist even if go $2\times$ to $5\times$ more coarser.} in the presence of these hyperparameters and otherwise.

\begin{figure*}[h!]
	\centering
\vskip -0.2in
	\subfigure[$\angle(\theta_{t+1}-\theta_t,\theta_t-\theta_{t-1})$ ]{\label{fig:updates_angle_rn1}
	\includegraphics[width=0.23\textwidth]{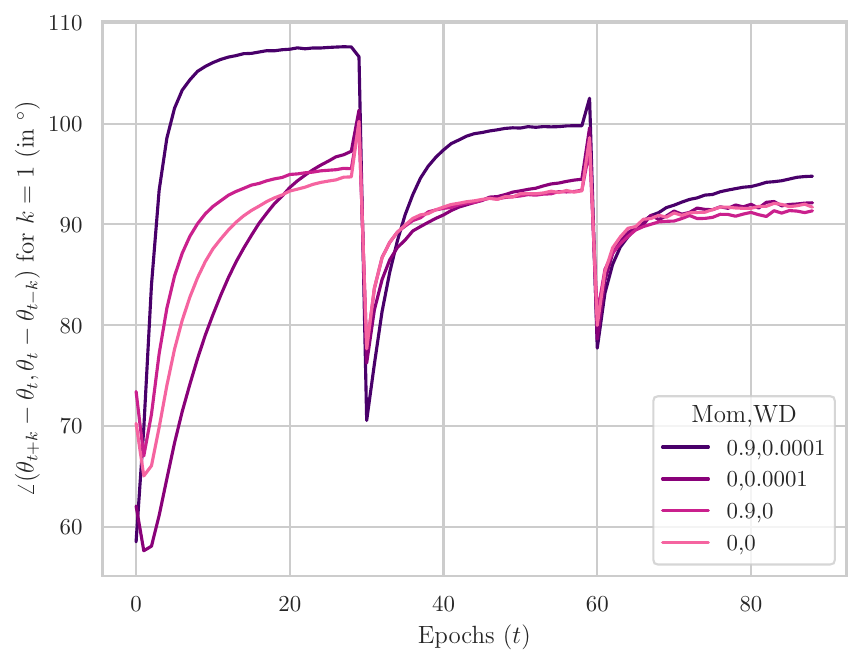}
	\vspace{-4mm}}
			\subfigure[$\angle(\theta_{t+1}-\theta_t,\theta_t)$]{\label{fig:update_loc_angle_rn1}
		\includegraphics[width=0.23\textwidth]{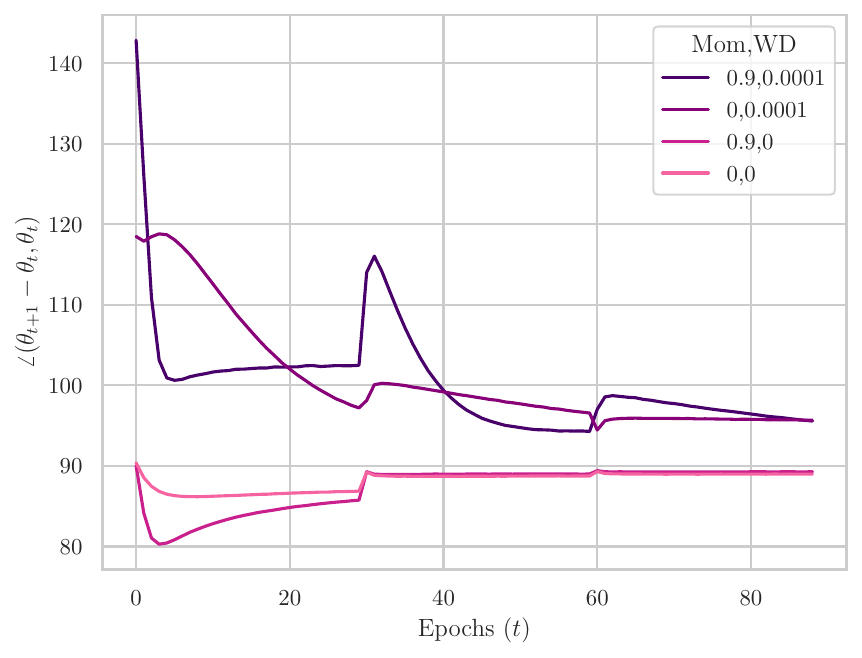}}
	\subfigure[ $\angle(\theta_t,\theta_0)$]{\label{fig:rn-apex-origin}
	\includegraphics[width=0.23\textwidth]{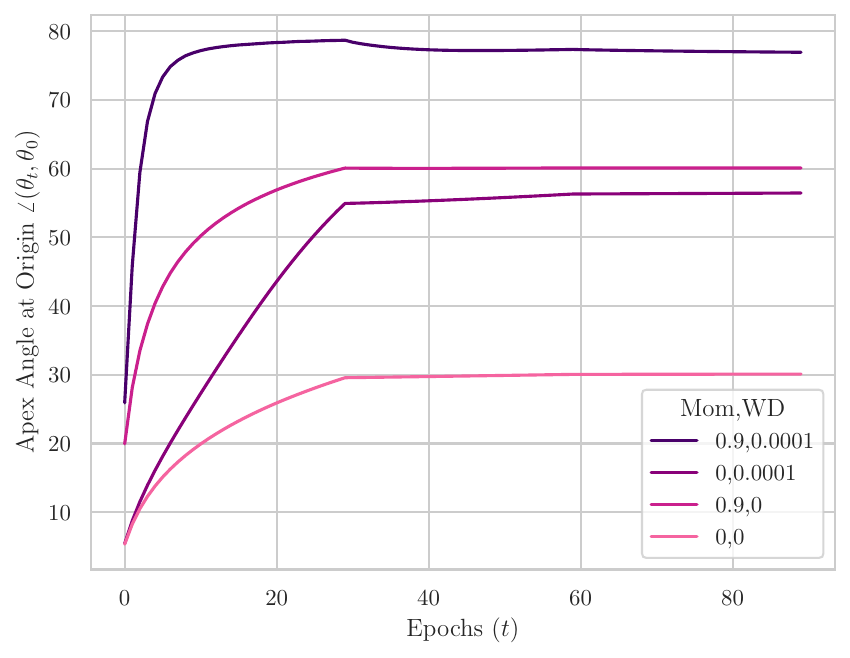}
	\vspace{-4mm}}
	\subfigure[ $\angle(\theta_t-\theta_0,\theta_1-\theta_0)$ ]{\label{fig:rn-apex-init}
		\includegraphics[width=0.23\textwidth]{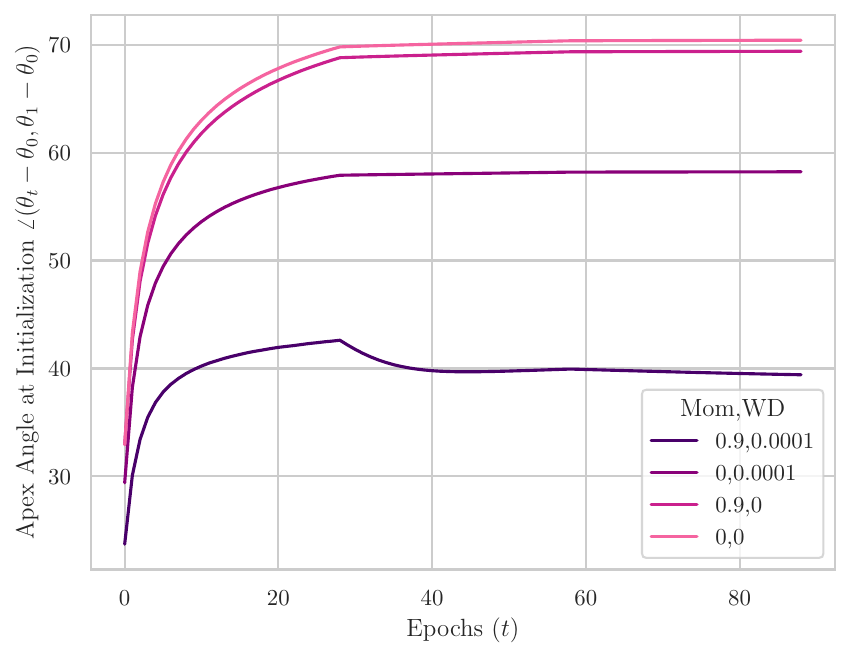}
		\vspace{-7mm}}
				\subfigure[Illustration of EoS with Momentum]{\label{fig:eos}
			\includegraphics[trim=0 5 0 40, clip,width=0.2\textwidth]{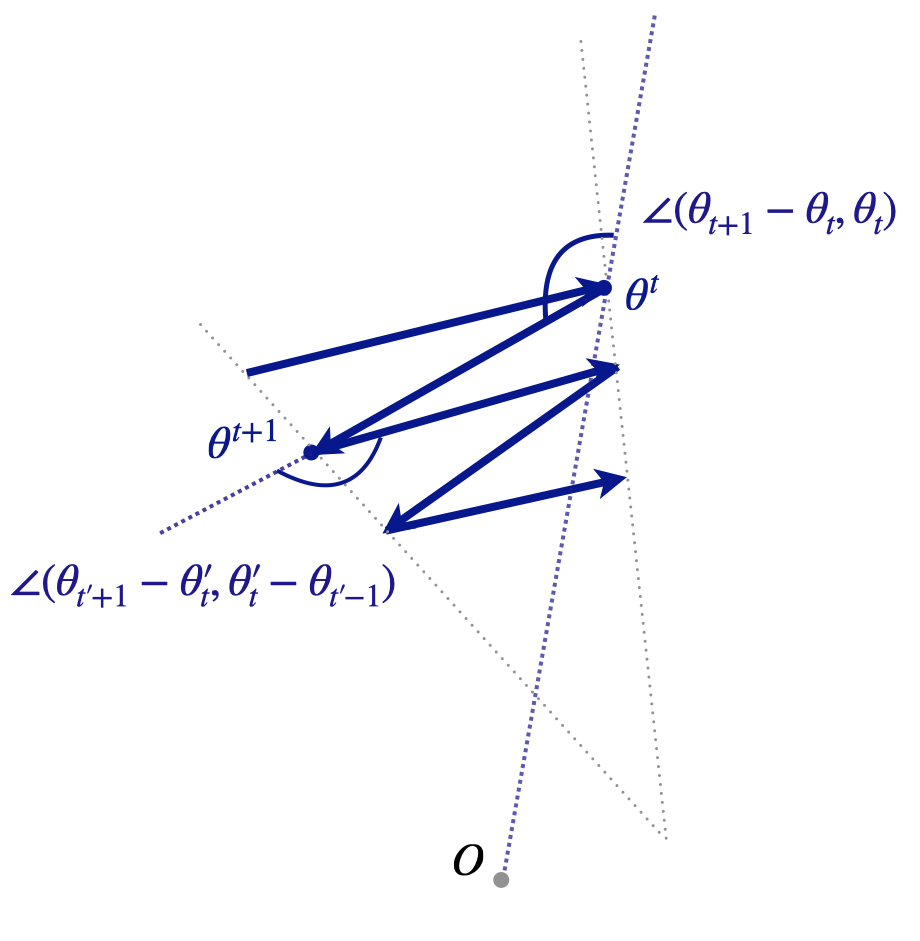}
			\vspace{-4mm}}
			\subfigure[$\|\theta_t\|_2$]{\label{fig:rn-par-norm}
			\includegraphics[width=0.23\textwidth]{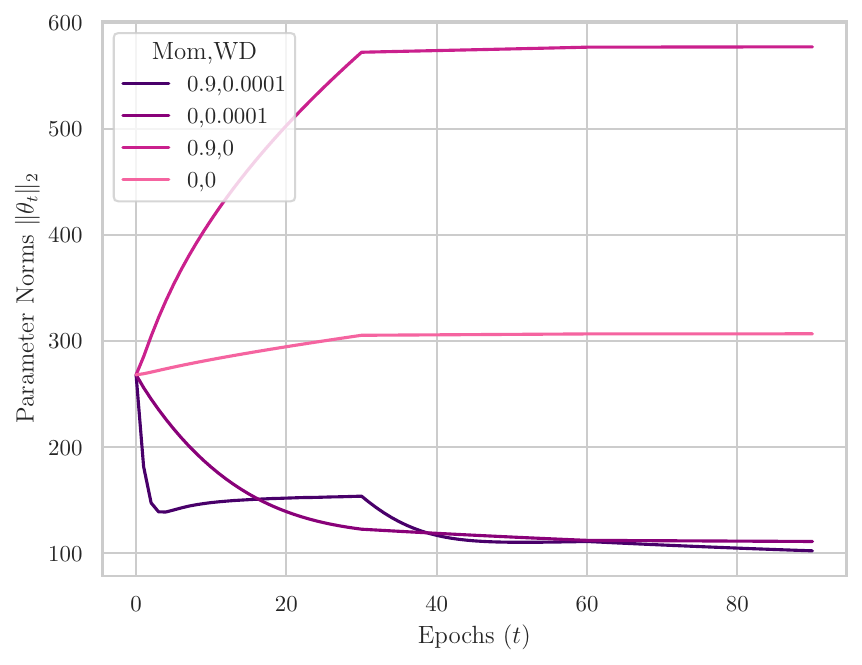}}
		\subfigure[$\|\theta_{t+k}-\theta_t\|_2$ ]{\label{fig:rn-update-norm}
	\includegraphics[width=0.23\textwidth]{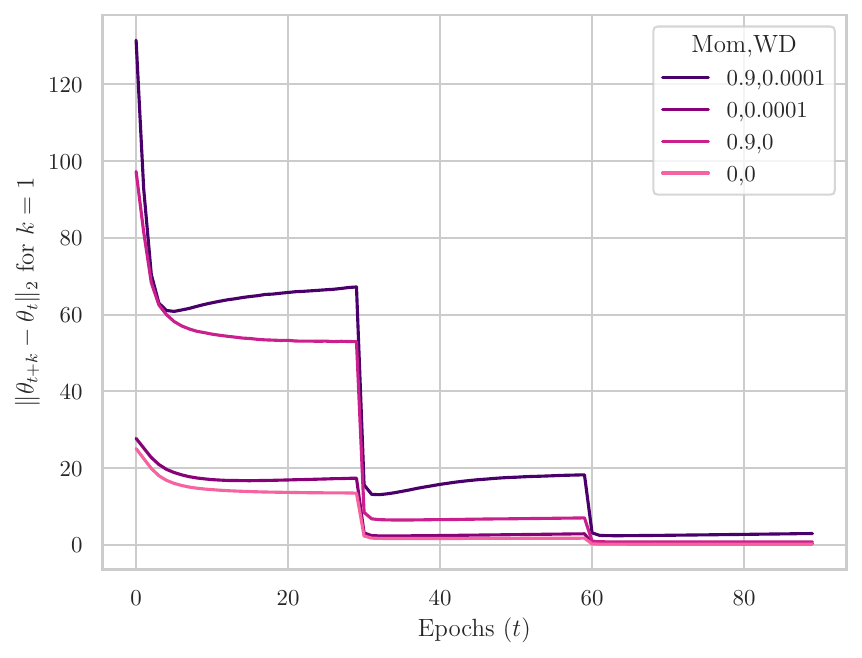}}
		\subfigure[$\|\theta_t-\theta_0\|_2$ ]{\label{fig:rn-dist-init}
	\includegraphics[width=0.23\textwidth]{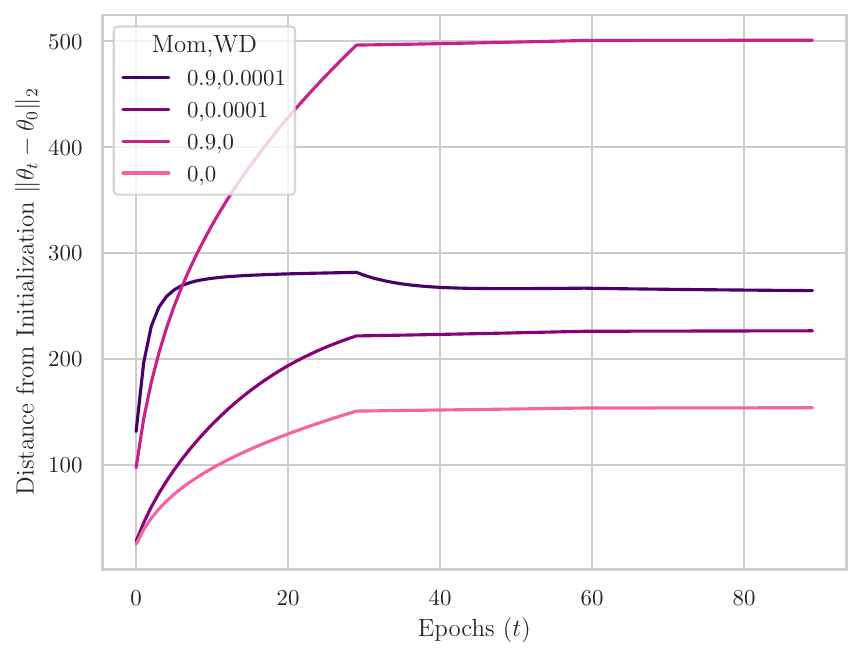}
			\vspace{-7mm}}
	\caption{Angular and Norm-based measures: ResNet50, ImageNet, Momentum and Weight Decay} 
\end{figure*}

\subsection{Momentum and the Angle between Updates} 
In particular, in Figure~\ref{fig:updates_angle_rn1}, we plot the angle between consecutive epochs, i.e, $\angle(\theta_{t+1}-\theta_t,\theta_t-\theta_{t-1})$. Interestingly, we find that this angle becomes obtuse a short while into the training process, and further, this angle is larger when momentum is turned on versus when off. Moreover, rather visibly, this increase in angle is larger when weight decay is also enabled, suggesting that weight decay and momentum are closely intertwined.  Taken as such, this observation would point as to how the MDS increases when these hyperparameters are switched off.  To gain a somewhat better understanding of this mechanism,
we turn to the simplest and oft-employed model of a quadratic problem. \looseness=-1

\begin{lemma}\label{lemma:angles} Given a quadratic problem with $\ell_2$ regularization of strength $\alpha>0$, namely, $\min_{\btheta\in\Reals{d}} \frac{1}{2} \btheta^\top \Mm \, \btheta + \frac{1}{2} \alpha \|\btheta\|^2\,,$ with $\Mm\in\Reals{d\times d}$ symmetric with eigenvalues $\lambda_1\geq\cdots \geq\lambda_d$, the angle between successive steps $\Delta_{t}=\btheta_{t} - \btheta_{t-1}, \Delta_{t+1}=\btheta_{t+1}-\btheta_{t}$, when using gradient descent with a one-step momentum ($\mu>0$) and learning rates $\eta_{t}, \eta_{t+1}$, can be upper and lower bounded as follows:\looseness=-1
	\begin{align*}
&  \langle\Delta_{t}, \Delta_{t+1}\rangle   \leq \eta_{t} \eta_{t+1} (1- \eta_{t}(\mu +\alpha + \lambda_d)) (\lambda_d + \alpha)^2  \|\btheta_{t-1}\|^2  \\
&  \langle\Delta_{t}, \Delta_{t+1}\rangle  \geq \eta_{t} \eta_{t+1}(1- \eta_{t}(\mu+\alpha + \lambda_1)) (\lambda_1 + \alpha)^2 \|\btheta_{t-1}\|^2
	\end{align*}

\end{lemma}

The proof, in Appendix~\ref{app:proof}, inherently considers the solution at $\opt=\mathbf{0}$, but if that is not the case, we can substitute it in the objective and our derived bounds would scale in the squared distance to the solution, i.e. $\|\btheta_{t-1}-\opt\|^2$. Besides, in the above proof, we consider a one-step momentum, which inherently means resetting the momentum after every $2$ steps. This is done for convenience, as our main purpose is to anyways gain insights into the phenomenon and not provide its ultimate proof. 

Turning to the bounds themselves, notice that if the learning rate $\eta_{t}\geq 1/(\lambda_1 + \mu + \alpha)$, the lower bound will turn negative and will be multiplied by a factor of $(\lambda_1+\alpha)^2 \|\btheta_{t-1}\|^2$. On the other hand, although the first term of the upper bound might still be positive, importantly, it is scaled by a factor of $(\lambda_d + \alpha)^2 \approx \alpha^2$ for matrices $\Mm$ which are close to degenerate ($\lambda_d\rightarrow0$). \looseness=-1

\paragraph{Low-rank Hessian and Edge of Stability.} In our context, the Hessian of the loss with respect to the parameters will play the role of the matrix $\Mm$, since we can assume a second-order Taylor series will hold across the two steps. But it is also known through prior empirical work that the Hessian is significantly degenerate~\citep{sagun2017empirical}, which has also been proven rigorously for deep linear fully-connected and convolutional networks~\citep{singh2021analytic}. Furthermore, this requirement on the learning rate $\eta_{t}$ is actually looser than the adaptivity of the largest eigenvalue of the Hessian to the learning rate $\lambda\approx \frac{2}{\eta}$, as shown in the recent work on Edge of Stability (EoS)~\citep{cohen2022gradient}.

\paragraph{Explaining the Obtuse angles.} Owing to these facts, we will have that $\lambda_d(\Mm)\approx0$, and which further implies that the upper bound on the inner-product between the updates will be approximately zero, and the lower-bound will be large in absolute value but negative. Therefore, this explains how the angles between consecutive epochs can be obtuse.  More broadly, the obtuse angle indeed implies that there are oscillations, especially along the direction of the largest Hessian eigenvector. Further, from Lemma~\ref{lemma:angles}, we see that the magnitude of the inner-product of the updates scales in proportion to $\eta_{t+1}$. Hence, a way to dampen the oscillations\footnote{These oscillations need not necessarily translate into gross instabilities at the level of the loss, since as can be seen in Figure~\ref{fig:rn-update-norm}, the update norms progressively shrink in each of the three learning rate phases. } is to decrease the learning rate, and as can be seen in Figure~\ref{fig:updates_angle_rn1}, the learning rate decay at epochs $30$ and $60$ is followed right after with the angles turning from obtuse to acute. Lastly, here in the constraint on the learning rate (the additive terms $\alpha$ and $\mu$), we can also see momentum and weight decay go hand-in-hand, each accentuating the effect of the other.\looseness=-1

\paragraph{Towards a holistic picture of Momentum.} Besides, in Figure~\ref{fig:rn-apex-origin} and~\ref{fig:rn-apex-init}, we find that in the presence of momentum, a larger angle is traced at the origin by the trajectory, suggesting a more directional exploration, while the angle traced at initialization is smaller. The latter can also be seen from Figure~\ref{fig:rn-dist-init}, since with momentum, the trajectory moves further away from the initialization. Apart from this, in the absence of weight decay, the updates seem to be strengthening with momentum and the parameter norm rises~\ref{fig:rn-par-norm} as well, giving rise to a mental picture of a trajectory similar to that left purple trajectory in Figure~\ref{fig:angular}, at least until the training hits EoS. 

With weight decay, as there is a decrease in parameter norm Figure~\ref{fig:rn-par-norm} alongside the EoS process, as well as due to the presence of larger obtuse angles, we expect a reasonable affinity with our illustration in Figure~\ref{fig:eos}, where we see the updates oscillating and slowly drifting towards the origin $O$ below.\looseness=-1

\subsection{More than just Weight Decay}

\begin{figure*}[h!]
	\centering
	\includegraphics[trim=5 7 10 3, clip,width=0.9\textwidth]{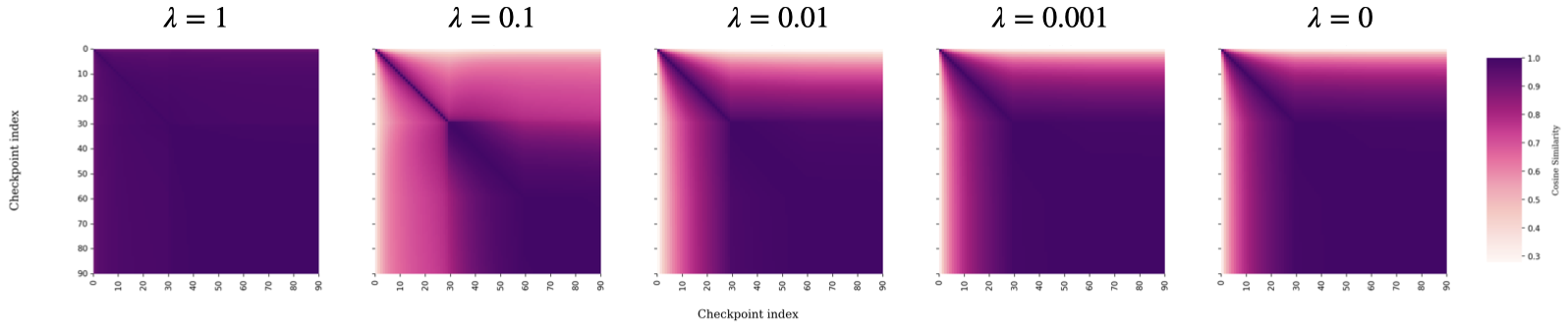}
	\caption{Relative Trajectory Maps (wrt. initialization) of ResNet50 models for different weight decay.\looseness=-1} 
 \label{fig:tm-wd}
\end{figure*}
\begin{wrapfigure}{R}{0.35\textwidth}
	\centering    
	\includegraphics[trim=0 4 4 0, clip,width=0.3\textwidth]{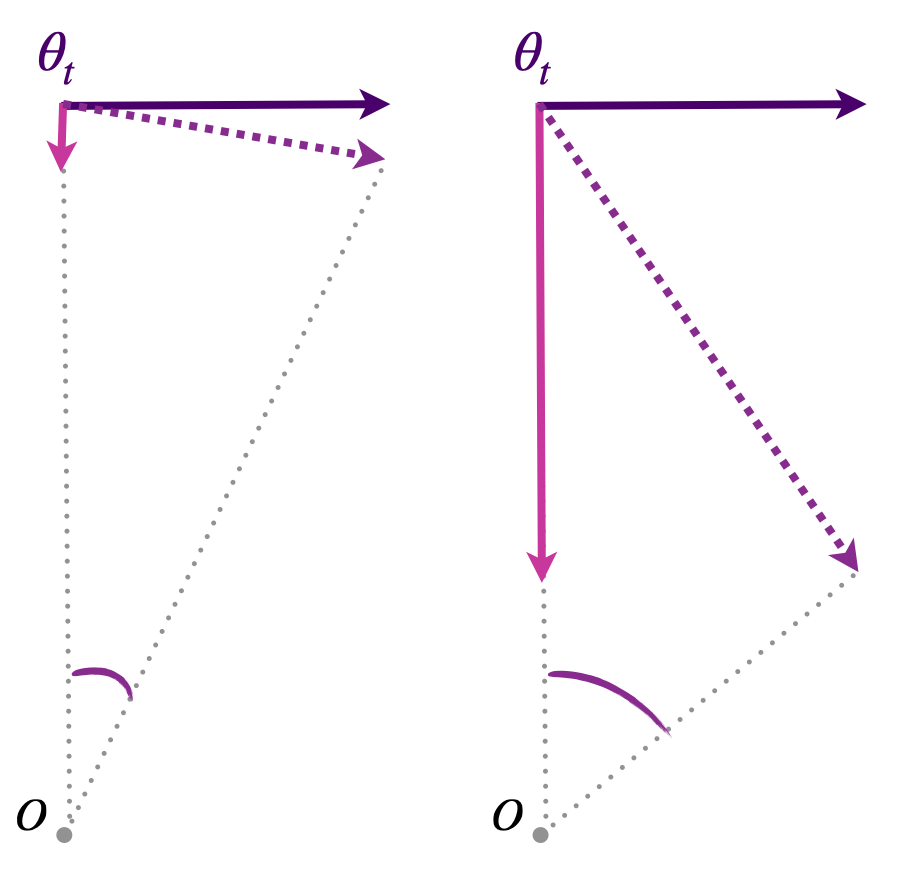}
	\caption{\textbf{Directional exploration effect of Weight Decay. }The downward vector represents the pull towards the origin (O) due to weight decay, while the rightward vector the force due to the loss \looseness=-1.\label{fig:dir-pic-wd}} 
\end{wrapfigure}

 Now that we have gained a richer understanding of momentum and its interaction with weight decay, let us turn to weight decay alone and understand its directional effect. We have already noticed the increase in mean directional similarity (MDS) when weight decay is disabled for ResNet50 trained with SGD on ImageNet. In fact, we find a similar effect with an adaptive optimizer, like AdamW~\citep{loshchilov2017decoupled} --- the trajectory maps for which are shown in Figure~\ref{fig:tm-wd}. Here, we used regularization constants from $\lambda=0$ until the first value where we witness a decrease in test performance, which in this case was $\lambda=1$. Specifically, we analyze the weight decay coefficients in $\lambda\in\lbrace1, 0.1, 0.01, 0.001, 0 \rbrace$. The corresponding MDS come out to be, $\omega=0.731, 0.679, 0.844, 0.882, 0.885$. We notice that, as before, increasing weight decay leads to a heightened directional exploration, or lower MDS; except $\lambda=1$ being the seeming anomaly.\looseness=-1

But we find that this can be remedied simply by looking at the relative trajectory maps (Figure~\ref{fig:tm-wd}), and computing the relative MDS, i.e.,  $\omega_0$ is $0.985, 0.807, 0.862, 0.897, 0.900$ for $\lambda=1, 0.1, 0.01, 0.001, 0$ respectively. This occurs since such a high weight decay $\lambda=1$, causes this particular network to underfit (train/test top-1 accuracy are $54.63\%, 50.52\%$).  The performance for the rest of the networks improves, more or less, as expected with weight decay, and in particular, achieve accuracies of $75.45\%, 73.38\%, 71.03\%, 71.41\%$.

Having reaffirmed our results extensively about the directional exploration due to weight decay, we can understand it through a simple physics-based intuition, as shown in the Figure~\ref{fig:dir-pic-wd}. In particular, we can think of the loss gradient pulling the network parameters rightwards, while the force exerted by weight decay tries to pull the network downwards. The relative strengths of these two `forces' have been represented by the lengths of the two vector arrows. We notice that as the weight decay strength is increased, from the left subfigure to the right, the angle traced at the origin (O) also increases. This explains how weight decay can contribute towards directional exploration.\looseness=-1

\section{Hauling Trajectory Hallmarks for LLMs and their model internals}

\begin{figure*}[h!]
	\centering    
	\includegraphics[trim=0 5 0 0, clip, width=0.95\textwidth]{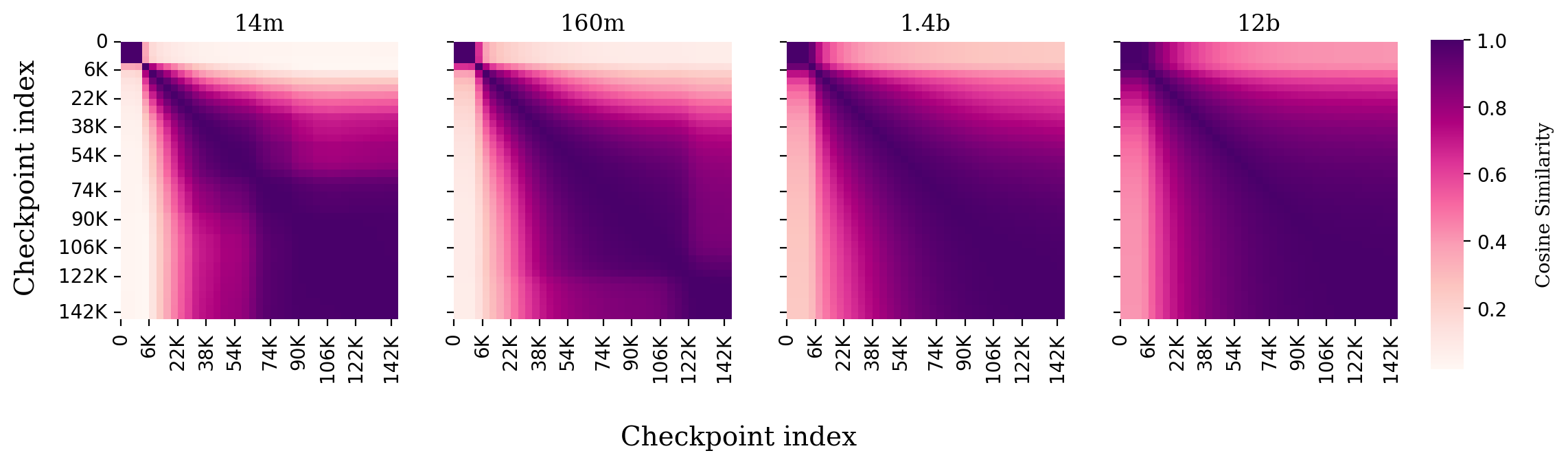}
	\caption{Trajectory Maps of Pythia GPT-NeoX models across two orders of scales trained on Pile. The corresponding MDS $\omega=0.650, 0.678, 0.759, 0.815$.} \vspace{-3mm} 
    \label{fig:cospythia}
\end{figure*}

With all recent interest in Large Language Models (LLMs) and their scaling, a natural question is whether the structure of trajectories exhibited before is shared in the case of language modelling tasks and across models of different sizes. %
Besides, while we expect increasing parameter count to provide new directions for learning, does scale make the optimization trajectories complex or does it instead regularize them? 

Thanks to Pythia's~\citep{biderman2023pythia} publically released model checkpoints over training, for GPT-NeoX~\citep{black2022gpt} models --- ranging in sizes from $14$ Million (M) to $12$ Billion (B) --- we can provide answers to the above questions. Given that processing all the available checkpoints would require several terabytes of cache, we select every fourth checkpoint, resulting in $39$ checkpoints that we analyze for models of sizes: $14$M, $70$M, $160$M, $410$M, $1.4$B, $2.8$B, $6.9$B, $12$B. The results for a shortlist of these experiments can be found in Figure~\ref{fig:cospythia} (for more, see Figure~\ref{fig:cospythia-full}).\looseness=-1

First, we note that there is a tiny square grid in the upper-left corner, which delineates precisely the learning rate warmup phase. Next, the subsequent larger grid starts out, for $14$M, with distinct subgrids but then with increasing model scale takes a funnel-like shape around $1.4$B, before becoming rather homogeneous by $12$B parameters. Even the horizontal and vertical slivers corresponding to warmup and the rest of the epochs start to assume a higher cosine similarity with scale. Overall, increasing scale lends an intense dark hue to the trajectory maps, which seems to suggest that the inductive bias of scale might be related to regularizing the trajectories.

\paragraph{Why do parameters become aligned with scale? A Theoretical Argument.}
 We prove, in Appendix~\ref{sec:arg}, that this surprising finding about the progressive increase in cosine similarity with scale has a relatively simple explanation, at least in the case of the large-width limit of deep networks. The gist of our argument is that \textit{in the large width limit, any parameter updates that lead to stable feature updates must necessarily yield updated parameters that are identically aligned with their initialisation. }This is a well-known fact for lazy learning regimes like the Neural Tangent Kernel~\citep{jacot2020neural} or standard parameterisations, where no feature learning occurs. What may be surprising is that this is necessarily true for feature learning regimes like $\mu$P~\citep{yang2022tensor}. \looseness=-1

\begin{figure*}[h!]
{\centering
  \includegraphics[trim=0 0 10 0, clip, width=0.95\textwidth]{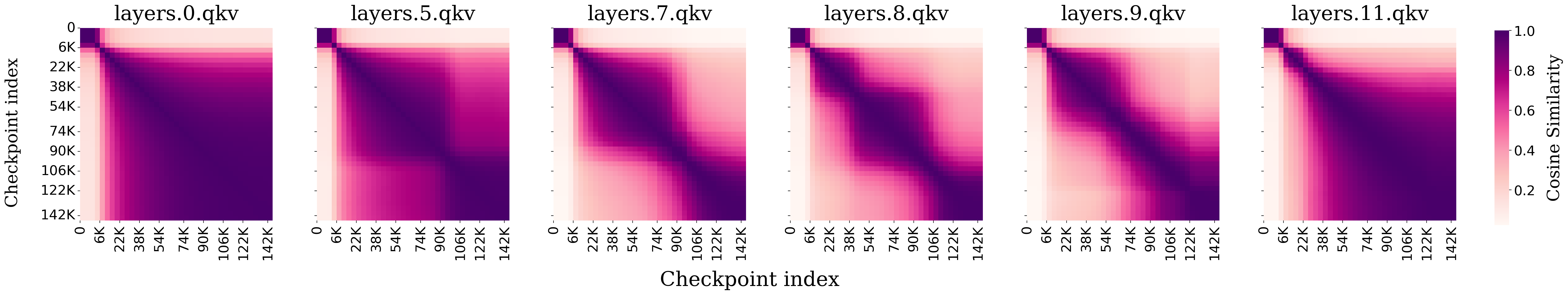}
	\caption{Layerwise Trajectory Maps for Q,K,V weights across depth fot GPT-NeoX trained on the Pile for the $160$M model. This suggests that the Q,K,V parameters of the middle layers seem to be converging last, and  differ from earlier and later layers in this regard.}
 \label{fig:mech}}\vspace{-3mm}
\end{figure*}

\paragraph{Layerwise Q,K,V dynamics homogenize over depth at $ \sim 1$B scale and over.}
Another interesting aspect of the structure of trajectory maps for the $14$M and $160$M cases is their relative heterogeneity (distinct subgrids), as compared to the billion parameter models. This heterogeneity cannot be explained away through the settings of a learning rate scheduler (which is just a cosine scheduler), so we inspect the layerwise trajectory maps (i.e., by building it over parameters of each layer separately). In particular, we find an intriguing heterogeneous structure for these models that is most starkly present in the query-key-value (Q,K,V) parameters of the attention layers, especially the middle layers, as shown in Figure~\ref{fig:mech}. 

\begin{figure}
    \centering
    \subfigure[$14$ M]{\includegraphics[width=0.9\textwidth]{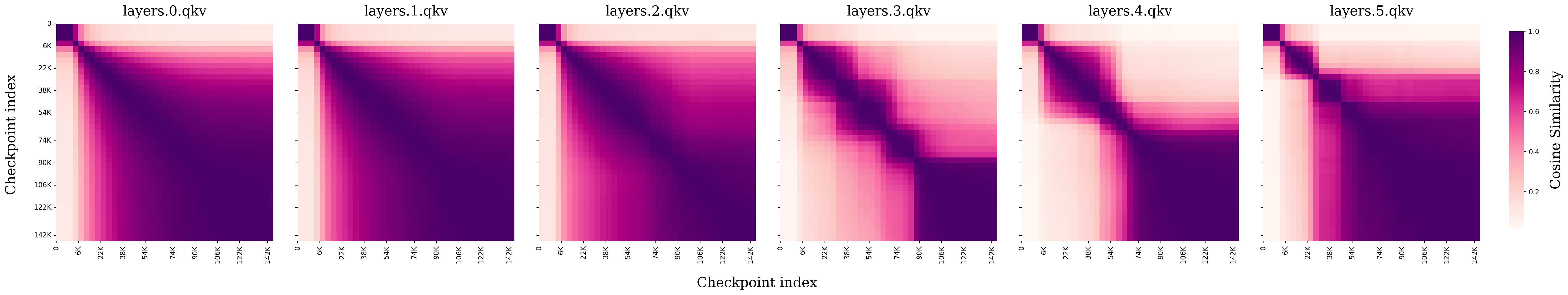}}
    \subfigure[$160$ M]{\includegraphics[width=0.9\textwidth]{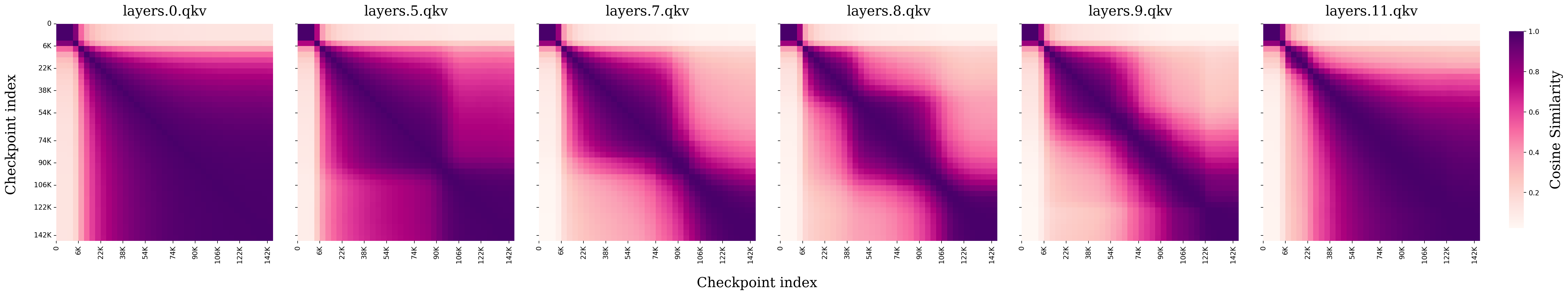}}
    \subfigure[$1.4$ B]{\includegraphics[width=0.9\textwidth]{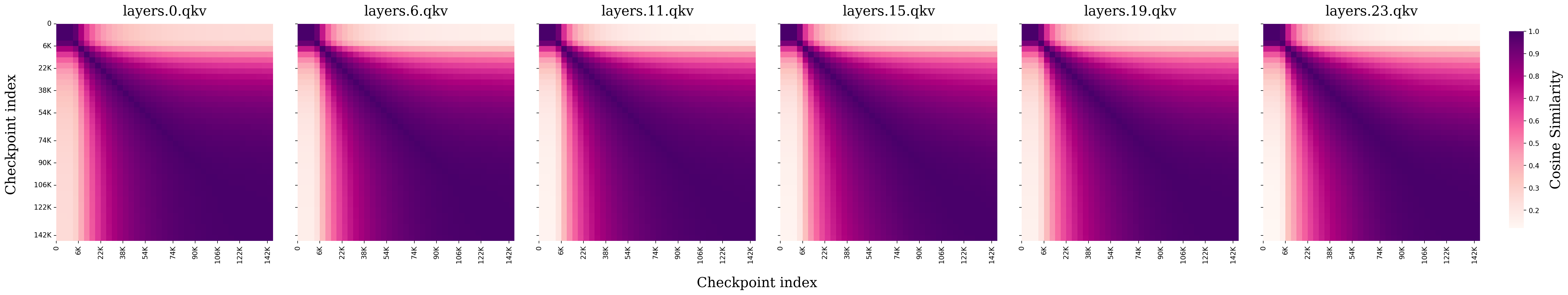}}
    \subfigure[$12$ B]{\includegraphics[width=0.9\textwidth]{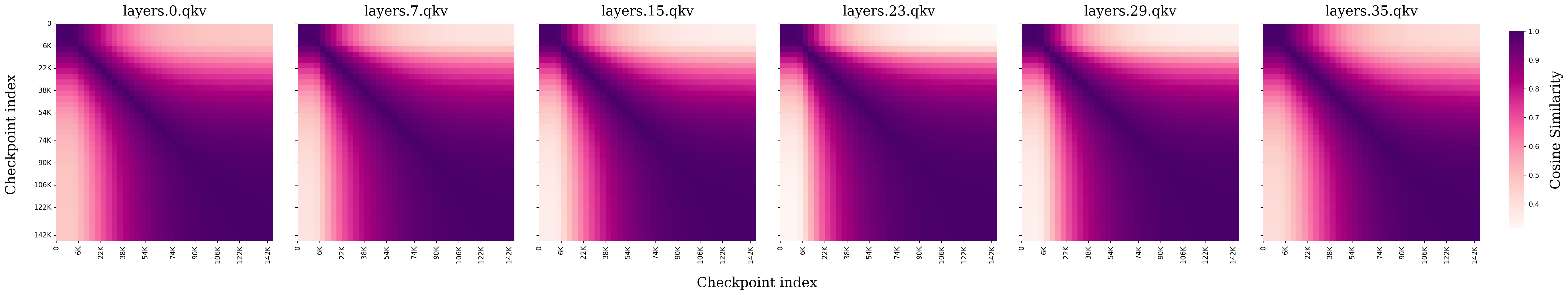}}
    \caption{Trajectory maps of Q,K,V layers become homogenized over increasing scale.}
    \label{fig:qkv-scales}
\end{figure}

This indicates that the Q,K,V parameter dynamics converge at different timescales, with the middle layers converging the last directionally  in contrast to earlier and later layers. Moreover, we find that \textit{scale has the striking effect of homogenizing 
the Q,K,V dynamics}, as shown in Figure~\ref{fig:qkv-scales}, with the layerwise trajectory maps essentially resembling the network-wide trajectory map structure in Figure~\ref{fig:cospythia}. \looseness=-1

\section{Putting Directional Redundancy to the Test}
Our trajectory map analyses reveal significant directional redundancy in the optimization trajectories of neural networks, which is especially prominent later into training. This raises the question of whether this redundancy can be leveraged for creating efficient hybrid optimization schemes or if it conceals minute but crucial directional changes.
Therefore, to test the true nature of directional redundancy, we consider optimizing only some layerwise scalar parameters after a little while into training. Further, instead of attaching extra scalar parameters per neuron or channel and tuning them, we will repurpose the scalar weight and bias parameters present in the ubiquitous (batch/layer) normalization layers in modern network architectures. We are inspired by the prior work of~\citet{frankle2021training} who demonstrated the remarking expressivity of training just the scalar parameters in batch-normalization (BN) layers in the entire network.%

\begin{figure}[h!]
	\centering
	\includegraphics[trim=5 7 10 3, clip,width=0.85\textwidth]{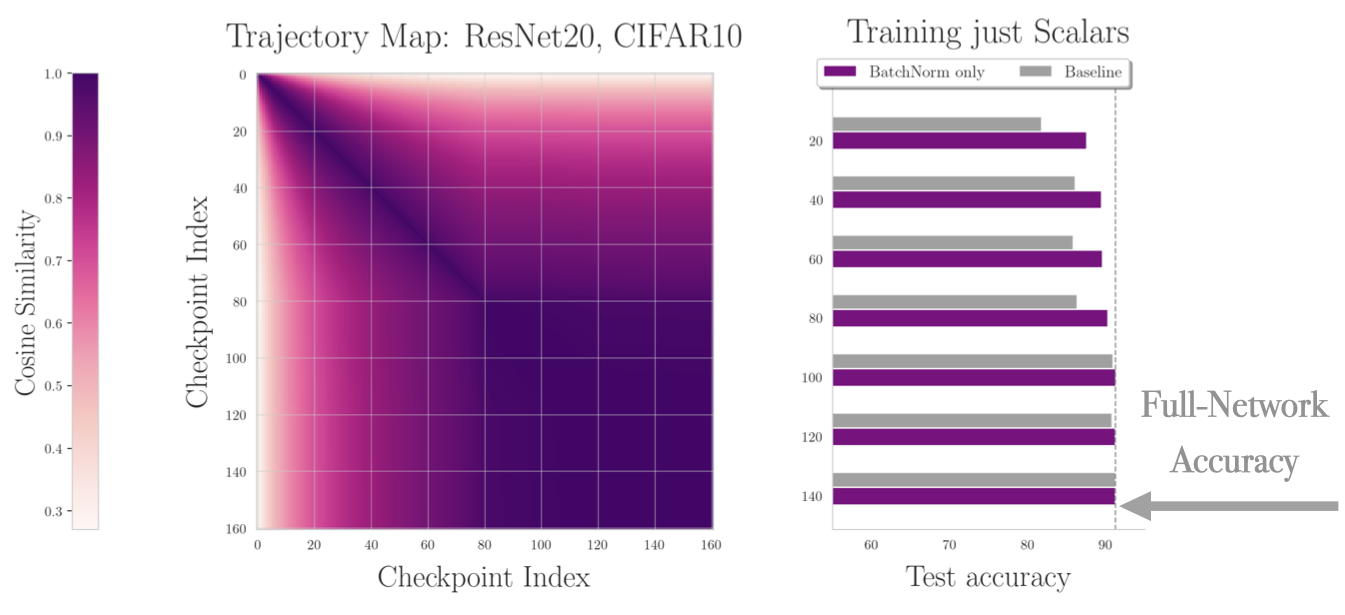}%
	\caption{\textit{Directional Redundancy put to test by training only the BN scalar parameters.} The trajectory map of the original, full network, training is shown on the left and, in the bar chart to the right, the bars denoting the performance achieved by training BN parameters from a particular epoch are horizontally aligned with the corresponding trajectory map rows. `Baseline' denotes the network accuracy just before switching to optimizing the BN parameters only\looseness=-1.}  \vspace{-2mm}
    \label{fig:direc-redun}
\end{figure}

\paragraph{CIFAR10 experiments. } We experiment with ResNet20 on CIFAR10 using SGD over 160 epochs, freezing non-BN layers at different points and training only the $1,376$ scalar parameters in the BN layers. Figure~\ref{fig:direc-redun} presents the results alongside the trajectory map of the full network training.

When right from the initialization, just the BN layers are trained, as claimed in~\citet{frankle2021training}, the resulting network achieves a test accuracy of $54.8\%$, which although quite interesting falls nevertheless considerably short of the $91.2\%$ test accuracy obtained by training the entire network. In a way, from the horizontal strip of trajectory map around $0$, which pales as we move from the left to right, we see that the full network parameters at the end of training deviate significantly in their directionality as opposed to the parameters at initialization (the cosine similarity being $\sim 0.3$). However soon after, from around $40$ epochs, where the trajectory map starts developing a dark hue (and the cosine similarity to the final parameters climbs to about $\sim 0.9$), training BN parameters alone brings us to within $2\%$  of the full-network accuracy; 
and from around $80$ epoch gets to within $1\%$, and completely matches thereafter. Notably, this feat is \textit{remarkably achieved by training only $0.5\%$ of the overall parameters,} and this fares even better than training all the parameters in the bulkier last layer as shown in Figure~\ref{fig:tm-bn-last}. 

\paragraph{ImageNet experiments. }Likewise, in our experiments on ResNet50 trained on ImageNet as in Figure~\ref{fig:rnteaser}, we found that although training BN layers right from initialization gets us to an accuracy of just $\sim6.4\%$ (a paltry amount, given the full-network's accuracy of $\sim76\%$), but training it from epoch $30$ gets to within $\sim 10\%$ of the full-network's accuracy and from epoch $60$ to within about $\sim 2\%$ of it, i.e, to $\sim73.2\%$ top-1 test accuracy.  This is again a striking result, since when training the BN parameters of ResNet50, only $0.18\%$ of the parameters are being used and $99.82\%$ of the parameters are kept frozen. 

It should be mentioned that in the above BN training procedure, although the forward and backward propagation runtime costs end up being similar to training the entire network, this still leads to savings in the GPU memory consumption as the optimization buffers (like the momentum buffer or those used for preconditioning in adaptive methods like Adam) now need to be of significantly smaller size (typically $< 1\%$, i.e.,  $99\%$ savings) to accommodate only the BN parameters and could, in turn, make larger batch sizes feasible. More broadly, we foresee that this idea can be readily adapted into a hybrid optimization scheme, where regular training can be interleaved with memory-light training of normalization layer parameters, and where the latter is done on cheaper GPU instances in the cloud, {hence contributing to non-trivial cost savings. }All in all, these findings pointedly show that the observed directional redundancy from the trajectory maps truly manifests during training and \textit{can be potentially utilized to reap practical gains. }

\section{Related Work}
\paragraph{Directional Convergence.} Prior work has theoretically noted a notion of directional convergence~\citep{ji2020directional}, wherein the parameters of simple networks and classifiers converge quickly, in terms of their direction. Likewise~\citep{merrill2020effects} have observed that cosine similarity between subsequent parameter checkpoints during T5~\citep{raffel2023exploring} pre-training rapidly approaches one. Our analysis of the trajectory maps can thus be seen as related to this aspect, however, we carry out a comprehensive analysis of how directional similarity of the parameters evolves over the course of training, how it behaves in a variety of relevant settings triggered by enabling or disabling key hyperparameters, and how it varies across models of unprecedented scales. Moreover, our analysis of the trajectory via the quantitative and qualitative hallmarks provides a more refined and insightful picture of the nature of optimization trajectory, instead of the mere fact about the rapid directional convergence noted in prior works. 

\paragraph{Implicit Effects of Hyperparameters.} 
Implicit bias~\citep{gunasekar2018characterizing,li2019algorithmic,li2020explaining,moroshko2020implicit} has emerged as one of the main contenders for explaining the success of deep neural networks. This principle has also inspired several works which seek to uncover the implicit effects that might be latent in the regular working of hyperparameters. To name a few,  \citet{andriushchenko2023need} for instance suggest a loss stabilization mechanism behind weight decay, while~\citep{liu2023implicit} attempt to characterize the implicit bias of large learning rates in terms of resulting in a flatter solution, and~\cite{jelassi2022towards,cao2023implicit} explain the implicit bias of momentum and batch normalization with regards to margin. In contrast,  we take a broad and general perspective by adopting a full trajectory view. In particular, we explore more deeply an intriguing interaction of various optimization knobs, such as that between momentum and weight decay, and show its link with Edge of Stability~\citep{cohen2022gradient}.

\paragraph{Mechanistic Understanding of Neural Networks and LLMs.} Besides, our analysis can also pave the way for a novel data-free mechanistic understanding of LLMs (as shown in Figure~\ref{fig:mech}. This holistic optimization path perspective can complement the top-down approach of mechanistic understanding of LLMs via influence functions~\citep{grosse2023studying} and the bottom-up circuit view of Transformers~\citep{elhage2021mathematical}.

\section{Conclusion}

Overall, we have merely scratched the surface of this trajectory perspective into understanding optimization behaviour in neural networks. We genuinely believe that there is a lot to be understood about the complex, intermingled behaviour of optimization in deep learning, and hopefully, this work will bring further nuance in these areas and contribute towards hybrid optimization schemes that can exploit the showcased directional redundancy.%

\textbf{Limitations and Future Work.} While we have focused on the directional aspects of trajectories, the length of the trajectories also holds relevance, especially when the experiments under comparison may have unequal number of sampled checkpoints. We expect a version of MDS, which is weighted by the step lengths to be useful in such a scenario. Besides, our experiments with training scalar parameters are currently based on vision models, given the excessive resources involved in testing this for LLMs. However, we hope that the wider community can seize on this observation and exploit it for training LLMs more efficiently. 

\begin{ack}
We would like to thank Tiago Pimentel for reading a draft of the paper and providing useful comments. Sidak Pal Singh would also like to acknowledge the financial support from Max Planck ETH Center for Learning Systems.
\end{ack}

\bibliography{example_paper
}

\bibliographystyle{plainnat}

\newpage
\appendix
\onecolumn

\section{Omitted Proofs}\label{app:proof}
\begin{lemma}
Given a quadratic problem with $\ell_2$ regularization of strength $\alpha>0$, namely, $\min_{\btheta\in\Reals{d}} \frac{1}{2} \btheta^\top \Mm \, \btheta + \frac{1}{2} \alpha \|\btheta\|^2\,,$ with $\Mm\in\Reals{d\times d}$ symmetric with eigenvalues $\lambda_1\geq\cdots \geq\lambda_d$, the angle between successive steps $\Delta_{t}=\btheta_{t} - \btheta_{t-1}, \Delta_{t+1}=\btheta_{t+1}-\btheta_{t}$, when using gradient descent with a one-step momentum ($\mu>0$) and learning rates $\eta_{t}, \eta_{t+1}$, can be upper and lower bounded as follows:\looseness=-1
	\begin{align*}
&  \langle\Delta_{t}, \Delta_{t+1}\rangle   \leq \eta_{t} \eta_{t+1} (1- \eta_{t}(\mu +\alpha + \lambda_d)) (\lambda_d + \alpha)^2  \|\btheta_{t-1}\|^2  \\
&  \langle\Delta_{t}, \Delta_{t+1}\rangle  \geq \eta_{t} \eta_{t+1}(1- \eta_{t}(\mu+\alpha + \lambda_1)) (\lambda_1 + \alpha)^2 \|\btheta_{t-1}\|^2
	\end{align*}

\end{lemma}
\begin{proof}
	Given function $f(\btheta) = \frac{1}{2} \btheta^\top \Mm \btheta + \frac{1}{2} \alpha \|\btheta\|^2 $, the gradient at $\btheta$ will be $\nabla f(\btheta) = (\Mm+\alpha \Im) \btheta$. Then at the first optimization step, we do
	$$\btheta_{t} = \btheta_{t-1} - \eta_{t} (\Mm+\alpha\Im) \btheta_{t-1}$$
	The particular update being $\Delta_{t} := \btheta_{t} - \btheta_{t-1} = -\eta_{t} (\Mm+\alpha\Im) \btheta_{t-1}$. The next update is similar, but now we also have to factor in the momentum, 
	\begin{align*}
		\btheta_{t+1} &= \btheta_{t} - \eta_{t+1}\left(\nabla f(\btheta_{t}) - \mu \eta_{t} (\Mm +\alpha\Im)\btheta_{t-1}\right)
	\end{align*}
	\begin{align*}
		\Delta_{t+1} & := \btheta_{t+1} - \btheta_{t} = -\eta_{t+1} \left((\Mm+\alpha\Im) \btheta_{t} - \mu \eta_{t} (\Mm +\alpha\Im)\btheta_{t-1}\right) \\
  & = -\eta_{t+1} \left((\Mm+\alpha\Im) \btheta_{t-1} - \eta_{t} (\Mm+\alpha\Im)^2\theta_{t-1} - \mu \eta_{t} (\Mm +\alpha\Im)\btheta_{t-1}\right) \\
		& = -\eta_{t+1}\left((1-\mu\eta_{t}-\alpha\eta_{t})\Im - \eta_{t}\Mm\right)  (\Mm+\alpha\Im) \btheta_{t-1}
	\end{align*}
	
	Now, let us evaluate the inner-product $\langle\Delta_{t}, \Delta_{t+1}\rangle$, 
	\begin{align*}
		\langle\Delta_{t}, \Delta_{t+1}\rangle &= \eta_{t} \eta_{t+1} \btheta_{t-1}^\top \underbrace{(\Mm+\alpha\Im) \left((1- \mu\eta_{t}-\eta_{t}\alpha ) \Im - \eta_{t} \Mm\right)(\Mm+\alpha\Im)}_{\Zm}\btheta_{t-1}
	\end{align*}
	
	Now without loss of generality we can consider $\Zm$ to be a diagonal matrix, as $\Zm$ is symmetric since $\Mm$ is symmetric, we can consider its spectral decomposition $\Zm=\Um\Dm\Um^\top$ and project $\btheta_0$ onto its eigenvectors contained in $\Um$. With this the matrices in the middle are diagonal and we can commute them, which yields us the following matrix:

	$$\Zm=  \diag \begin{pmatrix}
		(1- \mu\eta_{t}-\eta_{t}\alpha - \eta_{t} \lambda_1) (\lambda_1 + \alpha)^2\\
		\vdots\\
		(1- \mu\eta_{t}-\eta_{t}\alpha - \eta_{t} \lambda_d) (\lambda_d + \alpha)^2
	\end{pmatrix}$$
	
	where, we have denoted the eigenvalues of $\Mm$ as $\lambda_1 \geq \cdots \geq \lambda_d$.
	
	Since the inner product of the updates is a quadratic form, we can upper and lower bound it based on the maximum and minimum eigenvalues of $\Zm$, thus giving:
	
	\begin{align*}
		\eta_{t}\eta_{t+1}\lambda_{\min}(\Zm) \|\btheta_{t-1}\|^2 \leq \langle\Delta_{t}, \Delta_{t+1}\rangle \leq \eta_{t}\eta_{t+1}\lambda_{\max}(\Zm) \|\btheta_{t-1}\|^2 
	\end{align*}
	
	Because of the above form of eigenvalues of $\Zm$ (diagonal matrices have their eigenvalues as their diagonal entries), we will have:
	
	$\lambda_{\max}(\Zm) = (1- \mu\eta_{t}-\eta_{t}\alpha - \eta_{t} \lambda_d) (\lambda_d + \alpha)^2$ and $\lambda_{\min}(\Zm) = (1- \mu\eta_{t}-\eta_{t}\alpha - \eta_{t} \lambda_1) (\lambda_1 + \alpha)^2$
\end{proof}

\clearpage

\section{Why cosine similarities increase with scale?}\label{sec:arg}
Note, we assume that the majority of the parameter norm lies in the square hidden matrices, and not the input or output layers. Moreover, we use $o, O, \btheta$ to denote standard mathematical notation with regards to scaling in the limit width $n\rightarrow \infty$. For vectors, this notation is entry-wise.

Suppose we have a hidden layer with input $x_0\in\mathbb{R}^n$ for width $n$, that is acted on by (without loss of generality) a square matrix $W_0\in\mathbb{R}^{n\times n}$ to give:

$$h_0 = W_0 x_0$$

We suppose $x$ has $\btheta(1)$ entries, as is the case with standard initialisations/parameterisations \cite{he2015delving}. We suppose $W_0$ has i.i.d. elements with initialisation that is $O(1/\sqrt{n})$ in order to ensure that each element of the features $h$ has entries $\btheta(1)$.

Now, if we take a gradient update with learning rate $\eta$ on some downstream loss $L$ that depends on $h$ (and not $W$ or $x$), we get:

$$W_1 = W_0 - \eta \text{d}h \cdot x_0^{\top}$$

where $\text{d}h = \frac{\partial L}{\partial h}\in\mathbb{R}^{n\times 1}$ is our feature derivative.

Then if we have new input $x_1$ (wlog $x_1=x_0$), we have new features:

$$h_1 = x_1 W_1 = h_0 - n \eta \text{d}h \cdot \frac{x_0^{\top} x_0}{n}$$

For our features to be stable (i.e. $\btheta(1)$) after the update, we need $n\eta \text{d}h$ to be $O(1)$, because $\frac{x_0^{\top} x_0}{n}=\btheta(1)$ by assumption on $x$. NB: if $n\eta \text{d}h = o(1)$ we have no feature learning (ie NTK regime \cite{jacot2018neural}), and if $n\eta \text{d}h=\btheta(1)$ we have feature learning (ie $\mu$P \cite{yang2022tensor}).

In any case, $\eta \text{d}h = O(1/n)$ entry-wise, which means that $W_1 - W_0 = -\eta \text{d}h \cdot x_0^{\top} $ has $O(1/n)$ entries, again by assumption on the scale of elements of $x_0$.

But because $W_0=\btheta(1/\sqrt n)$, the initialisation will elementwise-dominate the $O(1/n)$ update for the first training step (and more training steps follows by induction). As a result, the update $W_T - W_0$ will always be an order of at least $\sqrt{n}$ smaller than the initialisation, and hence the new parameters $W_T$ will be exactly aligned with the initialisation $W_0$ for all $T$ in the large width limit, i.e. the cosine similarities will be $1$.

\clearpage
\section{Comparing Gradient Trajectories with Random Walks}\label{app:random}
The structure that we observe in trajectory maps following gradient trajectories raises the question of if we would observe similar structure in a random walk. 

If we have $T$ timesteps or epochs, with parameter space $\btheta\in\mathbb{R}^p$ and ``learning rate'' schedule $(\eta_t)_{t=1}^T$, we can consider a random walk with updates:

$$\btheta_t - \btheta_{t-1}\overset{\text{ind.}}{\sim}\mathcal{N}(0,\eta_t^2 I_p)$$

which is to say that at time step $t$, each parameter coordinate in the parameter vector is updated independently with a Gaussian of variance $\eta_t^2$, and the updates are independent across different time steps.

Then, if $\theta^i$ denotes a \textit{single} parameter coordinate for a dimension $i\leq p$, for two time steps $s<t$, we have:

$$(\theta^i_s, \theta^i_t) \sim \mathcal{N}(0,  \bigl( \begin{smallmatrix}H_{s} & H_{s} \\ H_{s} & H_{t}\end{smallmatrix}\bigr))$$
where $H_{u} = \sum_{t'=1}^u \eta_{t'}^2 $ is the cumulative squared learning rate from $t'=1$ to $t'=u$.

Then, by the strong law of large numbers we have for the large parameter space $p\rightarrow \infty$ limit:

$$\frac{1}{p} \lVert \btheta_s \rVert_2^2  = \frac{1}{p}\sum_{i=1}^p (\theta^i_s)^2 \overset{a.s.}{\rightarrow} H_{s}\,,\quad \frac{1}{p} \lVert \btheta_t \rVert_2^2  = \frac{1}{p}\sum_{i=1}^p (\theta^i_t)^2 \overset{a.s.}{\rightarrow} H_{t}$$

$$\frac{1}{p} \langle \btheta_s, \btheta_t \rangle  = \frac{1}{p}\sum_{i=1}^p \theta^i_t \theta_s^i \overset{a.s.}{\rightarrow} H_{s}$$

and by the property of composing almost sure limits, we also have almost sure convergence in the cosine similarity:
$$\frac{\langle \btheta_s, \btheta_t \rangle}{\lVert \btheta_s \rVert_2 \lVert \btheta_t \rVert_2} \overset{a.s.}{\rightarrow} \frac{H_{s}}{\sqrt{H_{s}H_{t}}} = \sqrt{\frac{H_s}{H_t}}$$

in the large parameter space limit, which we use as an approximation to give analytic formulas for the trajectory map and MDS that we can compare to gradient trajectories.

One thing to note, is that this cosine similarity $\sqrt{\frac{H_s}{H_t}}$ becomes invariant to the \textit{scale} of the learning rates $\eta$, and instead it is the relative rate of decay in the learning rate schedule that matters.
\subsection{Experimental Results}

\begin{figure}[!h]
    \centering
\includegraphics[width=0.8\textwidth]{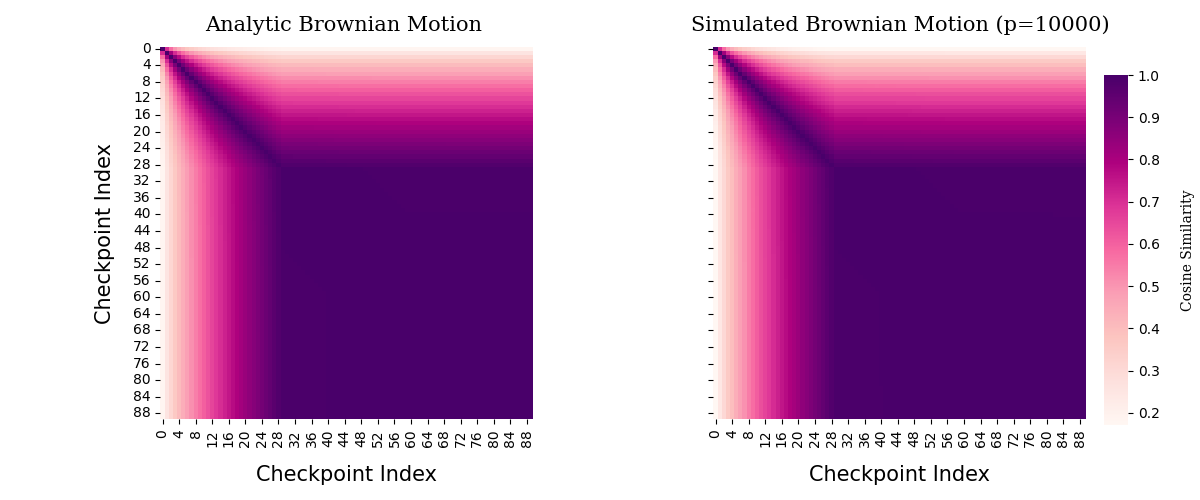}
    \caption{With step size decay: Relative Trajectory Map for a Random Walk/Brownian motion in both analytic and empirically simulated settings.}
    \label{fig:brownian}
\end{figure}

\begin{figure*}[h!]
	\centering
	\includegraphics[width=\textwidth]{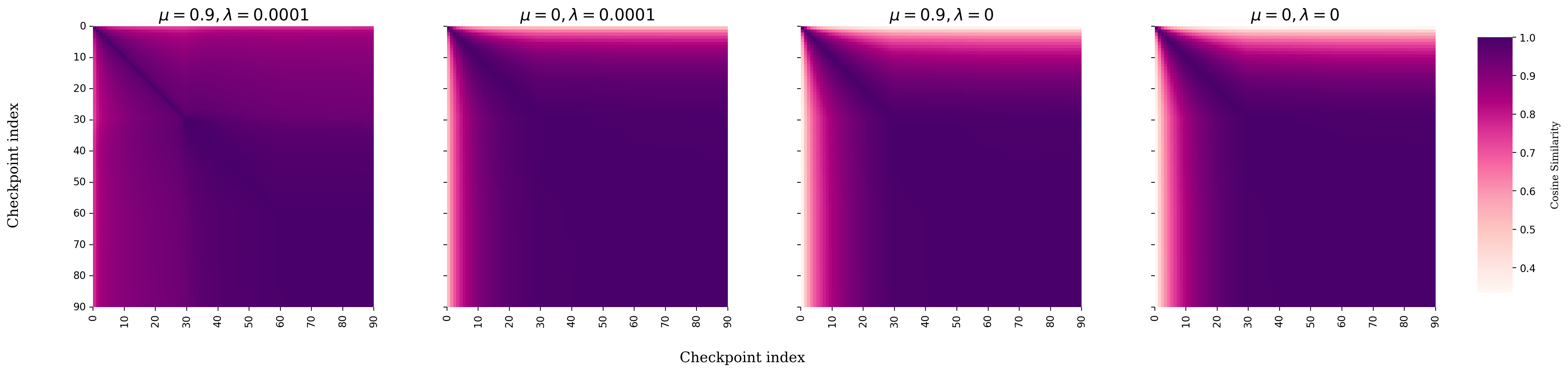}
	\caption{Relative Trajectory Maps, with respect to initialization, of ResNet50 models for different amounts of momentum and weight decay.} 
 \label{fig:rel-tmap-rn50}
\end{figure*}

The first thing to note is that our empirical simulation of the random walk matches the theoretical limit described in the section above, for a finite parameter count (such as $10,000$). Next, comparing the these relative trajectory maps with those for ResNet50 (Figure~\ref{fig:rel-tmap-rn50}) we find that the latter reveal a much more directional redundancy component to their trajectories as opposed to random walks. This further lends support to the thesis that optimization trajectories ensued when training  neural networks are highly structured and have significant directional redudndancy. 

\begin{figure}[!h]
    \centering
\includegraphics[width=0.8\textwidth]{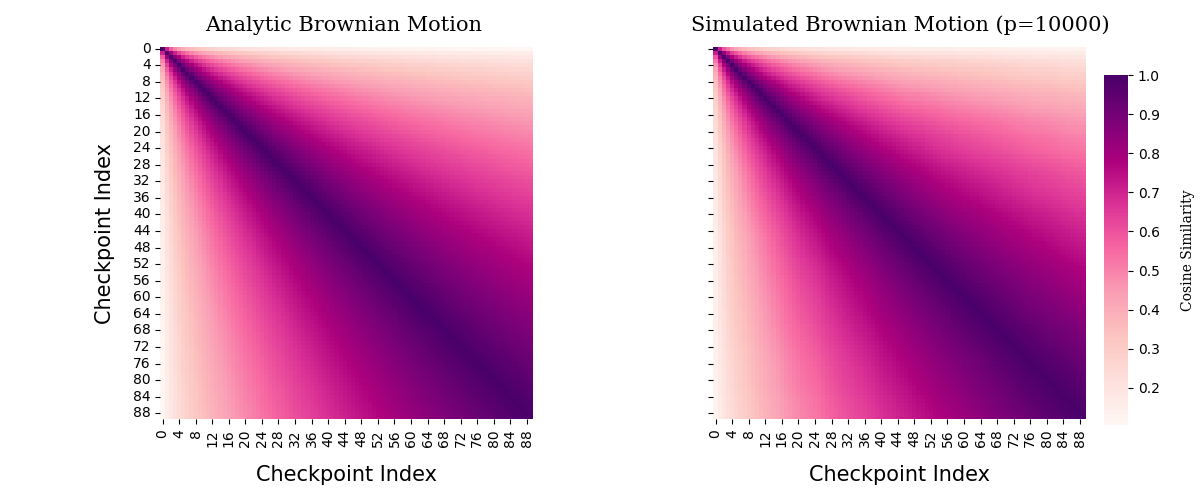}
    \caption{\textit{No Step size decay}: Relative Trajectory Map for a Random Walk/Brownian motion in both analytic and empirically simulated settings.}
    \label{fig:brownian-noDecay}
\end{figure}

An additional thing to note is that the above relative trajectory map for random walks covers the setting of decreasing the step size to mirror how the optimization procedure is setup for ResNet50. In the case of no such step size decay, the analytic and empirical versions of the relative trajectory map are depicted in the Figure~\ref{fig:brownian-noDecay}.

\clearpage
\section{Detailed Experimental Results}

\subsection{ResNet50: Switching off the hyperparameters}
\begin{figure*}[h!]
	\centering
	\includegraphics[width=0.9\textwidth]{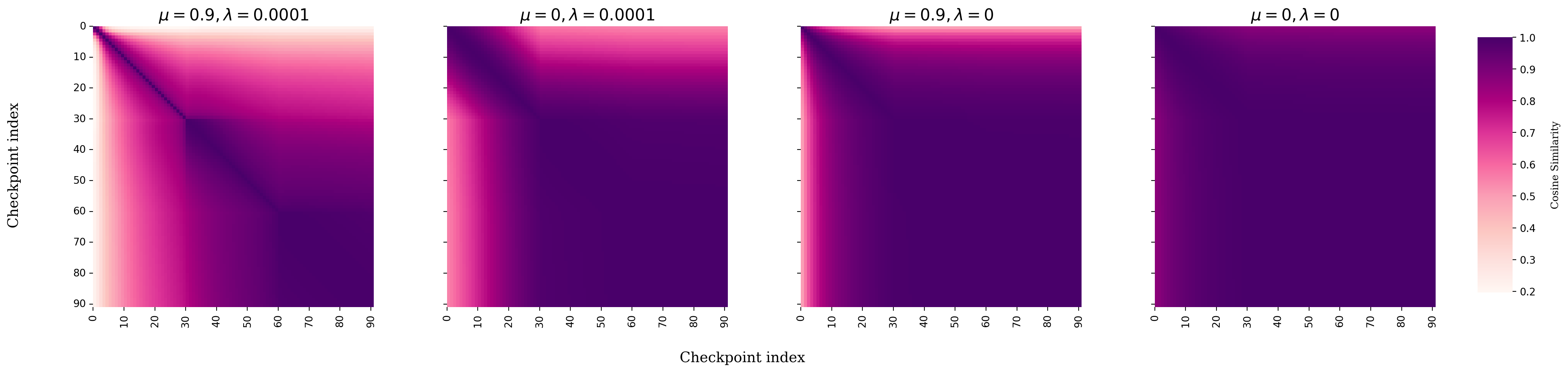}
	\caption{Trajectory Maps of ResNet50 models across different amounts of momentum and weight decay} 
\end{figure*}

The relative trajectory maps can be found in Figure~\ref{fig:rel-tmap-rn50}.

\begin{figure*}[h!]
	\centering
	\subfigure[$\angle(\theta_{t+1}-\theta_t,\theta_t)$]{\label{fig:}
		\includegraphics[width=0.34\textwidth]{figures/icml/Mom,WD/ckpt_freq-1_heatmap_from_multi-4_resnet50_imagenet_ep-90_lr-0.1_bsz-256_mom-0.9_wdecay-0.0001_seed-0_2023-02-21_10-45-30_983469_2024-02-01_02-26-16_172063/figures/pdf/angle_theta__t+1_-theta_t,theta_t__vs_Epochs__t__across_Mom,WD.pdf}}
	\subfigure[$\angle(\theta_{t+1}-\theta_t,\theta_T-\theta_0)$]{\label{fig:}
		\includegraphics[width=0.34\textwidth]{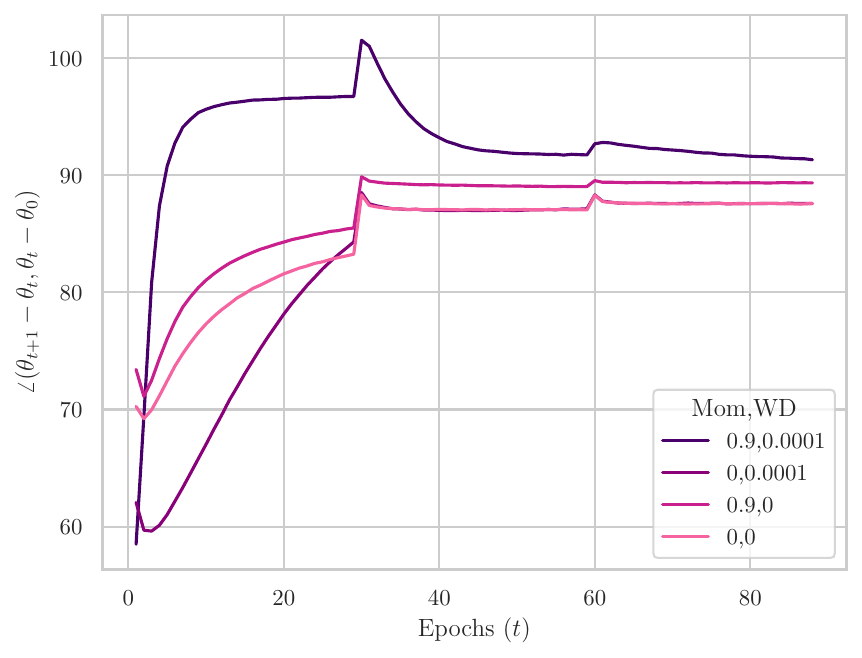}}
	\subfigure[$\angle(\theta_{t+k}-\theta_t,\theta_t-\theta_{t-k})$, for $k=1$ ]{\label{fig:}
		\includegraphics[width=0.34\textwidth]{figures/icml/Mom,WD/ckpt_freq-1_heatmap_from_multi-4_resnet50_imagenet_ep-90_lr-0.1_bsz-256_mom-0.9_wdecay-0.0001_seed-0_2023-02-21_10-45-30_983469_2024-02-01_02-26-16_172063/figures/pdf/angle_theta__t+k_-theta_t,theta_t-theta__t-k____for_k=1__in__circ__vs_Epochs__t__across_Mom,WD.pdf}
		\vspace{-2mm}}
	\subfigure[$\angle(\theta_{t}-\theta_0,\theta_T-\theta_0)$]{\label{fig:}
		\includegraphics[width=0.34\textwidth]{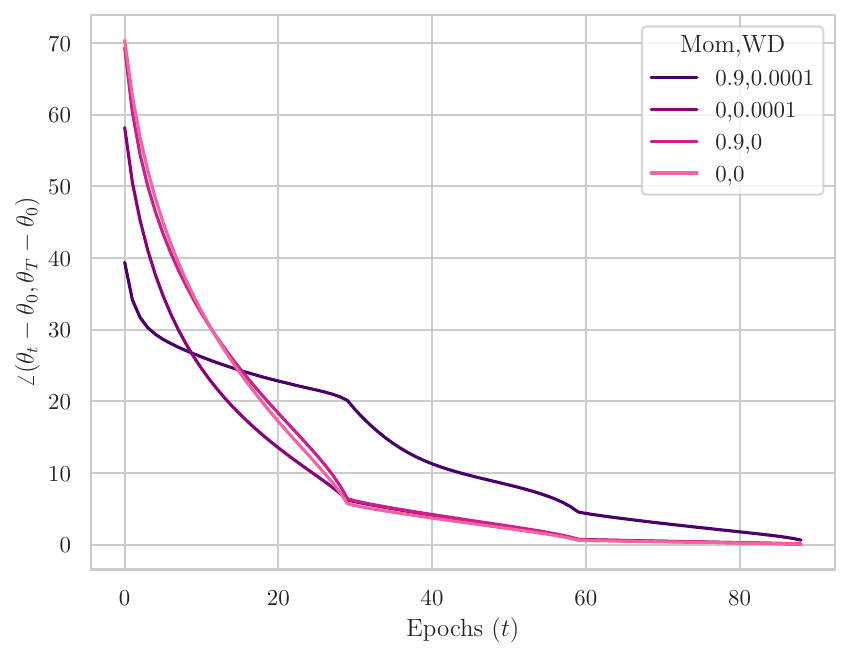}
		\vspace{-2mm}}
	\subfigure[$\angle(\theta_{t+1}-\theta_t, \theta_T-\theta_0)$]{\label{fig:}
		\includegraphics[width=0.34\textwidth]{figures/icml/Mom,WD/ckpt_freq-1_heatmap_from_multi-4_resnet50_imagenet_ep-90_lr-0.1_bsz-256_mom-0.9_wdecay-0.0001_seed-0_2023-02-21_10-45-30_983469_2024-02-01_02-26-16_172063/figures/pdf/angle_theta__t+1_-theta_t,theta_T-theta_0__vs_Epochs__t__across_Mom,WD.pdf}}
	\subfigure[Apex Angle at Initialization $\angle(\theta_t-\theta_0,\theta_1-\theta_0)$ ]{\label{fig:}
		\includegraphics[width=0.34\textwidth]{figures/icml/Mom,WD/ckpt_freq-1_heatmap_from_multi-4_resnet50_imagenet_ep-90_lr-0.1_bsz-256_mom-0.9_wdecay-0.0001_seed-0_2023-02-21_10-45-30_983469_2024-02-01_02-26-16_172063/figures/pdf/Apex_Angle_at_Initialization_angle_theta_t-theta_0,theta_1-theta_0__vs_Epochs__t__across_Mom,WD.pdf}
		\vspace{-2mm}}
	\subfigure[Apex Angle at Origin $\angle(\theta_t,\theta_0)$]{\label{fig:}
		\includegraphics[width=0.34\textwidth]{figures/icml/Mom,WD/ckpt_freq-1_heatmap_from_multi-4_resnet50_imagenet_ep-90_lr-0.1_bsz-256_mom-0.9_wdecay-0.0001_seed-0_2023-02-21_10-45-30_983469_2024-02-01_02-26-16_172063/figures/pdf/Apex_Angle_at_Origin_angle_theta_t,theta_0__vs_Epochs__t__across_Mom,WD.pdf}
		\vspace{-2mm}}
	\caption{Angular measures of the Trajectory for ResNet50 trained on ImageNet} 
\end{figure*}

\begin{figure*}[h!]
	\centering
	\subfigure[$\|\theta_t\|_2$]{\label{fig:}
	\includegraphics[width=0.3\textwidth]{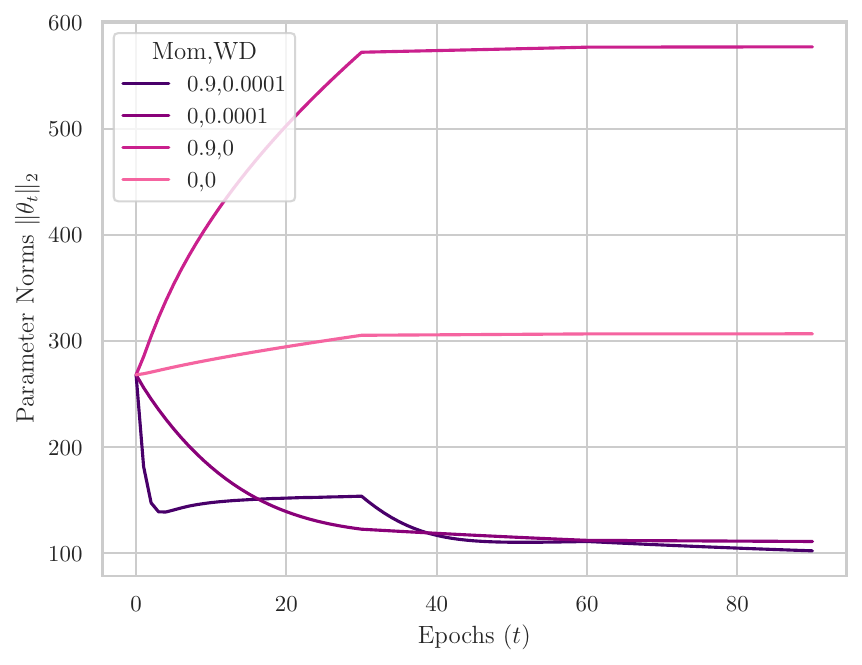}}
	\subfigure[$\|\theta_{t+k}-\theta_t\|_2$ ]{\label{fig:}
		\includegraphics[width=0.3\textwidth]{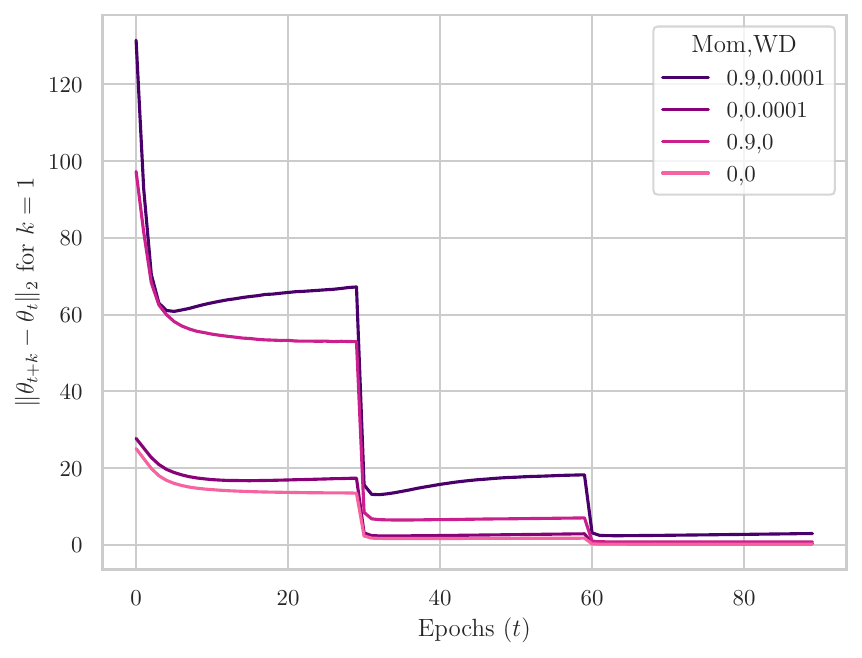}}
	\subfigure[$\|\theta_t-\theta_0\|_2$ ]{\label{fig:}
		\includegraphics[width=0.3\textwidth]{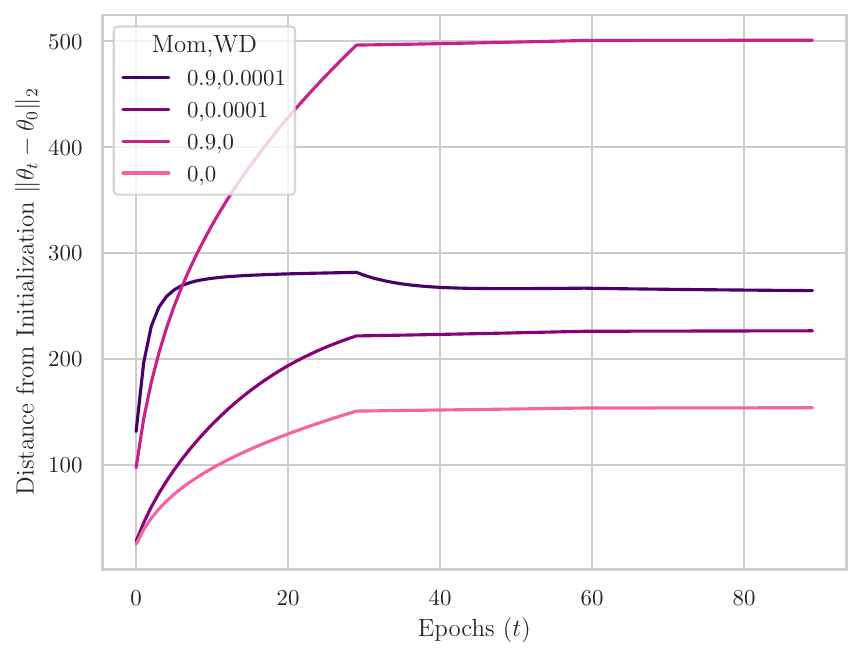}
		\vspace{-2mm}}
	\caption{Norm-based measures of the Trajectory for ResNet50 trained on ImageNet} 
\end{figure*}

\begin{figure*}[h!]
	\centering
	\subfigure[Eigenvalues: $\Km$]{\label{fig:}
		\includegraphics[width=0.3\textwidth]{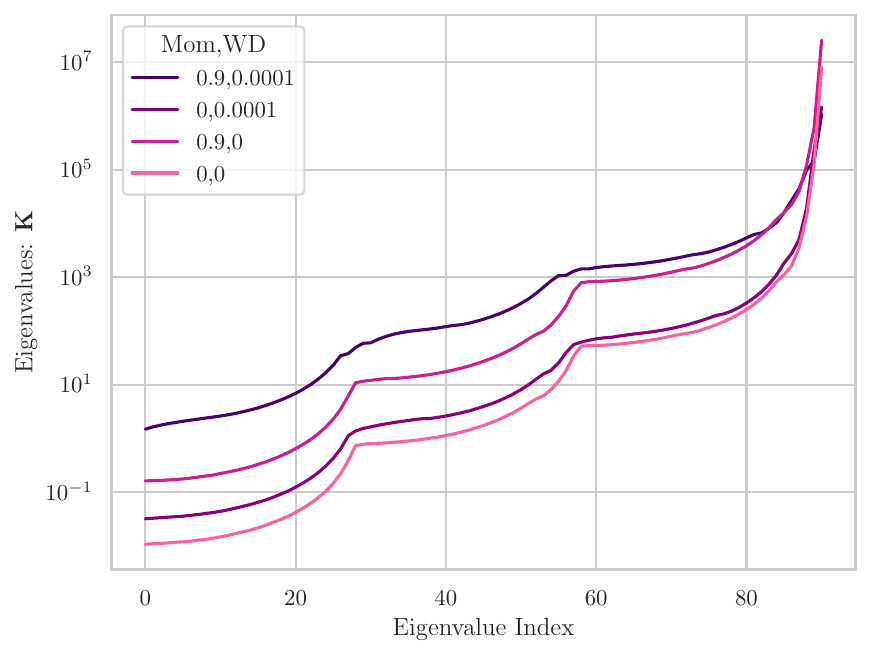}}
	\subfigure[Eigenvalues: $\Km_0$]{\label{fig:}
		\includegraphics[width=0.3\textwidth]{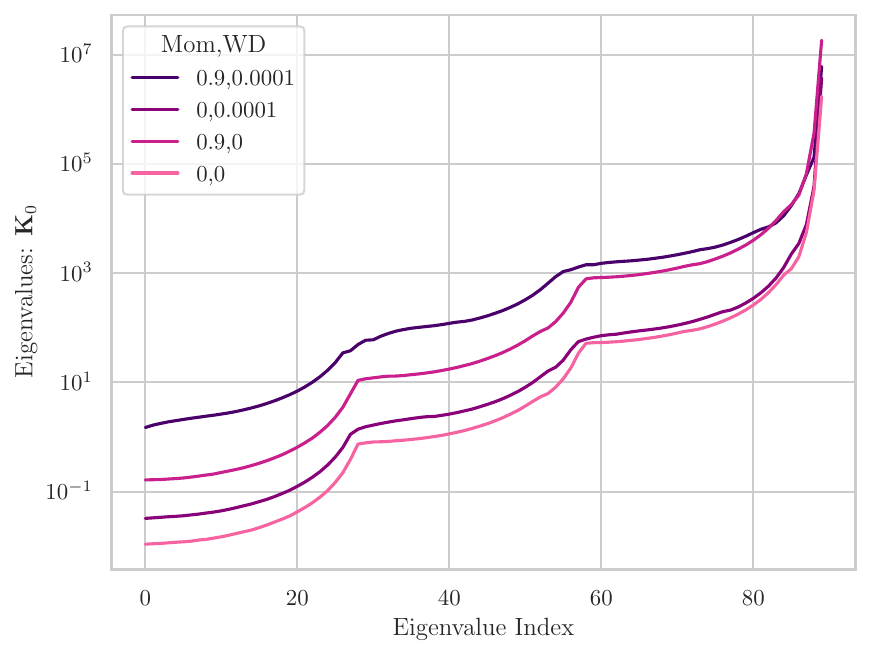}}

	\subfigure[Eigenvalues: $\Cm$ ]{\label{fig:}
		\includegraphics[width=0.3\textwidth]{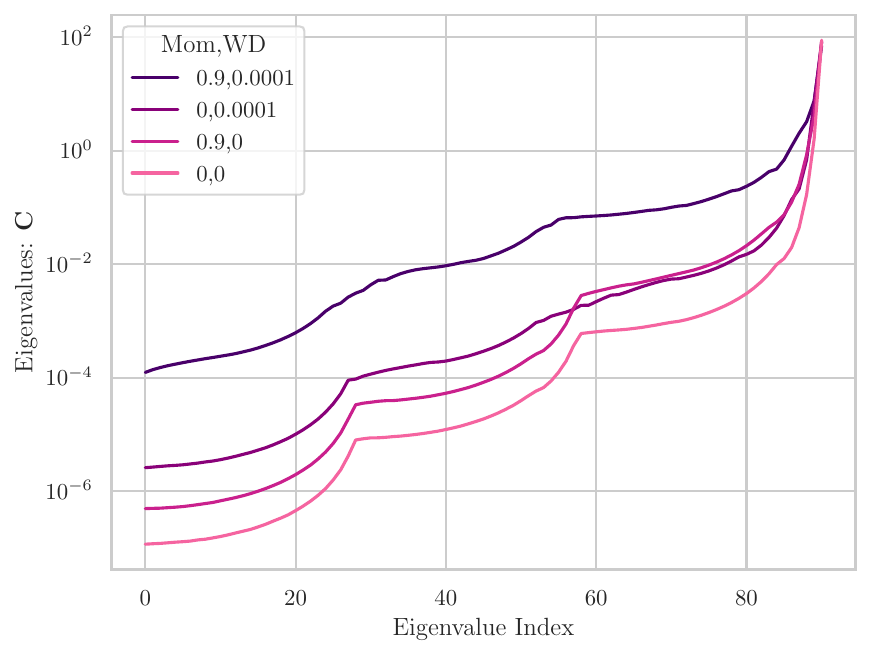}
		\vspace{-2mm}}
	\subfigure[Eigenvalues: $\Cm_0$]{\label{fig:}
		\includegraphics[width=0.3\textwidth]{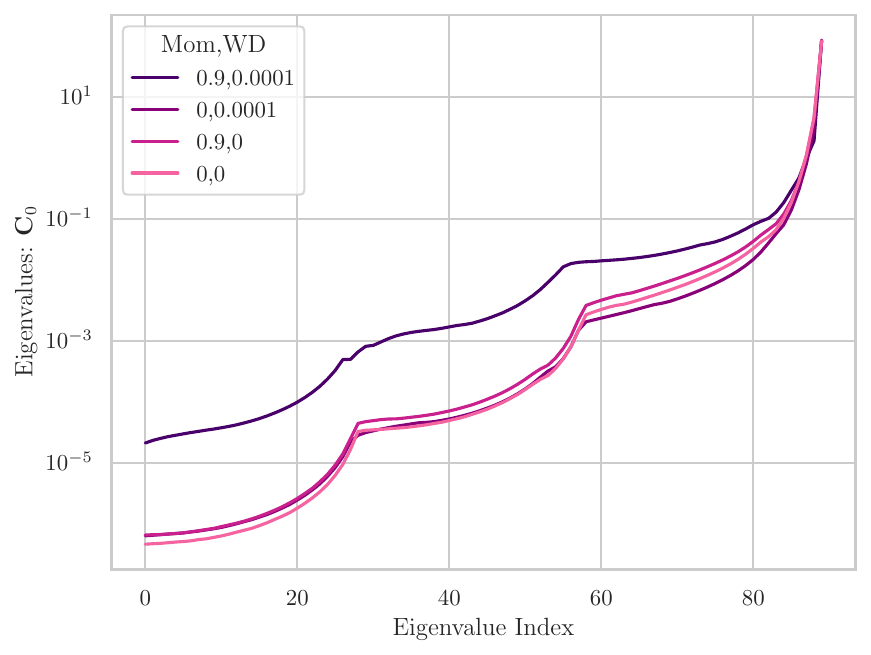}
		\vspace{-2mm}}
	\caption{Spectral measures of the Trajectory for ResNet50 trained on ImageNet} 
\end{figure*}

\clearpage

\subsection{ResNet50: Weight Decay, AdamW}\label{app:wd-adam-rn}

\begin{figure*}[h!]
	\centering
	\includegraphics[width=0.9\textwidth]{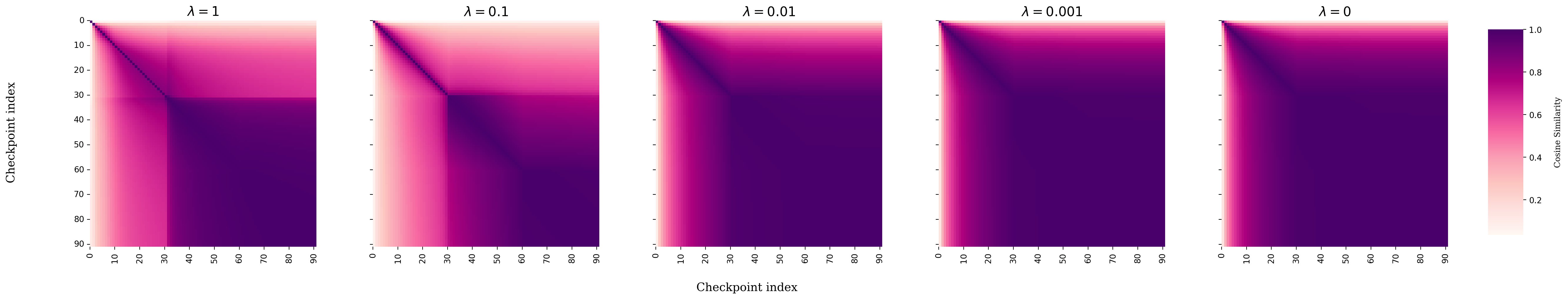}
	\caption{Trajectory Maps of ResNet50 models across different amounts of weight decay} 
\end{figure*}

\begin{figure*}[h!]
	\centering
	\includegraphics[width=0.9\textwidth]{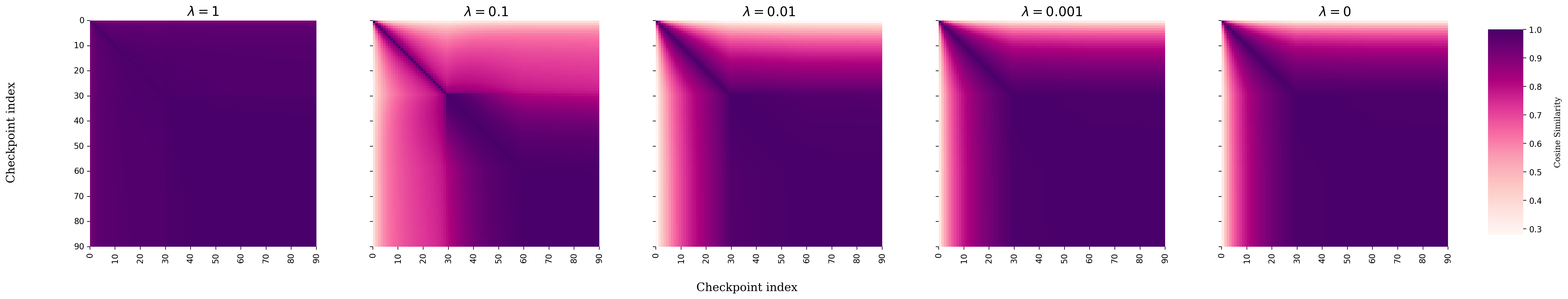}
	\caption{Relative Trajectory Maps, with respect to initialization, of ResNet50 models across different amounts of weight decay} 
\end{figure*}

\clearpage
\begin{figure*}[h!]
	\centering
	\subfigure[$\angle(\theta_{t+1}-\theta_t,\theta_t)$]{\label{fig:}
		\includegraphics[width=0.34\textwidth]{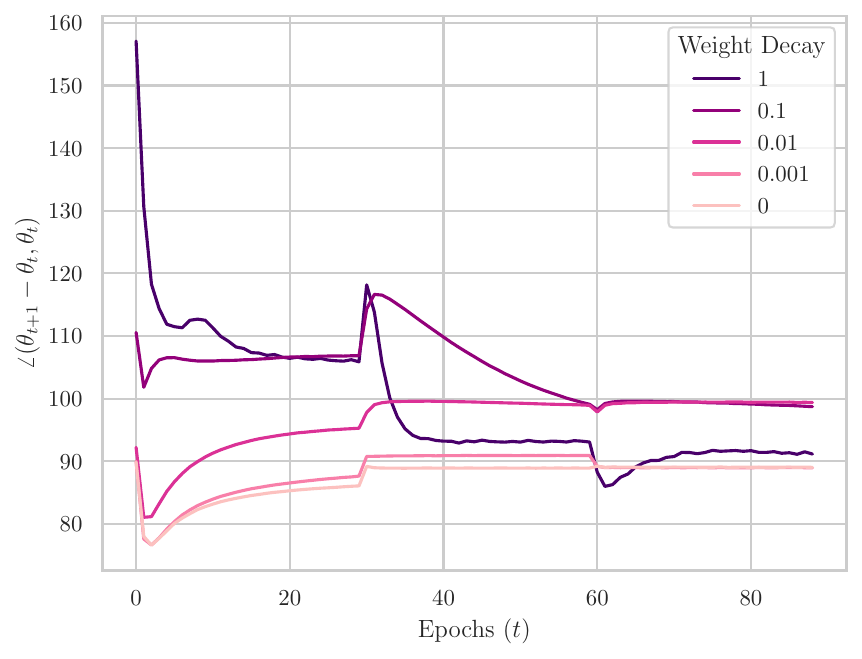}}
	\subfigure[$\angle(\theta_{t+1}-\theta_t,\theta_T-\theta_0)$]{\label{fig:}
		\includegraphics[width=0.34\textwidth]{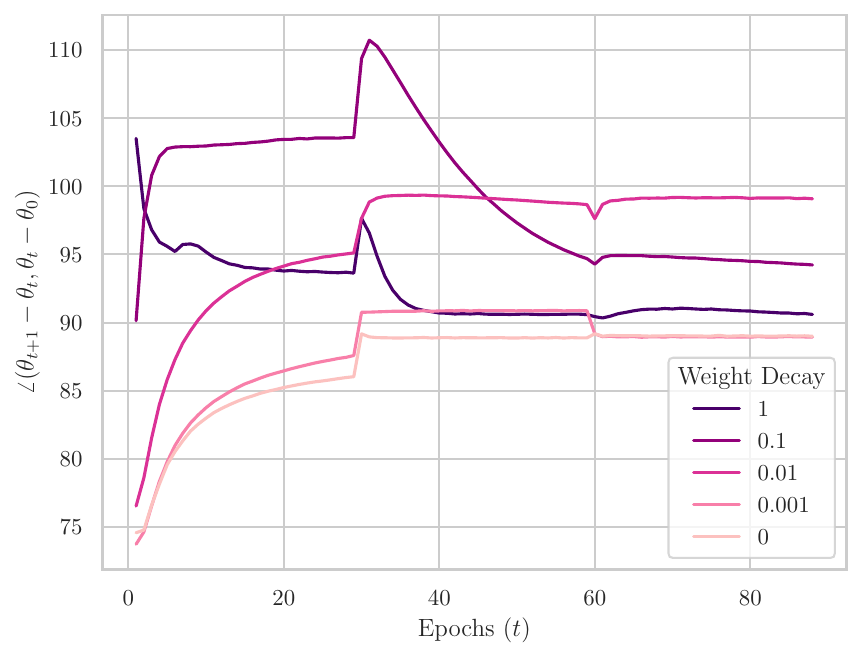}}
	\subfigure[$\angle(\theta_{t+k}-\theta_t,\theta_t-\theta_{t-k})$, for $k=1$ ]{\label{fig:}
		\includegraphics[width=0.34\textwidth]{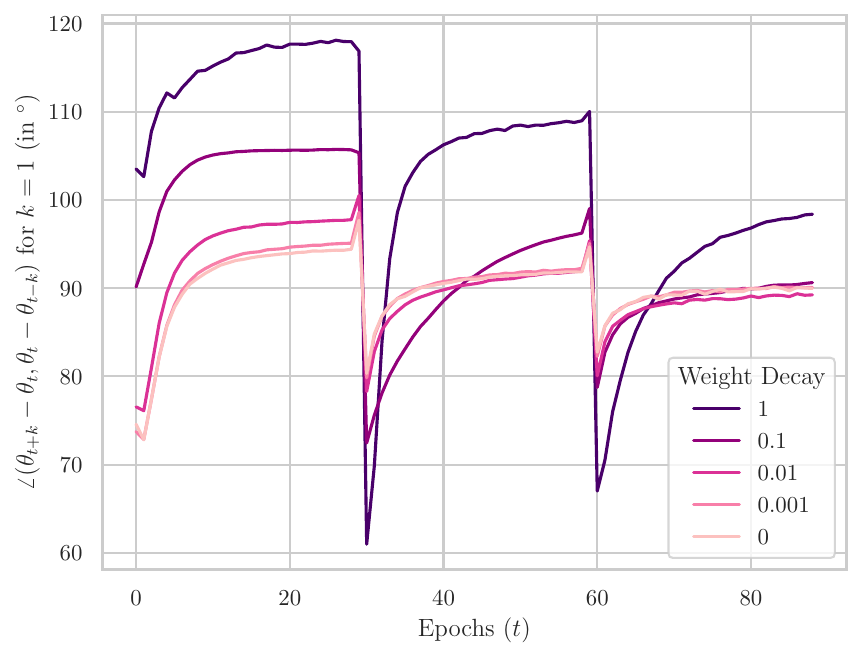}
		\vspace{-2mm}}
	\subfigure[$\angle(\theta_{t}-\theta_0,\theta_T-\theta_0)$]{\label{fig:}
		\includegraphics[width=0.34\textwidth]{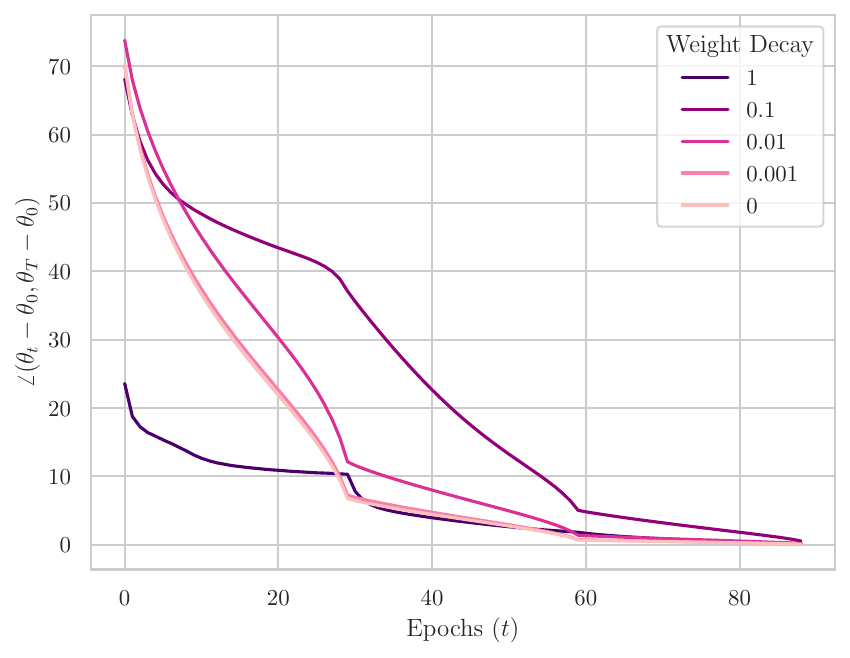}
		\vspace{-2mm}}
	\subfigure[$\angle(\theta_{t+1}-\theta_t, \theta_T-\theta_0)$]{\label{fig:}
		\includegraphics[width=0.34\textwidth]{figures/icml/Weight_Decay/ckpt_freq-1_heatmap_from_multi-5_resnet50_opt-adamw_imagenet_ep-90_lr-0.001_bsz-256_mom-0.5_wdecay-1.0_aug-True_seed-0_2024-01-20_13-07-01_747341_2024-02-01_02-26-21_884644/figures/pdf/angle_theta__t+1_-theta_t,theta_T-theta_0__vs_Epochs__t__across_Weight_Decay.pdf}}
	\subfigure[Apex Angle at Initialization $\angle(\theta_t-\theta_0,\theta_1-\theta_0)$ ]{\label{fig:}
		\includegraphics[width=0.34\textwidth]{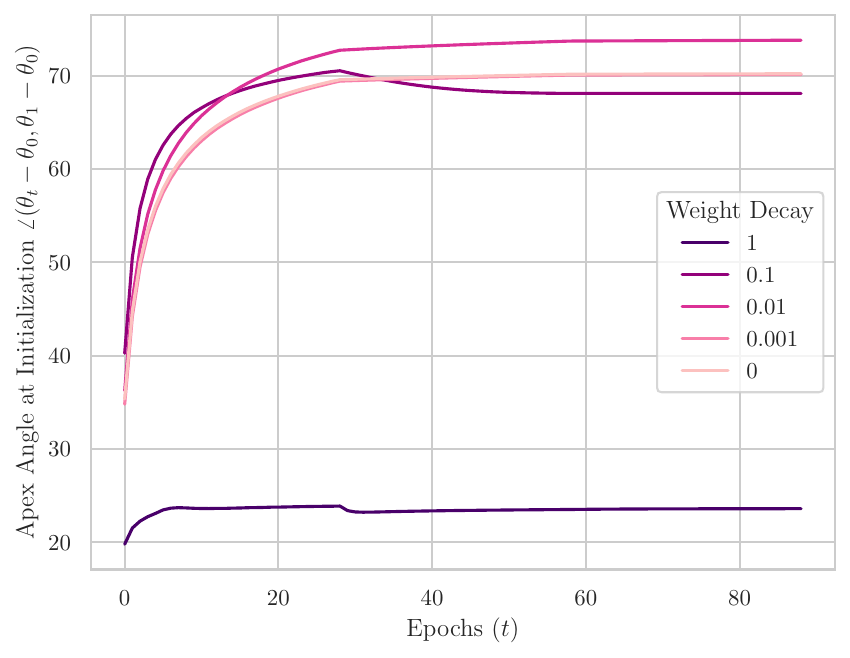}
		\vspace{-2mm}}
	\subfigure[Apex Angle at Origin $\angle(\theta_t,\theta_0)$]{\label{fig:}
		\includegraphics[width=0.34\textwidth]{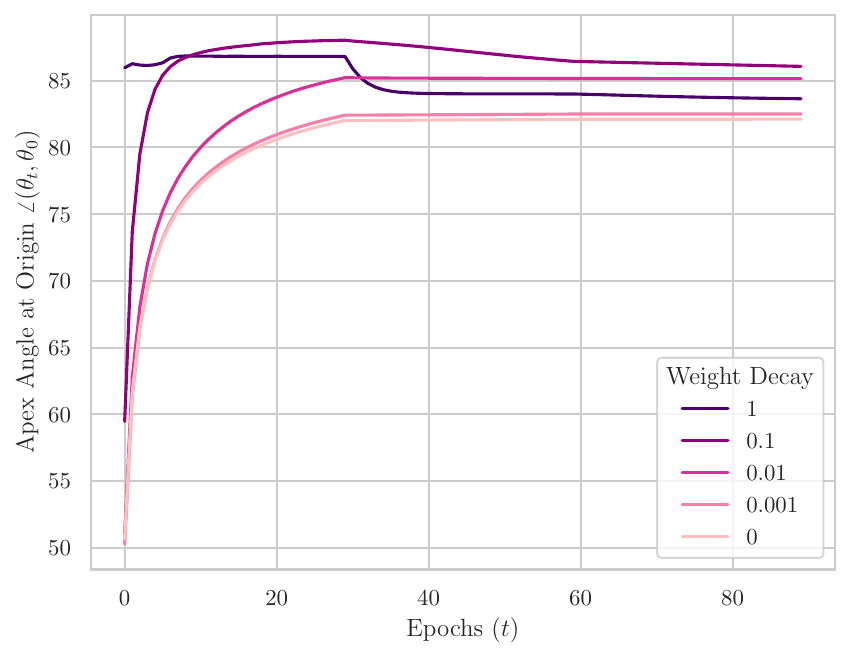}
		\vspace{-2mm}}
	\caption{Angular measures of the Trajectory for ResNet50 trained on ImageNet} 
\end{figure*}

\begin{figure*}[h!]
	\centering
	\subfigure[$\|\theta_t\|_2$]{\label{fig:}
	\includegraphics[width=0.3\textwidth]{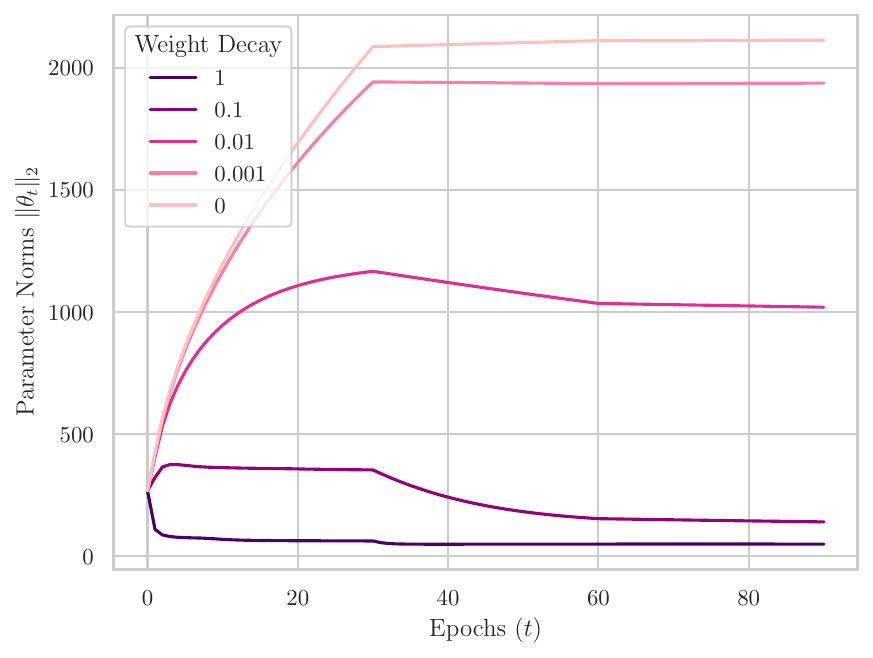}}
	\subfigure[$\|\theta_{t+k}-\theta_t\|_2$ ]{\label{fig:}
		\includegraphics[width=0.3\textwidth]{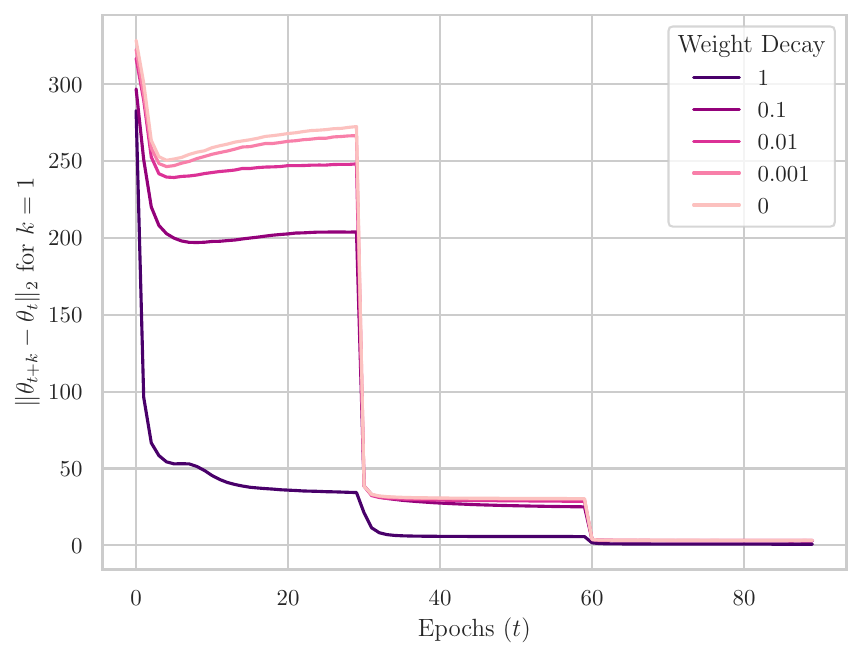}}
	\subfigure[$\|\theta_t-\theta_0\|_2$ ]{\label{fig:}
		\includegraphics[width=0.3\textwidth]{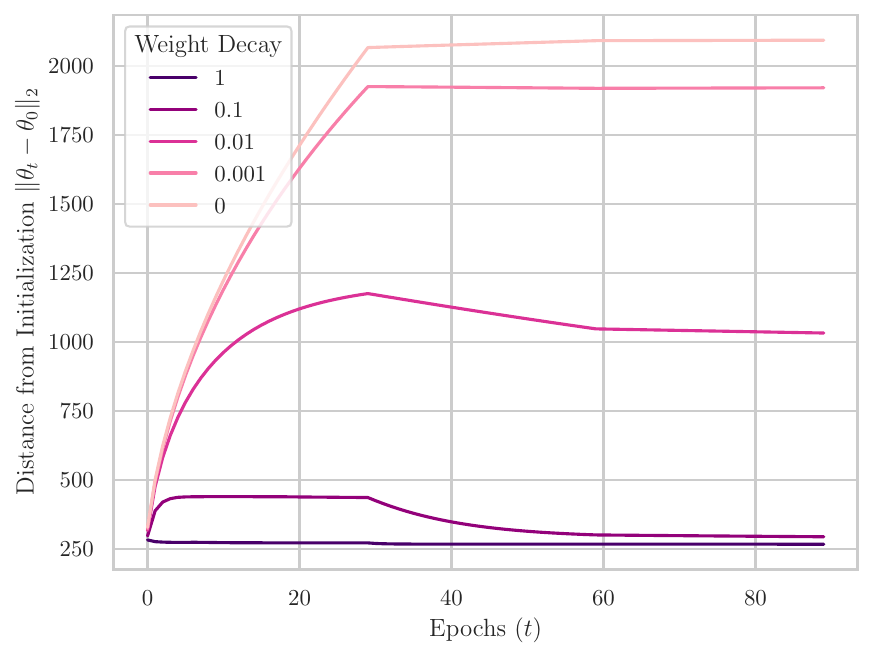}
		\vspace{-2mm}}
	\caption{Norm-based measures of the Trajectory for ResNet50 trained on ImageNet} 
\end{figure*}

\begin{figure*}[h!]
	\centering
	\subfigure[Eigenvalues: $\Km$]{\label{fig:}
		\includegraphics[width=0.3\textwidth]{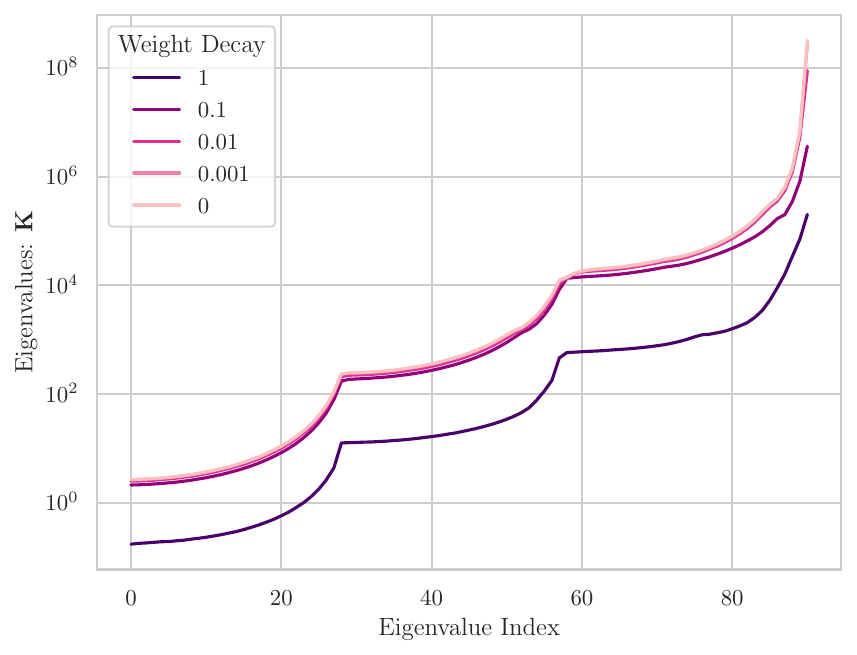}}
	\subfigure[Eigenvalues: $\Km_0$]{\label{fig:}
		\includegraphics[width=0.3\textwidth]{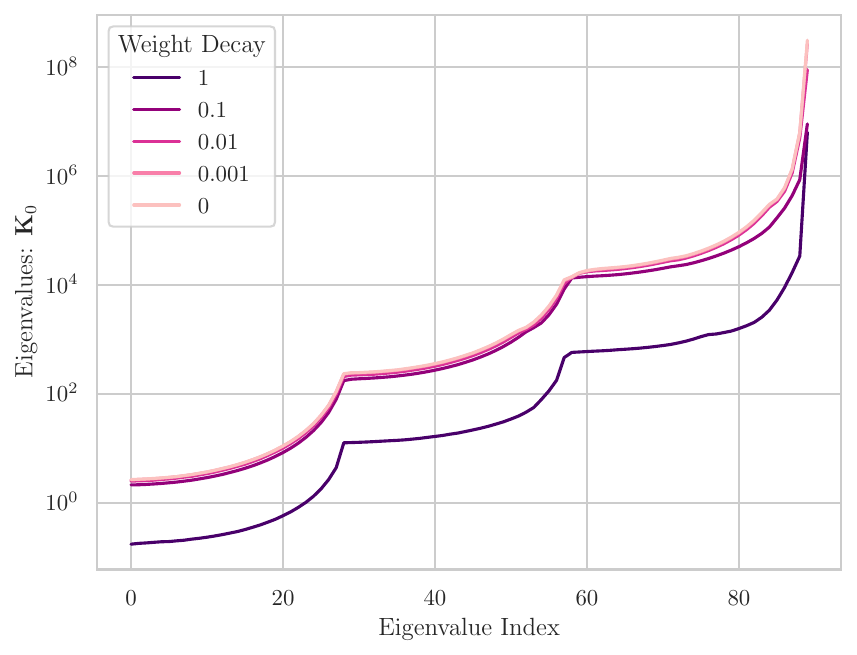}}

	\subfigure[Eigenvalues: $\Cm$ ]{\label{fig:}
		\includegraphics[width=0.3\textwidth]{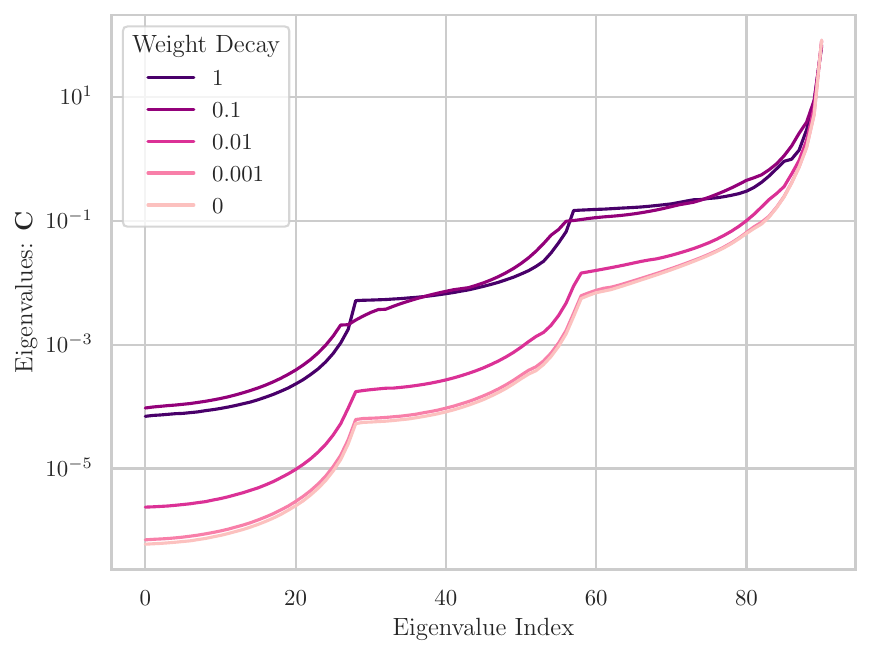}
		\vspace{-2mm}}
	\subfigure[Eigenvalues: $\Cm_0$]{\label{fig:}
		\includegraphics[width=0.3\textwidth]{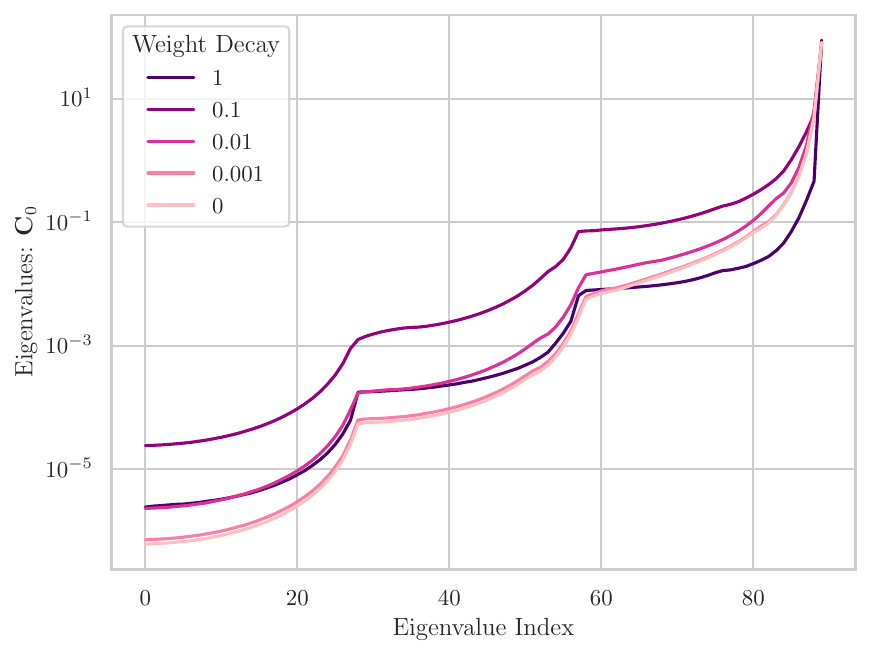}
		\vspace{-2mm}}
	\caption{Spectral measures of the Trajectory for ResNet50 trained on ImageNet} 
\end{figure*}

\clearpage

\section{Directional Effects in Other Key Settings}\label{app:direc-effects}

\begin{figure*}[h!]
	\centering
	\includegraphics[width=0.9\textwidth]{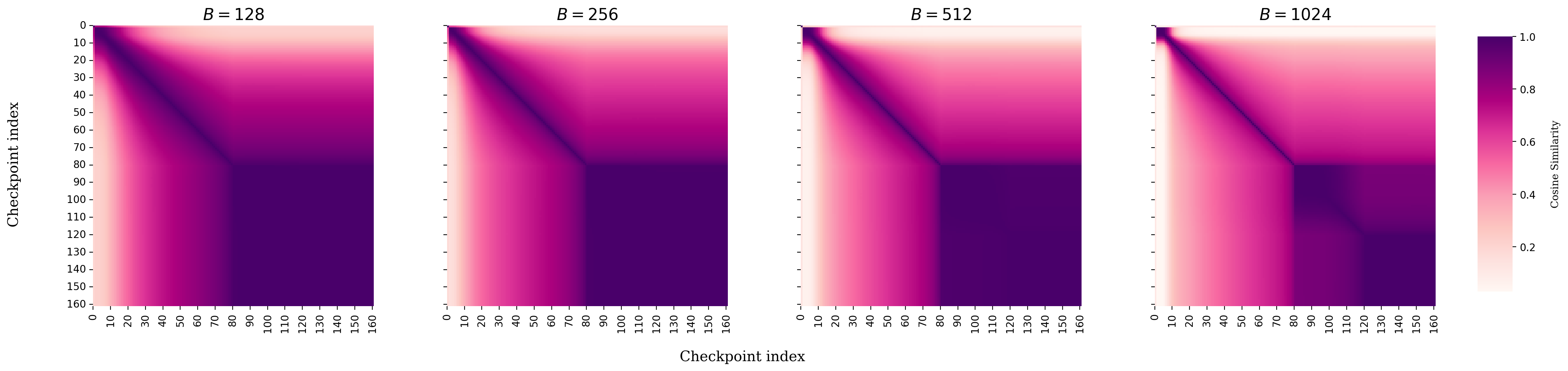}
	\caption{Trajectory Maps of VGG16 models across different batch sizes. The learning rates have been scaled in proportion to the batch size, and the training schedule was adjusted to ensure an equal number of steps (and not simply epochs) for all the runs. We also adjusted the learning rate schedule to drop learning rates at a corresponding number of steps across the experiments. The respective MDS values are $\omega=0.753, 0.723, 0.660, 0.619$ and the test accuracies are $91.63\%, 91.82\%, 92.44\%, 92.39\%$.} 
 \label{fig:tm-bsz}
\end{figure*}

In addition to the momentum and weight decay, there are other crucial hyperparameters, such as learning rate and batch size, whose directional effects warrant a mention. We carry out additional experiments in these settings, and from where, the key findings are that the \textbf{learning rate}, as it would be easy to guess, indeed encourages directional exploration leading to low MDS scores. But, somewhat more interestingly, we find that increasing the \textbf{batch size} also helps further exploration and thereby decreases the MDS scores. The trajectory maps can be found in the Figure~\ref{fig:tm-bsz} While we encourage the curious reader to have a look at the Appendix~\ref{app:bsz}, we find that with increased batch size, the angle between the updates as well as the angle between the update and the current location become increasingly obtuse, and thus making room for a wider directional exploration. In contrast, for smaller batch sizes these angles are closer to $90^\circ$. We hypothesize that a similar mutual interaction, as observed with weight decay and momentum, also occurs with batch size is considered. A detailed analysis, however, remains outside the current scope.

Lastly, we also experimented with \textbf{Sharpness-Aware Minimization}~\citep{foret2021sharpnessaware} (SAM), where we found that a higher value of the SAM regularization coefficient leads to a slightly increased directional similarity, which could potentially be related to SAM directing optimisation to flatter basins wherein the individual points are more directionally alike and have higher cosine similarities. The detailed results can be found in the Appendix~\ref{app:sam}.

\paragraph{Other Settings and Datasets.}
As a final remark for this section, we would like to emphasize that similar results for weight decay as well as momentum, can be found under different hyperparameter settings in the supplementary material. In particular, we analyze the qualitative and quantitative hallmarks for multiple values of learning rate, weight decay, and momentum for VGG16 on CIFAR10 as well as other values for momentum and weight decay in the case of ResNet50 trained with SGD, and even Vision Transformer trained with AdamW on ImageNet across varying weight decay, but these have to be omitted here due to space constraints. \looseness=-1

\subsection{ViT: Weight Decay, AdamW}\label{app:wd-adam-vit}

\begin{figure*}[h!]
	\centering
	\includegraphics[width=0.9\textwidth]{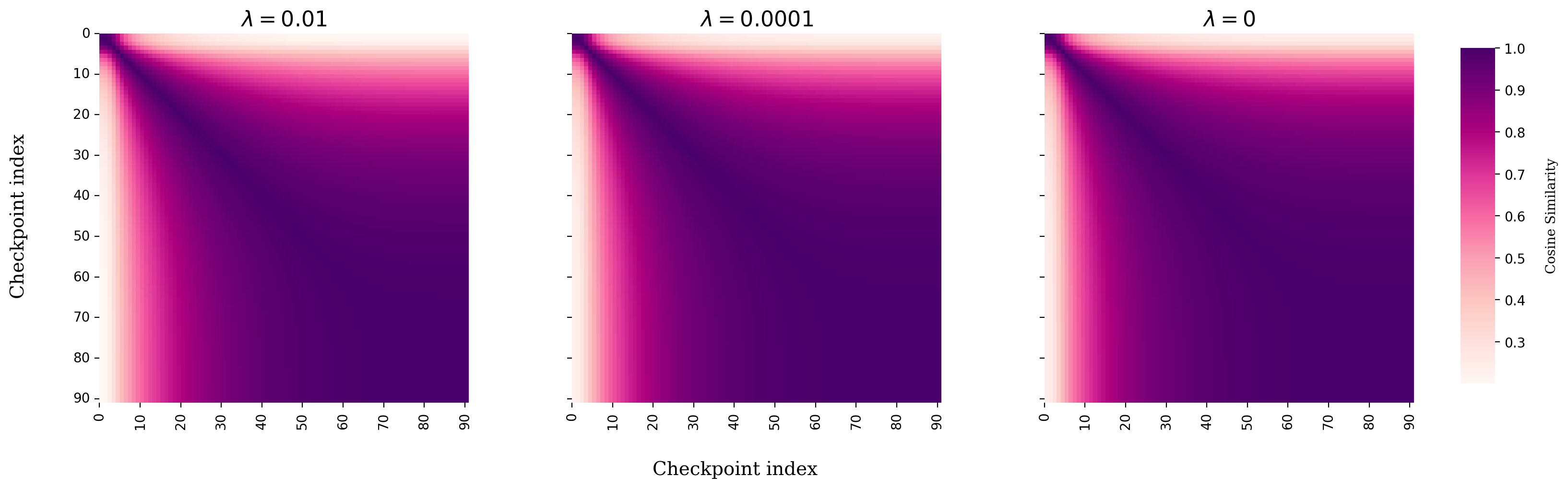}
	\caption{Trajectory Maps of ViT models across different amounts of weight decay} 
\end{figure*}

\begin{figure*}[h!]
	\centering
	\includegraphics[width=0.9\textwidth]{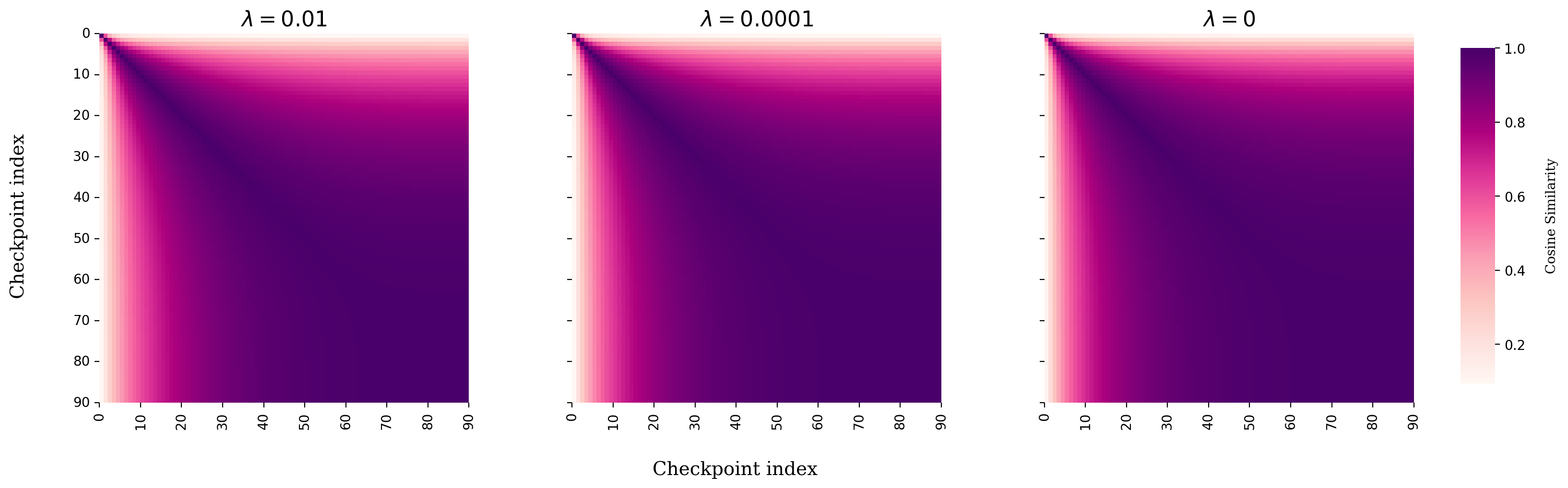}
	\caption{Relative Trajectory Maps, with respect to initialization, of ViT models across different amounts of weight decay} 
\end{figure*}

\clearpage
\begin{figure*}[h!]
	\centering
	\subfigure[$\angle(\theta_{t+1}-\theta_t,\theta_t)$]{\label{fig:}
		\includegraphics[width=0.34\textwidth]{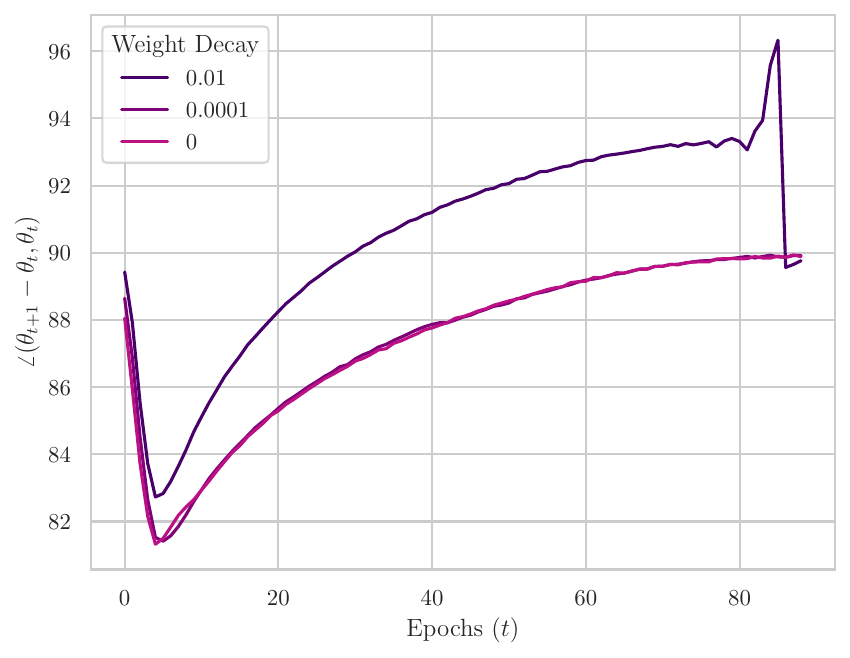}}
	\subfigure[$\angle(\theta_{t+1}-\theta_t,\theta_T-\theta_0)$]{\label{fig:}
		\includegraphics[width=0.34\textwidth]{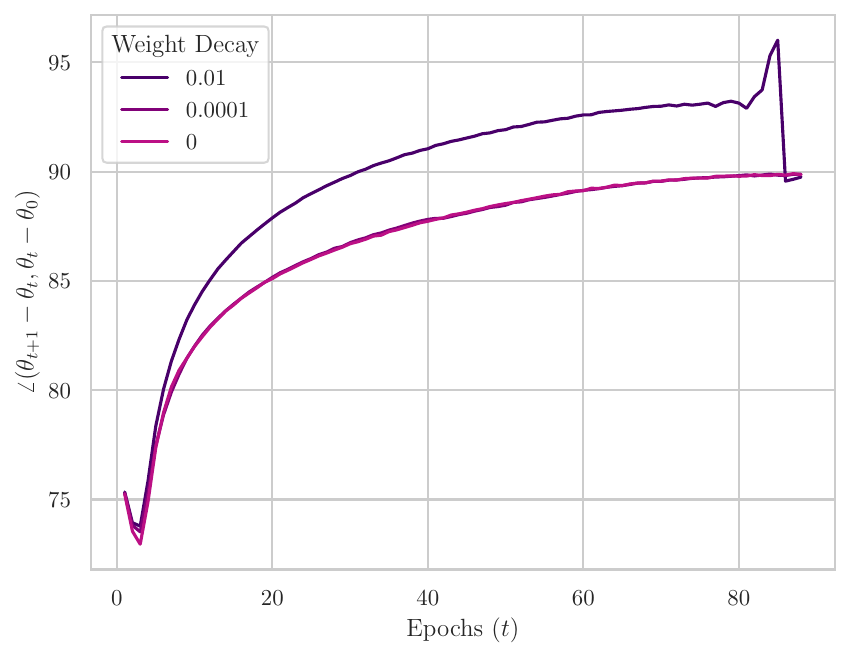}}
	\subfigure[$\angle(\theta_{t+k}-\theta_t,\theta_t-\theta_{t-k})$, for $k=1$ ]{\label{fig:}
		\includegraphics[width=0.34\textwidth]{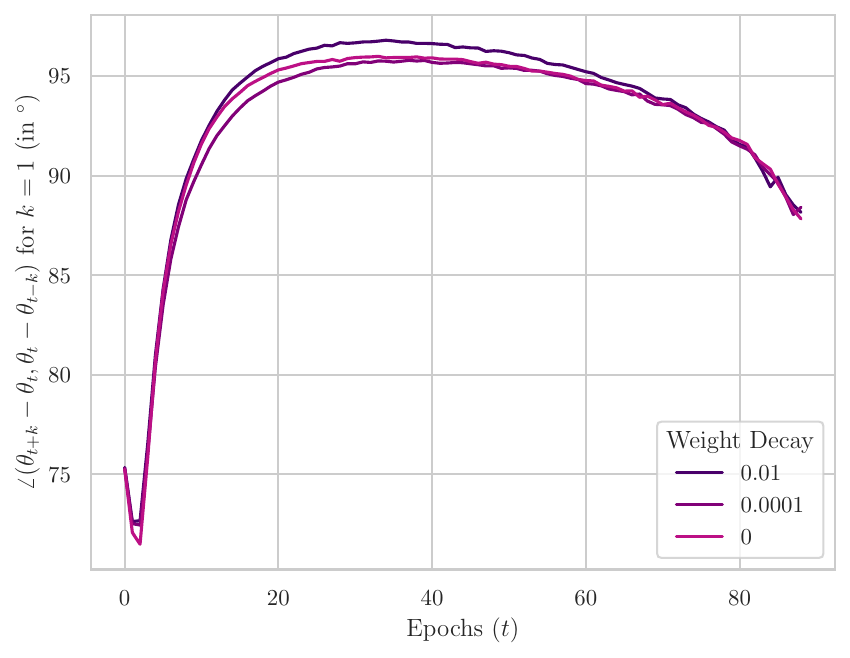}
		\vspace{-2mm}}
	\subfigure[$\angle(\theta_{t}-\theta_0,\theta_T-\theta_0)$]{\label{fig:}
		\includegraphics[width=0.34\textwidth]{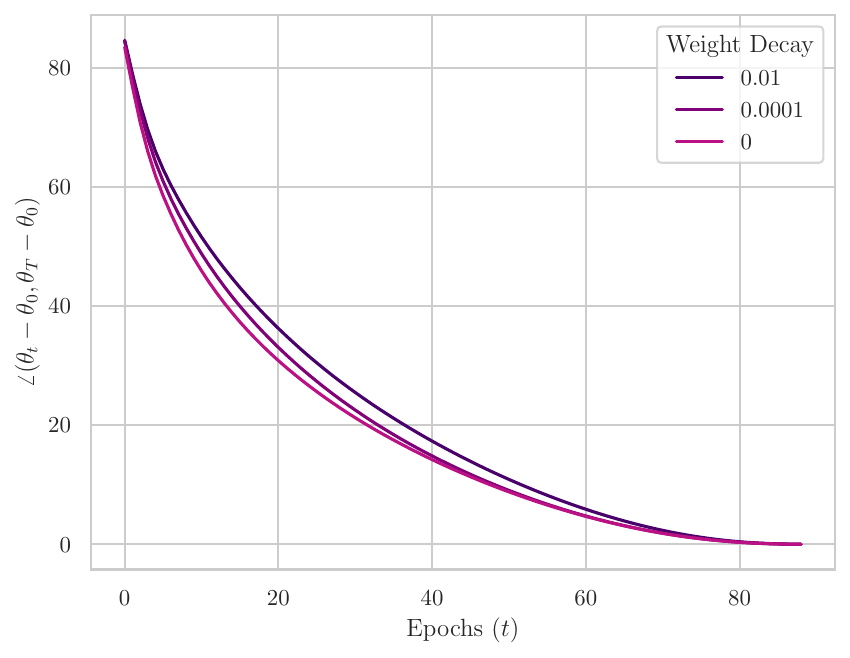}
		\vspace{-2mm}}
	\subfigure[$\angle(\theta_{t+1}-\theta_t, \theta_T-\theta_0)$]{\label{fig:}
		\includegraphics[width=0.34\textwidth]{figures/icml/Weight_Decay/ckpt_freq-1_heatmap_from_multi-3_simple_vit_opt-adamw_imagenet_ep-90_lr-0.0005_bsz-512_mom-0.5_wdecay-0.01_aug-True_seed-0_2024-01-26_16-22-24_024509_2024-02-02_01-26-09_479188/figures/pdf/angle_theta__t+1_-theta_t,theta_T-theta_0__vs_Epochs__t__across_Weight_Decay.pdf}}
	\subfigure[Apex Angle at Initialization $\angle(\theta_t-\theta_0,\theta_1-\theta_0)$ ]{\label{fig:}
		\includegraphics[width=0.34\textwidth]{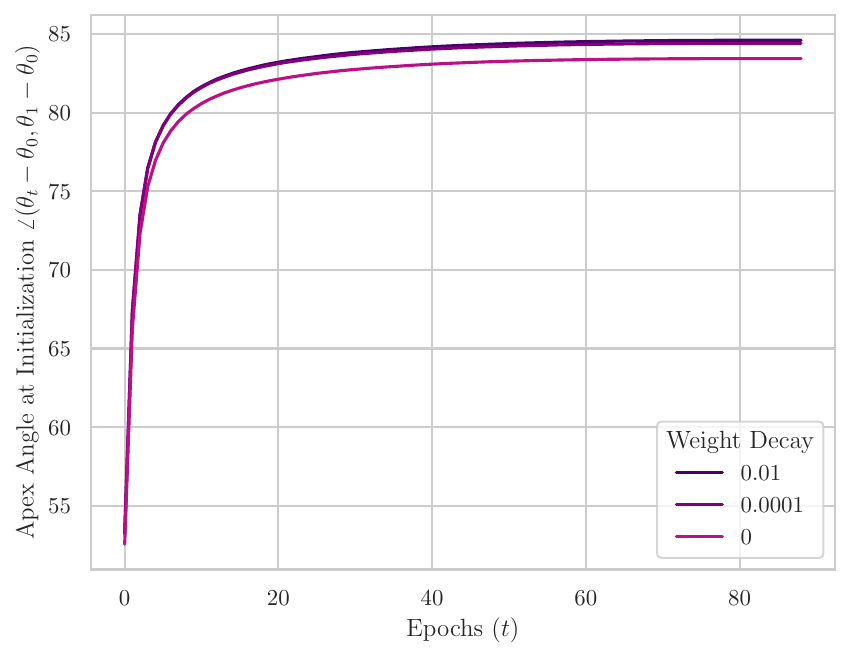}
		\vspace{-2mm}}
	\subfigure[Apex Angle at Origin $\angle(\theta_t,\theta_0)$]{\label{fig:}
		\includegraphics[width=0.34\textwidth]{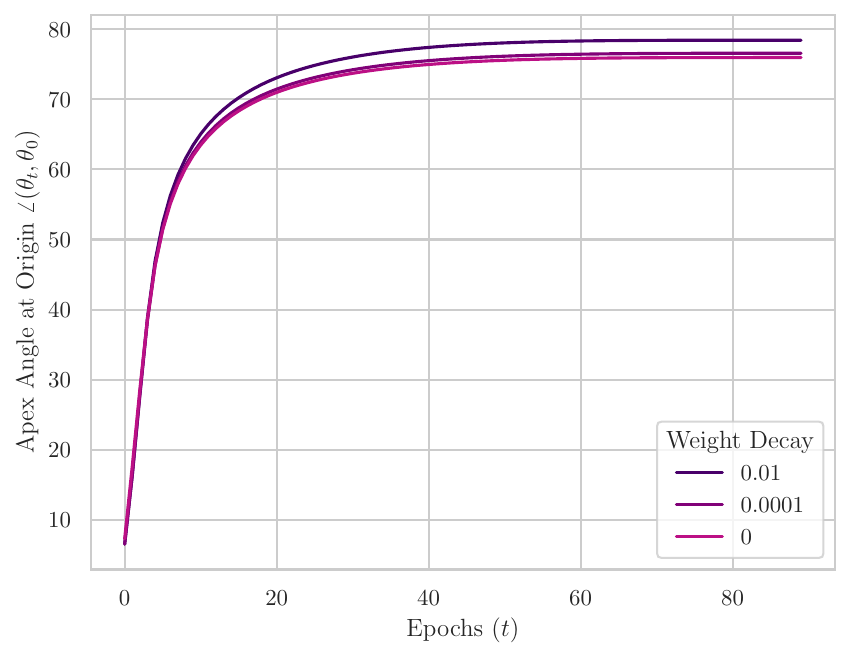}
		\vspace{-2mm}}
	\caption{Angular measures of the Trajectory for ViT trained on the ImageNet dataset} 
\end{figure*}

\begin{figure*}[h!]
	\centering
	\subfigure[$\|\theta_t\|_2$]{\label{fig:}
	\includegraphics[width=0.3\textwidth]{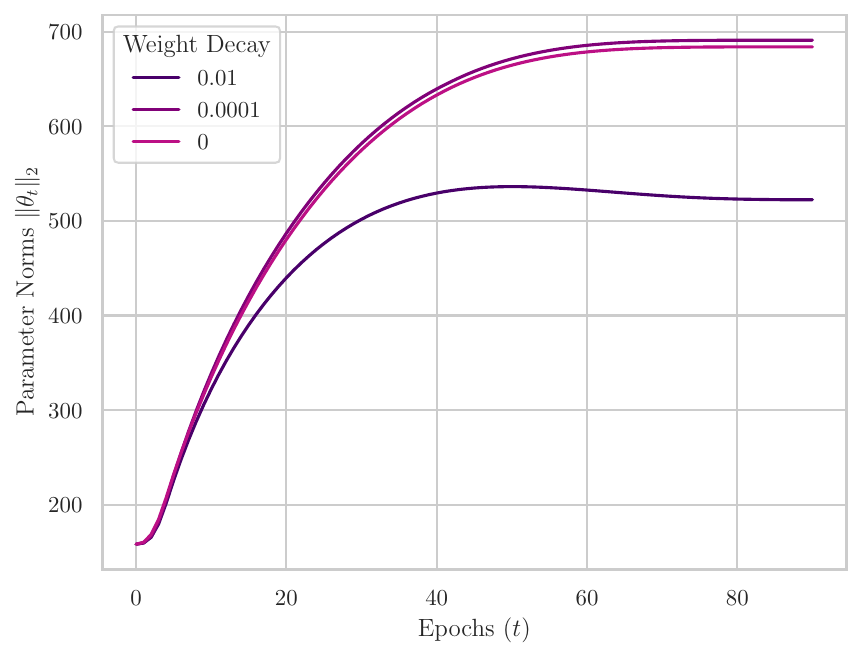}}
	\subfigure[$\|\theta_{t+k}-\theta_t\|_2$ ]{\label{fig:}
		\includegraphics[width=0.3\textwidth]{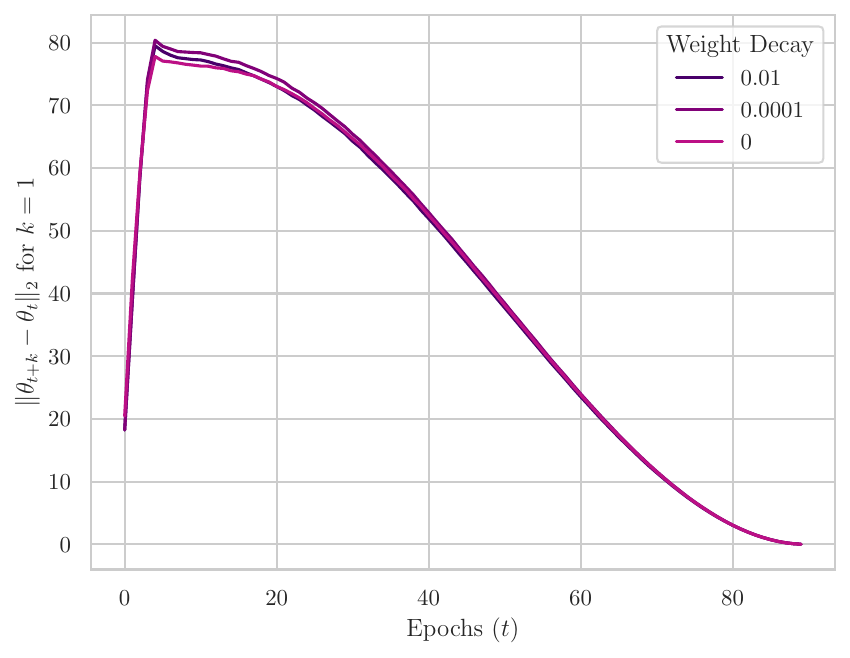}}
	\subfigure[$\|\theta_t-\theta_0\|_2$ ]{\label{fig:}
		\includegraphics[width=0.3\textwidth]{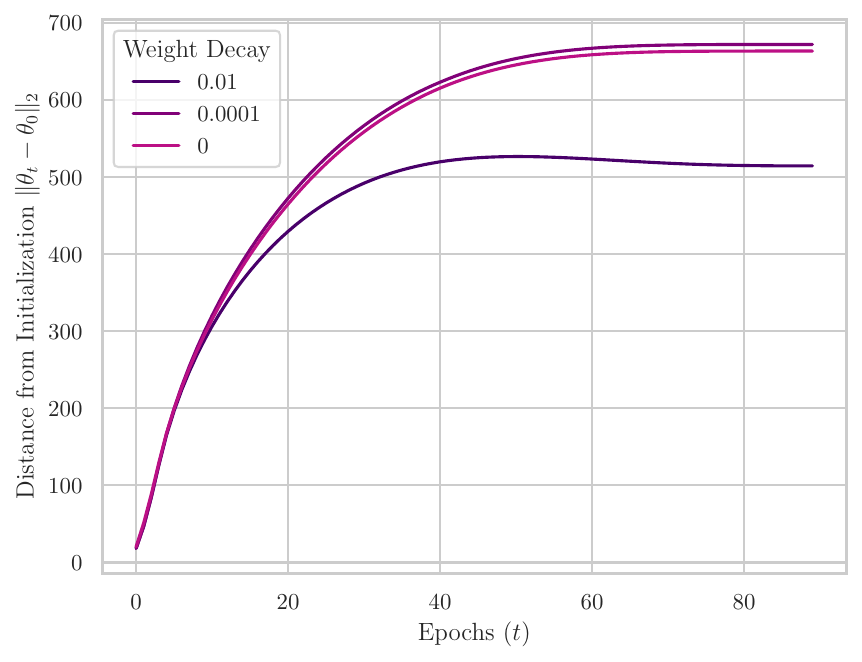}
		\vspace{-2mm}}
	\caption{Norm-based measures of the Trajectory for ViT trained on the ImageNet dataset} 
\end{figure*}

\begin{figure*}[h!]
	\centering
	\subfigure[Eigenvalues: $\Km$]{\label{fig:}
		\includegraphics[width=0.3\textwidth]{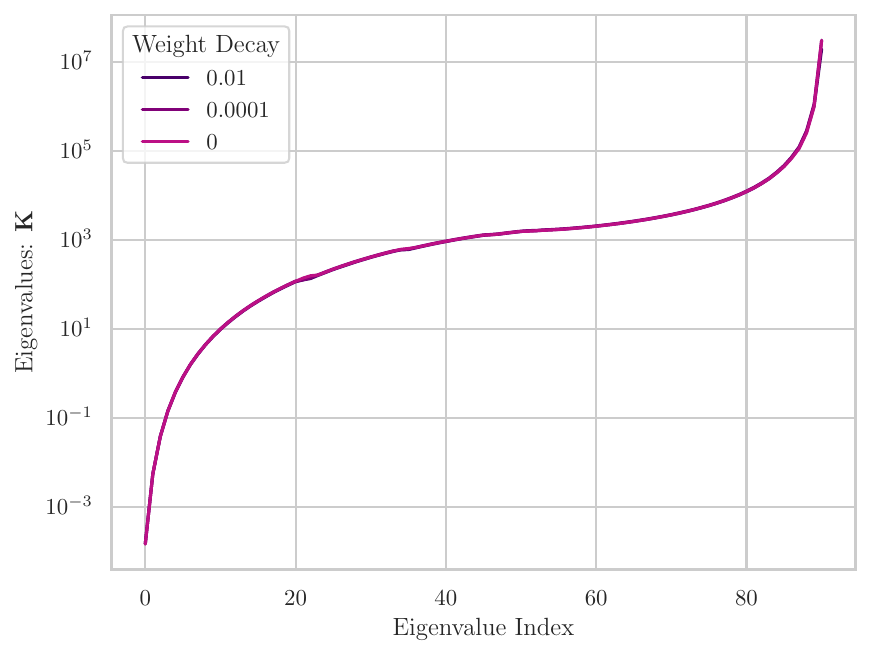}}
	\subfigure[Eigenvalues: $\Km_0$]{\label{fig:}
		\includegraphics[width=0.3\textwidth]{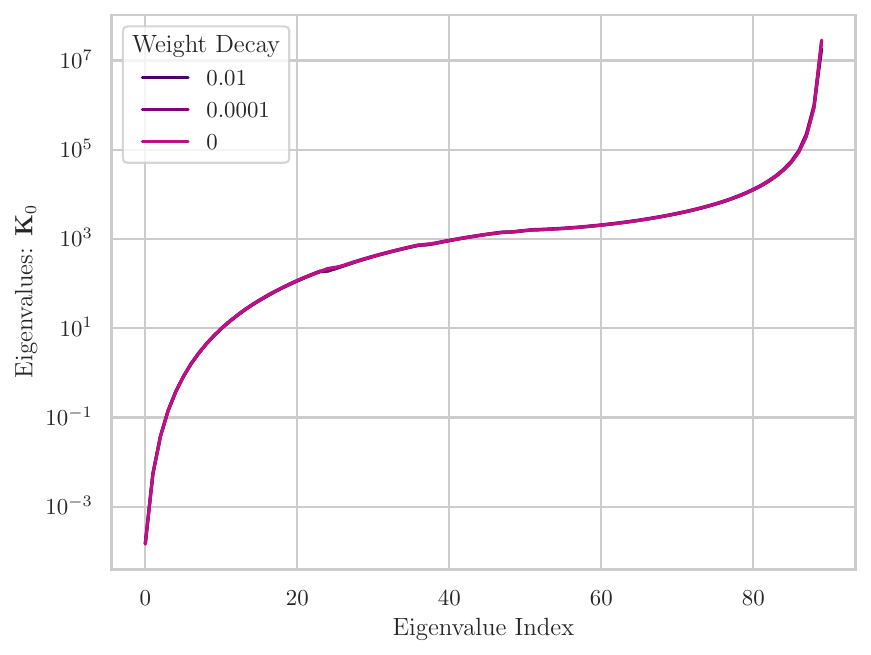}}

	\subfigure[Eigenvalues: $\Cm$ ]{\label{fig:}
		\includegraphics[width=0.3\textwidth]{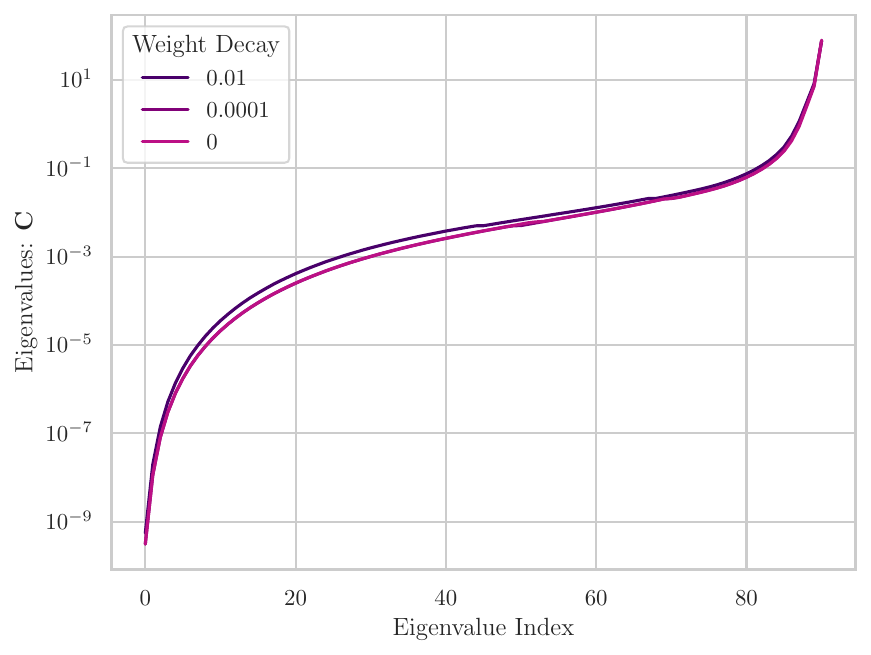}
		\vspace{-2mm}}
	\subfigure[Eigenvalues: $\Cm_0$]{\label{fig:}
		\includegraphics[width=0.3\textwidth]{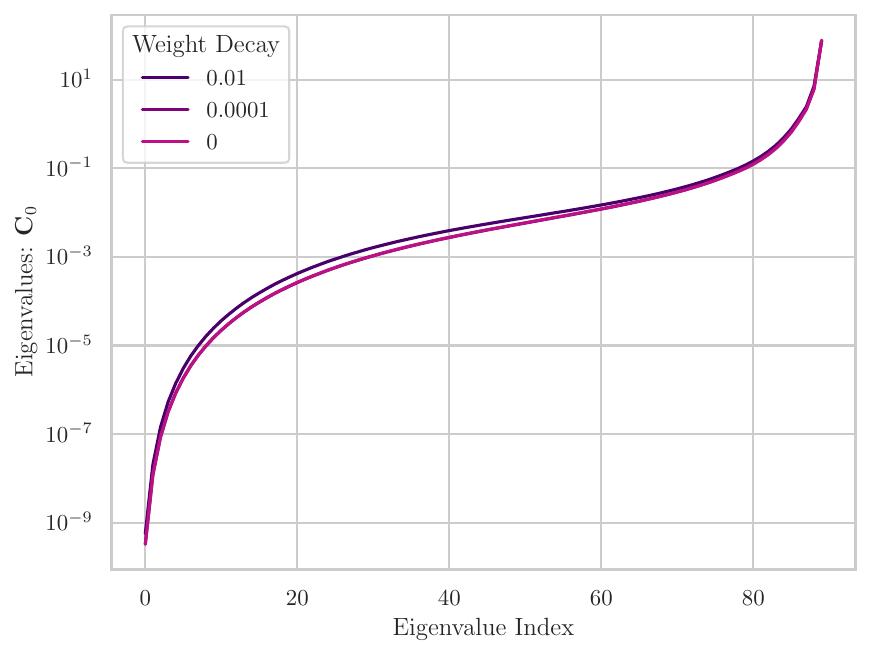}
		\vspace{-2mm}}
	\caption{Spectral measures of the Trajectory for ViT trained on ImageNet} 
\end{figure*}

\clearpage

\subsection{ResNet50: Weight Decay, SGD}

\begin{figure*}[h!]
	\centering
	\includegraphics[width=0.9\textwidth]{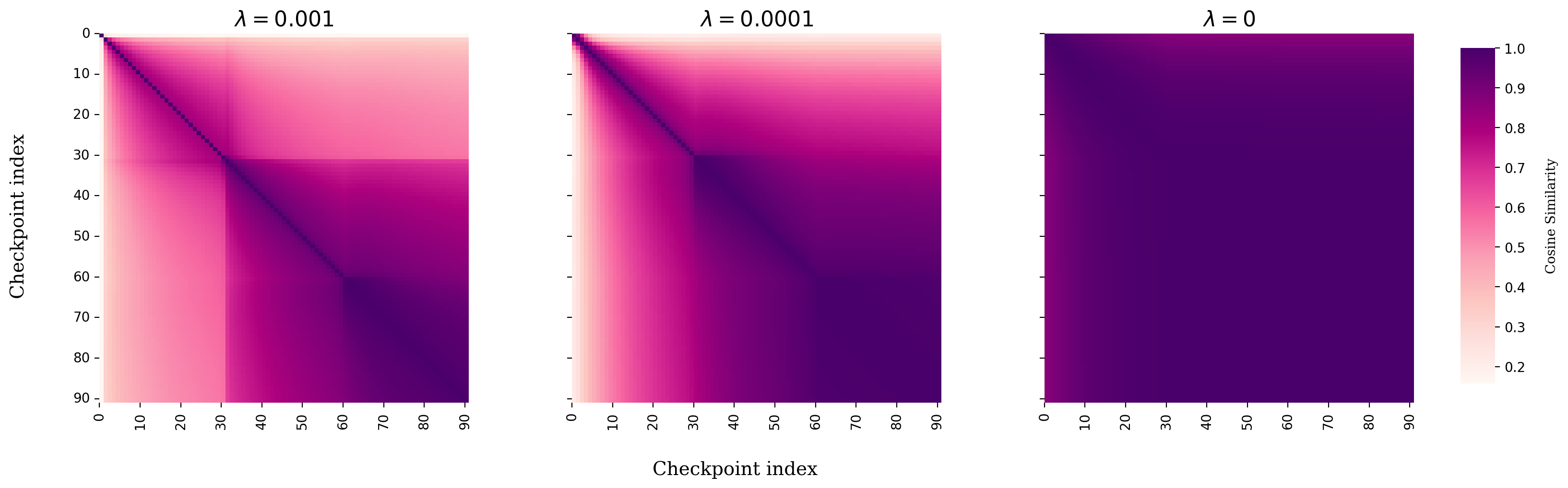}
	\caption{Trajectory Maps of ResNet50 models across different amounts of weight decay} 
\end{figure*}

\begin{figure*}[h!]
	\centering
	\includegraphics[width=0.9\textwidth]{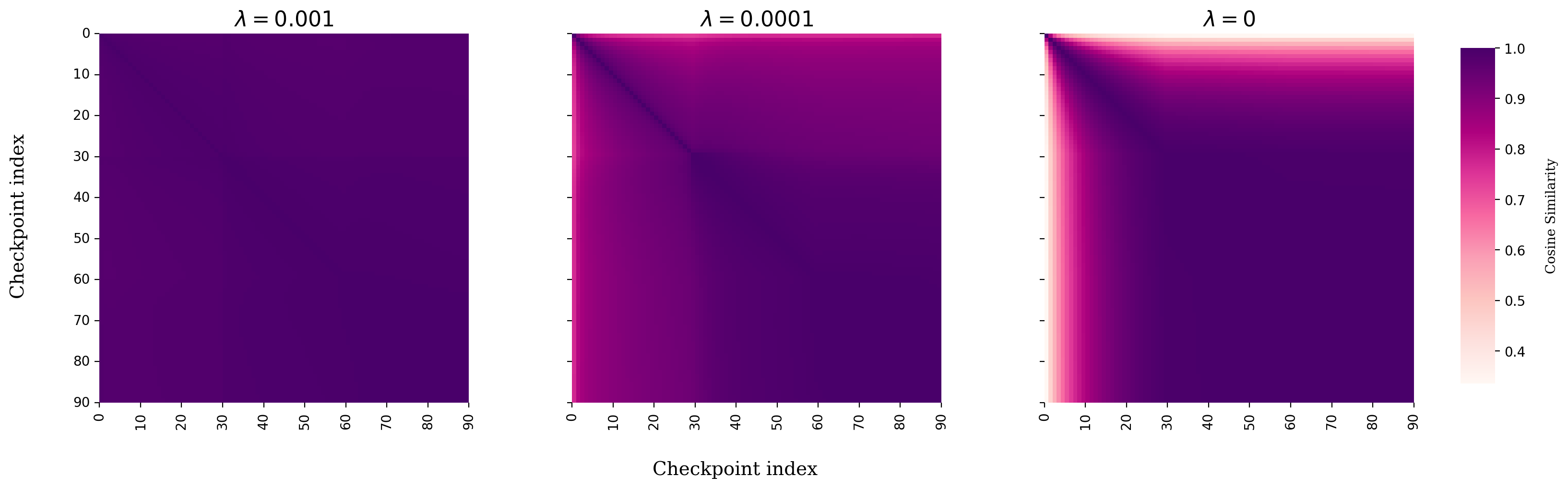}
	\caption{Relative Trajectory Maps, with respect to initialization, of ResNet50 models across different amounts of weight decay} 
\end{figure*}

\clearpage
\begin{figure*}[h!]
	\centering
	\subfigure[$\angle(\theta_{t+1}-\theta_t,\theta_t)$]{\label{fig:}
		\includegraphics[width=0.34\textwidth]{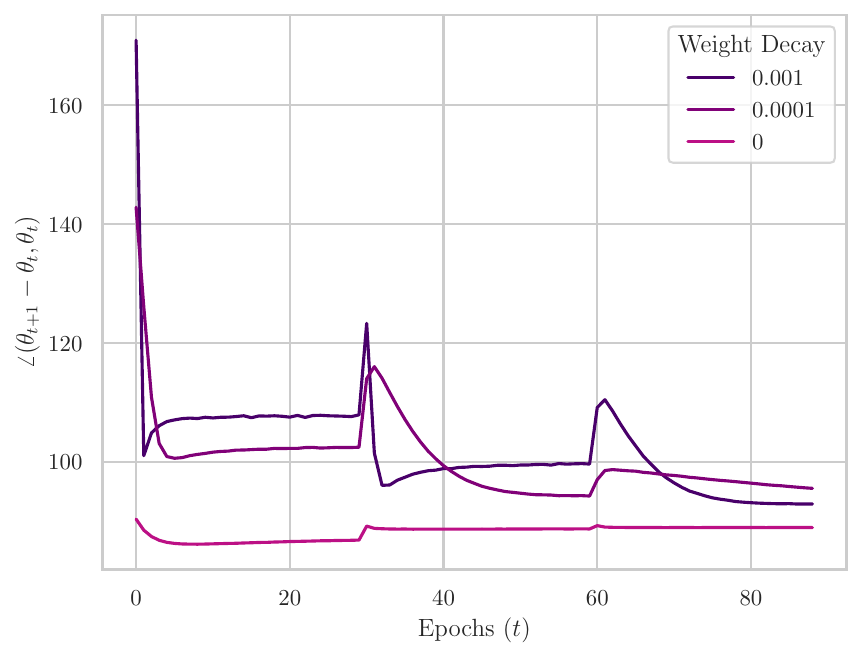}}
	\subfigure[$\angle(\theta_{t+1}-\theta_t,\theta_T-\theta_0)$]{\label{fig:}
		\includegraphics[width=0.34\textwidth]{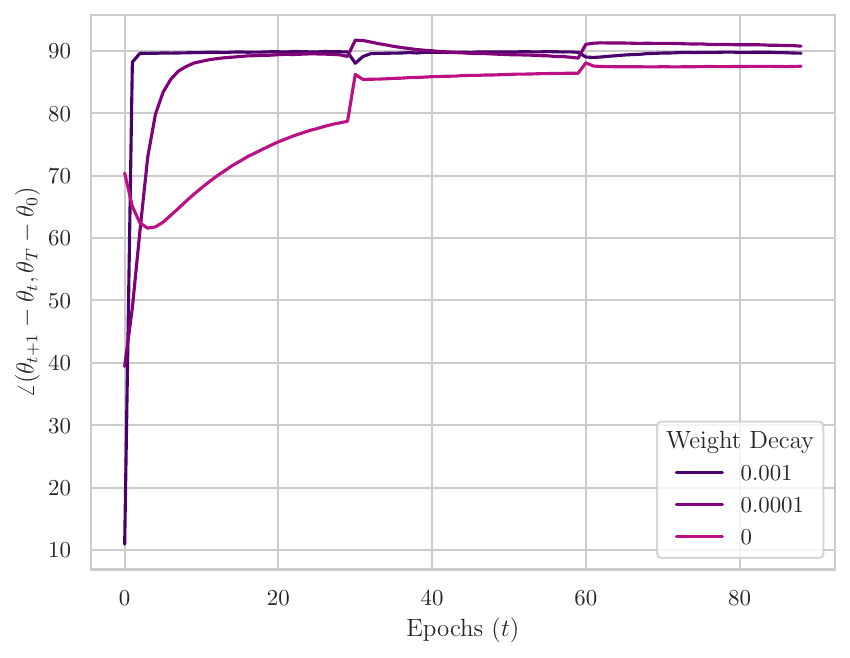}}
	\subfigure[$\angle(\theta_{t+k}-\theta_t,\theta_t-\theta_{t-k})$, for $k=1$ ]{\label{fig:}
		\includegraphics[width=0.34\textwidth]{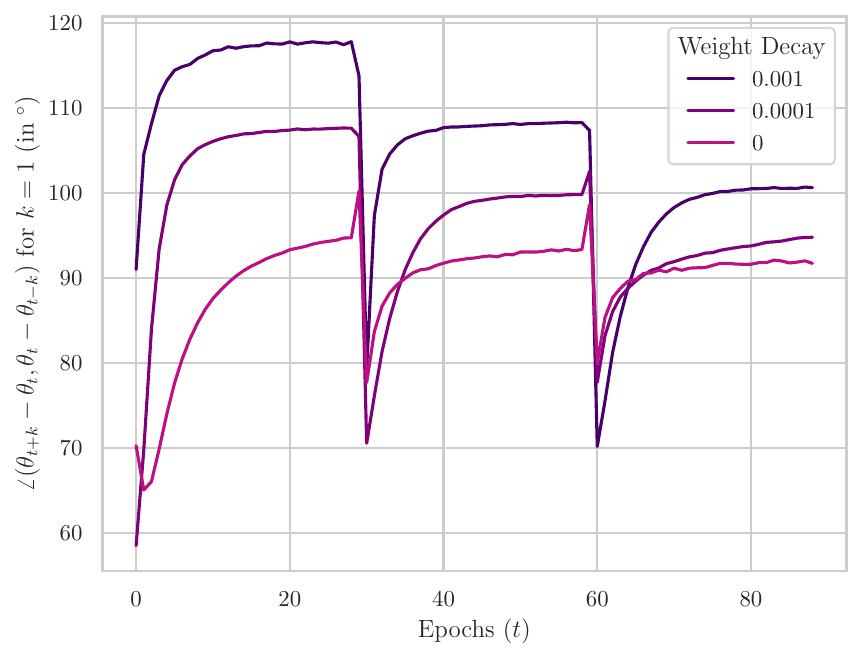}
		\vspace{-2mm}}
	\subfigure[$\angle(\theta_{t}-\theta_0,\theta_T-\theta_0)$]{\label{fig:}
		\includegraphics[width=0.34\textwidth]{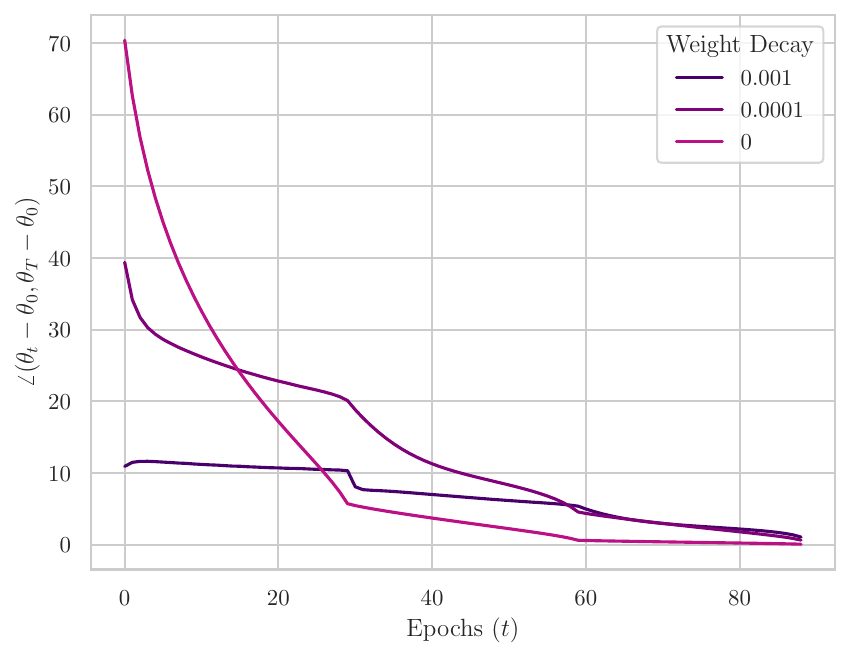}
		\vspace{-2mm}}
	\subfigure[$\angle(\theta_{t+1}-\theta_t, \theta_T-\theta_0)$]{\label{fig:}
		\includegraphics[width=0.34\textwidth]{figures/icml/Weight_Decay/ckpt_freq-1_heatmap_from_multi-3_resnet50_opt-sgd_imagenet_ep-90_lr-0.1_bsz-256_mom-0.9_wdecay-0.001_seed-0_2023-09-23_11-34-38_740477_2024-02-01_02-26-16_131005/figures/pdf/angle_theta__t+1_-theta_t,theta_T-theta_0__vs_Epochs__t__across_Weight_Decay.pdf}}
	\subfigure[Apex Angle at Initialization $\angle(\theta_t-\theta_0,\theta_1-\theta_0)$ ]{\label{fig:}
		\includegraphics[width=0.34\textwidth]{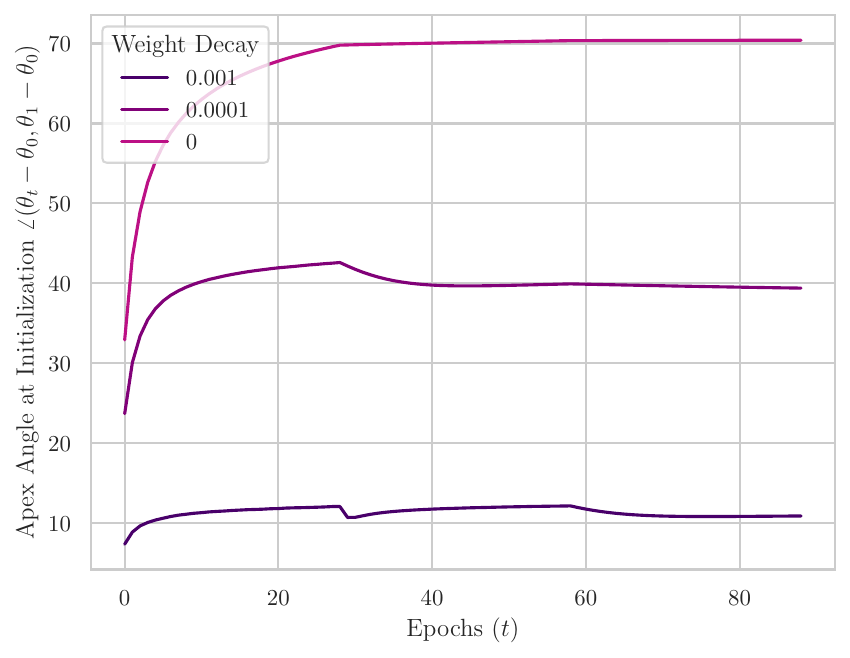}
		\vspace{-2mm}}
	\subfigure[Apex Angle at Origin $\angle(\theta_t,\theta_0)$]{\label{fig:}
		\includegraphics[width=0.34\textwidth]{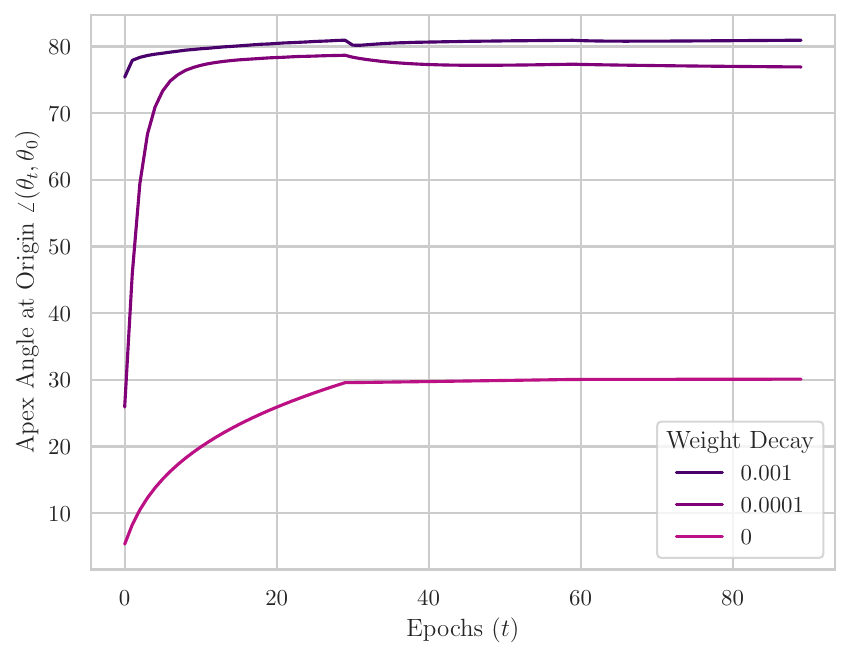}
		\vspace{-2mm}}
	\caption{Angular measures of the Trajectory for ResNet50 trained on ImageNet} 
\end{figure*}

\begin{figure*}[h!]
	\centering
	\subfigure[$\|\theta_t\|_2$]{\label{fig:}
	\includegraphics[width=0.3\textwidth]{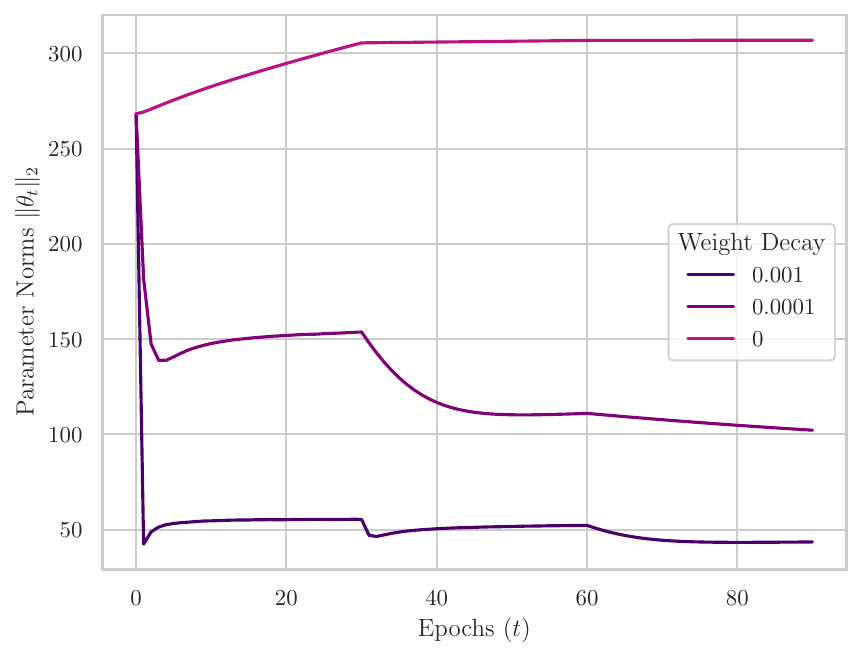}}
	\subfigure[$\|\theta_{t+k}-\theta_t\|_2$ ]{\label{fig:}
		\includegraphics[width=0.3\textwidth]{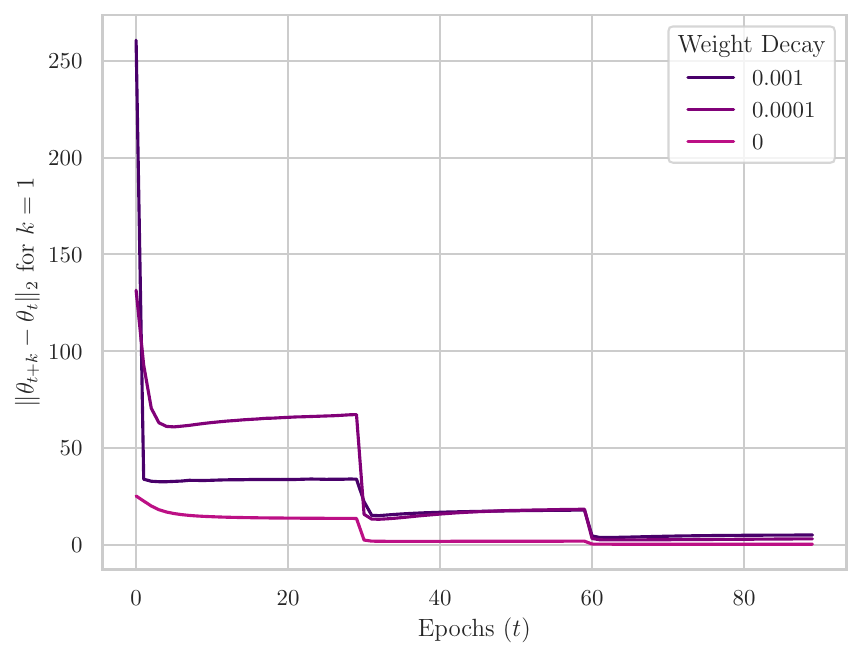}}
	\subfigure[$\|\theta_t-\theta_0\|_2$ ]{\label{fig:}
		\includegraphics[width=0.3\textwidth]{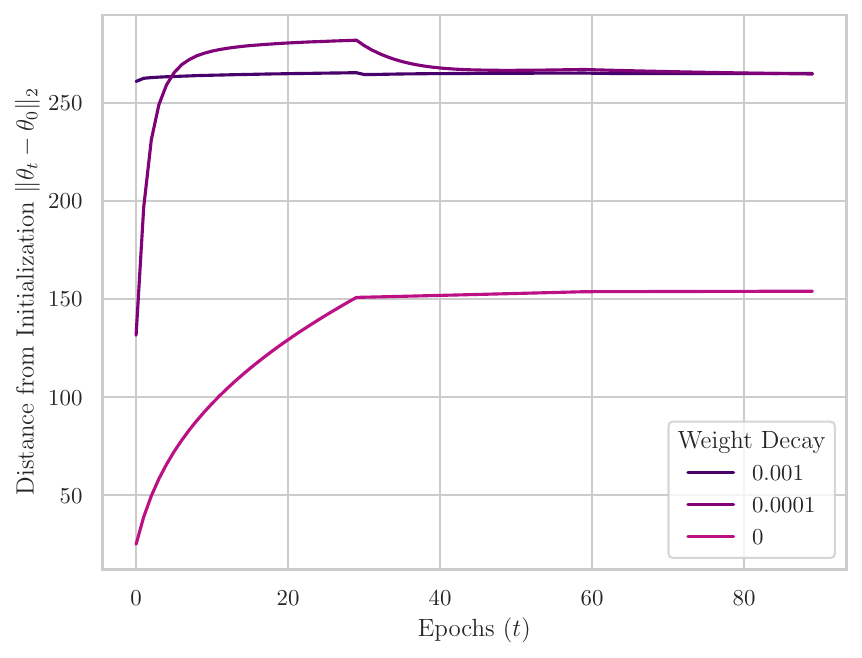}
		\vspace{-2mm}}
	\caption{Norm-based measures of the Trajectory for ResNet50 trained on ImageNet} 
\end{figure*}

\begin{figure*}[h!]
	\centering
	\subfigure[Eigenvalues: $\Km$]{\label{fig:}
		\includegraphics[width=0.3\textwidth]{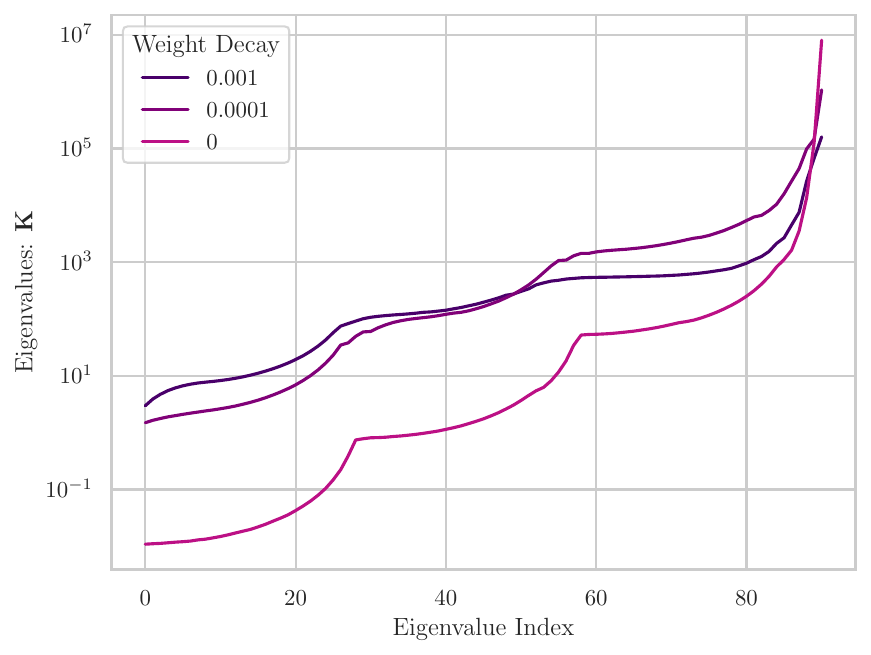}}
	\subfigure[Eigenvalues: $\Km_0$]{\label{fig:}
		\includegraphics[width=0.3\textwidth]{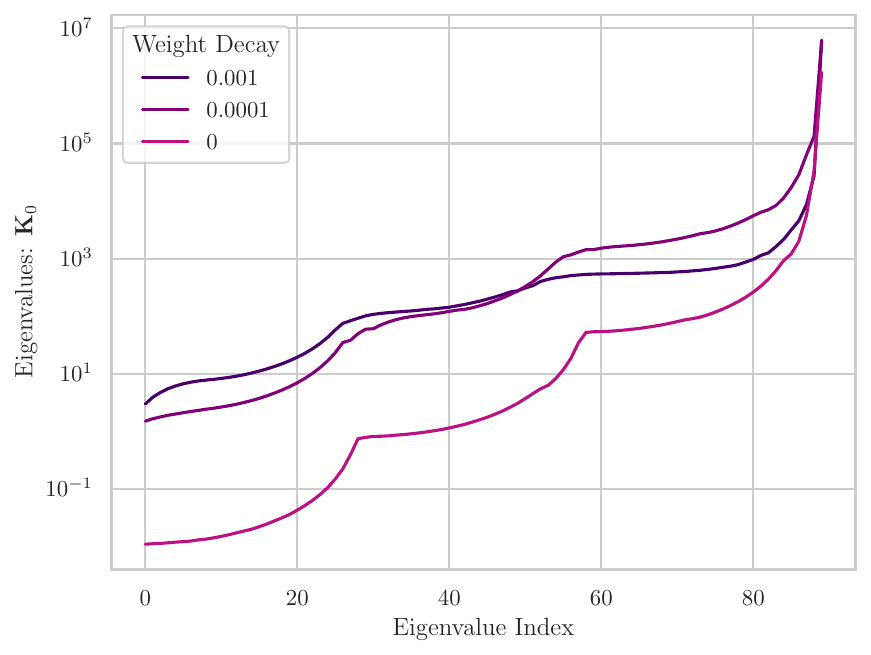}}

	\subfigure[Eigenvalues: $\Cm$ ]{\label{fig:}
		\includegraphics[width=0.3\textwidth]{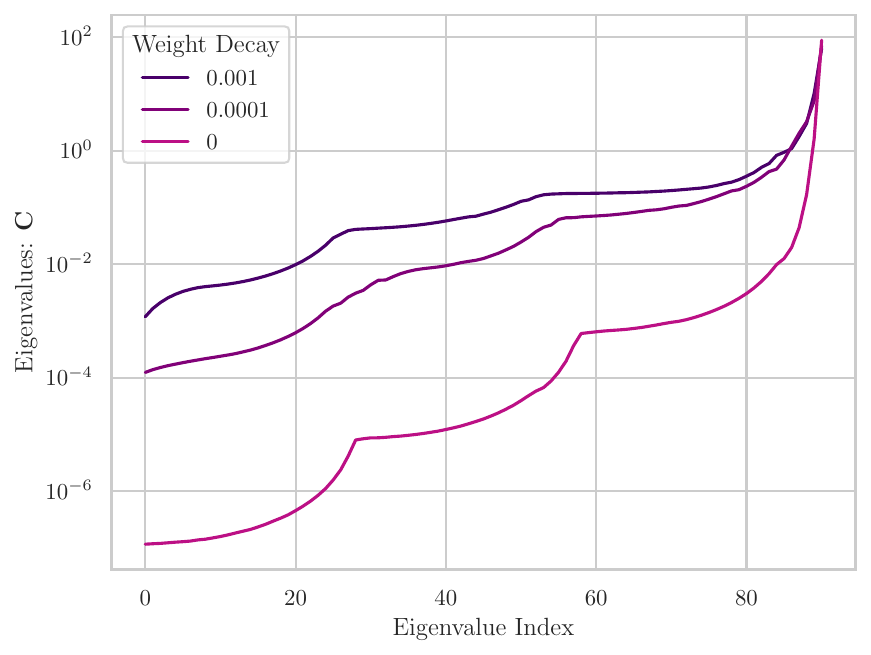}
		\vspace{-2mm}}
	\subfigure[Eigenvalues: $\Cm_0$]{\label{fig:}
		\includegraphics[width=0.3\textwidth]{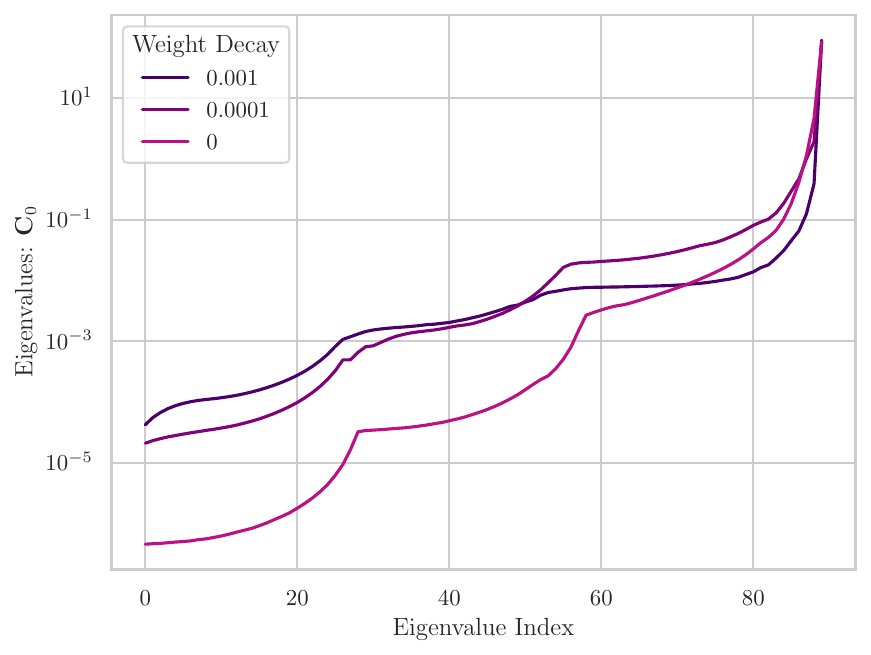}
		\vspace{-2mm}}
	\caption{Spectral measures of the Trajectory for ResNet50 trained on ImageNet} 
\end{figure*}

\clearpage

\subsection{ResNet50: Sharpness Aware Minimization analysis}\label{app:sam}

\begin{figure*}[h!]
	\centering
	\includegraphics[width=0.9\textwidth]{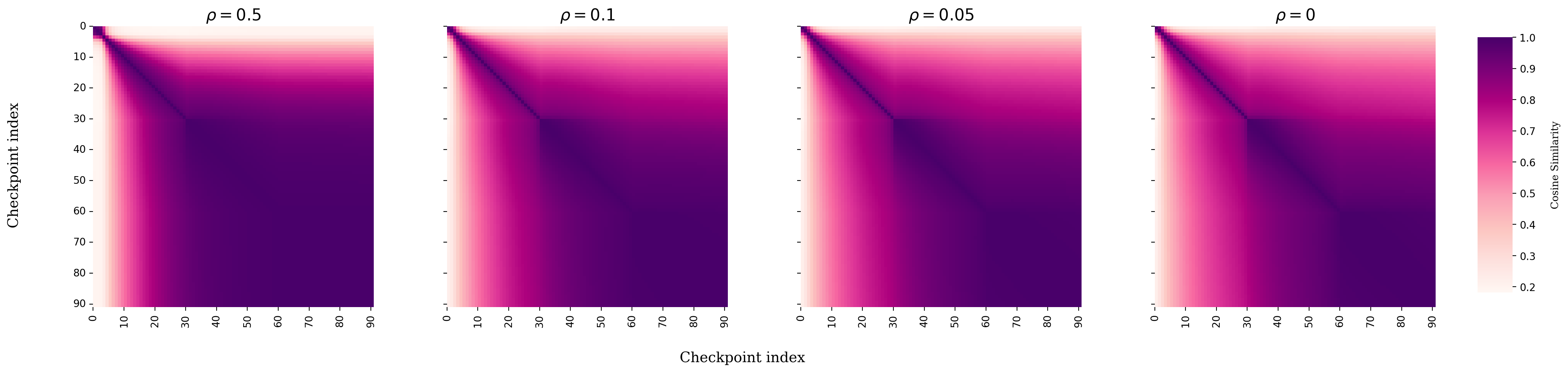}
	\caption{Trajectory Maps of ResNet50 models across different values of SAM regularization coefficient} 
\end{figure*}

\begin{figure*}[h!]
	\centering
	\includegraphics[width=0.9\textwidth]{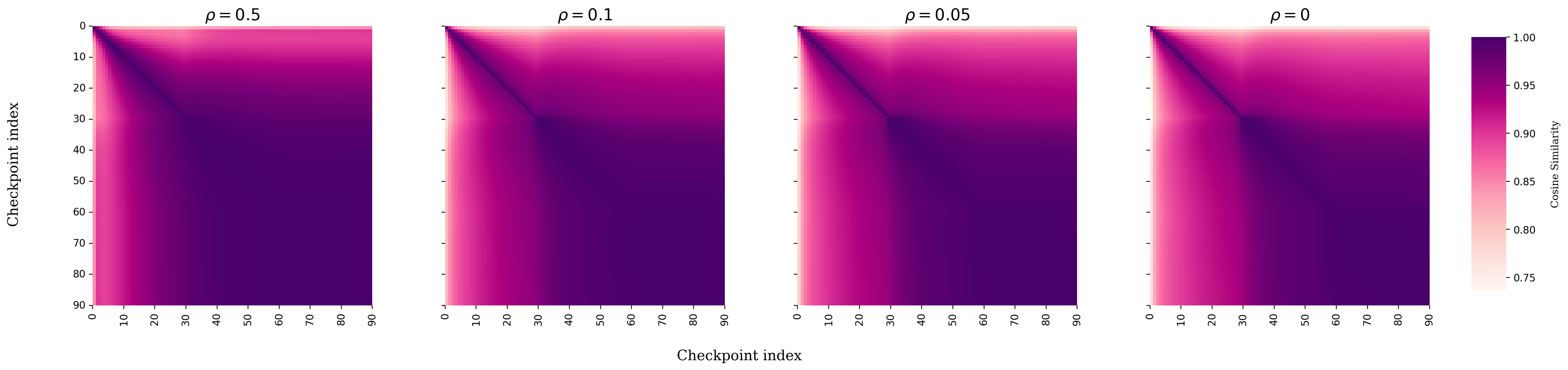}
	\caption{Relative Trajectory Maps, with respect to initialization, of ResNet50 models across different values of SAM regularization coefficient} 
\end{figure*}

\clearpage
\begin{figure*}[h!]
	\centering
	\subfigure[$\angle(\theta_{t+1}-\theta_t,\theta_t)$]{\label{fig:}
		\includegraphics[width=0.34\textwidth]{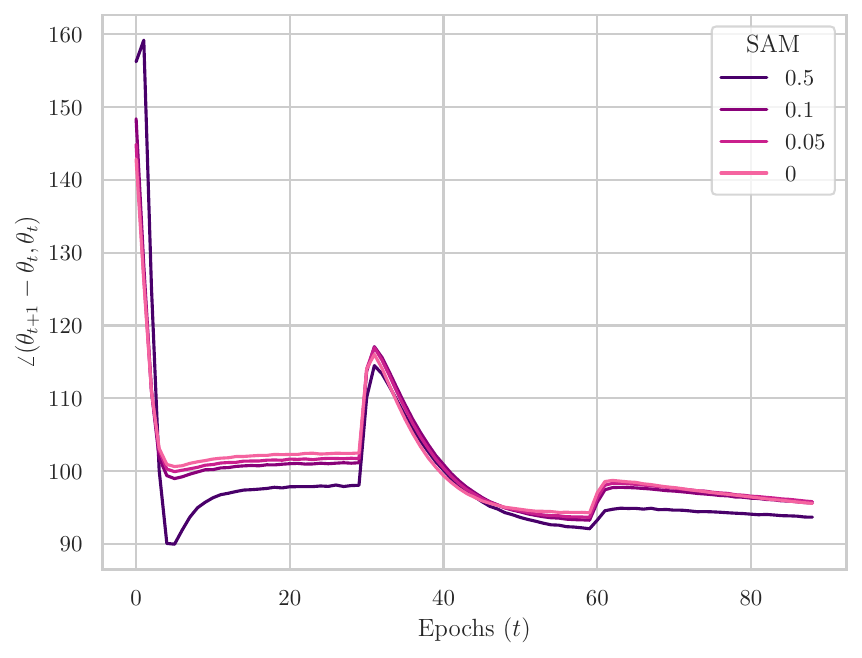}}
	\subfigure[$\angle(\theta_{t+1}-\theta_t,\theta_T-\theta_0)$]{\label{fig:}
		\includegraphics[width=0.34\textwidth]{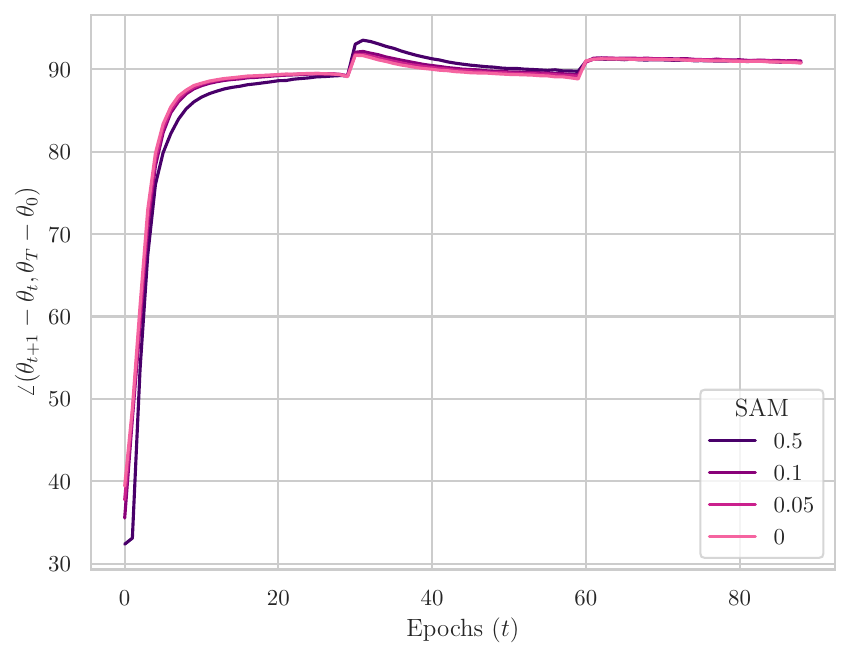}}
	\subfigure[$\angle(\theta_{t+k}-\theta_t,\theta_t-\theta_{t-k})$, for $k=1$ ]{\label{fig:}
		\includegraphics[width=0.34\textwidth]{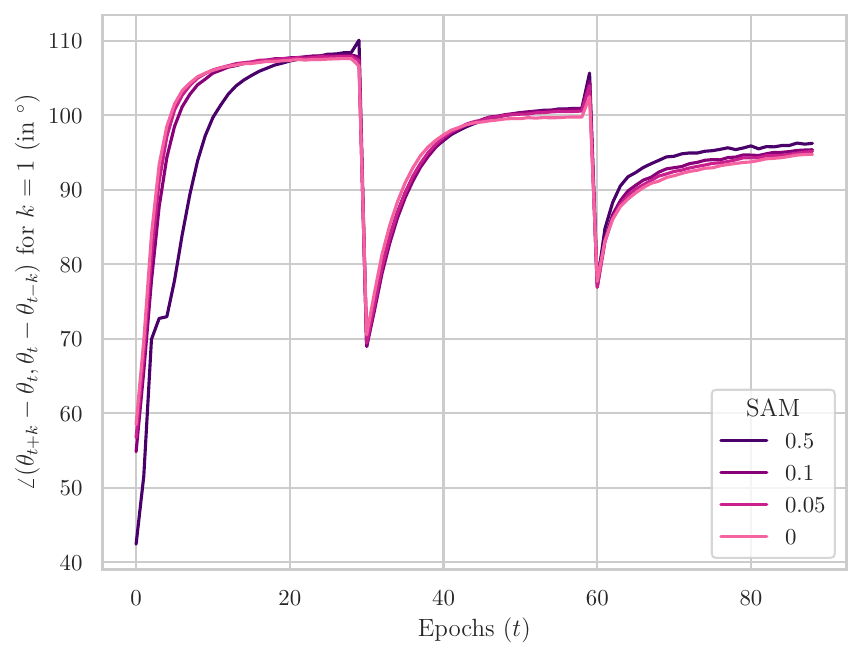}
		\vspace{-2mm}}
	\subfigure[$\angle(\theta_{t}-\theta_0,\theta_T-\theta_0)$]{\label{fig:}
		\includegraphics[width=0.34\textwidth]{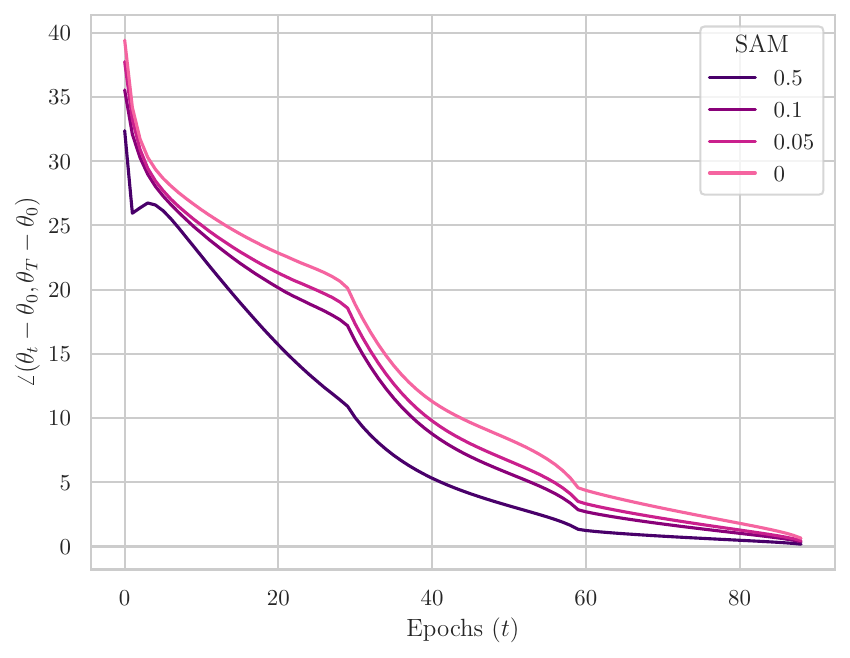}
		\vspace{-2mm}}
	\subfigure[$\angle(\theta_{t+1}-\theta_t, \theta_T-\theta_0)$]{\label{fig:}
		\includegraphics[width=0.34\textwidth]{figures/icml/SAM/ckpt_freq-1_heatmap_from_multi-4_resnet50_opt-sam_imagenet_ep-90_lr-0.1_bsz-256_mom-0.9_wdecay-0.0001_seed-0_2023-10-12_09-42-24_289304_2024-02-01_02-26-27_021523/figures/pdf/angle_theta__t+1_-theta_t,theta_T-theta_0__vs_Epochs__t__across_SAM.pdf}}
	\subfigure[Apex Angle at Initialization $\angle(\theta_t-\theta_0,\theta_1-\theta_0)$ ]{\label{fig:}
		\includegraphics[width=0.34\textwidth]{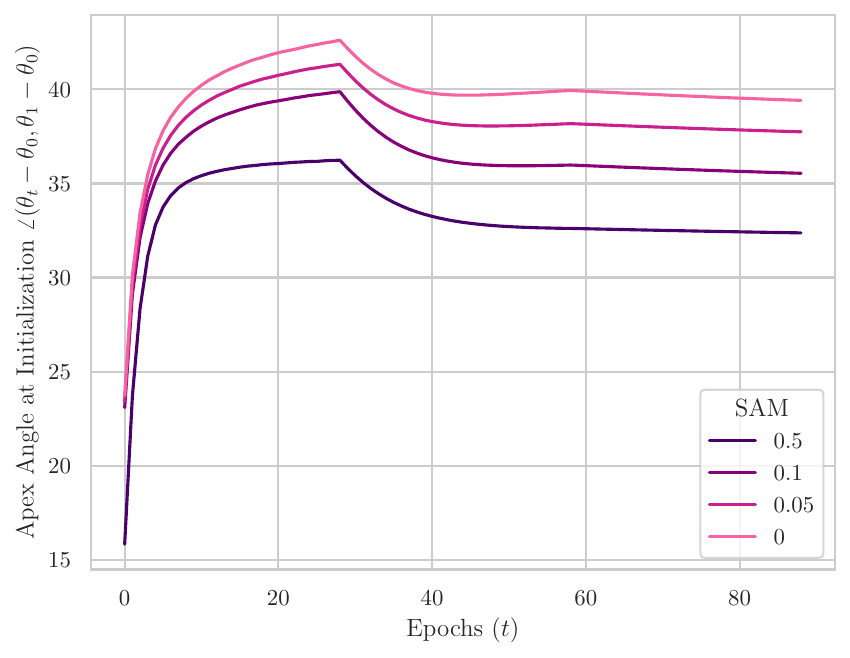}
		\vspace{-2mm}}
	\subfigure[Apex Angle at Origin $\angle(\theta_t,\theta_0)$]{\label{fig:}
		\includegraphics[width=0.34\textwidth]{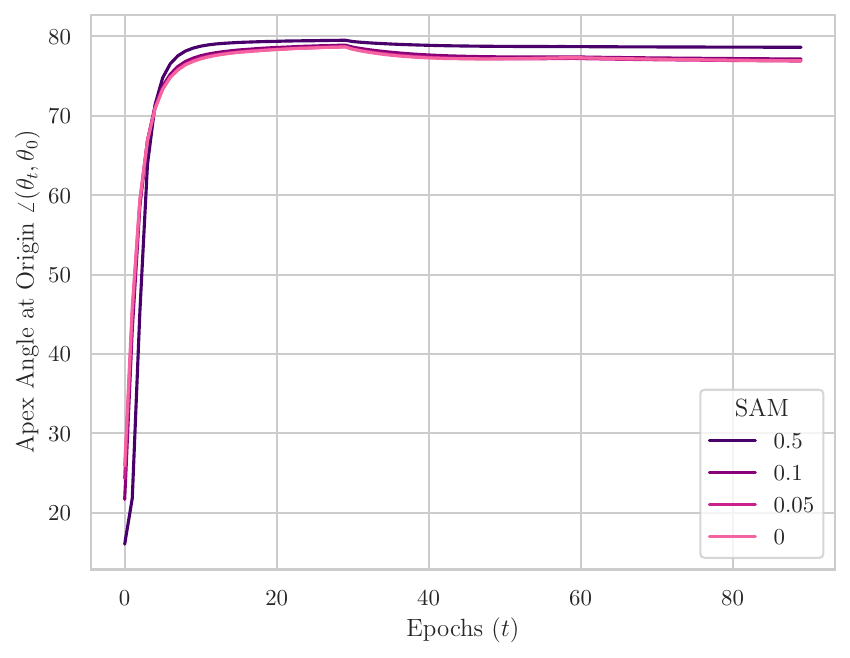}
		\vspace{-2mm}}
	\caption{Angular measures of the Trajectory for ResNet50 trained on ImageNet} 
\end{figure*}

\begin{figure*}[h!]
	\centering
	\subfigure[$\|\theta_t\|_2$]{\label{fig:}
	\includegraphics[width=0.3\textwidth]{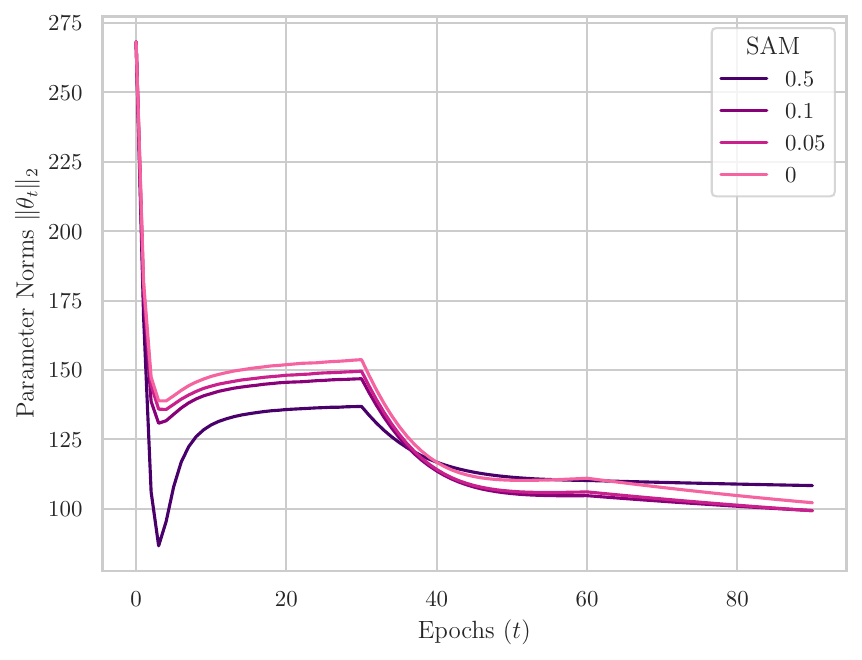}}
	\subfigure[$\|\theta_{t+k}-\theta_t\|_2$ ]{\label{fig:}
		\includegraphics[width=0.3\textwidth]{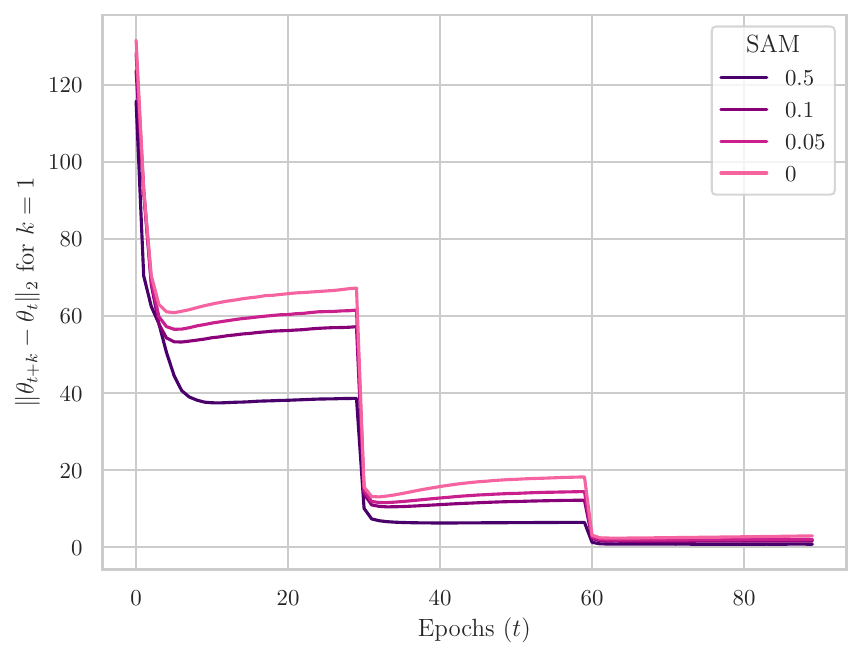}}
	\subfigure[$\|\theta_t-\theta_0\|_2$ ]{\label{fig:}
		\includegraphics[width=0.3\textwidth]{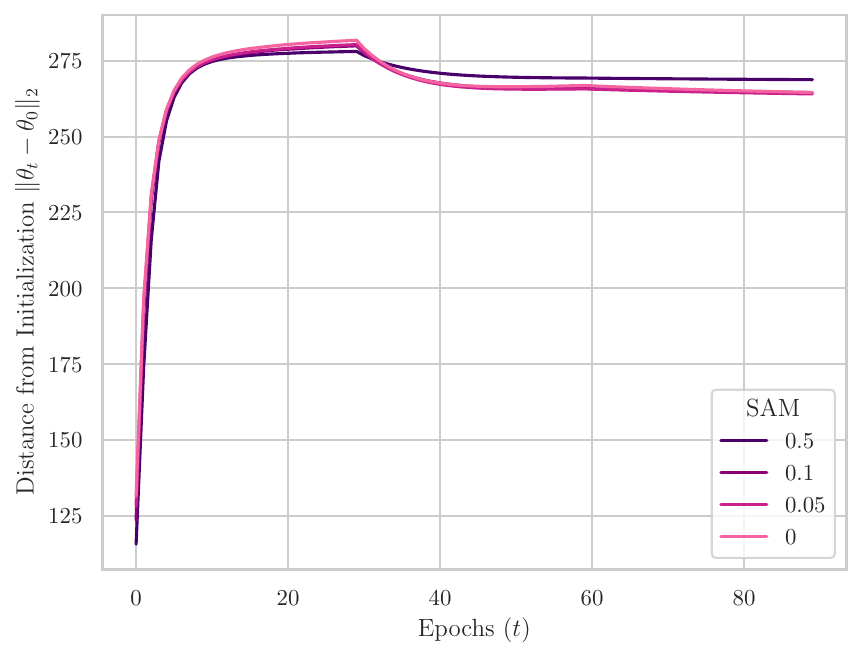}
		\vspace{-2mm}}
	\caption{Norm-based measures of the Trajectory for ResNet50 trained on ImageNet} 
\end{figure*}

\begin{figure*}[h!]
	\centering
	\subfigure[Eigenvalues: $\Km$]{\label{fig:}
		\includegraphics[width=0.3\textwidth]{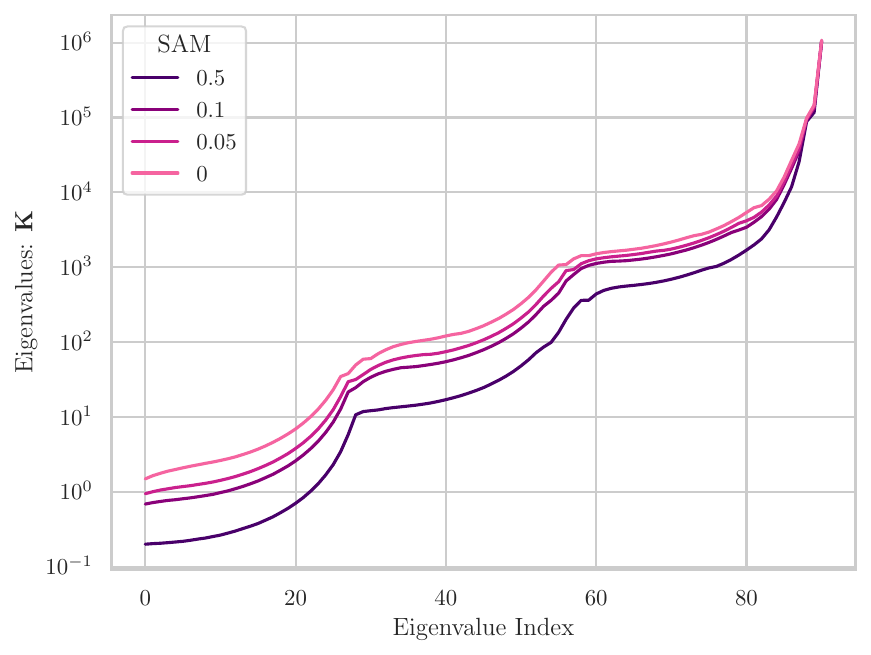}}
	\subfigure[Eigenvalues: $\Km_0$]{\label{fig:}
		\includegraphics[width=0.3\textwidth]{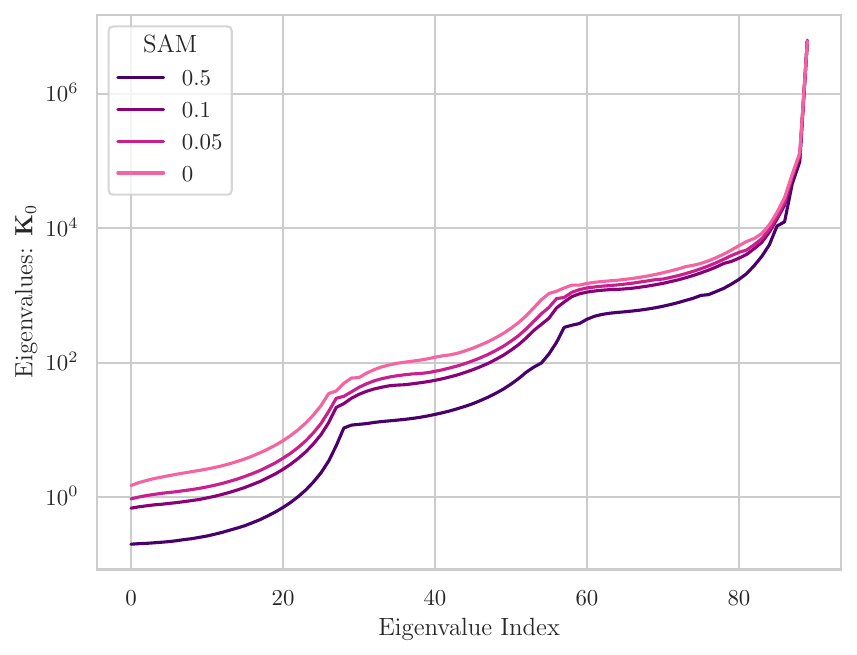}}

	\subfigure[Eigenvalues: $\Cm$ ]{\label{fig:}
		\includegraphics[width=0.3\textwidth]{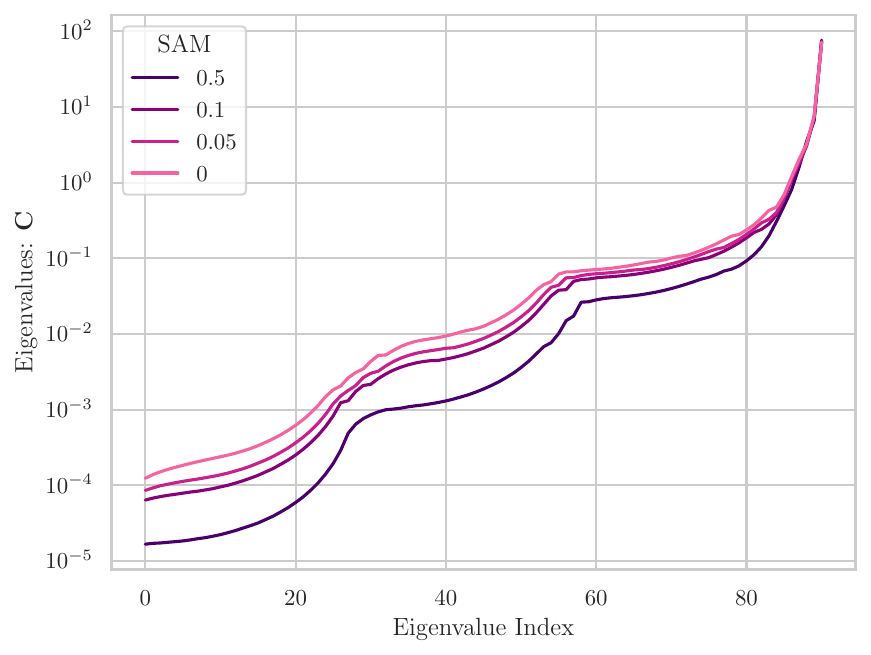}
		\vspace{-2mm}}
	\subfigure[Eigenvalues: $\Cm_0$]{\label{fig:}
		\includegraphics[width=0.3\textwidth]{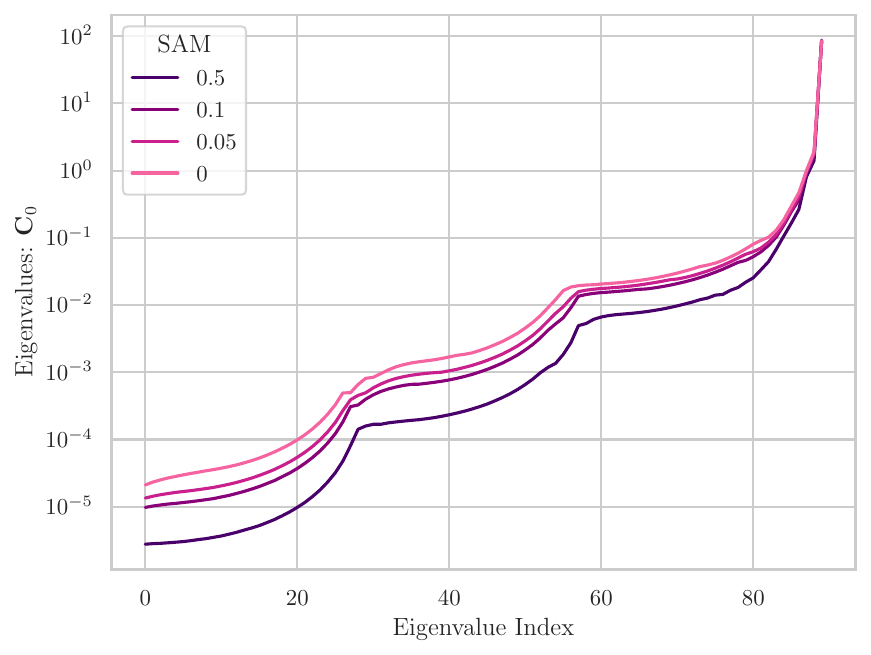}
		\vspace{-2mm}}
	\caption{Spectral measures of the Trajectory for ResNet50 trained on ImageNet} 
\end{figure*}

\clearpage

\subsection{ResNet50: Momentum Analysis, LR 0.1, WD 0.0001}

\begin{figure*}[h!]
	\centering
	\includegraphics[width=0.9\textwidth]{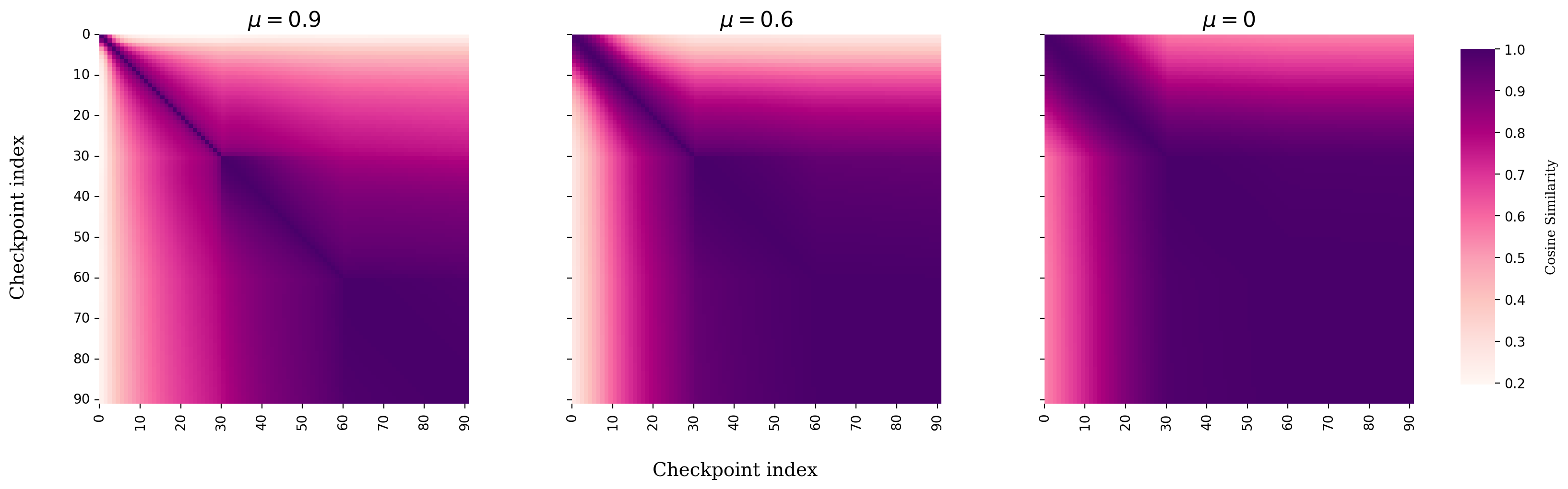}
	\caption{Trajectory Maps of ResNet50 models across different amounts of momentum} 
\end{figure*}

\begin{figure*}[h!]
	\centering
	\includegraphics[width=0.9\textwidth]{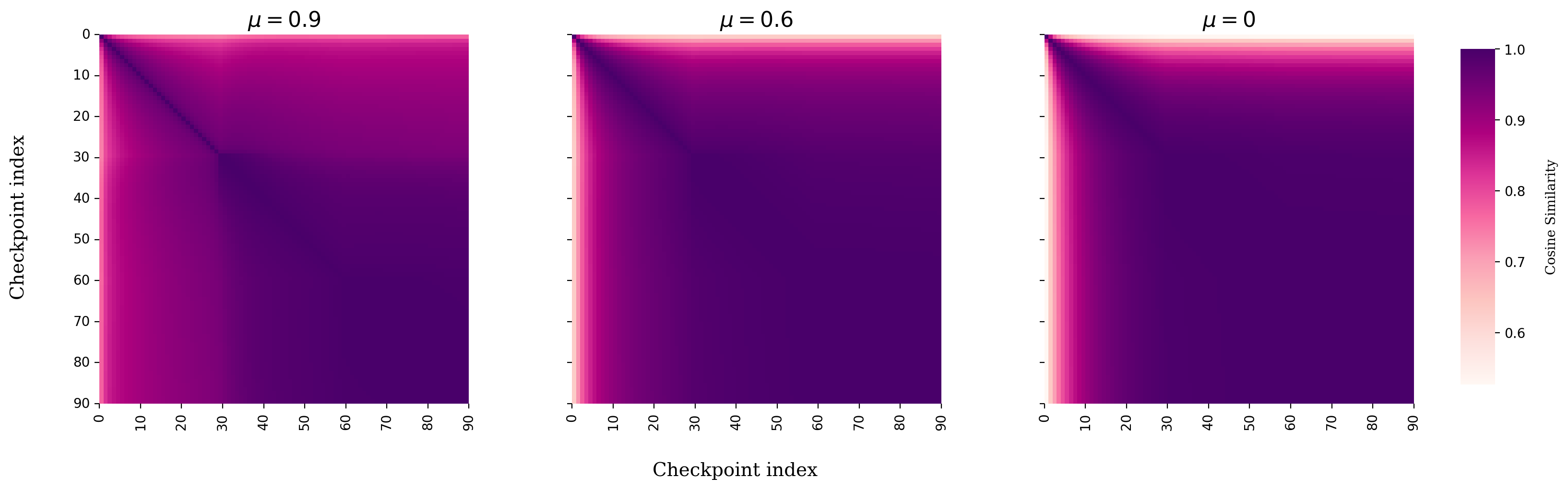}
	\caption{Relative Trajectory Maps, with respect to initialization, of ResNet50 models across different amounts of momentum} 
\end{figure*}

\clearpage
\begin{figure*}[h!]
	\centering
	\subfigure[$\angle(\theta_{t+1}-\theta_t,\theta_t)$]{\label{fig:}
		\includegraphics[width=0.34\textwidth]{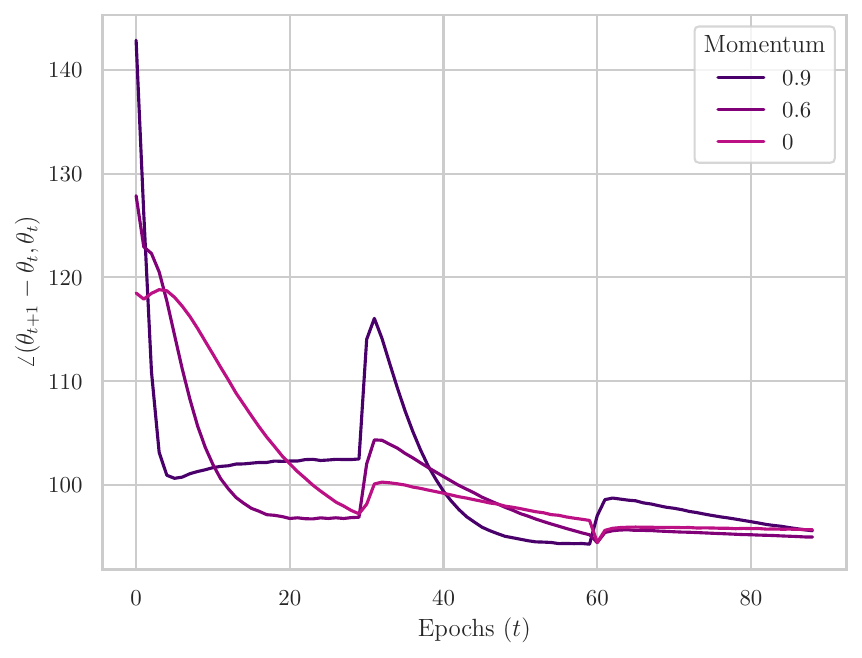}}
	\subfigure[$\angle(\theta_{t+1}-\theta_t,\theta_T-\theta_0)$]{\label{fig:}
		\includegraphics[width=0.34\textwidth]{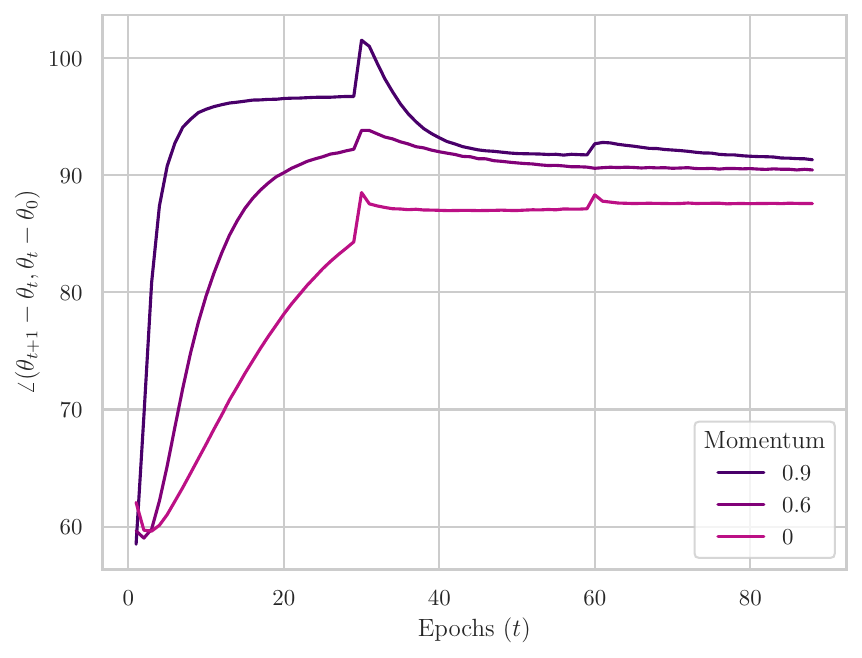}}
	\subfigure[$\angle(\theta_{t+k}-\theta_t,\theta_t-\theta_{t-k})$, for $k=1$ ]{\label{fig:}
		\includegraphics[width=0.34\textwidth]{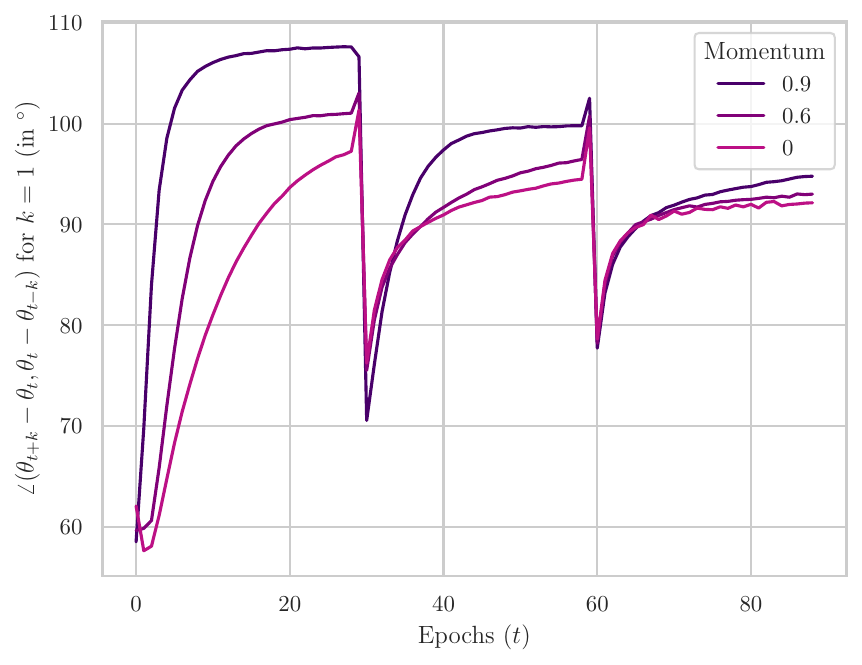}
		\vspace{-2mm}}
	\subfigure[$\angle(\theta_{t}-\theta_0,\theta_T-\theta_0)$]{\label{fig:}
		\includegraphics[width=0.34\textwidth]{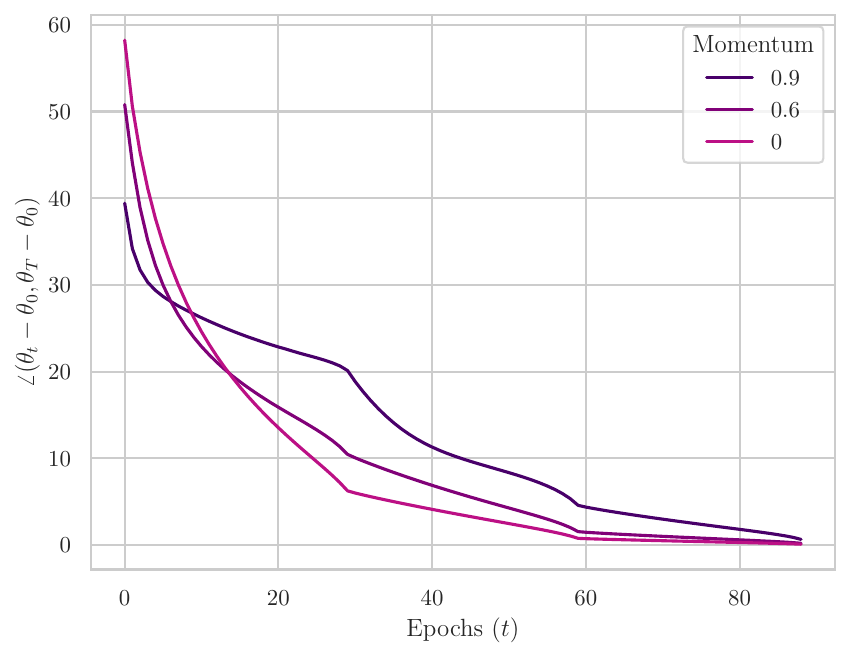}
		\vspace{-2mm}}
	\subfigure[$\angle(\theta_{t+1}-\theta_t, \theta_T-\theta_0)$]{\label{fig:}
		\includegraphics[width=0.34\textwidth]{figures/icml/Momentum/ckpt_freq-1_heatmap_from_multi-3_resnet50_imagenet_ep-90_lr-0.1_bsz-256_mom-0.9_wdecay-0.0001_seed-0_2023-02-21_10-45-30_983469_2024-02-01_02-26-16_131051/figures/pdf/angle_theta__t+1_-theta_t,theta_T-theta_0__vs_Epochs__t__across_Momentum.pdf}}
	\subfigure[Apex Angle at Initialization $\angle(\theta_t-\theta_0,\theta_1-\theta_0)$ ]{\label{fig:}
		\includegraphics[width=0.34\textwidth]{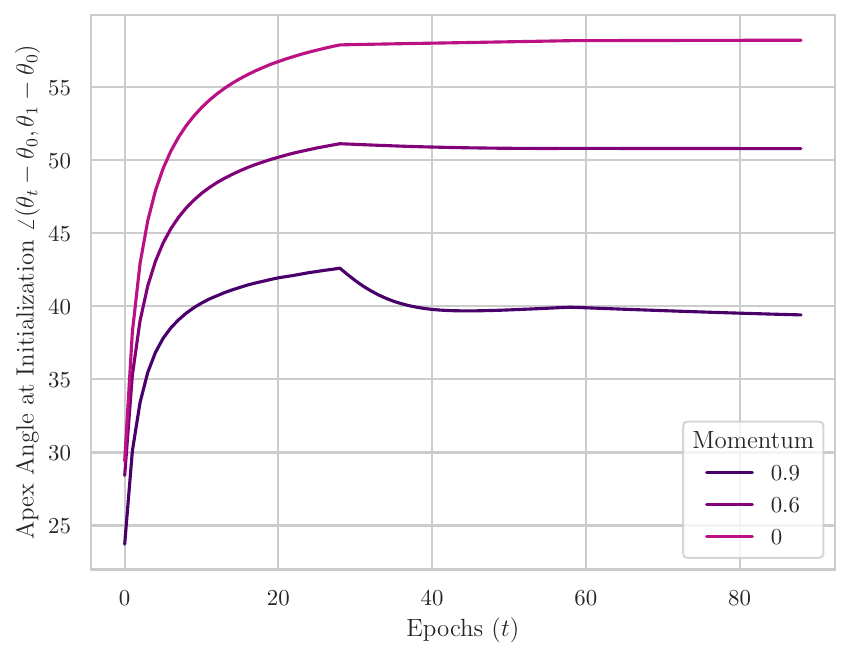}
		\vspace{-2mm}}
	\subfigure[Apex Angle at Origin $\angle(\theta_t,\theta_0)$]{\label{fig:}
		\includegraphics[width=0.34\textwidth]{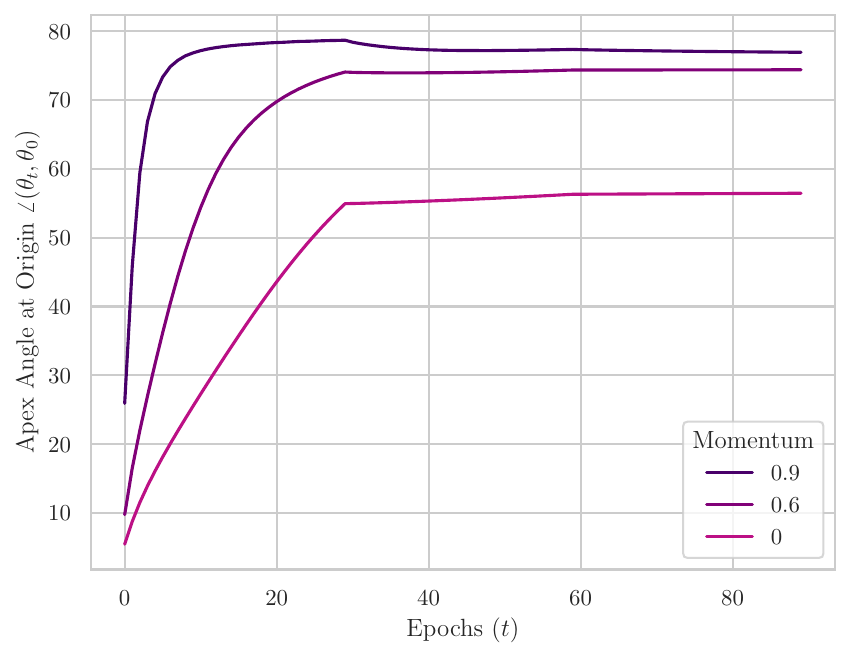}
		\vspace{-2mm}}
	\caption{Angular measures of the Trajectory for ResNet50 trained on ImageNet} 
\end{figure*}

\begin{figure*}[h!]
	\centering
	\subfigure[$\|\theta_t\|_2$]{\label{fig:}
	\includegraphics[width=0.3\textwidth]{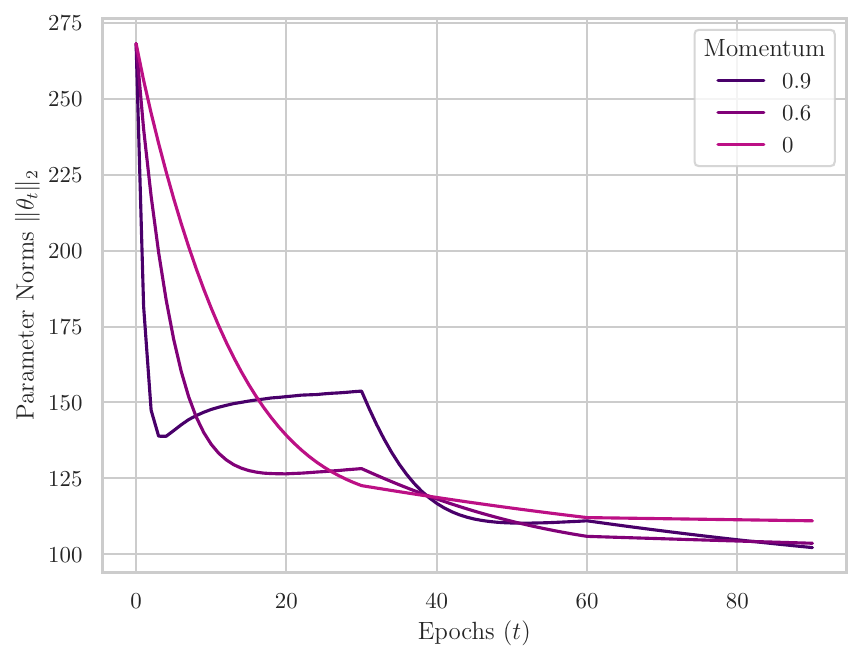}} \subfigure[$\|\theta_{t+k}-\theta_t\|_2$ ]{\label{fig:}
		\includegraphics[width=0.3\textwidth]{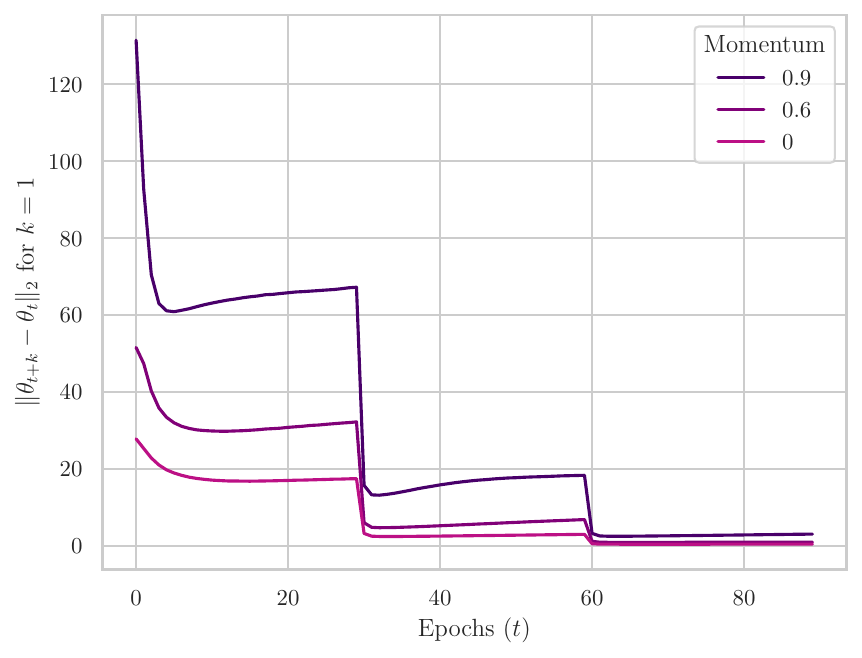}}
	\subfigure[$\|\theta_t-\theta_0\|_2$ ]{\label{fig:}
		\includegraphics[width=0.3\textwidth]{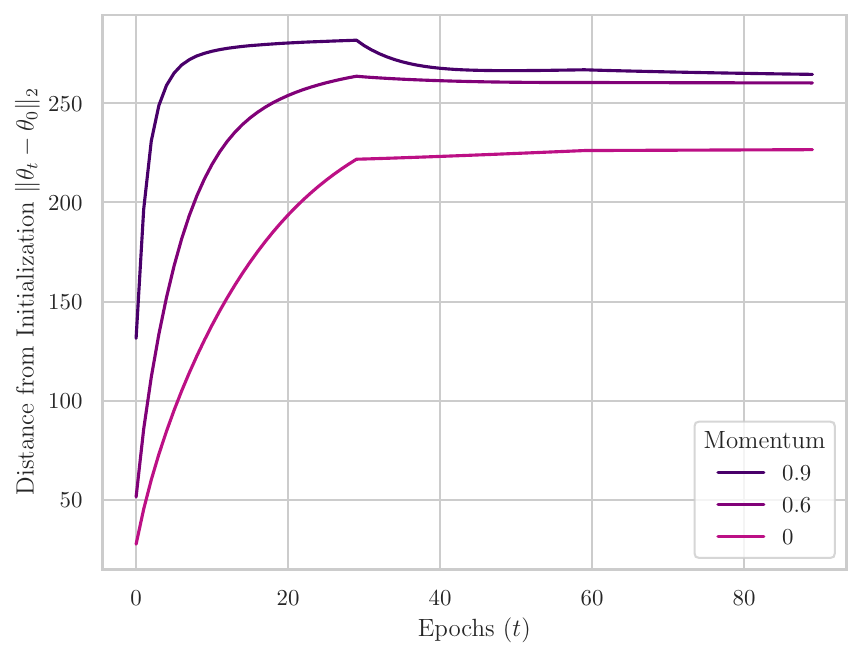}
		\vspace{-2mm}}
	\caption{Norm-based measures of the Trajectory for ResNet50 trained on ImageNet} 
\end{figure*}

\begin{figure*}[h!]
	\centering
	\subfigure[Eigenvalues: $\Km$]{\label{fig:}
		\includegraphics[width=0.3\textwidth]{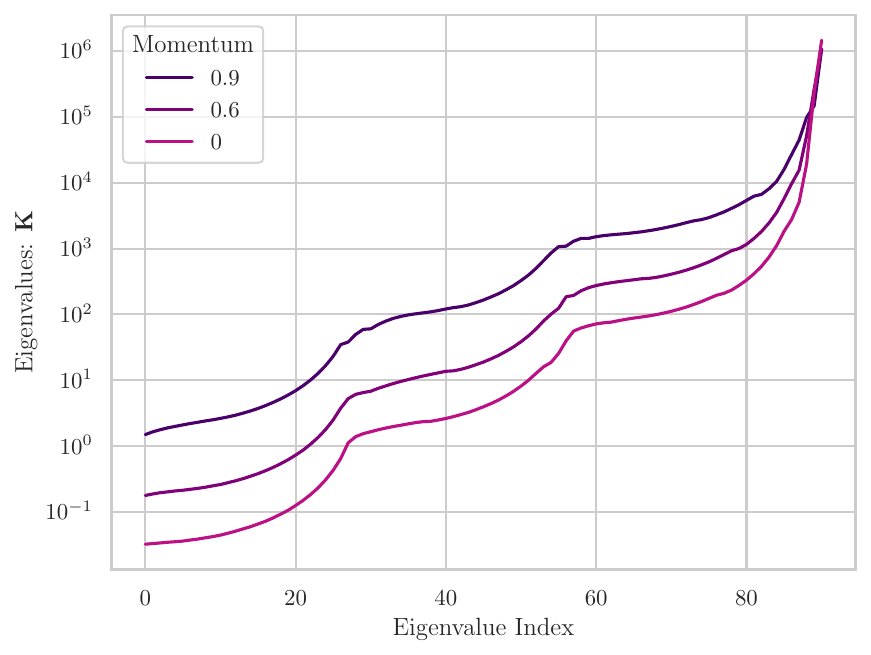}}
	\subfigure[Eigenvalues: $\Km_0$]{\label{fig:}
		\includegraphics[width=0.3\textwidth]{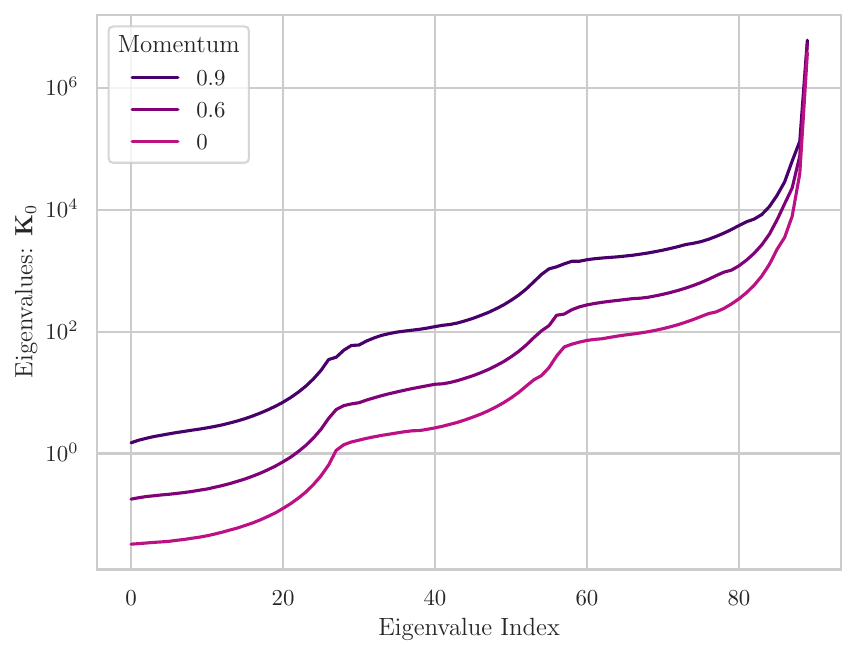}}

	\subfigure[Eigenvalues: $\Cm$ ]{\label{fig:}
		\includegraphics[width=0.3\textwidth]{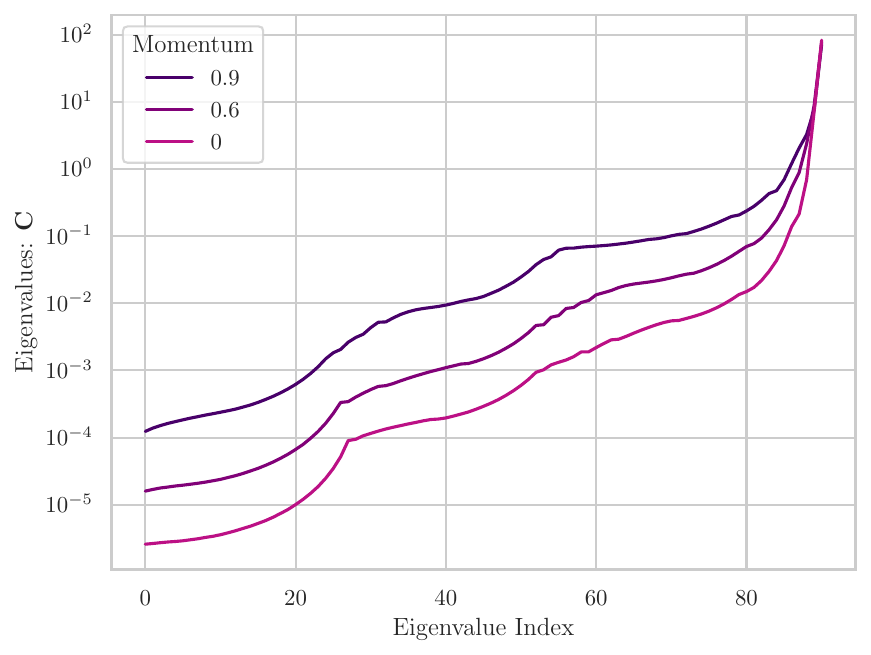}
		\vspace{-2mm}}
	\subfigure[Eigenvalues: $\Cm_0$]{\label{fig:}
		\includegraphics[width=0.3\textwidth]{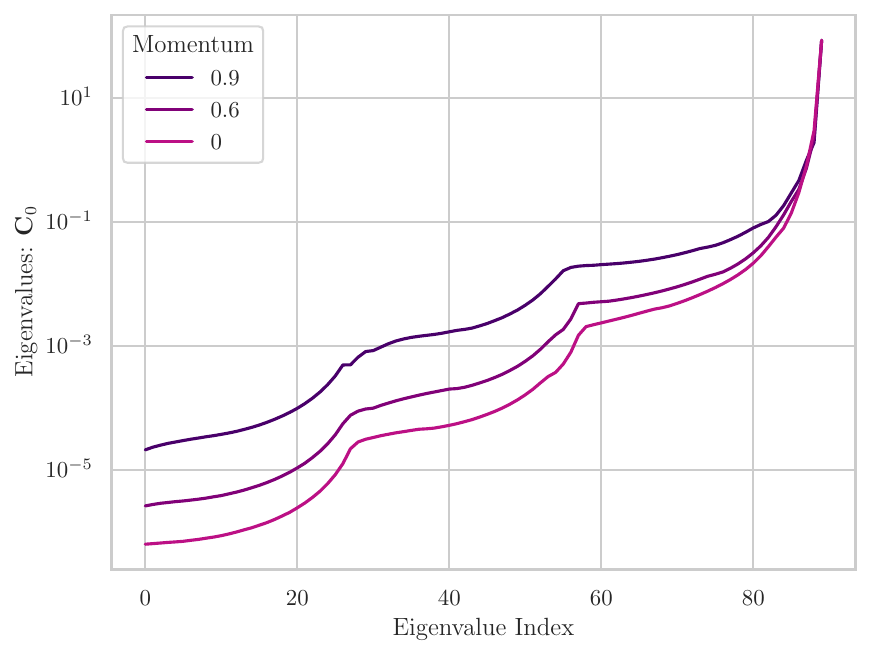}
		\vspace{-2mm}}
	\caption{Spectral measures of the Trajectory for ResNet50 trained on ImageNet} 
\end{figure*}

\clearpage

\subsection{VGG: Momentum Analysis, LR 0.1, WD 0.0001}

\begin{figure*}[h!]
	\centering
	\includegraphics[width=0.9\textwidth]{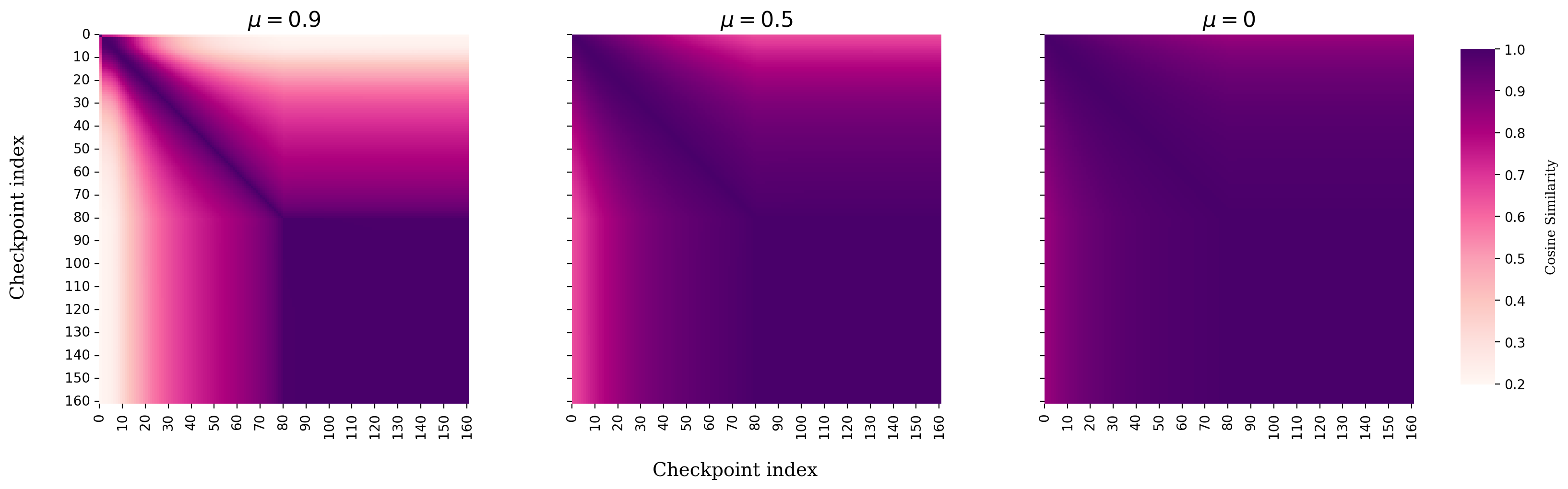}
	\caption{Trajectory Maps of VGG16 models across different amounts of momentum} 
\end{figure*}

\begin{figure*}[h!]
	\centering
	\includegraphics[width=0.9\textwidth]{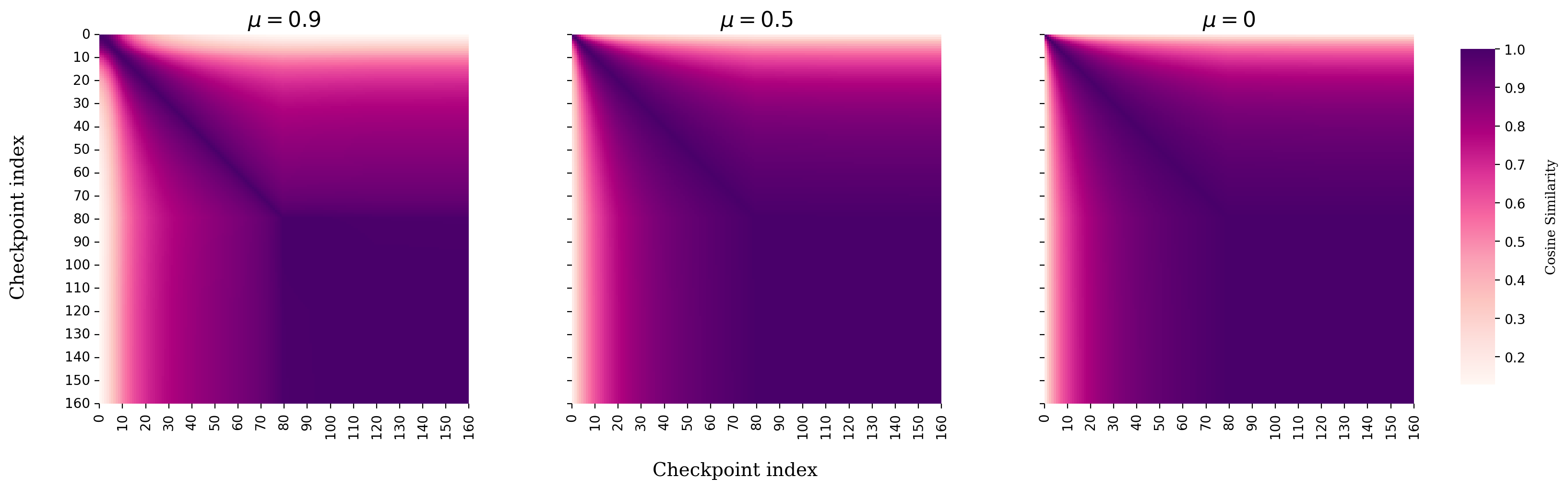}
	\caption{Relative Trajectory Maps, with respect to initialization, of VGG16 models across different amounts of momentum} 
\end{figure*}

\clearpage
\begin{figure*}[h!]
	\centering
	\subfigure[$\angle(\theta_{t+1}-\theta_t,\theta_t)$]{\label{fig:}
		\includegraphics[width=0.34\textwidth]{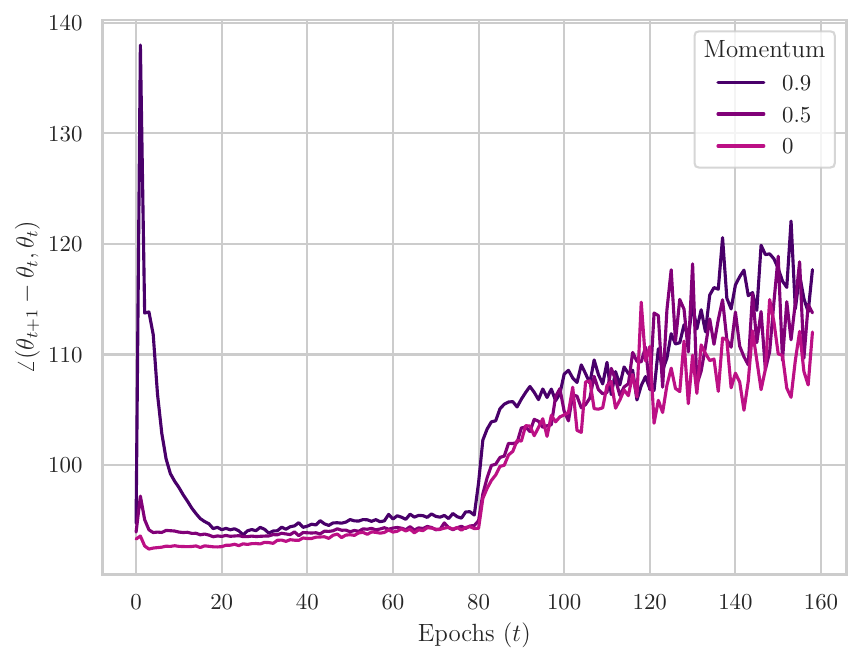}}
	\subfigure[$\angle(\theta_{t+1}-\theta_t,\theta_T-\theta_0)$]{\label{fig:}
		\includegraphics[width=0.34\textwidth]{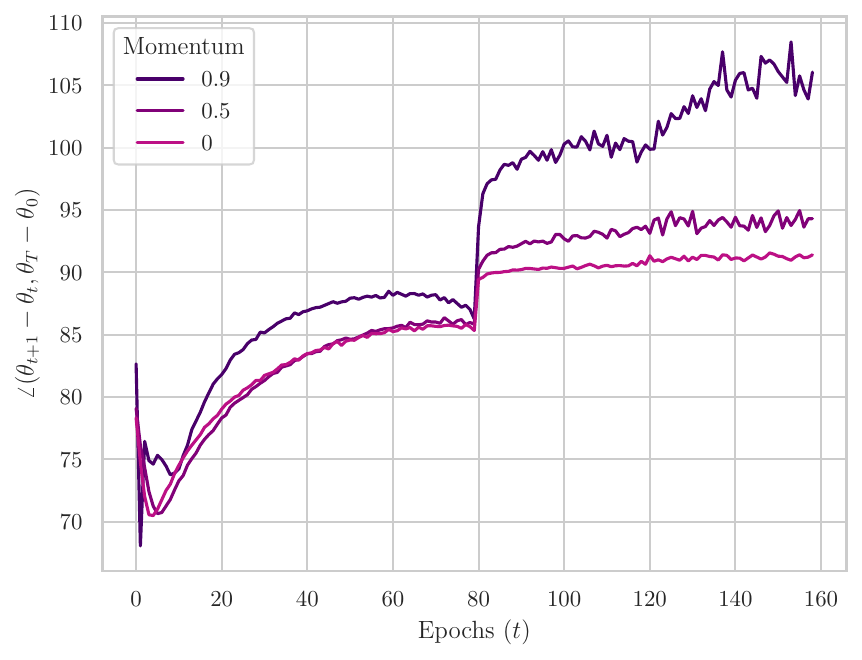}}
	\subfigure[$\angle(\theta_{t+k}-\theta_t,\theta_t-\theta_{t-k})$, for $k=1$ ]{\label{fig:}
		\includegraphics[width=0.34\textwidth]{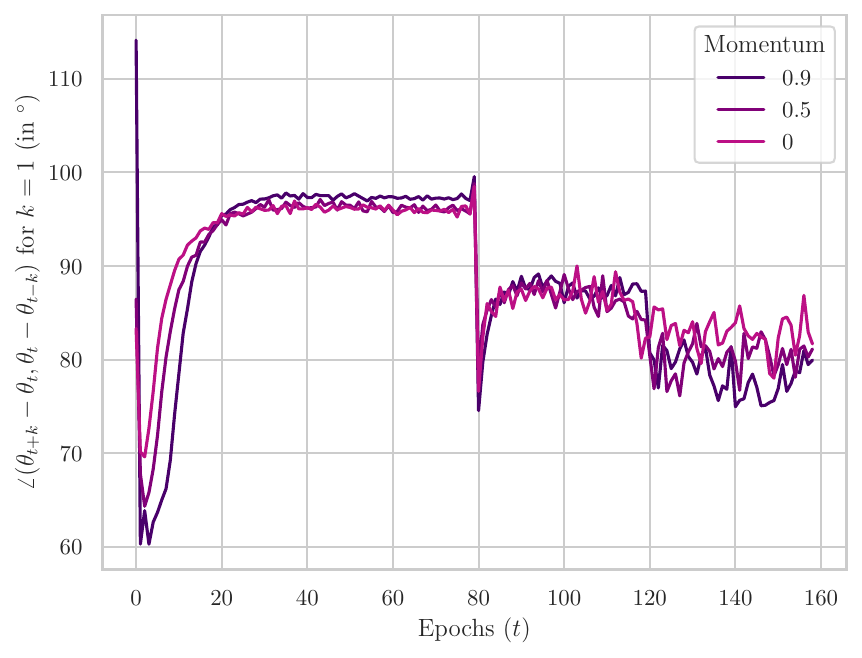}
		\vspace{-2mm}}
	\subfigure[$\angle(\theta_{t}-\theta_0,\theta_T-\theta_0)$]{\label{fig:}
		\includegraphics[width=0.34\textwidth]{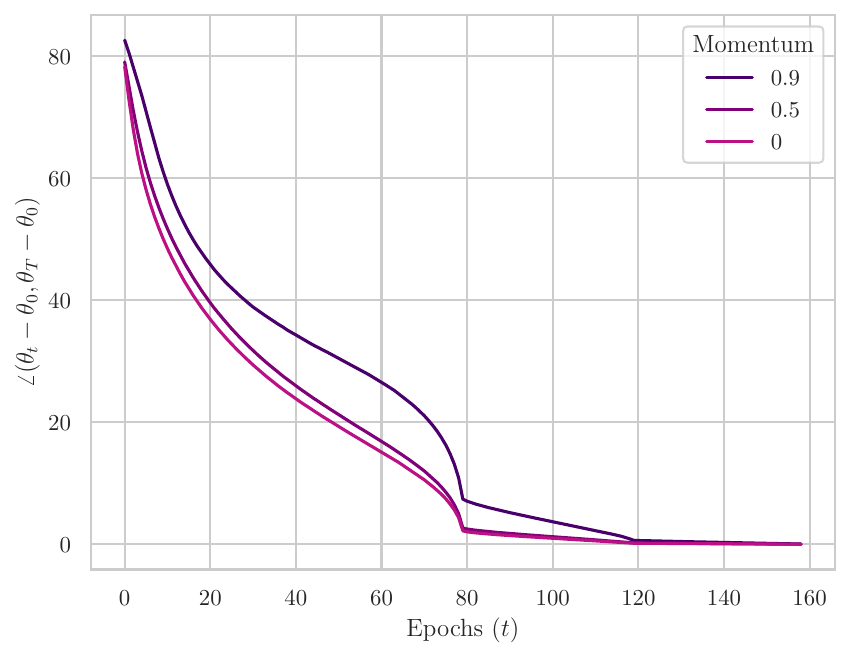}
		\vspace{-2mm}}
	\subfigure[$\angle(\theta_{t+1}-\theta_t, \theta_T-\theta_0)$]{\label{fig:}
		\includegraphics[width=0.34\textwidth]{figures/icml/Momentum/ckpt_freq-1_heatmap_from_multi-3_vgg16_bn_parents-_0__relu_cifar10_ln-0_ep-160_lr-0.1_mom-0.9_sched-0.1_0.25_wd-0.0001_2024-01-21_20-48-27_065553_2024-02-01_02-26-36_026670/figures/pdf/angle_theta__t+1_-theta_t,theta_T-theta_0__vs_Epochs__t__across_Momentum.pdf}}
	\subfigure[Apex Angle at Initialization $\angle(\theta_t-\theta_0,\theta_1-\theta_0)$ ]{\label{fig:}
		\includegraphics[width=0.34\textwidth]{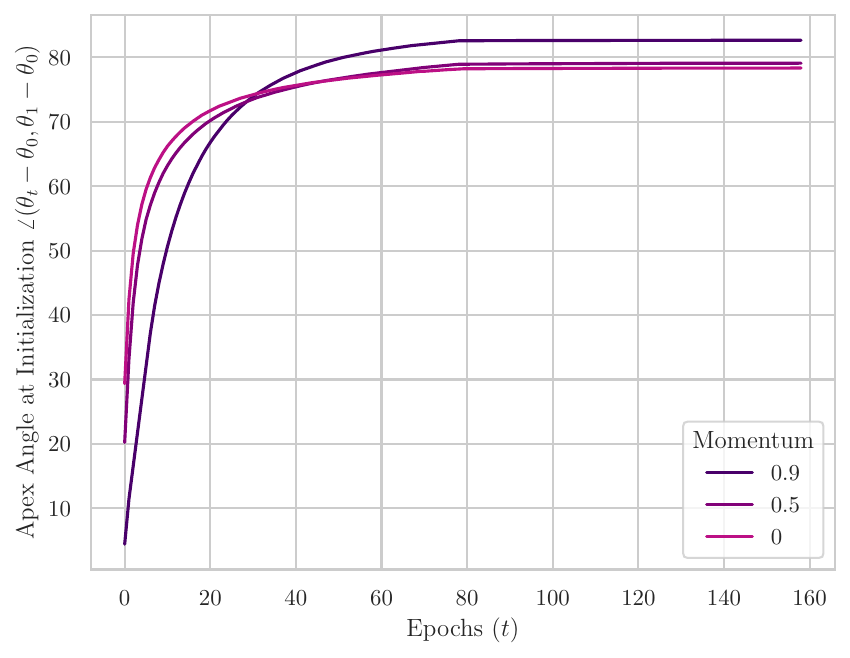}
		\vspace{-2mm}}
	\subfigure[Apex Angle at Origin $\angle(\theta_t,\theta_0)$]{\label{fig:}
		\includegraphics[width=0.34\textwidth]{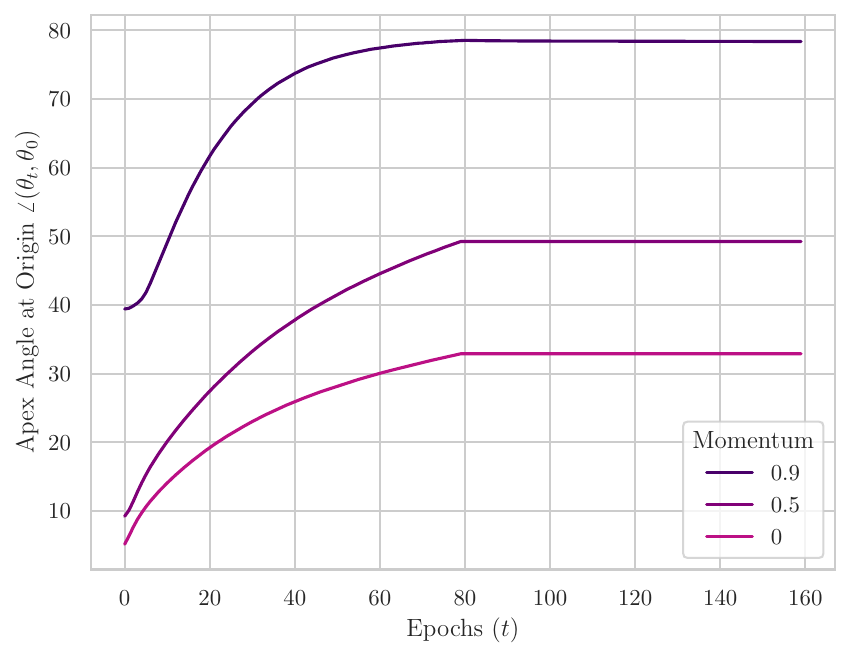}
		\vspace{-2mm}}
	\caption{Angular measures of the Trajectory for VGG16 models trained on CIFAR10.} 
\end{figure*}

\begin{figure*}[h!]
	\centering
	\subfigure[$\|\theta_t\|_2$]{\label{fig:}
	\includegraphics[width=0.3\textwidth]{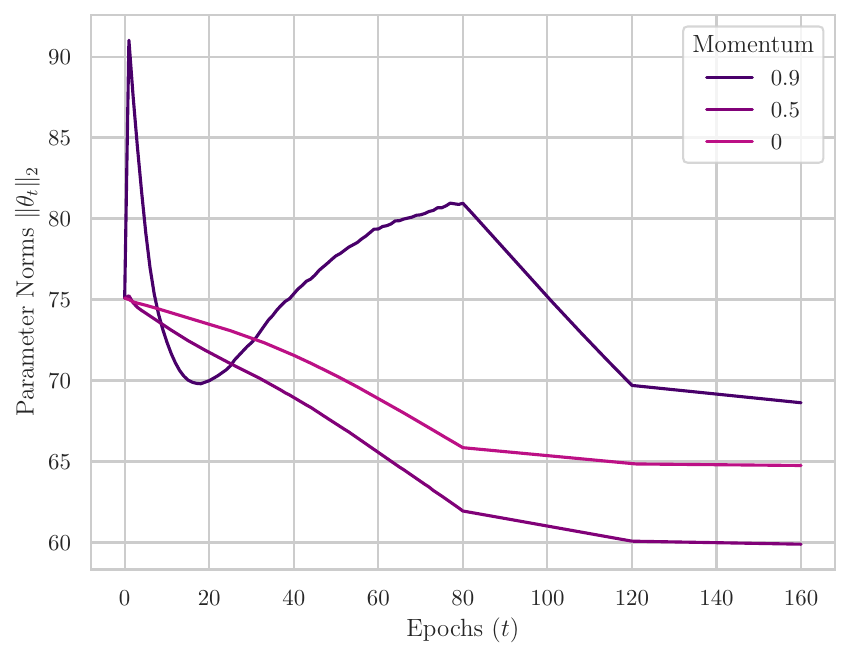}} \subfigure[$\|\theta_{t+k}-\theta_t\|_2$ ]{\label{fig:}
		\includegraphics[width=0.3\textwidth]{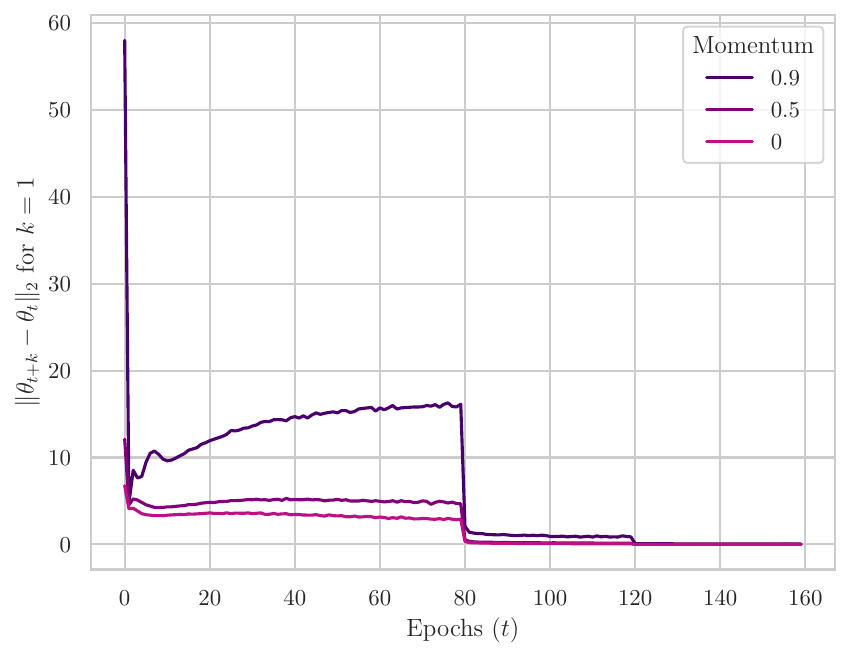}}
	\subfigure[$\|\theta_t-\theta_0\|_2$ ]{\label{fig:}
		\includegraphics[width=0.3\textwidth]{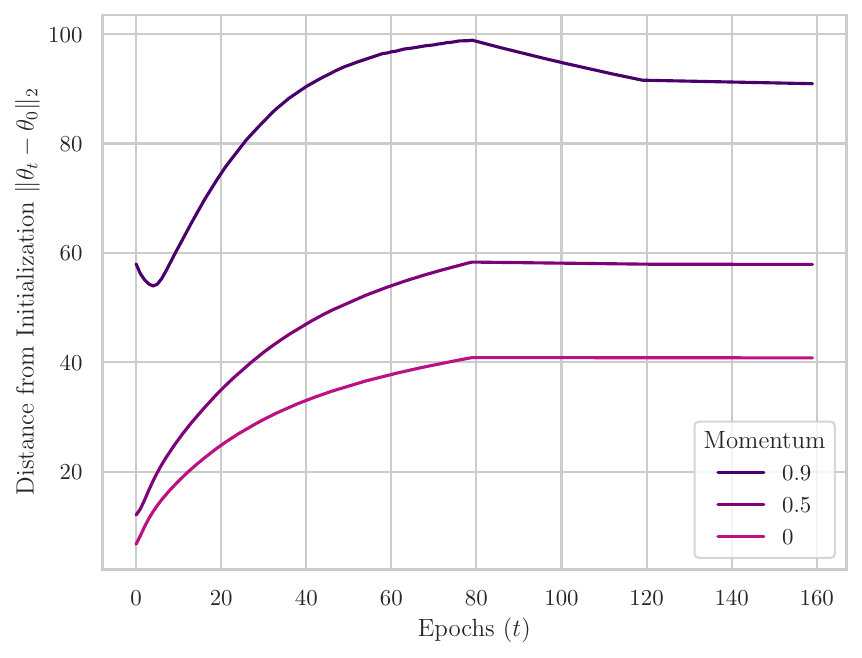}
		\vspace{-2mm}}
	\caption{Norm-based measures of the Trajectory for VGG16 models trained on CIFAR10.} 
\end{figure*}

\begin{figure*}[h!]
	\centering
	\subfigure[Eigenvalues: $\Km$]{\label{fig:}
		\includegraphics[width=0.3\textwidth]{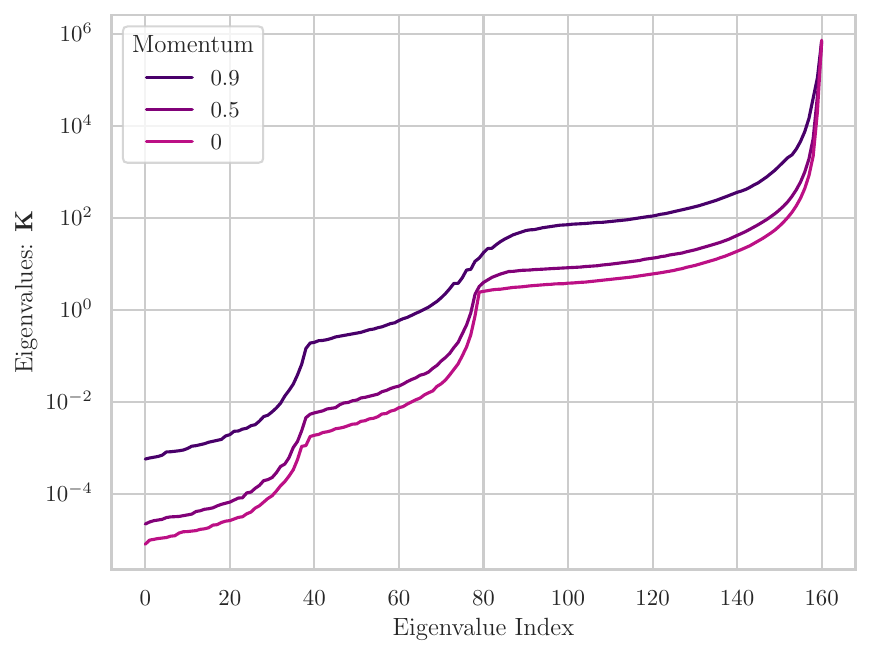}}
	\subfigure[Eigenvalues: $\Km_0$]{\label{fig:}
		\includegraphics[width=0.3\textwidth]{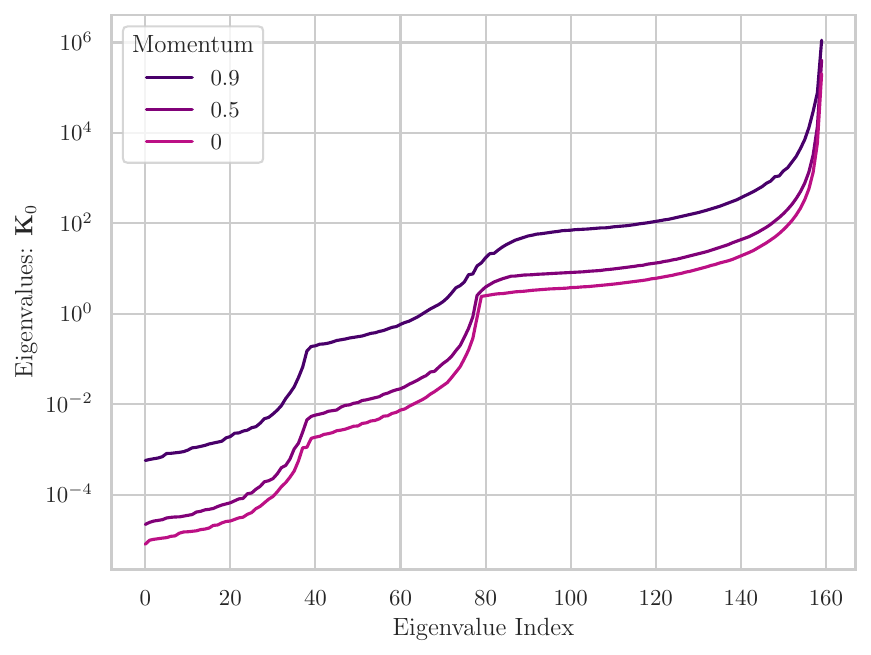}}

	\subfigure[Eigenvalues: $\Cm$ ]{\label{fig:}
		\includegraphics[width=0.3\textwidth]{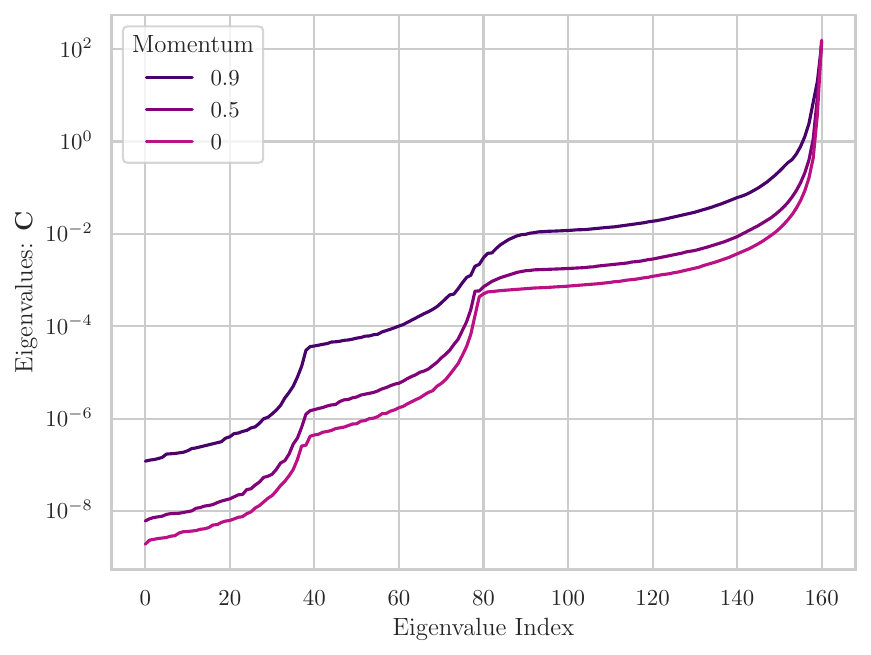}
		\vspace{-2mm}}
	\subfigure[Eigenvalues: $\Cm_0$]{\label{fig:}
		\includegraphics[width=0.3\textwidth]{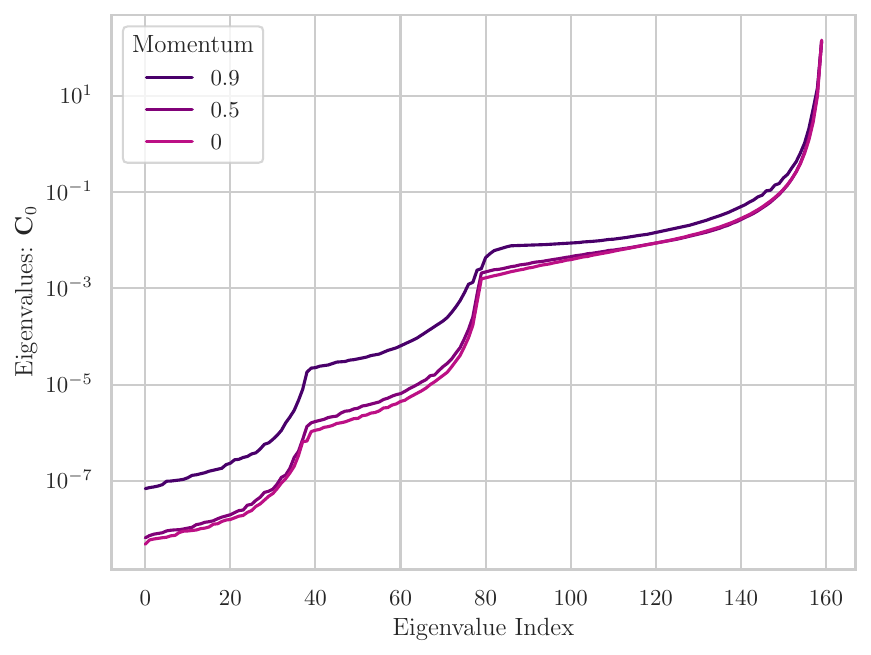}
		\vspace{-2mm}}
	\caption{Spectral measures of the Trajectory for VGG16 models trained on CIFAR10.} 
\end{figure*}

\clearpage

\subsection{VGG: Momentum Analysis, LR 0.1, WD 0}

\begin{figure*}[h!]
	\centering
	\includegraphics[width=0.9\textwidth]{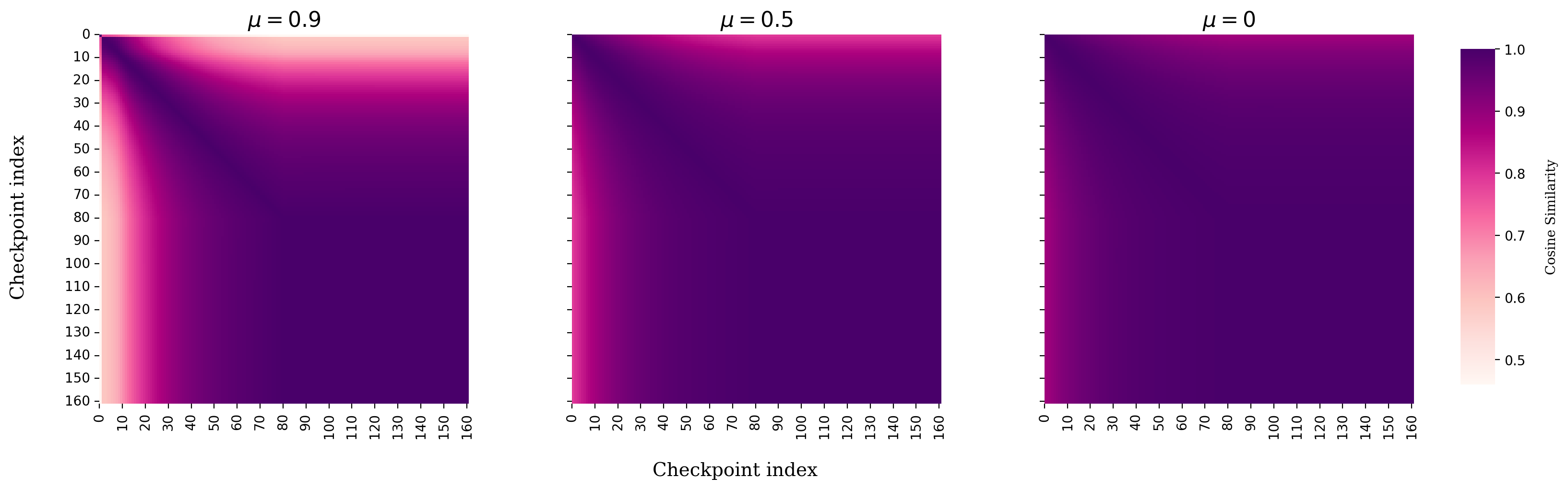}
	\caption{Trajectory Maps of VGG16 models across different amounts of momentum} 
\end{figure*}

\begin{figure*}[h!]
	\centering
	\includegraphics[width=0.9\textwidth]{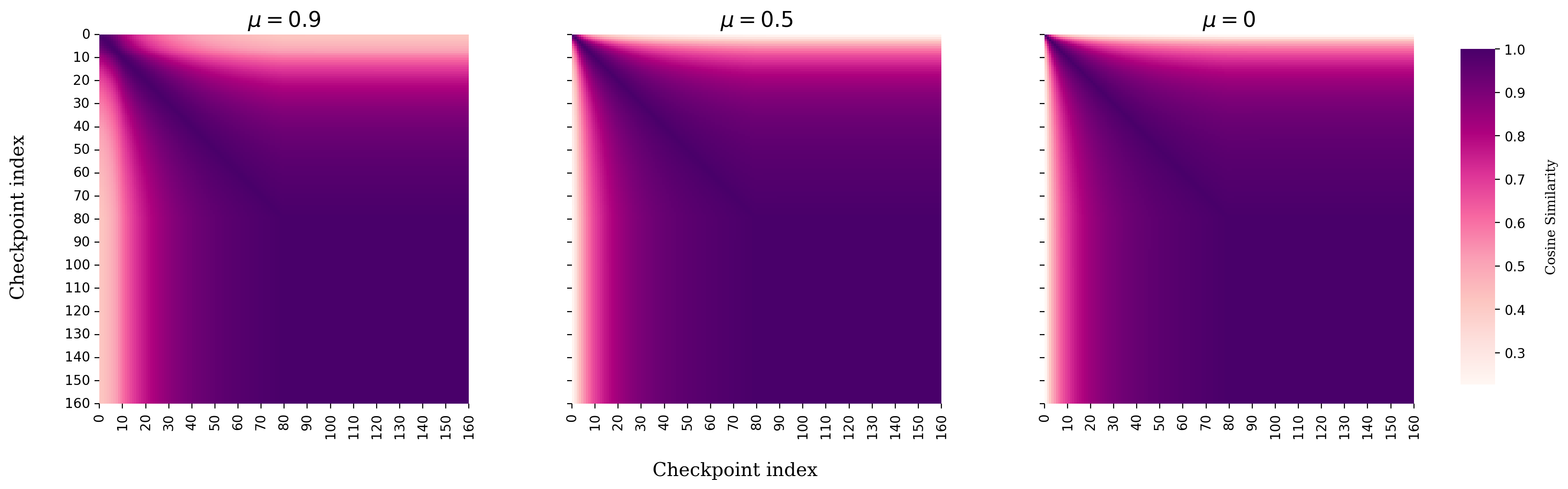}
	\caption{Relative Trajectory Maps, with respect to initialization, of VGG16 models across different amounts of momentum} 
\end{figure*}

\clearpage
\begin{figure*}[h!]
	\centering
	\subfigure[$\angle(\theta_{t+1}-\theta_t,\theta_t)$]{\label{fig:}
		\includegraphics[width=0.34\textwidth]{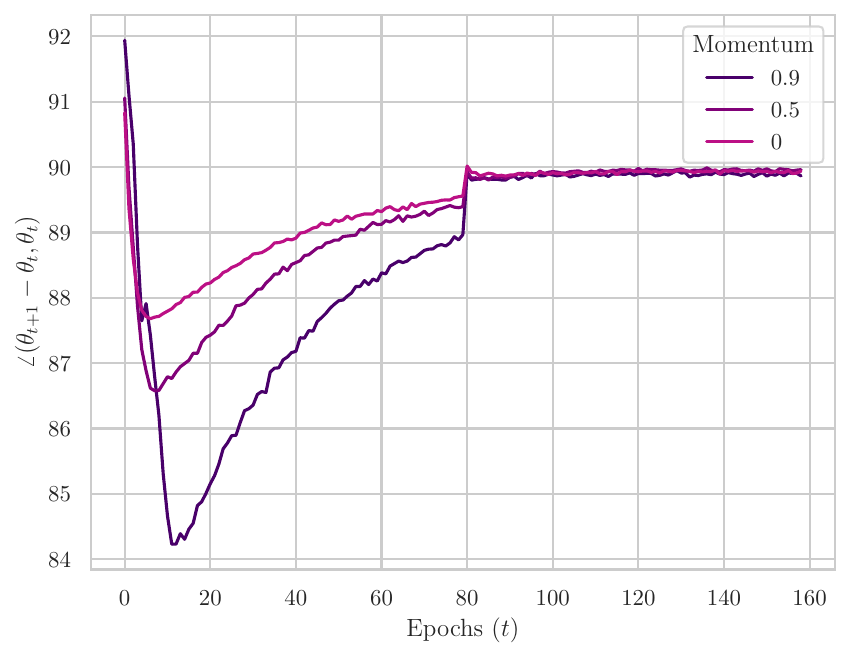}}
	\subfigure[$\angle(\theta_{t+1}-\theta_t,\theta_T-\theta_0)$]{\label{fig:}
		\includegraphics[width=0.34\textwidth]{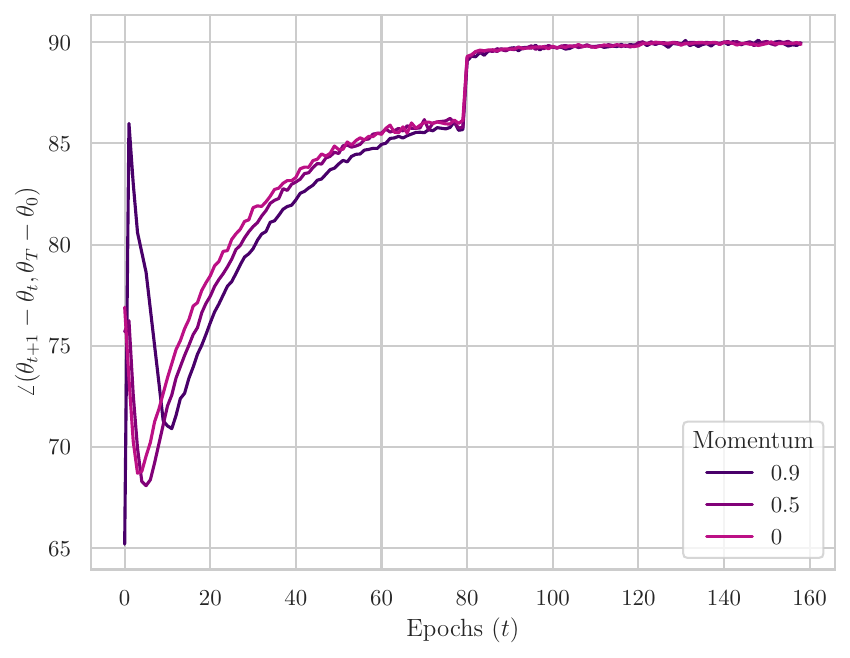}}
	\subfigure[$\angle(\theta_{t+k}-\theta_t,\theta_t-\theta_{t-k})$, for $k=1$ ]{\label{fig:}
		\includegraphics[width=0.34\textwidth]{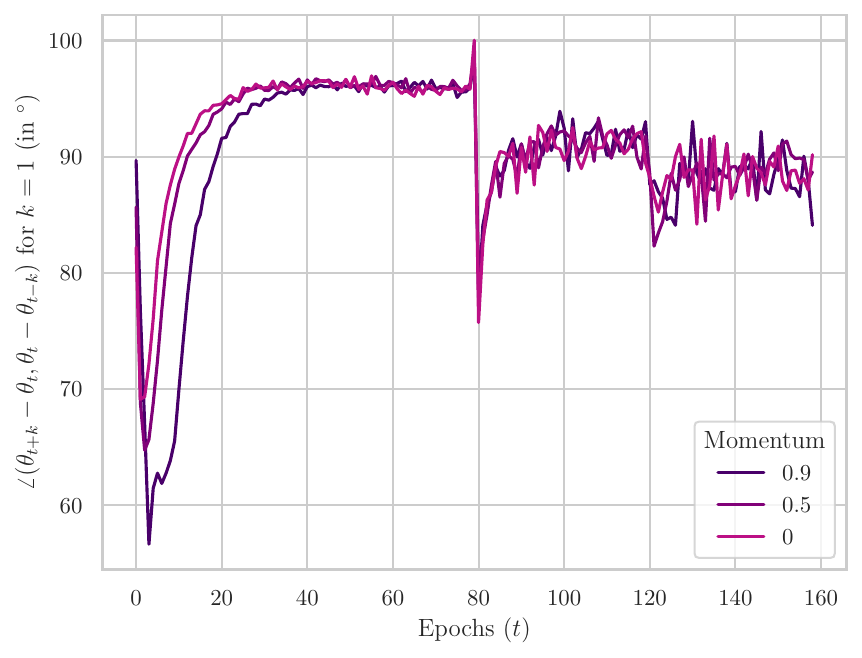}
		\vspace{-2mm}}
	\subfigure[$\angle(\theta_{t}-\theta_0,\theta_T-\theta_0)$]{\label{fig:}
		\includegraphics[width=0.34\textwidth]{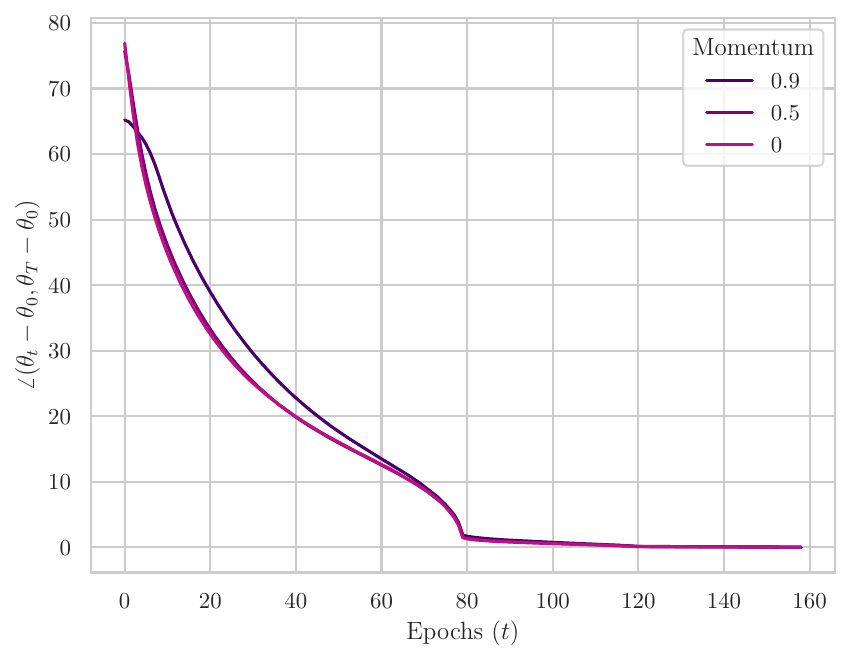}
		\vspace{-2mm}}
	\subfigure[$\angle(\theta_{t+1}-\theta_t, \theta_T-\theta_0)$]{\label{fig:}
		\includegraphics[width=0.34\textwidth]{figures/icml/Momentum/ckpt_freq-1_heatmap_from_multi-3_vgg16_bn_parents-_0__relu_cifar10_ln-0_ep-160_lr-0.1_mom-0.9_sched-0.1_0.25_wd-0.0_2024-01-21_20-48-27_065684_2024-02-01_02-26-35_557210/figures/pdf/angle_theta__t+1_-theta_t,theta_T-theta_0__vs_Epochs__t__across_Momentum.pdf}}
	\subfigure[Apex Angle at Initialization $\angle(\theta_t-\theta_0,\theta_1-\theta_0)$ ]{\label{fig:}
		\includegraphics[width=0.34\textwidth]{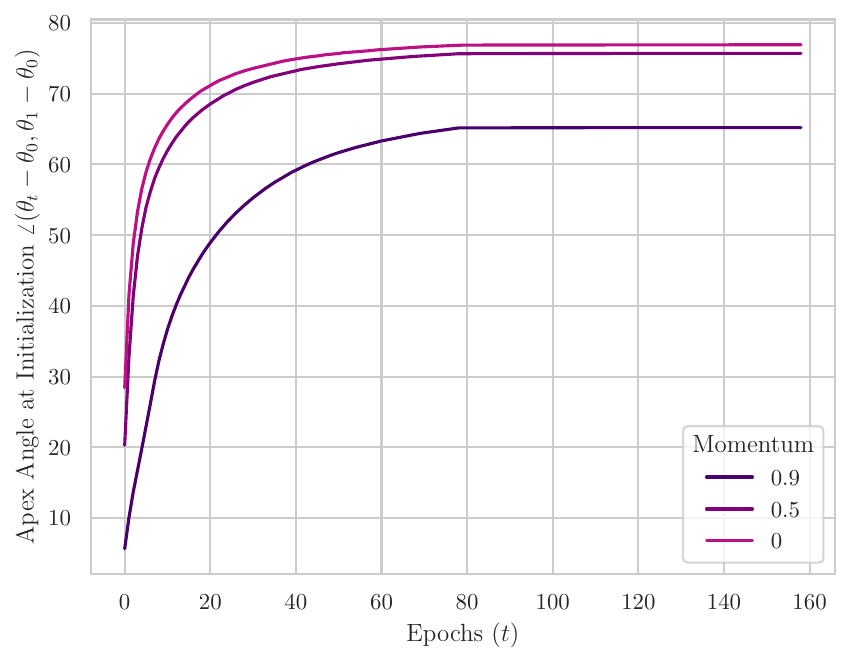}
		\vspace{-2mm}}
	\subfigure[Apex Angle at Origin $\angle(\theta_t,\theta_0)$]{\label{fig:}
		\includegraphics[width=0.34\textwidth]{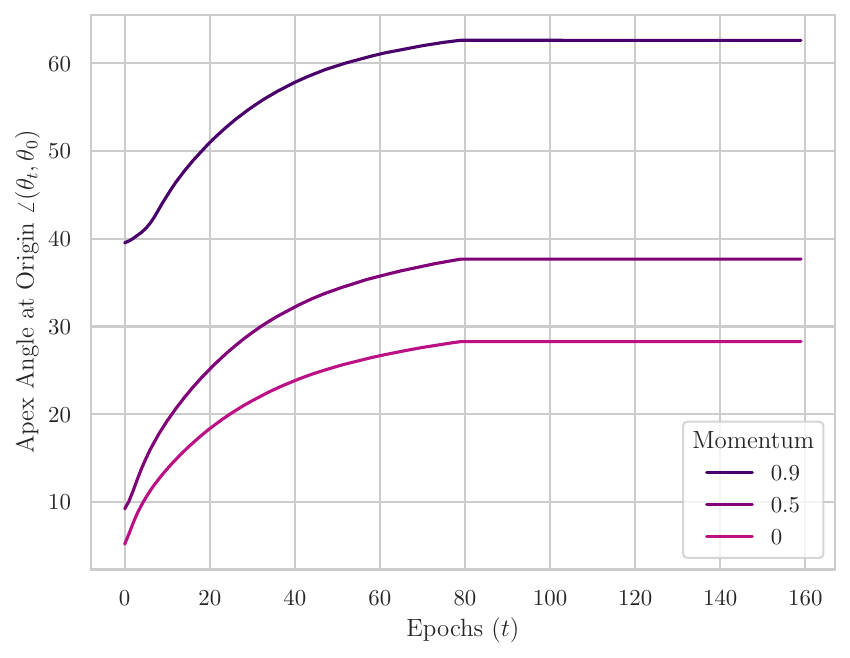}
		\vspace{-2mm}}
	\caption{Angular measures of the Trajectory for VGG16 models trained on CIFAR10.} 
\end{figure*}

\begin{figure*}[h!]
	\centering
	\subfigure[$\|\theta_t\|_2$]{\label{fig:}
	\includegraphics[width=0.3\textwidth]{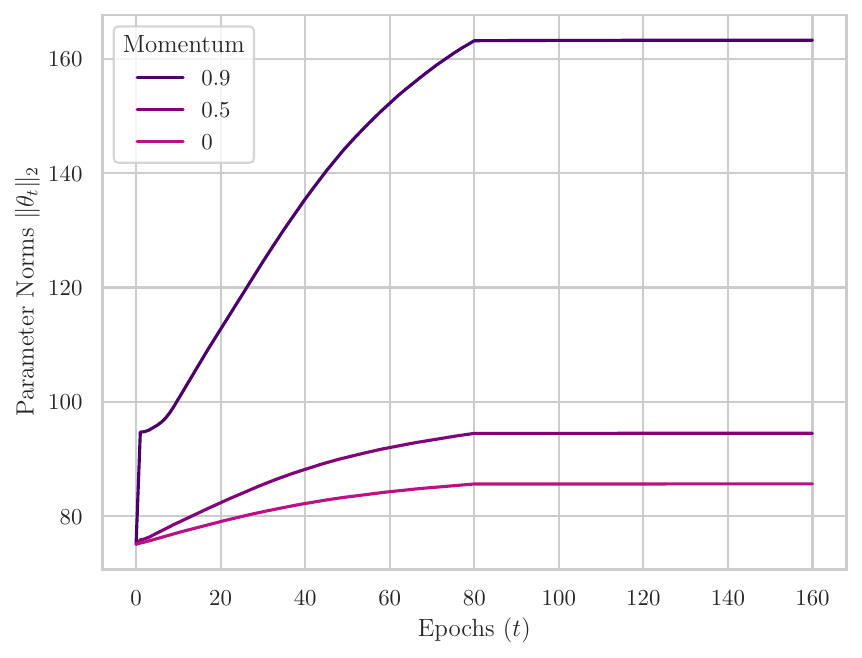}} \subfigure[$\|\theta_{t+k}-\theta_t\|_2$ ]{\label{fig:}
		\includegraphics[width=0.3\textwidth]{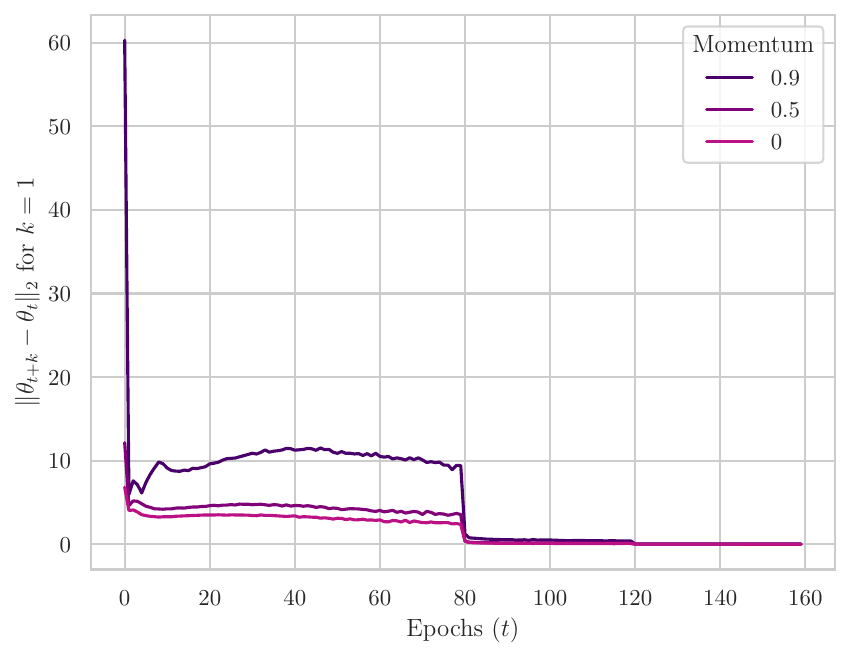}}
	\subfigure[$\|\theta_t-\theta_0\|_2$ ]{\label{fig:}
		\includegraphics[width=0.3\textwidth]{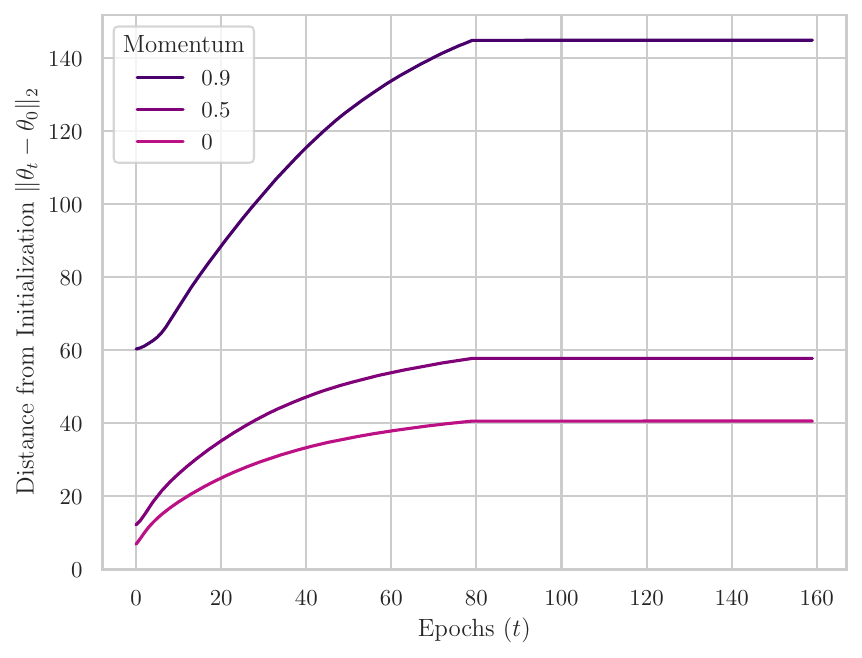}
		\vspace{-2mm}}
	\caption{Norm-based measures of the Trajectory for VGG16 models trained on CIFAR10.} 
\end{figure*}

\begin{figure*}[h!]
	\centering
	\subfigure[Eigenvalues: $\Km$]{\label{fig:}
		\includegraphics[width=0.3\textwidth]{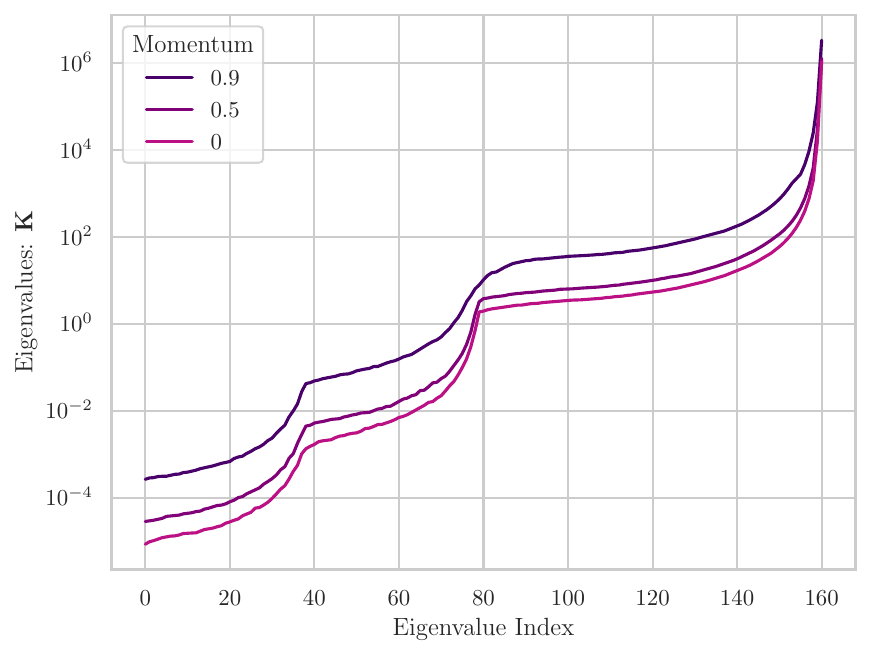}}
	\subfigure[Eigenvalues: $\Km_0$]{\label{fig:}
		\includegraphics[width=0.3\textwidth]{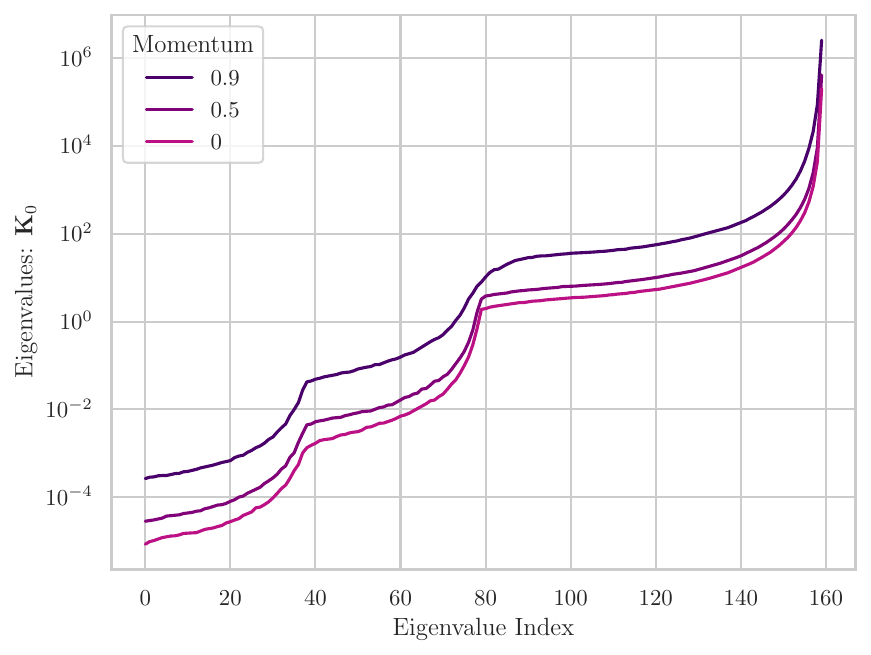}}

	\subfigure[Eigenvalues: $\Cm$ ]{\label{fig:}
		\includegraphics[width=0.3\textwidth]{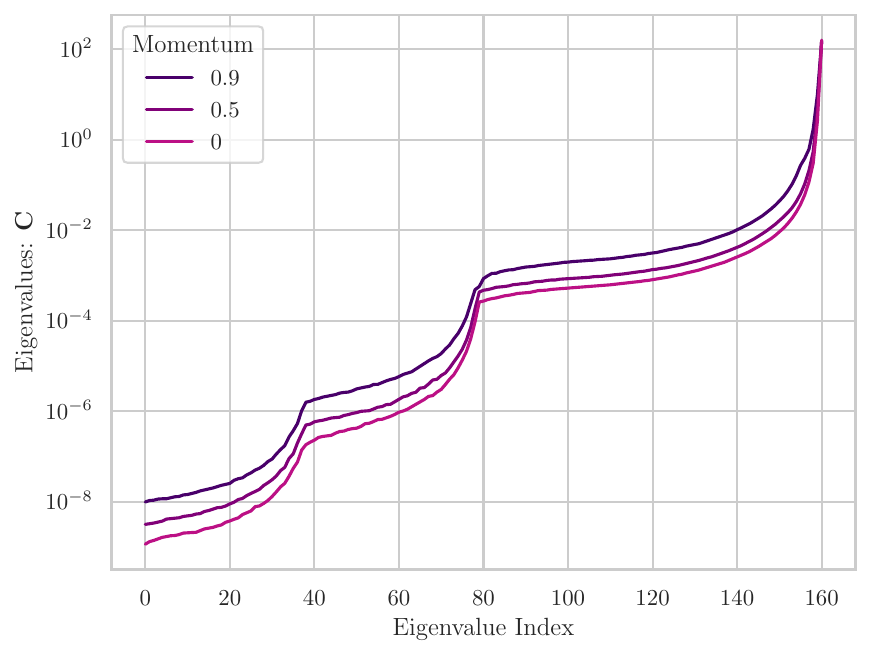}
		\vspace{-2mm}}
	\subfigure[Eigenvalues: $\Cm_0$]{\label{fig:}
		\includegraphics[width=0.3\textwidth]{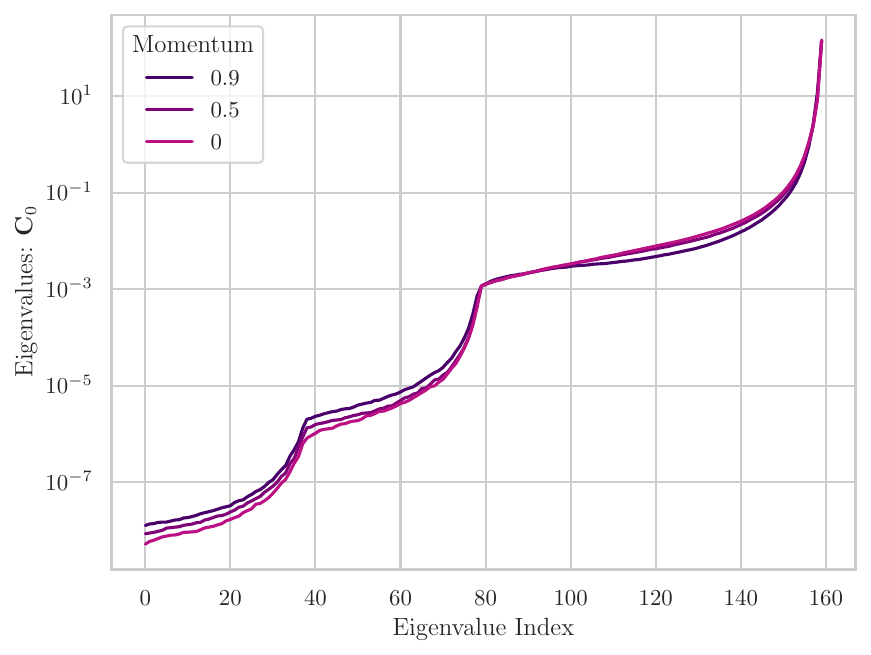}
		\vspace{-2mm}}
	\caption{Spectral measures of the Trajectory for VGG16 models trained on CIFAR10.} 
\end{figure*}

\clearpage

\subsection{VGG: Momentum Analysis, LR 0.01, WD 0.0001}

\begin{figure*}[h!]
	\centering
	\includegraphics[width=0.9\textwidth]{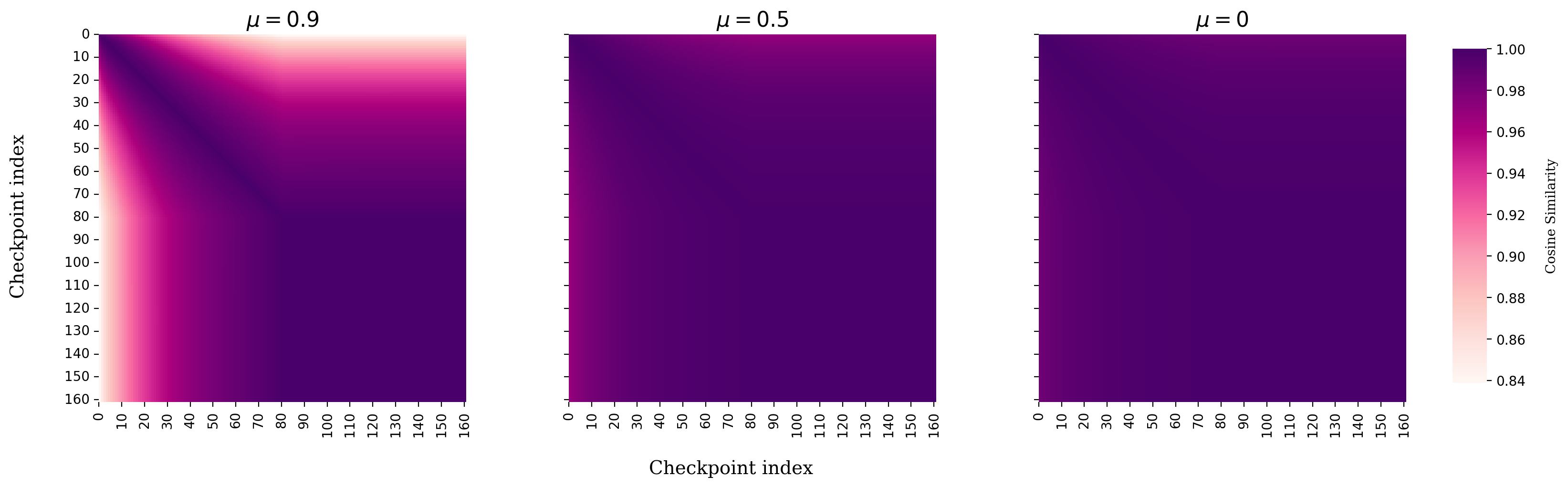}
	\caption{Trajectory Maps of VGG16 models across different amounts of momentum} 
\end{figure*}

\begin{figure*}[h!]
	\centering
	\includegraphics[width=0.9\textwidth]{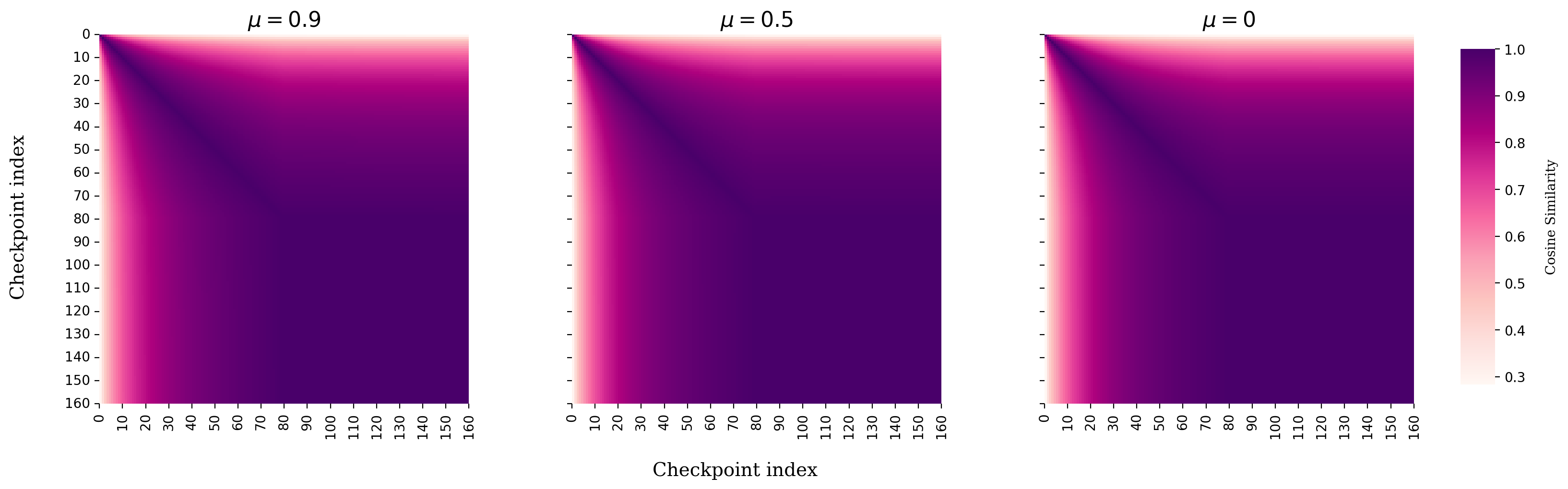}
	\caption{Relative Trajectory Maps, with respect to initialization, of VGG16 models across different amounts of momentum} 
\end{figure*}

\clearpage
\begin{figure*}[h!]
	\centering
	\subfigure[$\angle(\theta_{t+1}-\theta_t,\theta_t)$]{\label{fig:}
		\includegraphics[width=0.34\textwidth]{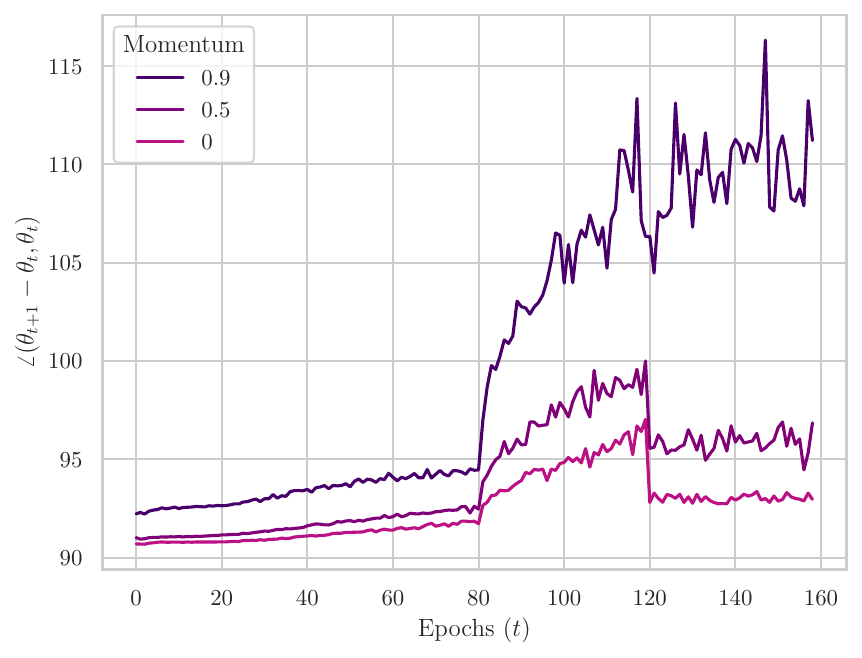}}
	\subfigure[$\angle(\theta_{t+1}-\theta_t,\theta_T-\theta_0)$]{\label{fig:}
		\includegraphics[width=0.34\textwidth]{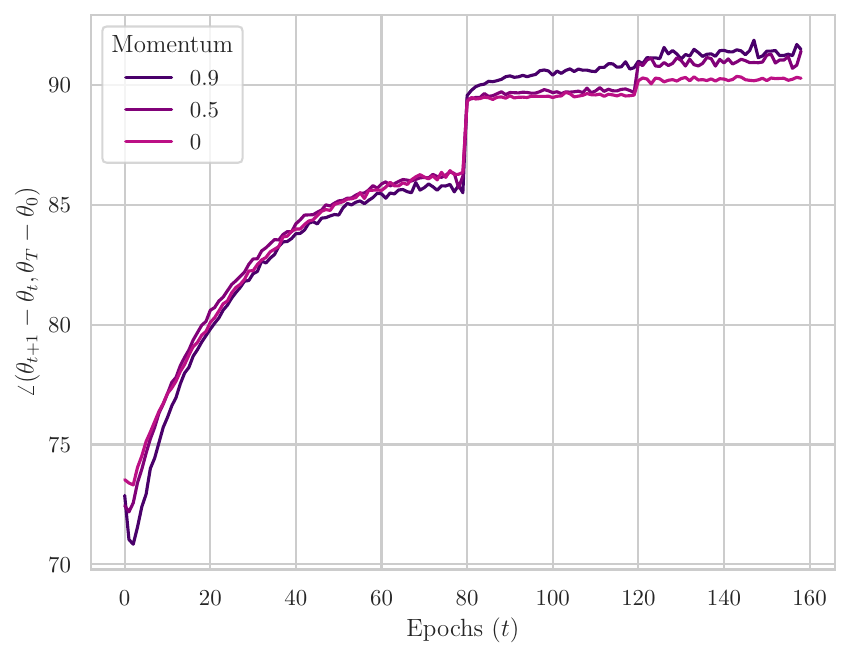}}
	\subfigure[$\angle(\theta_{t+k}-\theta_t,\theta_t-\theta_{t-k})$, for $k=1$ ]{\label{fig:}
		\includegraphics[width=0.34\textwidth]{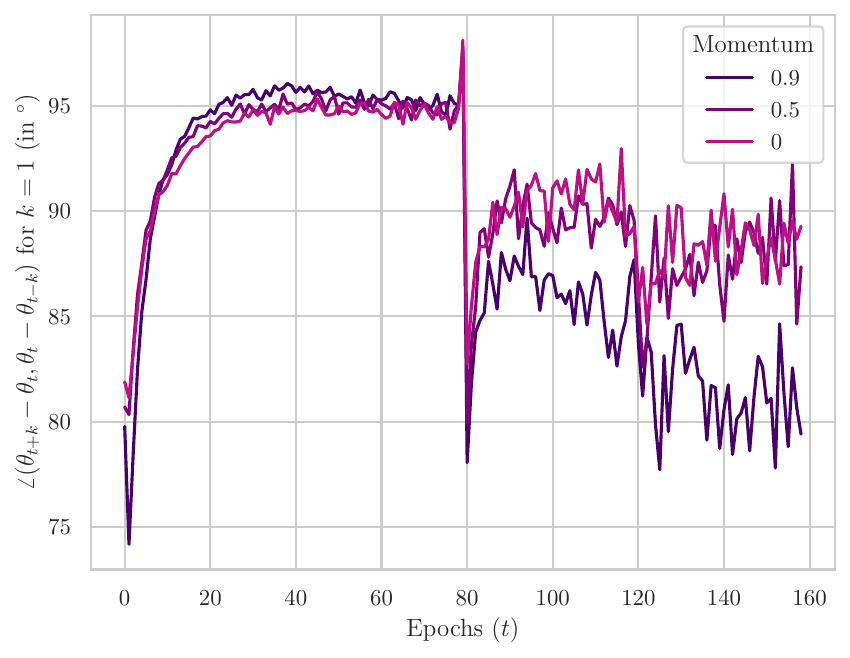}
		\vspace{-2mm}}
	\subfigure[$\angle(\theta_{t}-\theta_0,\theta_T-\theta_0)$]{\label{fig:}
		\includegraphics[width=0.34\textwidth]{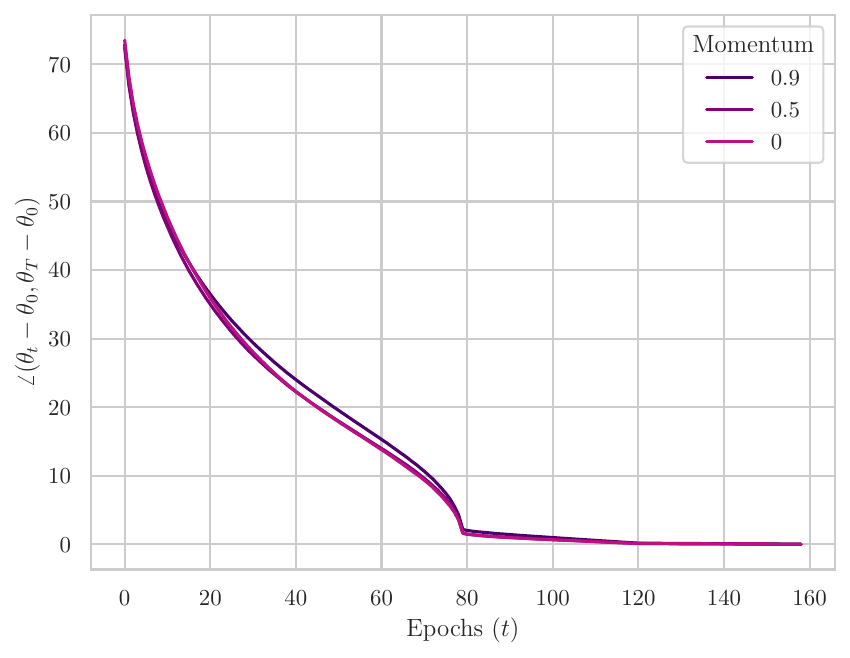}
		\vspace{-2mm}}
	\subfigure[$\angle(\theta_{t+1}-\theta_t, \theta_T-\theta_0)$]{\label{fig:}
		\includegraphics[width=0.34\textwidth]{figures/icml/Momentum/ckpt_freq-1_heatmap_from_multi-3_vgg16_bn_parents-_0__relu_cifar10_ln-0_ep-160_lr-0.01_mom-0.9_sched-0.1_0.25_wd-0.0001_2024-01-21_20-48-27_061932_2024-02-01_02-26-35_733233/figures/pdf/angle_theta__t+1_-theta_t,theta_T-theta_0__vs_Epochs__t__across_Momentum.pdf}}
	\subfigure[Apex Angle at Initialization $\angle(\theta_t-\theta_0,\theta_1-\theta_0)$ ]{\label{fig:}
		\includegraphics[width=0.34\textwidth]{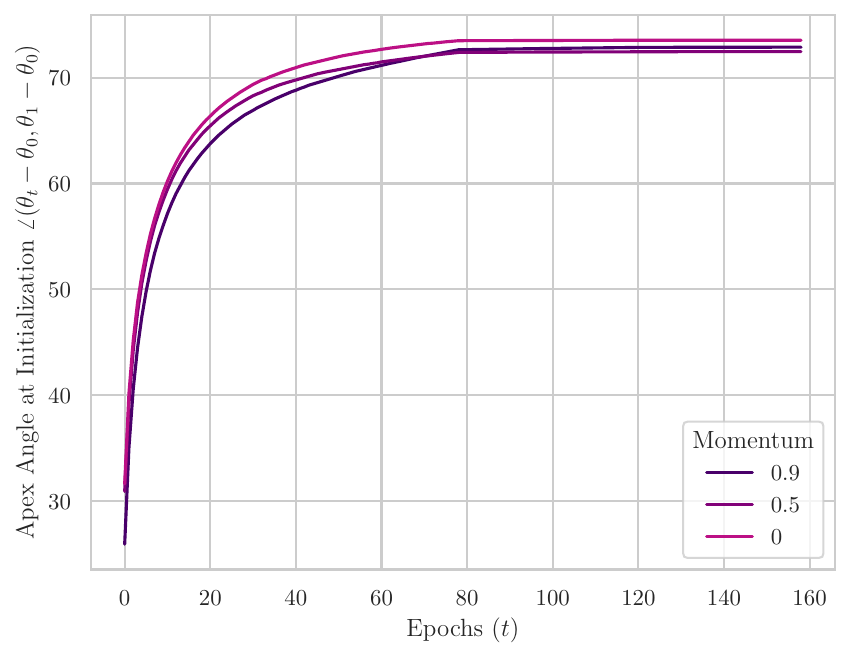}
		\vspace{-2mm}}
	\subfigure[Apex Angle at Origin $\angle(\theta_t,\theta_0)$]{\label{fig:}
		\includegraphics[width=0.34\textwidth]{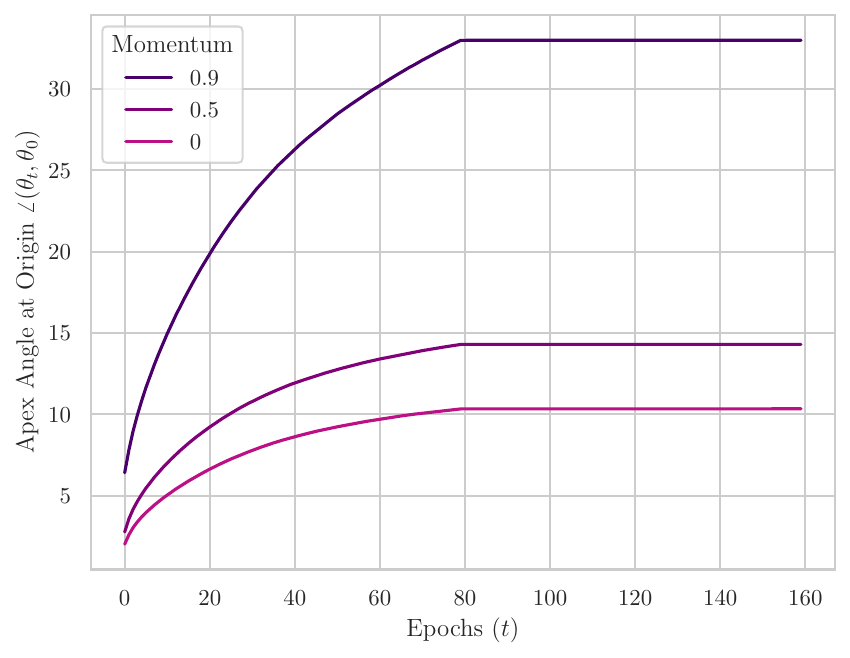}
		\vspace{-2mm}}
	\caption{Angular measures of the Trajectory for VGG16 models trained on CIFAR10.} 
\end{figure*}

\begin{figure*}[h!]
	\centering
	\subfigure[$\|\theta_t\|_2$]{\label{fig:}
	\includegraphics[width=0.3\textwidth]{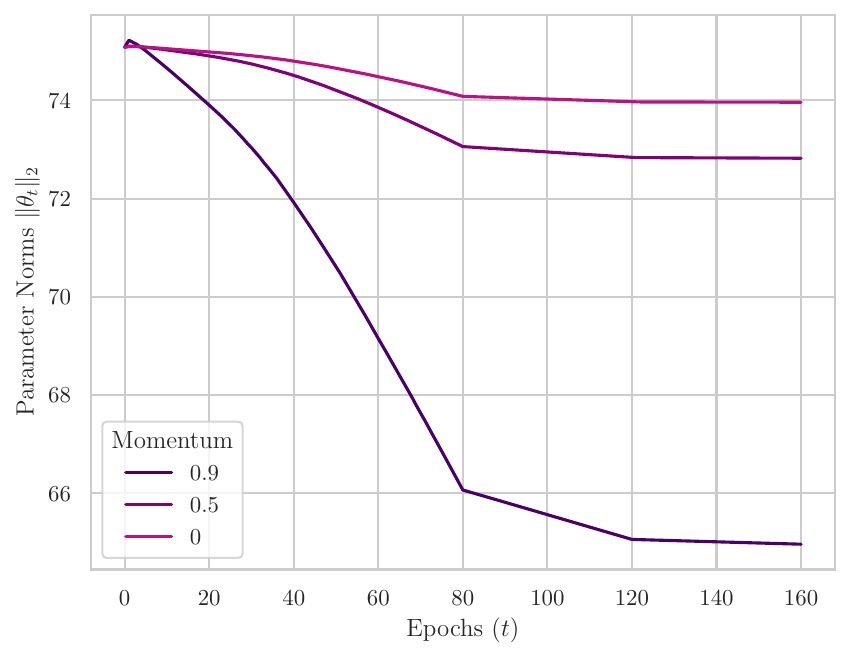}} \subfigure[$\|\theta_{t+k}-\theta_t\|_2$ ]{\label{fig:}
		\includegraphics[width=0.3\textwidth]{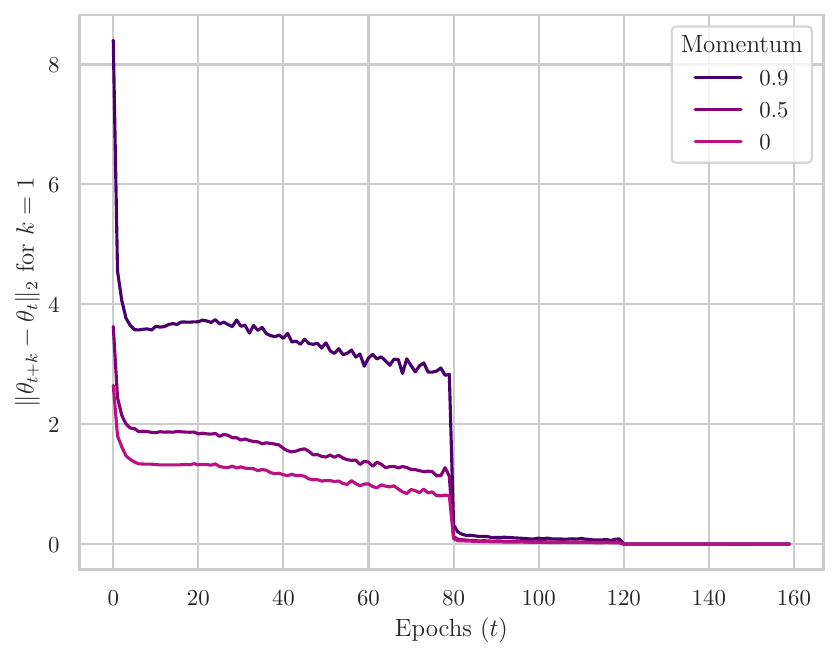}}
	\subfigure[$\|\theta_t-\theta_0\|_2$ ]{\label{fig:}
		\includegraphics[width=0.3\textwidth]{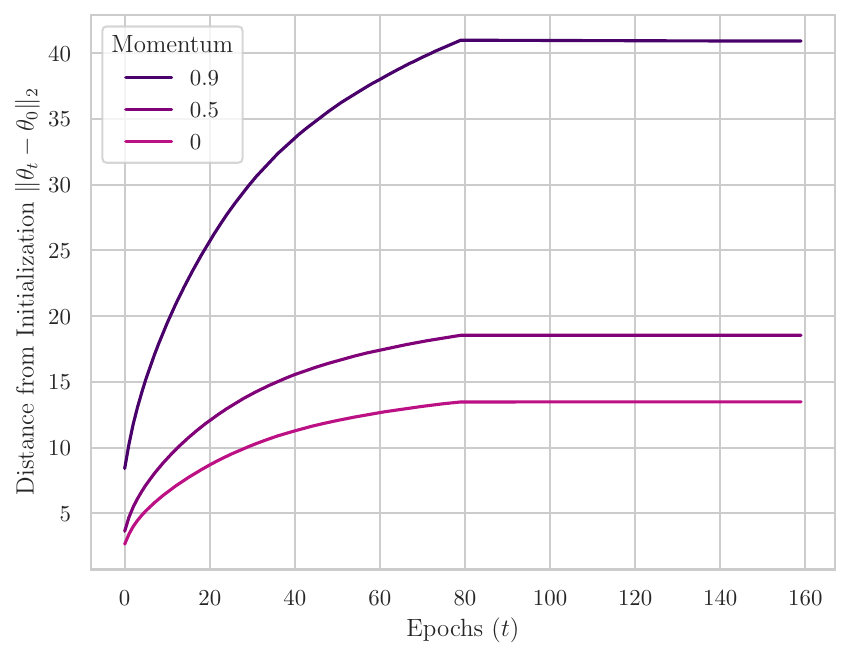}
		\vspace{-2mm}}
	\caption{Norm-based measures of the Trajectory for VGG16 models trained on CIFAR10.} 
\end{figure*}

\begin{figure*}[h!]
	\centering
	\subfigure[Eigenvalues: $\Km$]{\label{fig:}
		\includegraphics[width=0.3\textwidth]{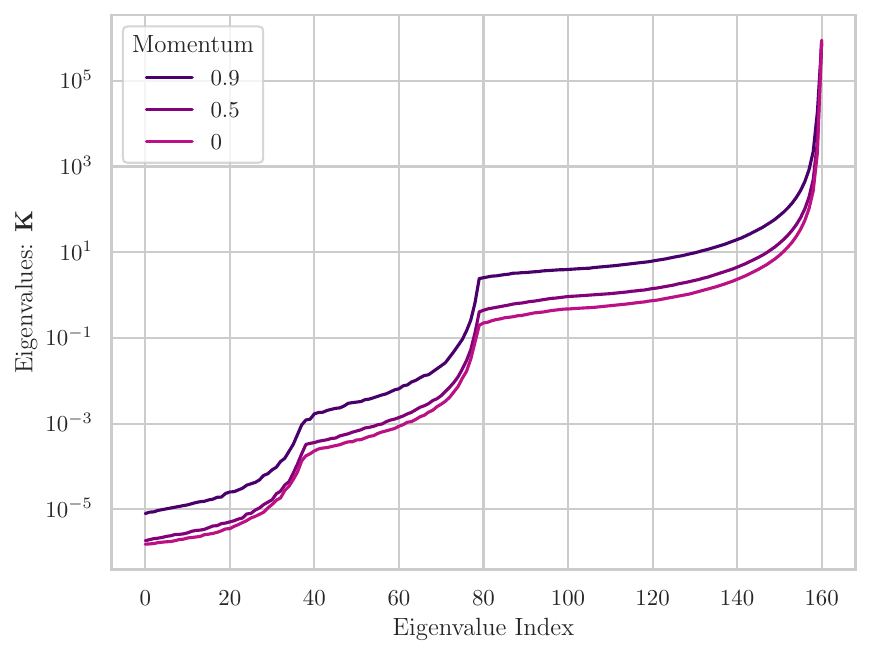}}
	\subfigure[Eigenvalues: $\Km_0$]{\label{fig:}
		\includegraphics[width=0.3\textwidth]{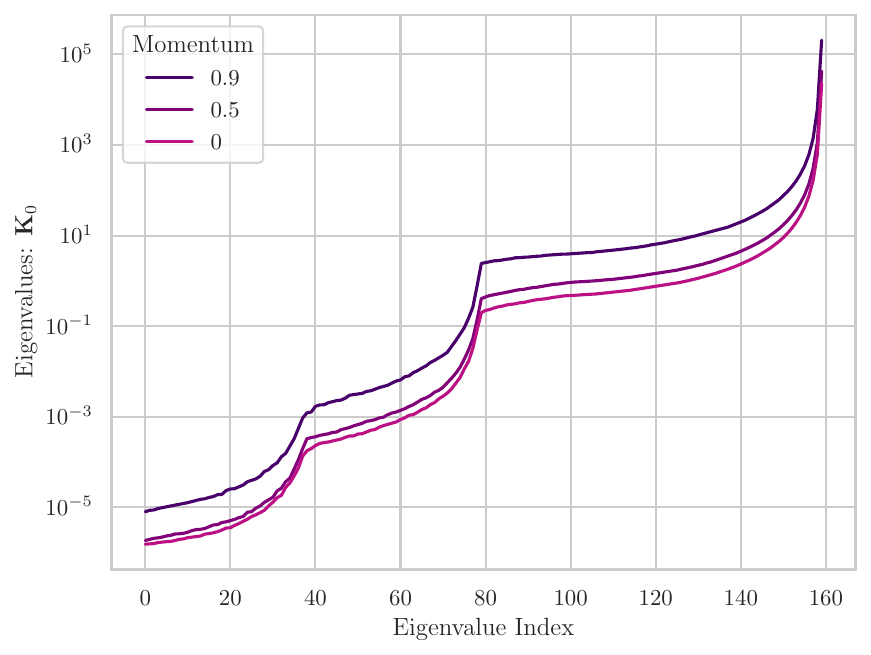}}

	\subfigure[Eigenvalues: $\Cm$ ]{\label{fig:}
		\includegraphics[width=0.3\textwidth]{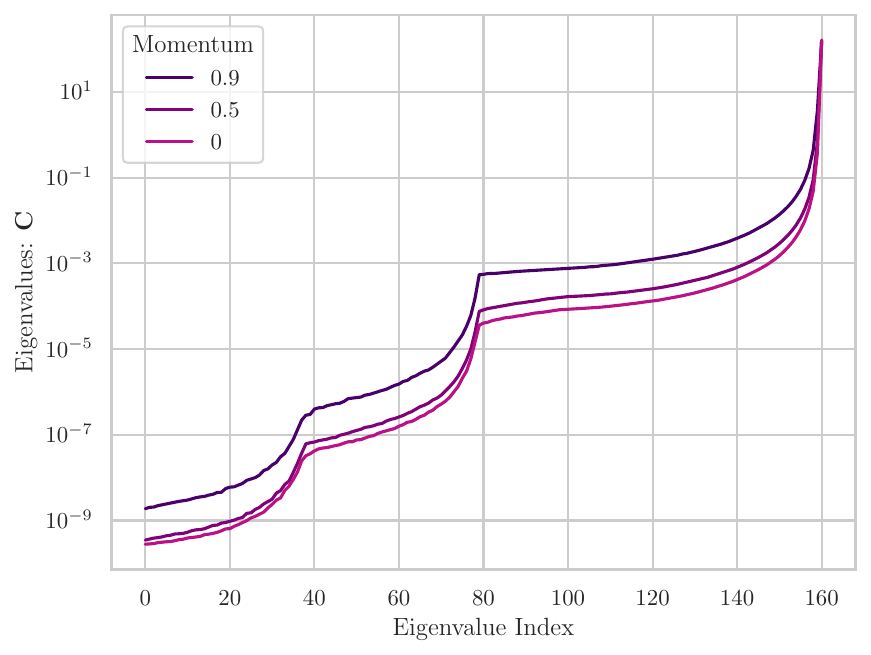}
		\vspace{-2mm}}
	\subfigure[Eigenvalues: $\Cm_0$]{\label{fig:}
		\includegraphics[width=0.3\textwidth]{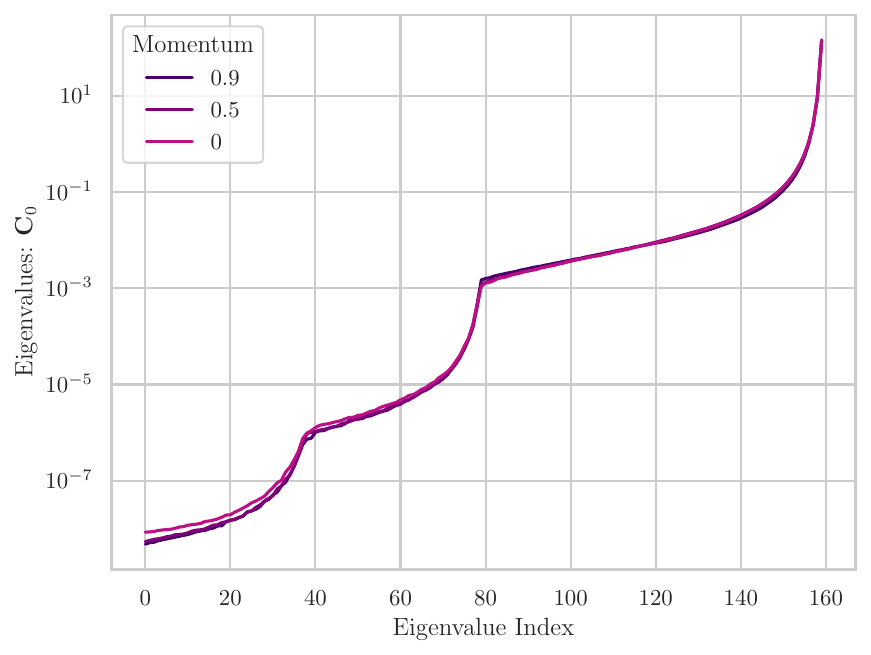}
		\vspace{-2mm}}
	\caption{Spectral measures of the Trajectory for VGG16 models trained on CIFAR10.} 
\end{figure*}

\clearpage
\subsection{VGG: Momentum Analysis, LR 0.01, WD 0}

\begin{figure*}[h!]
	\centering
	\includegraphics[width=0.9\textwidth]{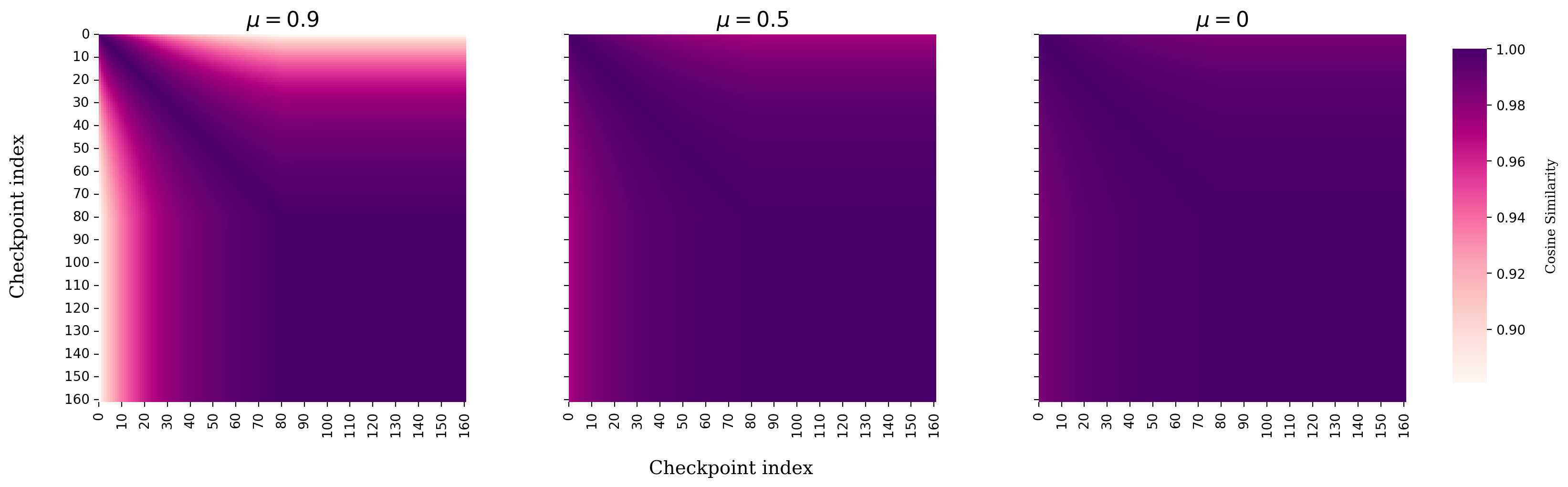}
	\caption{Trajectory Maps of VGG16 models across different amounts of momentum} 
\end{figure*}

\begin{figure*}[h!]
	\centering
	\includegraphics[width=0.9\textwidth]{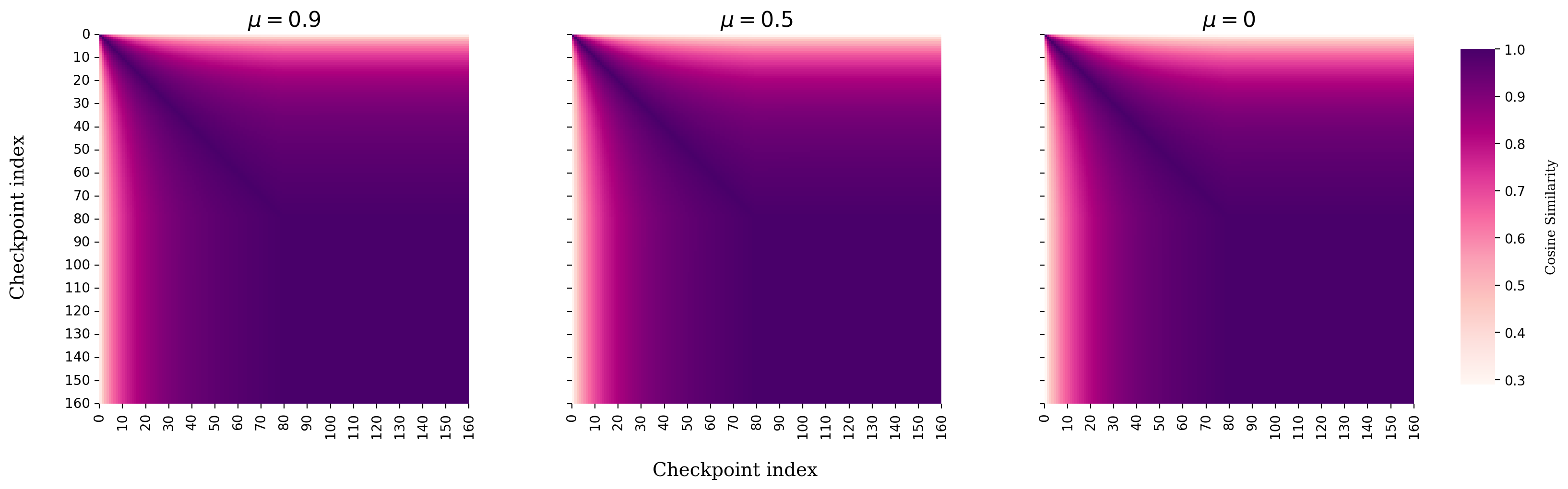}
	\caption{Relative Trajectory Maps, with respect to initialization, of VGG16 models across different amounts of momentum} 
\end{figure*}

\clearpage
\begin{figure*}[h!]
	\centering
	\subfigure[$\angle(\theta_{t+1}-\theta_t,\theta_t)$]{\label{fig:}
		\includegraphics[width=0.34\textwidth]{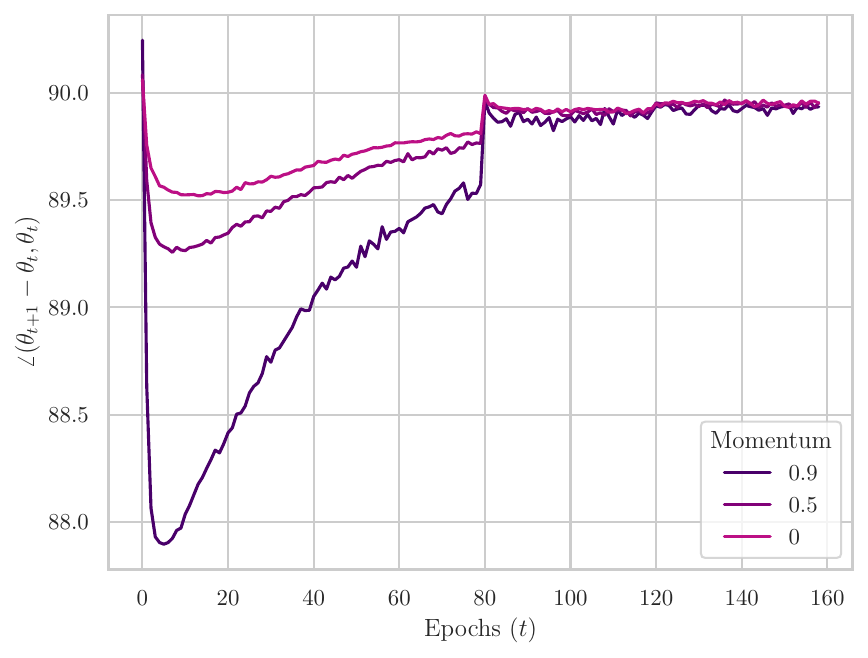}}
	\subfigure[$\angle(\theta_{t+1}-\theta_t,\theta_T-\theta_0)$]{\label{fig:}
		\includegraphics[width=0.34\textwidth]{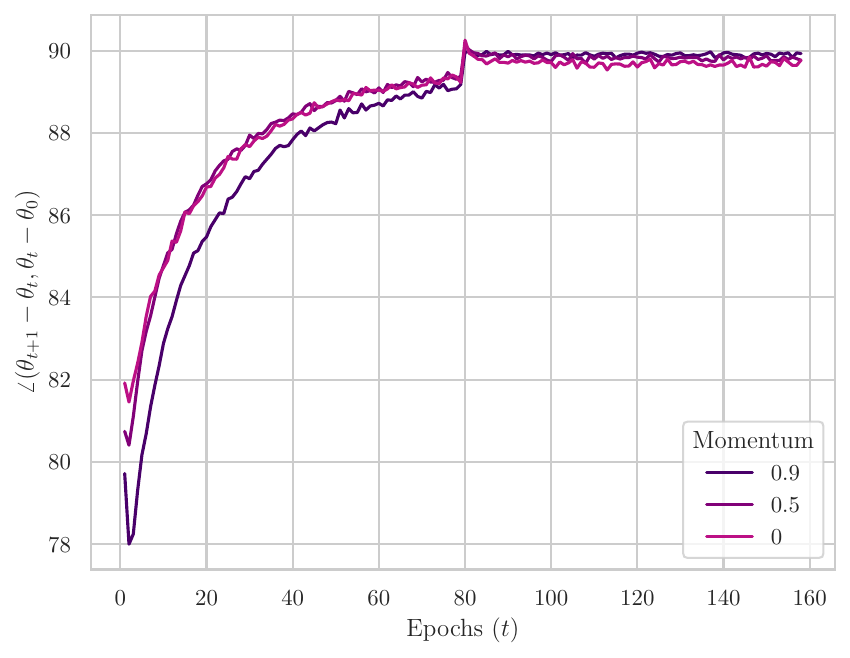}}
	\subfigure[$\angle(\theta_{t+k}-\theta_t,\theta_t-\theta_{t-k})$, for $k=1$ ]{\label{fig:}
		\includegraphics[width=0.34\textwidth]{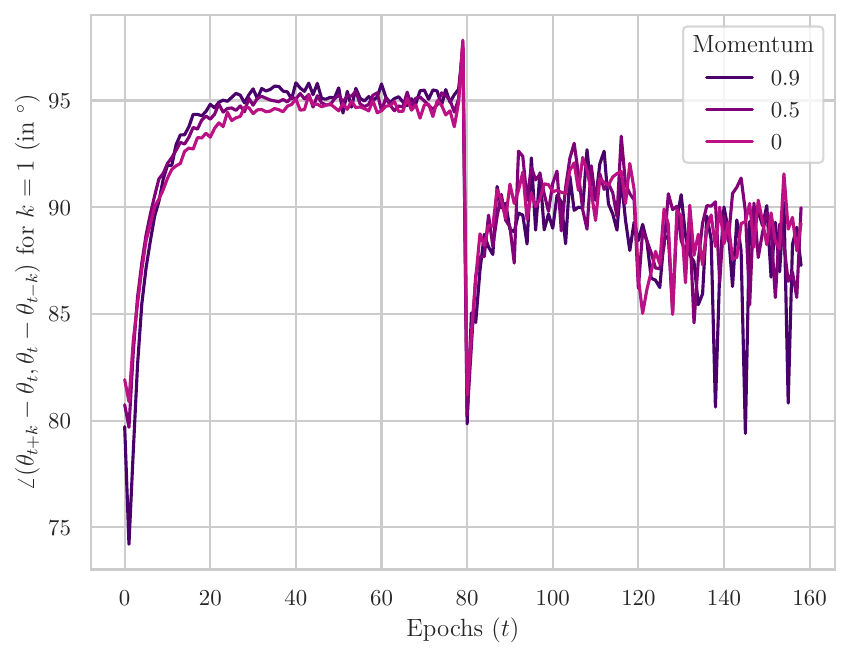}
		\vspace{-2mm}}
	\subfigure[$\angle(\theta_{t}-\theta_0,\theta_T-\theta_0)$]{\label{fig:}
		\includegraphics[width=0.34\textwidth]{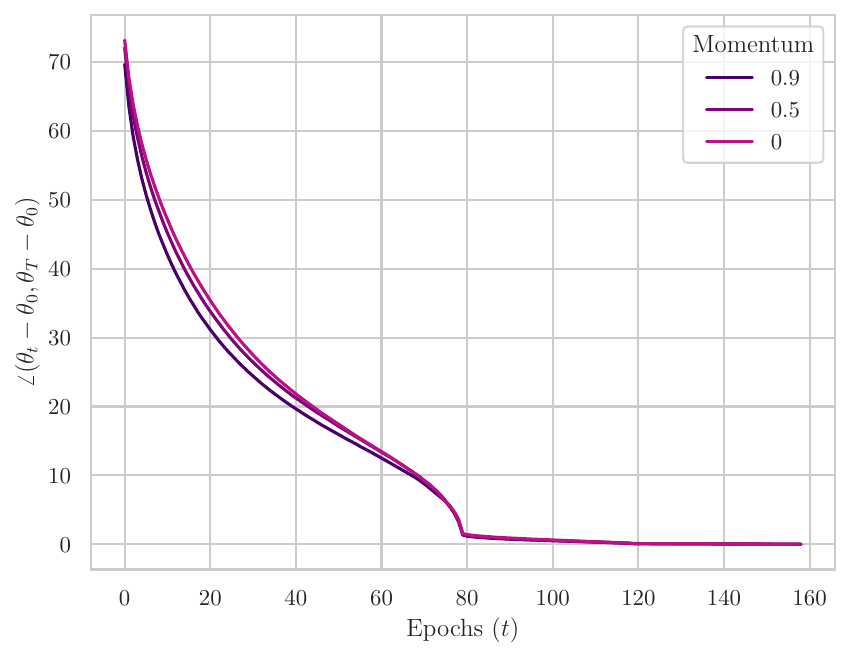}
		\vspace{-2mm}}
	\subfigure[$\angle(\theta_{t+1}-\theta_t, \theta_T-\theta_0)$]{\label{fig:}
		\includegraphics[width=0.34\textwidth]{figures/icml/Momentum/ckpt_freq-1_heatmap_from_multi-3_vgg16_bn_parents-_0__relu_cifar10_ln-0_ep-160_lr-0.01_mom-0.9_sched-0.1_0.25_wd-0.0_2024-01-21_20-48-27_064743_2024-02-01_02-26-36_288581/figures/pdf/angle_theta__t+1_-theta_t,theta_T-theta_0__vs_Epochs__t__across_Momentum.pdf}}
	\subfigure[Apex Angle at Initialization $\angle(\theta_t-\theta_0,\theta_1-\theta_0)$ ]{\label{fig:}
		\includegraphics[width=0.34\textwidth]{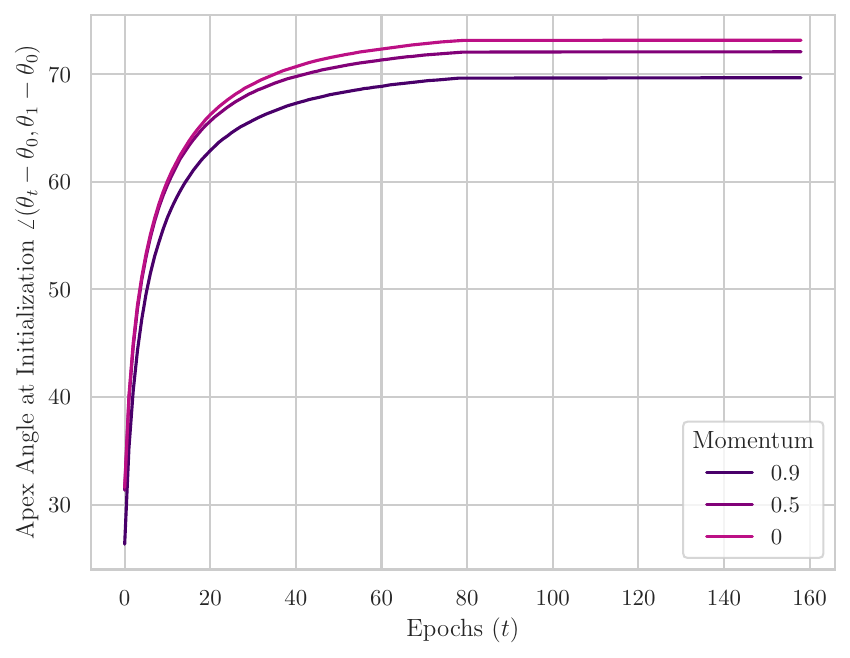}
		\vspace{-2mm}}
	\subfigure[Apex Angle at Origin $\angle(\theta_t,\theta_0)$]{\label{fig:}
		\includegraphics[width=0.34\textwidth]{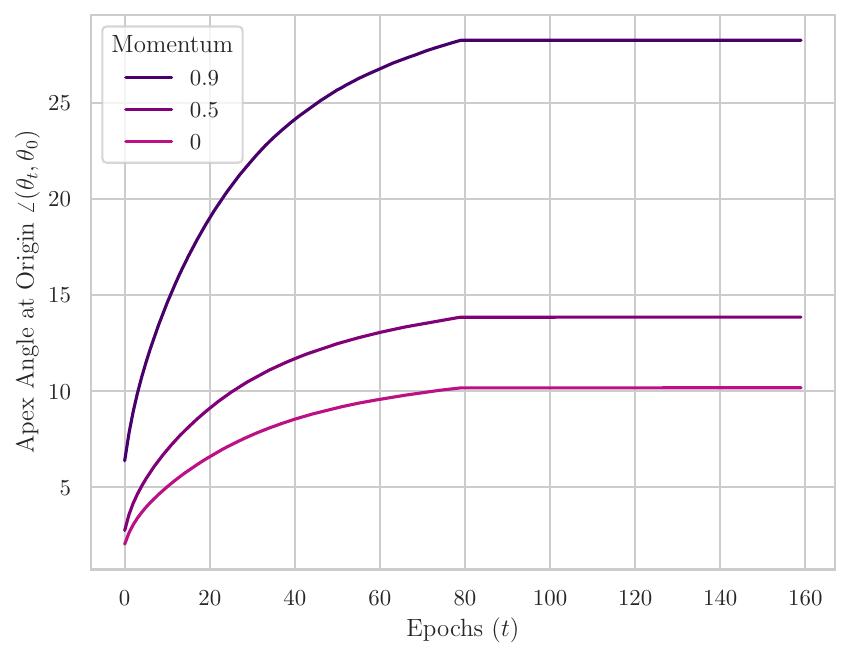}
		\vspace{-2mm}}
	\caption{Angular measures of the Trajectory for VGG16 models trained on CIFAR10.} 
\end{figure*}

\begin{figure*}[h!]
	\centering
	\subfigure[$\|\theta_t\|_2$]{\label{fig:}
	\includegraphics[width=0.3\textwidth]{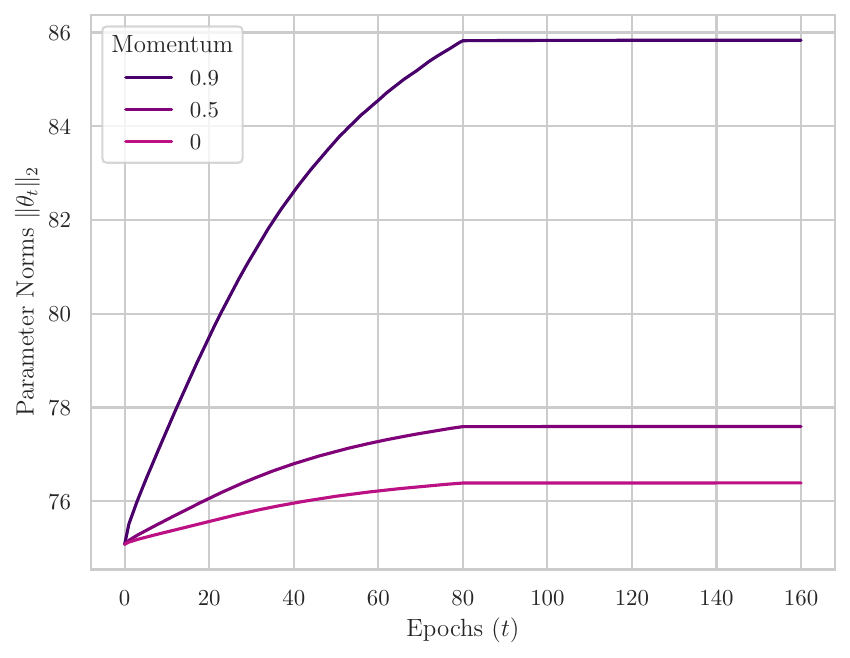}} \subfigure[$\|\theta_{t+k}-\theta_t\|_2$ ]{\label{fig:}
		\includegraphics[width=0.3\textwidth]{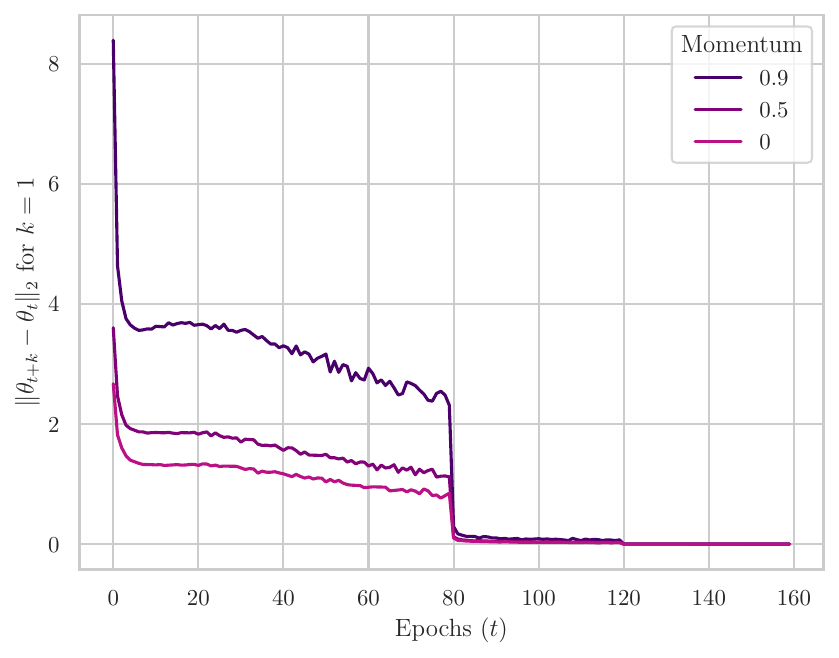}}
	\subfigure[$\|\theta_t-\theta_0\|_2$ ]{\label{fig:}
		\includegraphics[width=0.3\textwidth]{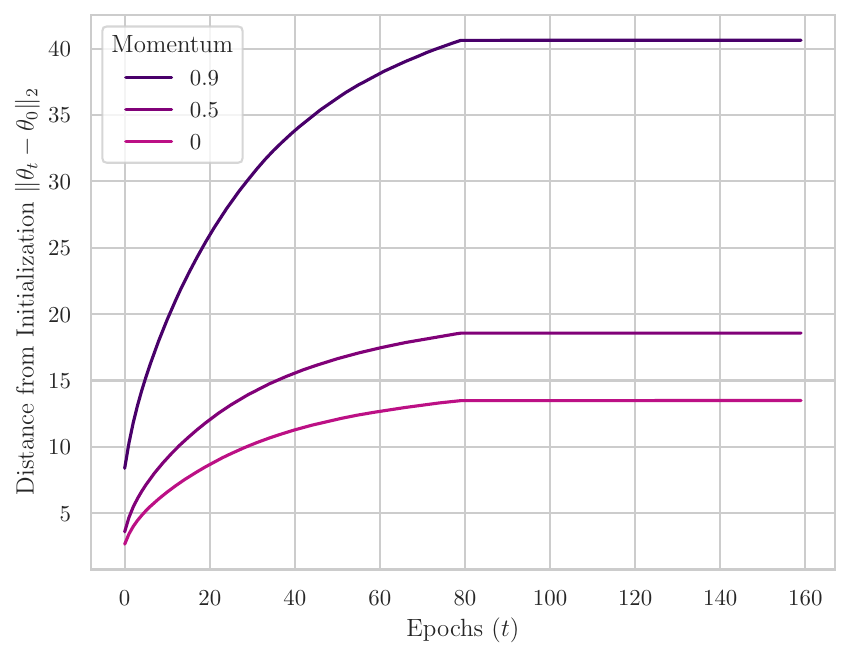}
		\vspace{-2mm}}
	\caption{Norm-based measures of the Trajectory for VGG16 models trained on CIFAR10.} 
\end{figure*}

\begin{figure*}[h!]
	\centering
	\subfigure[Eigenvalues: $\Km$]{\label{fig:}
		\includegraphics[width=0.3\textwidth]{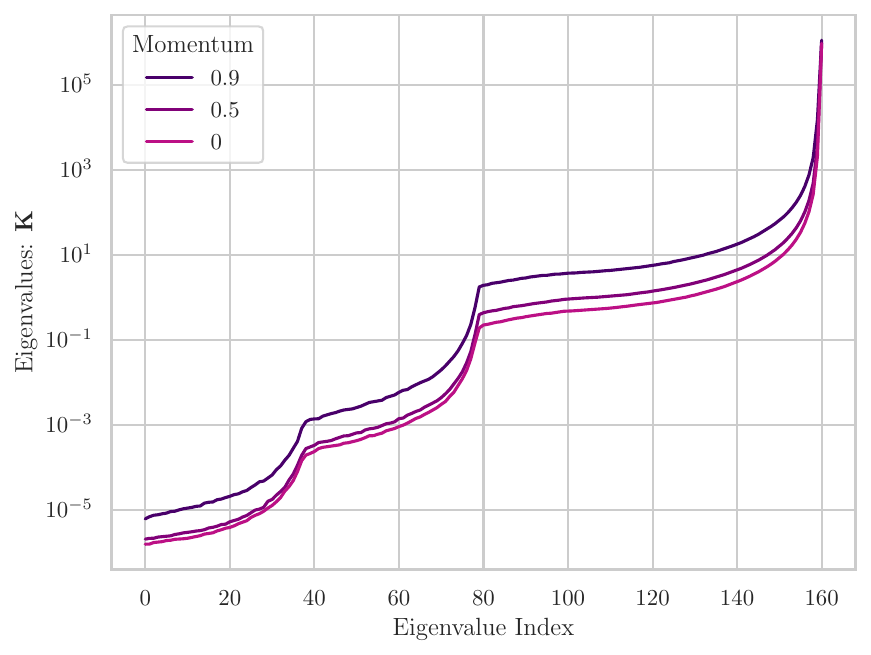}}
	\subfigure[Eigenvalues: $\Km_0$]{\label{fig:}
		\includegraphics[width=0.3\textwidth]{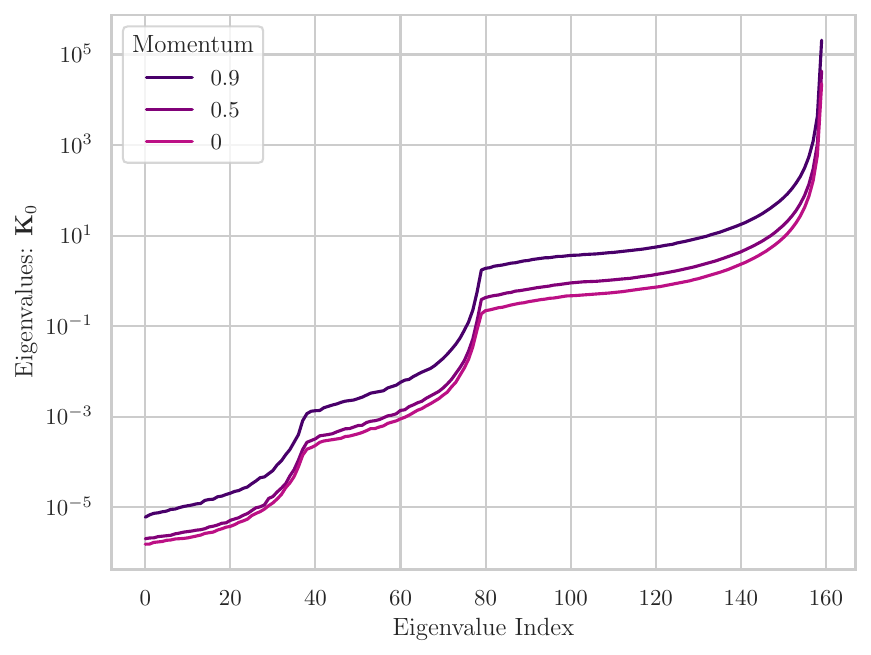}}

	\subfigure[Eigenvalues: $\Cm$ ]{\label{fig:}
		\includegraphics[width=0.3\textwidth]{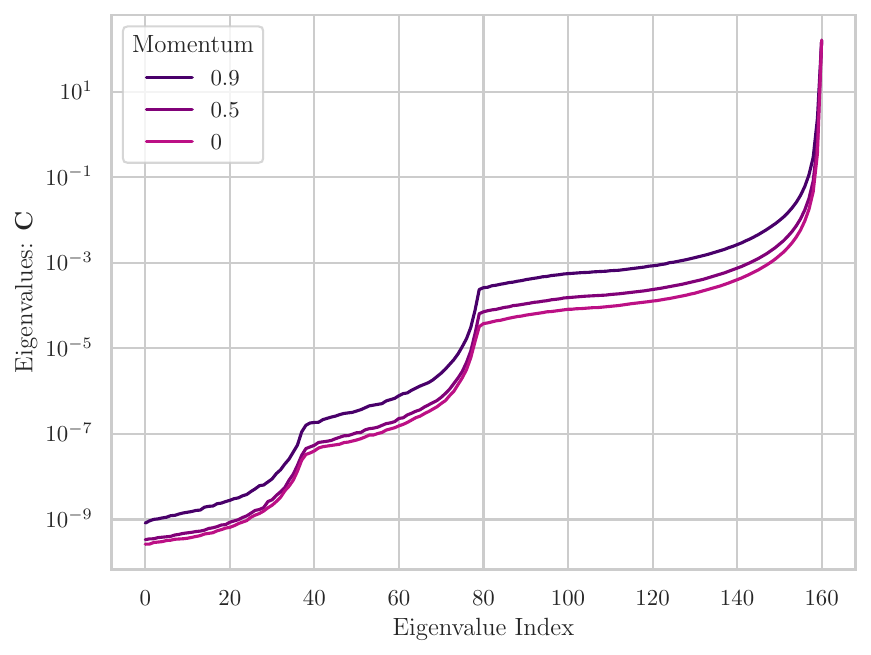}
		\vspace{-2mm}}
	\subfigure[Eigenvalues: $\Cm_0$]{\label{fig:}
		\includegraphics[width=0.3\textwidth]{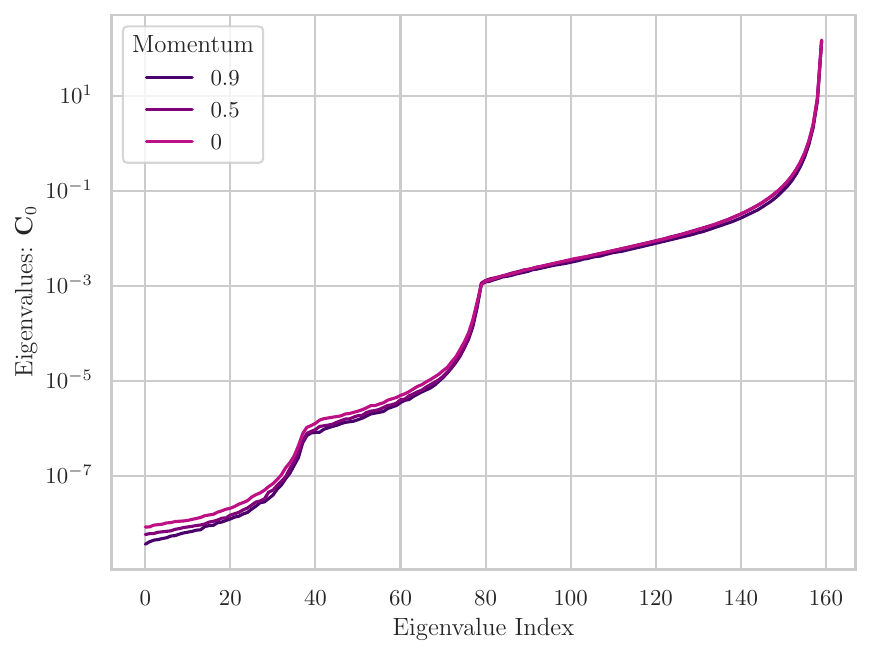}
		\vspace{-2mm}}
	\caption{Spectral measures of the Trajectory for VGG16 models trained on CIFAR10.} 
\end{figure*}

\clearpage

\subsection{VGG16 Batch Size Analysis}\label{app:bsz}
\begin{figure*}[h!]
	\centering
	\includegraphics[width=0.9\textwidth]{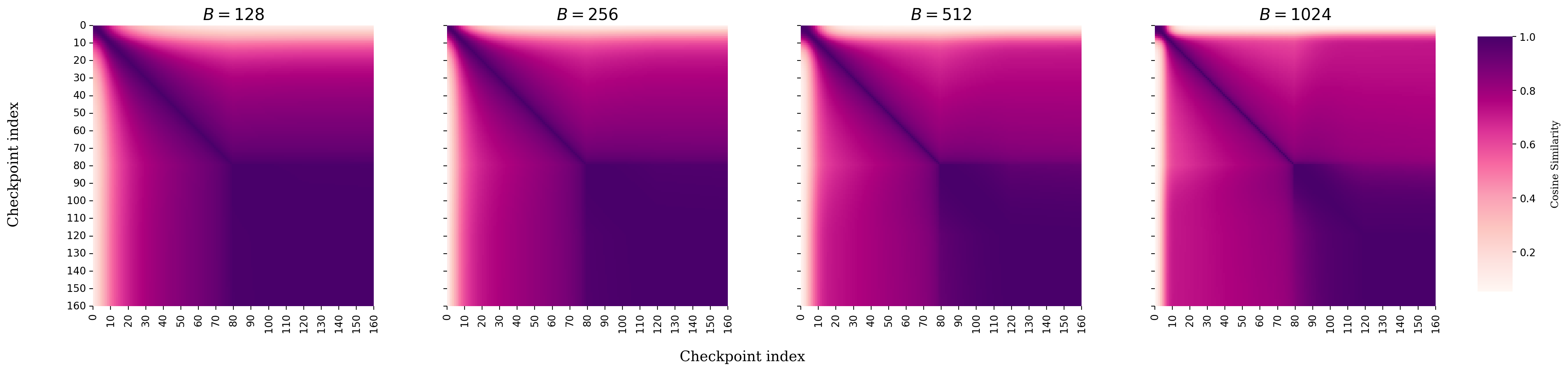}
	\caption{Relative Trajectory Maps, with respect to initialization, of VGG16 models across different batch sizes} 
\end{figure*}
\clearpage
\begin{figure*}[h!]
	\centering
	\subfigure[$\angle(\theta_{t+1}-\theta_t,\theta_t)$]{\label{fig:}
		\includegraphics[width=0.34\textwidth]{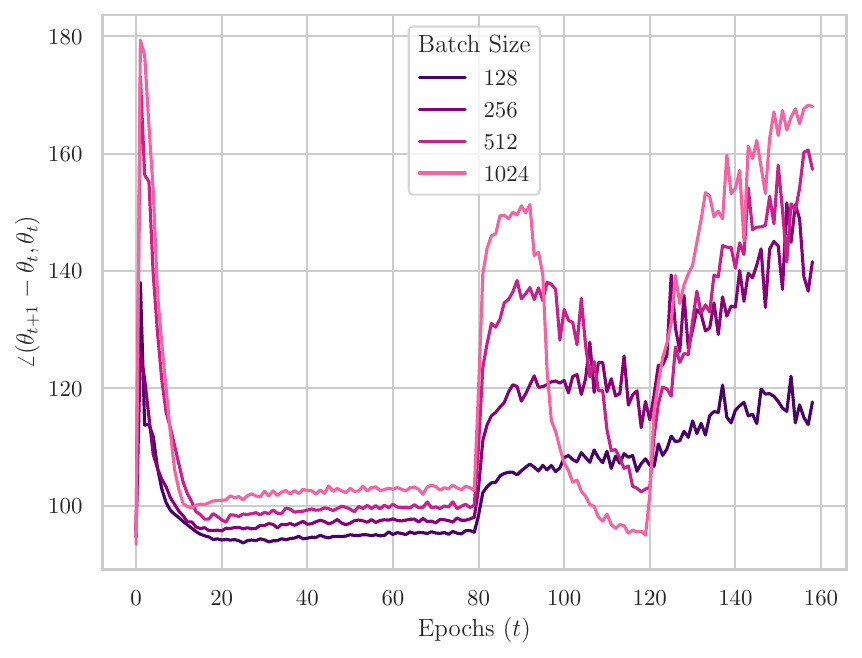}}
	\subfigure[$\angle(\theta_{t+1}-\theta_t,\theta_T-\theta_0)$]{\label{fig:}
		\includegraphics[width=0.34\textwidth]{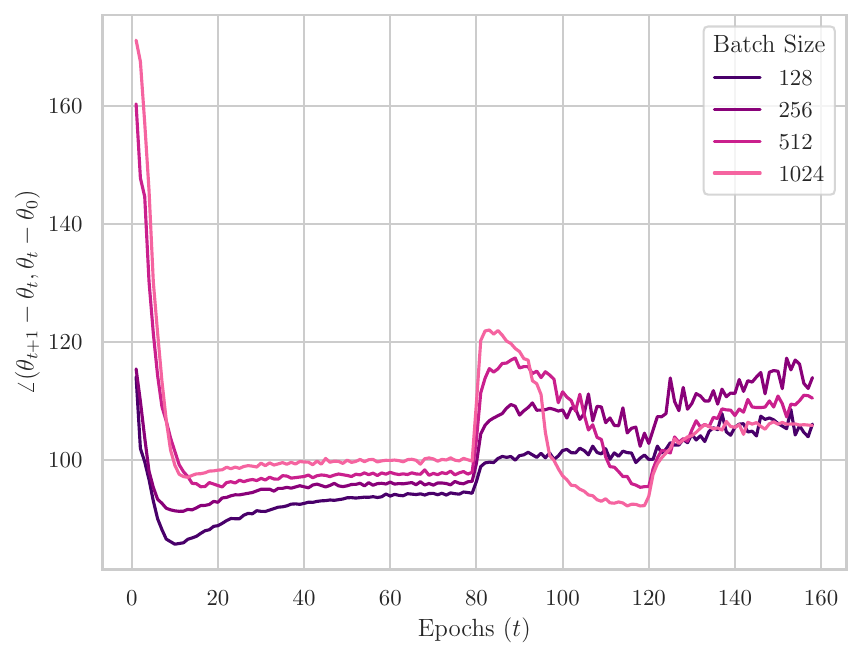}}
	\subfigure[$\angle(\theta_{t+k}-\theta_t,\theta_t-\theta_{t-k})$, for $k=1$ ]{\label{fig:}
		\includegraphics[width=0.34\textwidth]{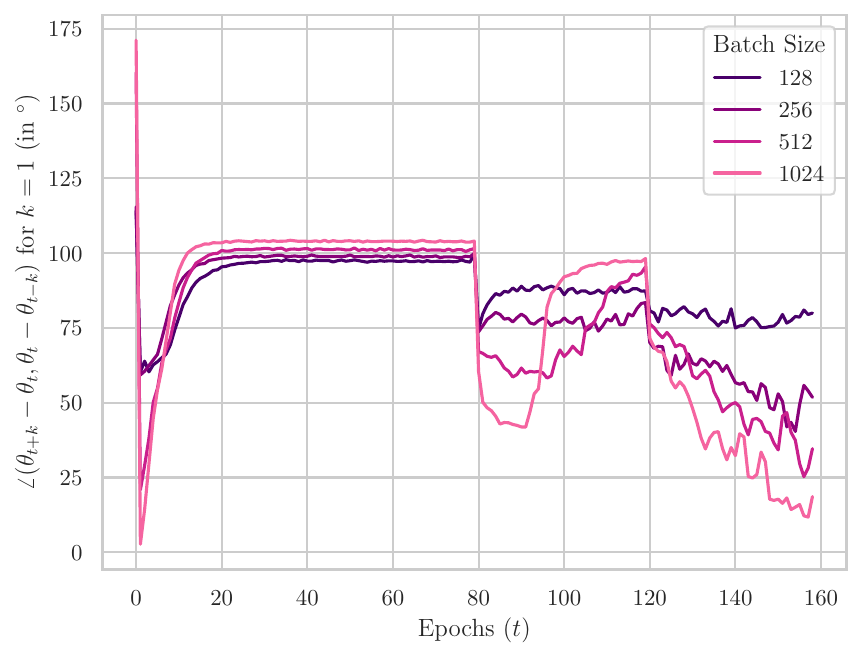}
		\vspace{-2mm}}
	\subfigure[$\angle(\theta_{t}-\theta_0,\theta_T-\theta_0)$]{\label{fig:}
		\includegraphics[width=0.34\textwidth]{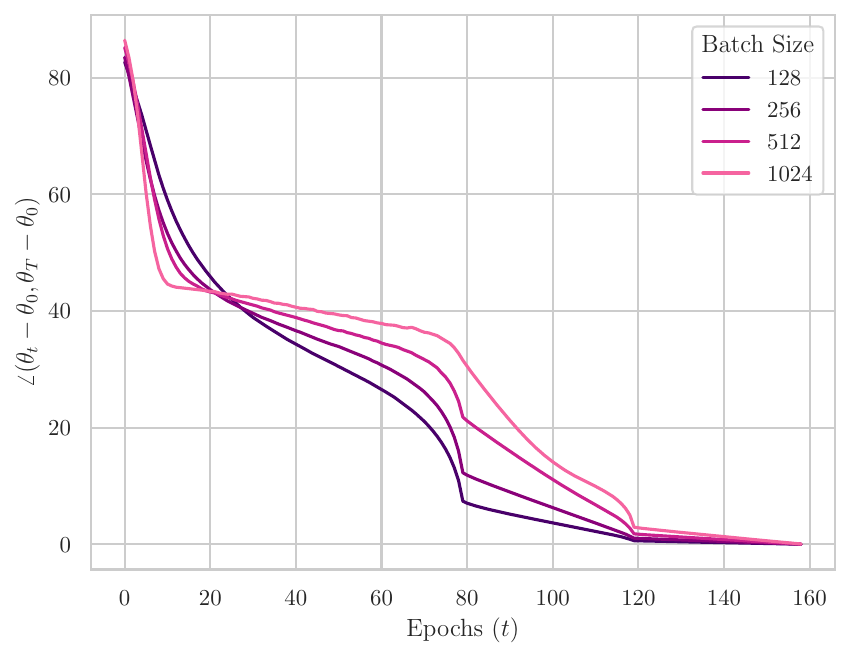}
		\vspace{-2mm}}
	\subfigure[$\angle(\theta_{t+1}-\theta_t, \theta_T-\theta_0)$]{\label{fig:}
		\includegraphics[width=0.34\textwidth]{figures/icml/Batch_Size/ckpt_freq-1_heatmap_from_multi-4_vgg16_bn_parents-_0__relu_cifar10_ln-0_ep-160_lr-0.1_mom-0.9_sched-0.1_0.25_wd-0.0001_2024-01-21_20-48-27_065553_2024-02-01_02-26-35_871133/figures/pdf/angle_theta__t+1_-theta_t,theta_T-theta_0__vs_Epochs__t__across_Batch_Size.pdf}}
	\subfigure[Apex Angle at Initialization $\angle(\theta_t-\theta_0,\theta_1-\theta_0)$ ]{\label{fig:}
		\includegraphics[width=0.34\textwidth]{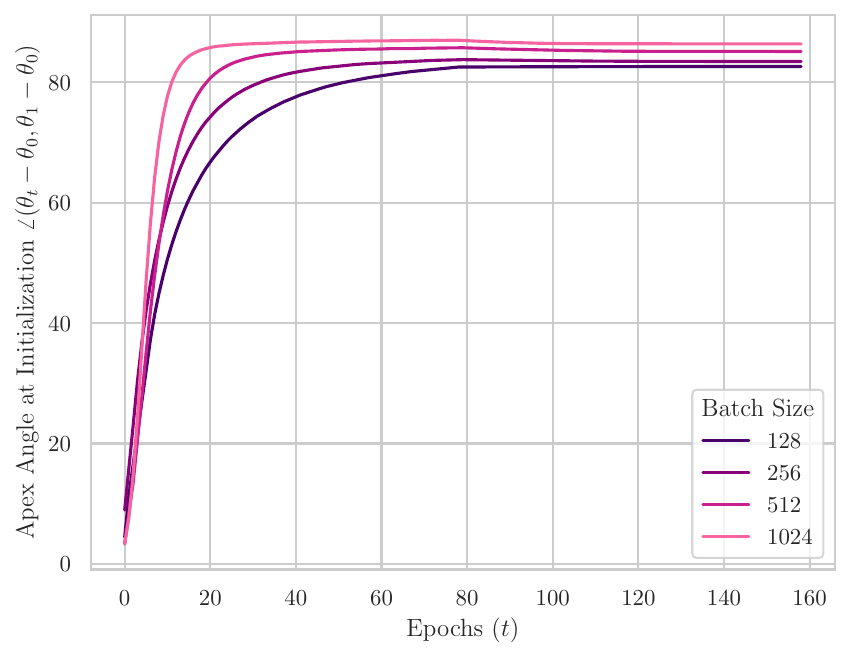}
		\vspace{-2mm}}
	\subfigure[Apex Angle at Origin $\angle(\theta_t,\theta_0)$]{\label{fig:}
		\includegraphics[width=0.34\textwidth]{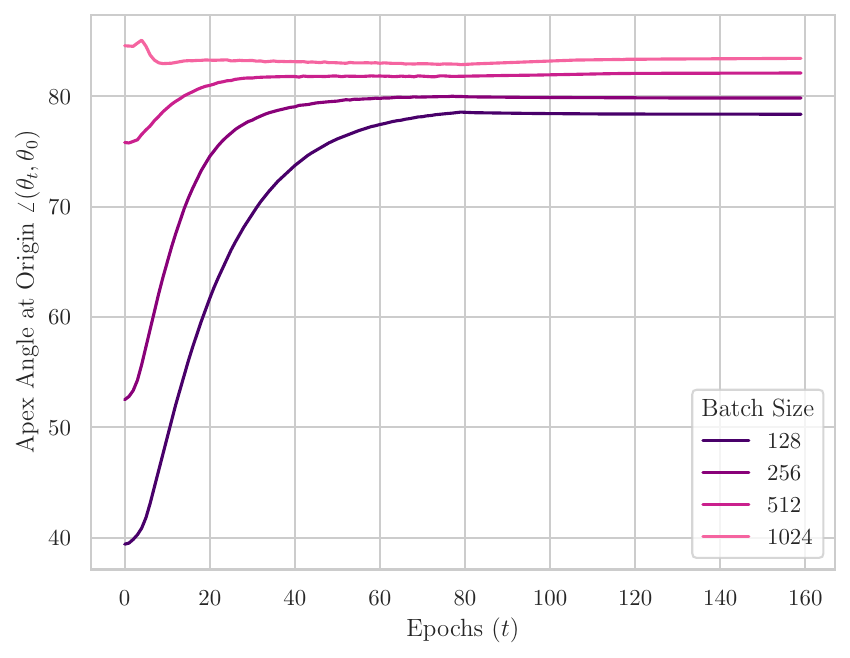}
		\vspace{-2mm}}
	\caption{Angular measures of the Trajectory for VGG16 models trained on CIFAR10.} 
\end{figure*}
\begin{figure*}[h!]
	\centering
	\subfigure[$\|\theta_t\|_2$]{\label{fig:}
	\includegraphics[width=0.3\textwidth]{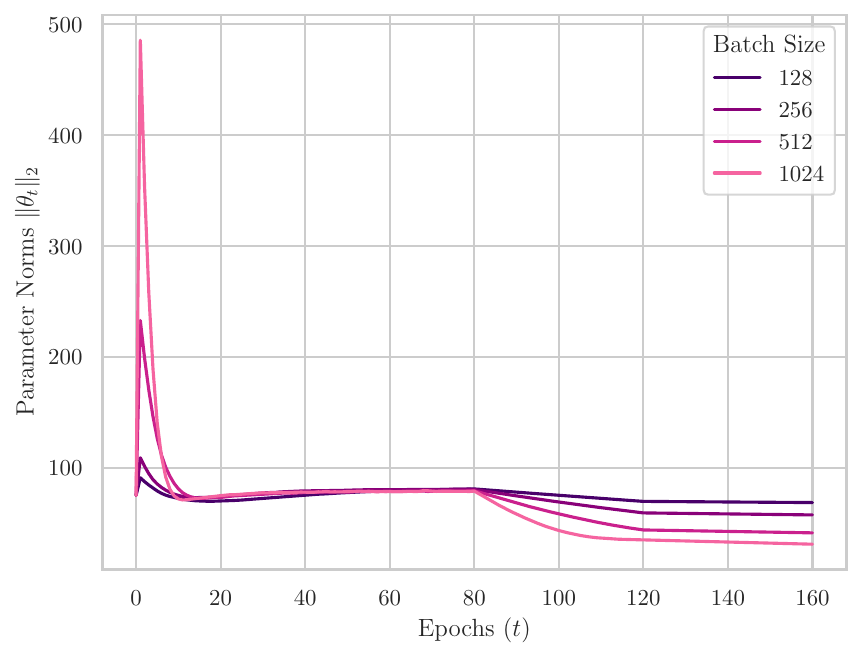}} \subfigure[$\|\theta_{t+k}-\theta_t\|_2$ ]{\label{fig:}
		\includegraphics[width=0.3\textwidth]{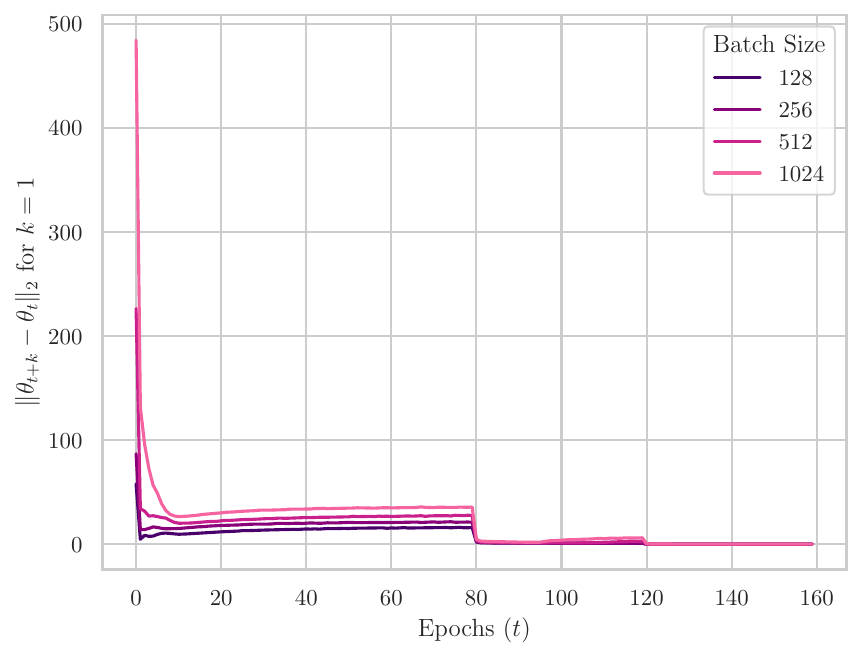}}
	\subfigure[$\|\theta_t-\theta_0\|_2$ ]{\label{fig:}
		\includegraphics[width=0.3\textwidth]{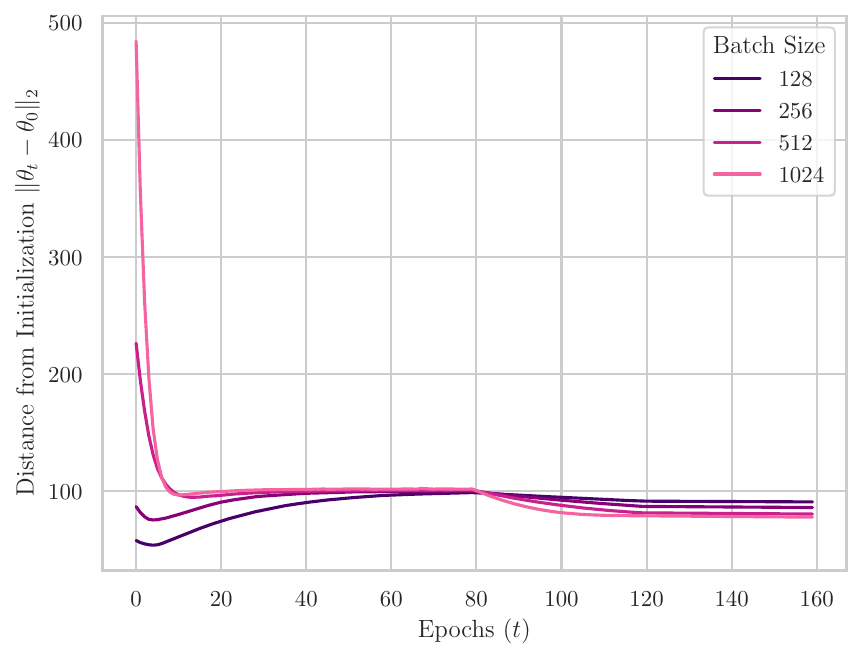}
		\vspace{-2mm}}
	\caption{Norm-based measures of the Trajectory for VGG16 models trained on CIFAR10.} 
\end{figure*}
\begin{figure*}[h!]
	\centering
	\subfigure[Eigenvalues: $\Km$]{\label{fig:}
		\includegraphics[width=0.3\textwidth]{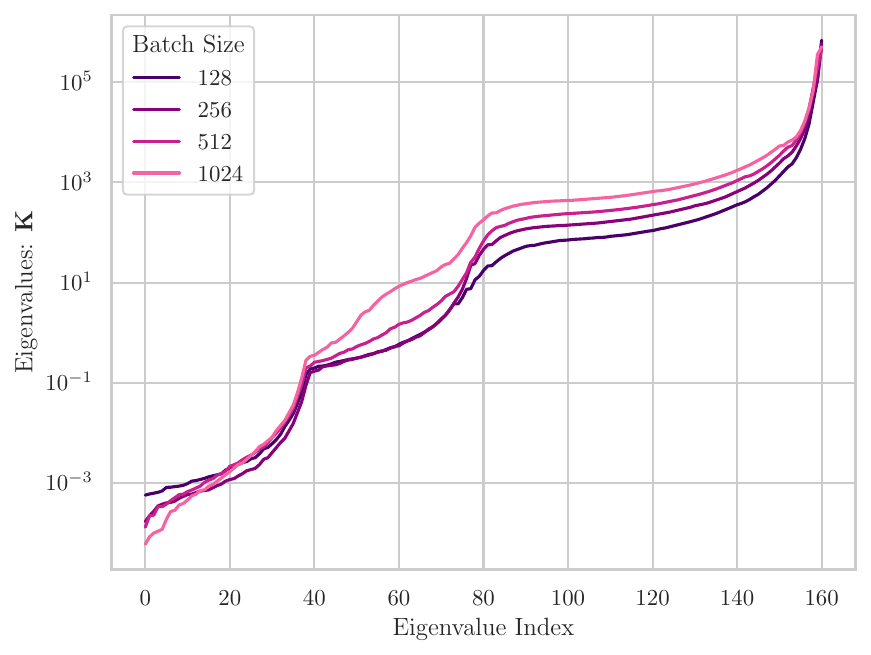}}
	\subfigure[Eigenvalues: $\Km_0$]{\label{fig:}
		\includegraphics[width=0.3\textwidth]{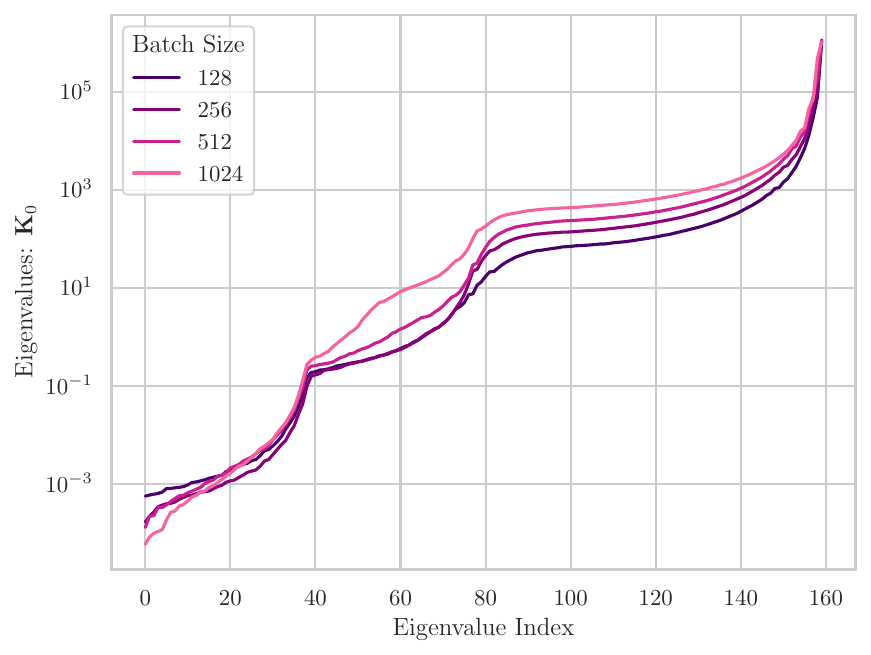}}
	\subfigure[Eigenvalues: $\Cm$ ]{\label{fig:}
		\includegraphics[width=0.3\textwidth]{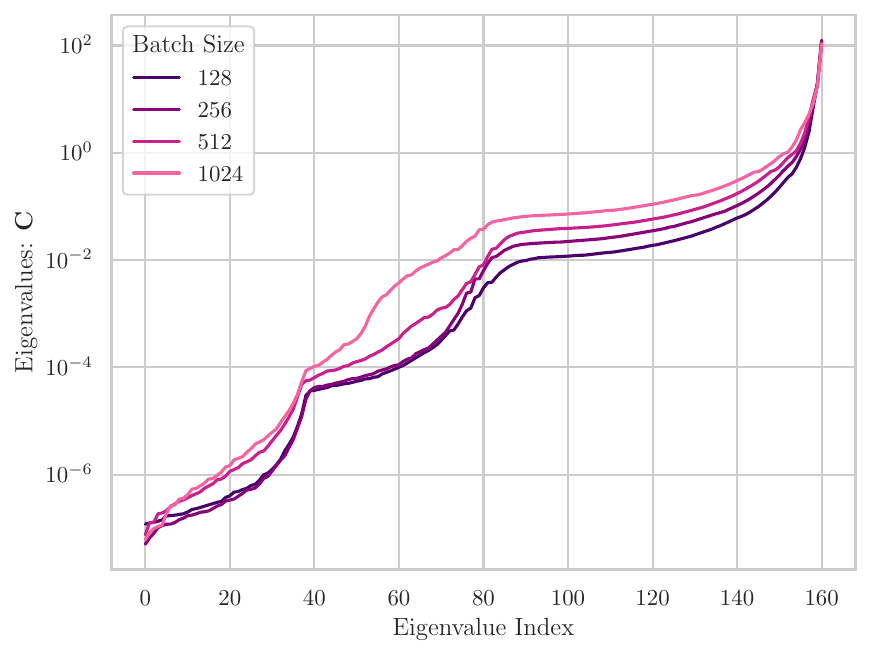}
		\vspace{-2mm}}
	\subfigure[Eigenvalues: $\Cm_0$]{\label{fig:}
		\includegraphics[width=0.3\textwidth]{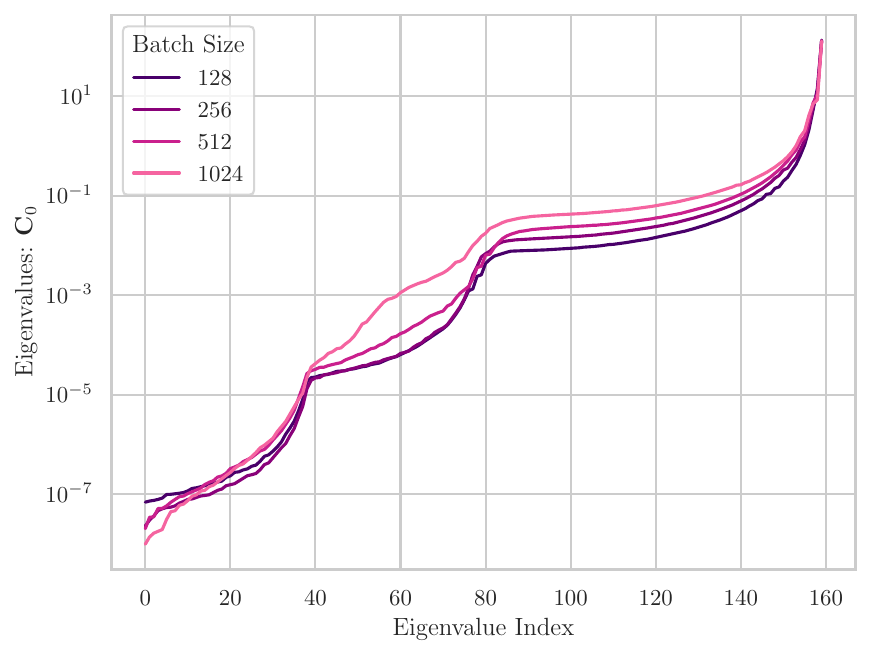}
		\vspace{-2mm}}
	\caption{Spectral measures of the Trajectory for VGG16 models trained on CIFAR10.} 
\end{figure*}

\subsection{Trajectory Maps in the presence of label noise}
\begin{figure}[!h]
    \centering
    \subfigure[Label Noise $0.4$]{\includegraphics[width=0.3\textwidth]{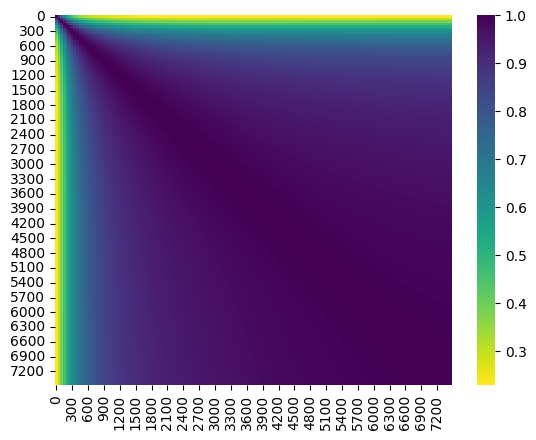}}
    \subfigure[Label Noise $0.7$]{\includegraphics[width=0.3\textwidth]{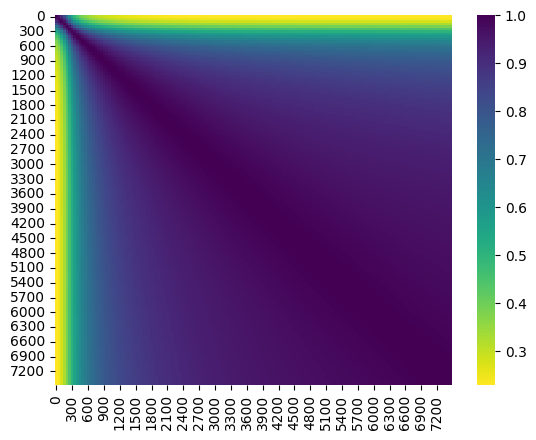}}
    \subfigure[Label Noise $1.0$]{\includegraphics[width=0.3\textwidth]{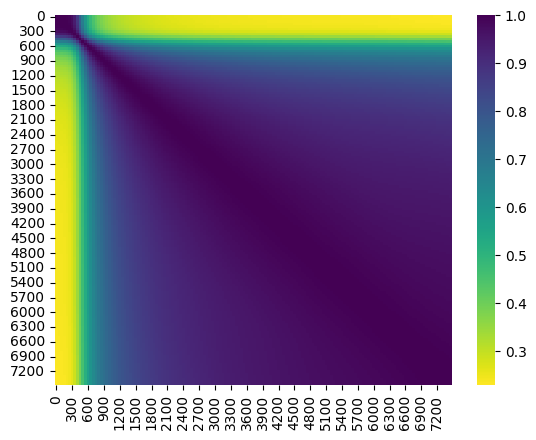}}
    \caption{Trajectory maps when a CNN is trained on CIFAR10 with different amounts of label noise, i.e., what fraction of samples have been assigned random labels.}
    \label{fig:label-noise}
\end{figure}

We observe that with increasing label noise, the network is required to undergo more directional exploration to find a solution that can interpolate the training set. The MDS scores decrease monotonically with increasing label noise. 

\clearpage

\section{GPT-NeoX Trajectory Analysis}

\begin{figure*}[h!]
	\centering    
	\includegraphics[width=0.85\textwidth]{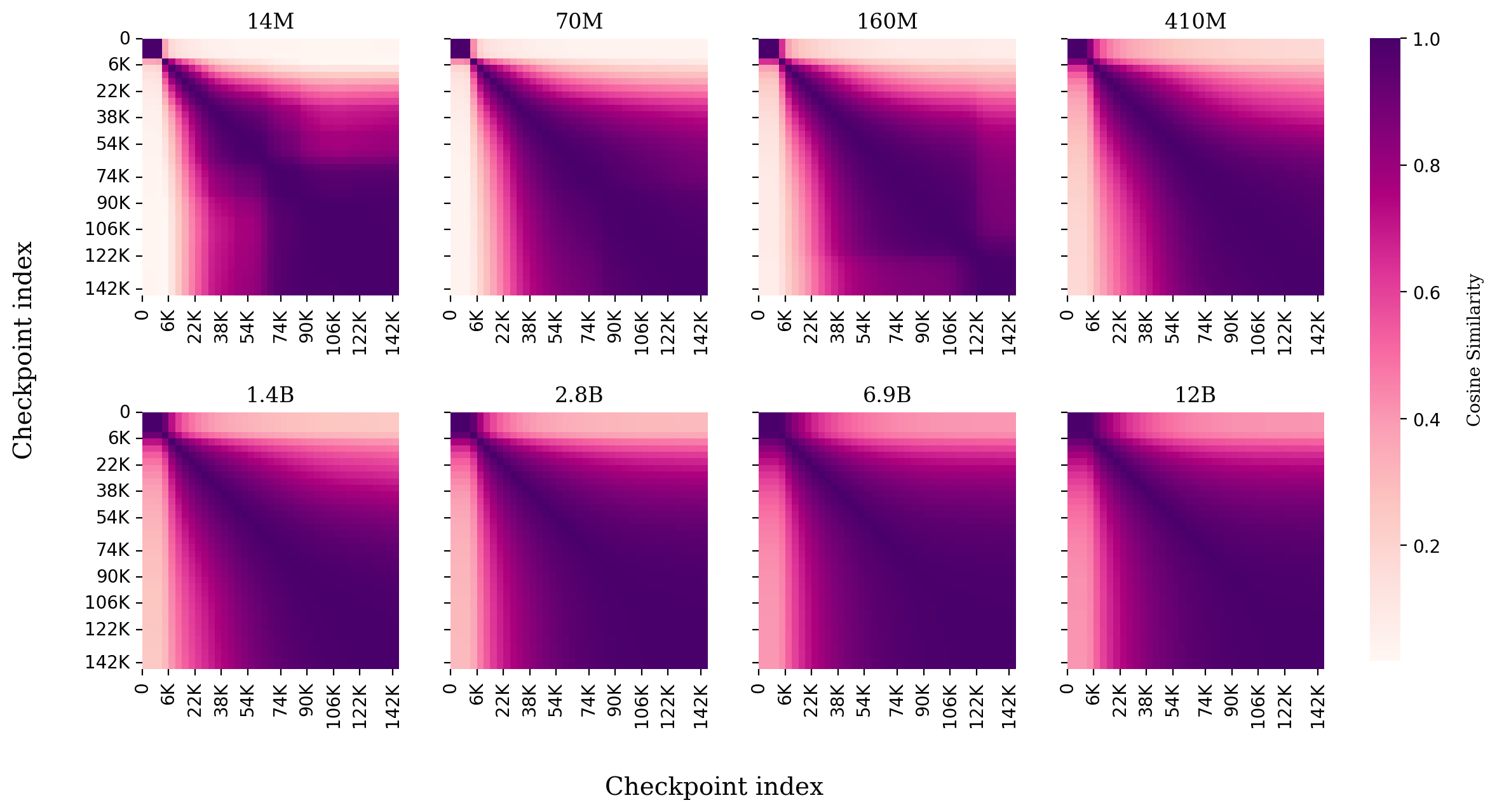}
	\caption{Trajectory Maps of Pythia GPT-NeoX models across two orders of model scales trained on Pile. The corresponding MDS values are  $\omega=0.650, 0.672, 0.678, 0.726, 0.759, 0.786, 0.818, 0.815$.} 
    \label{fig:cospythia-full}
\end{figure*}

\begin{figure*}[h!]
	\centering
	\includegraphics[width=0.9\textwidth]{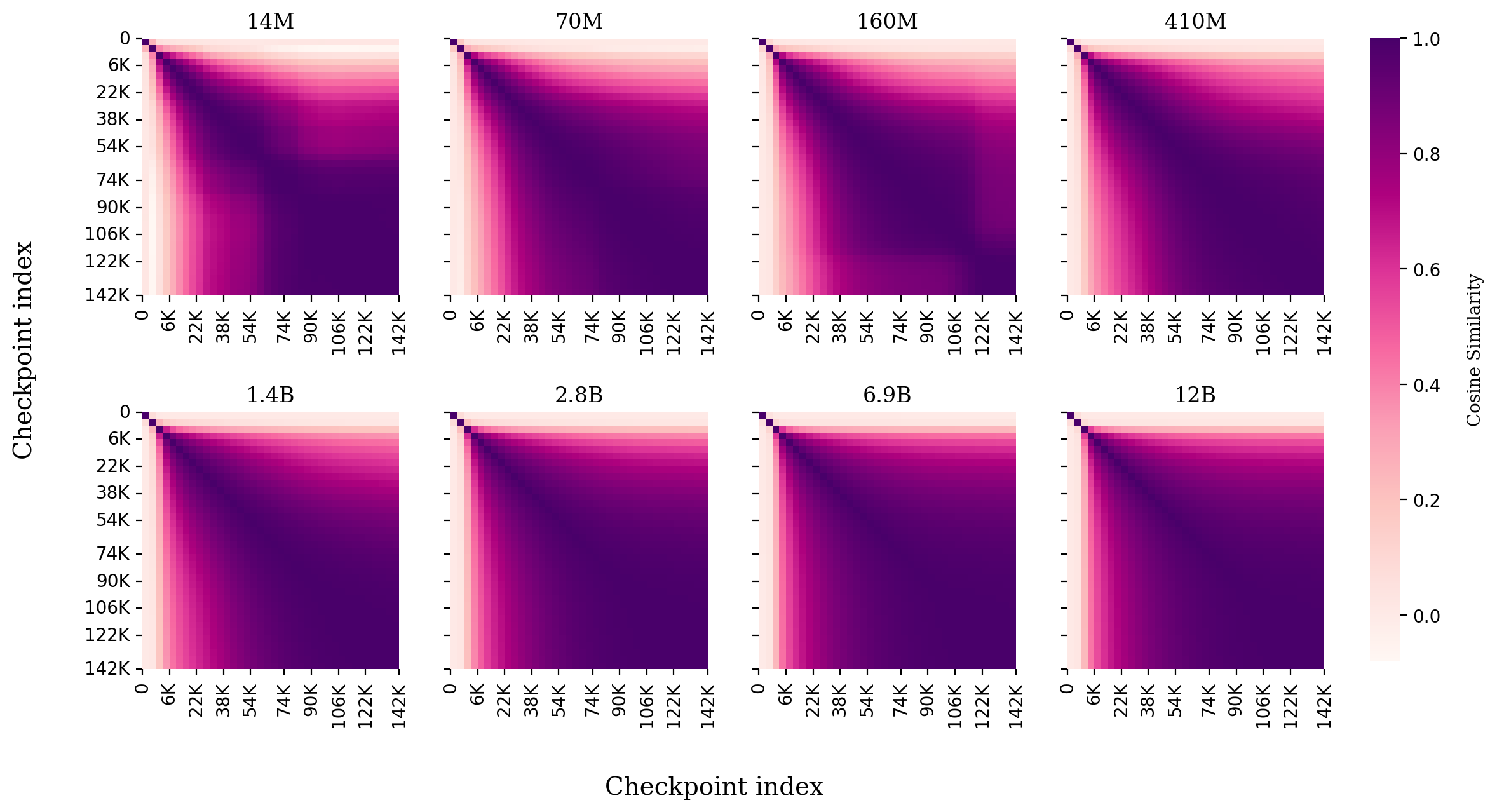}
	\caption{Relative Trajectory Maps, with respect to initialization, of Pythia GPT-NeoX models across two orders of model scales.} 
\end{figure*}

\begin{figure*}[h!]
	\centering
	\subfigure[$\|\theta_t\|_2$]{\label{fig:}
		\includegraphics[width=0.3\textwidth]{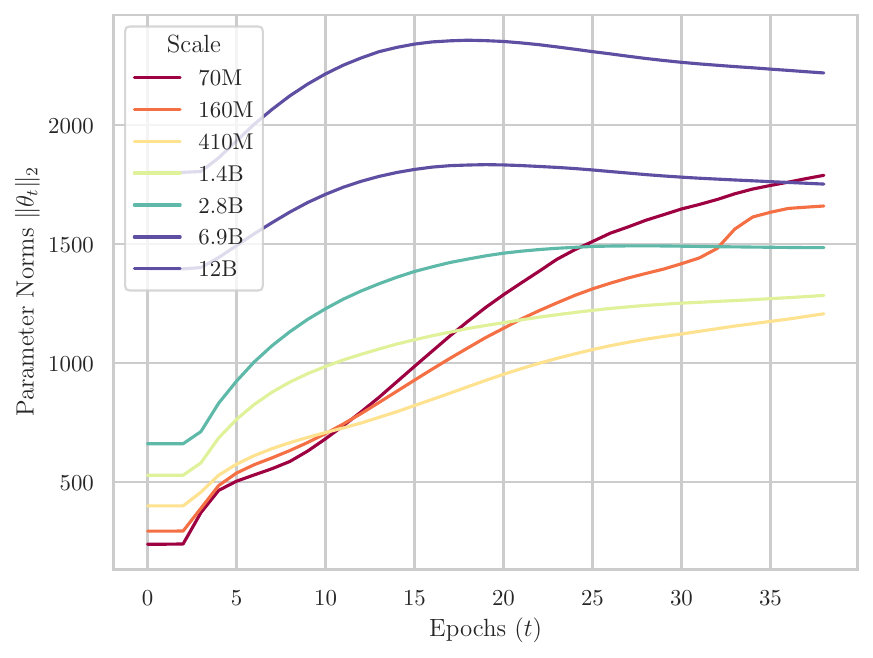}}
	\subfigure[$\|\theta_{t+k}-\theta_t\|_2$ ]{\label{fig:}
		\includegraphics[width=0.3\textwidth]{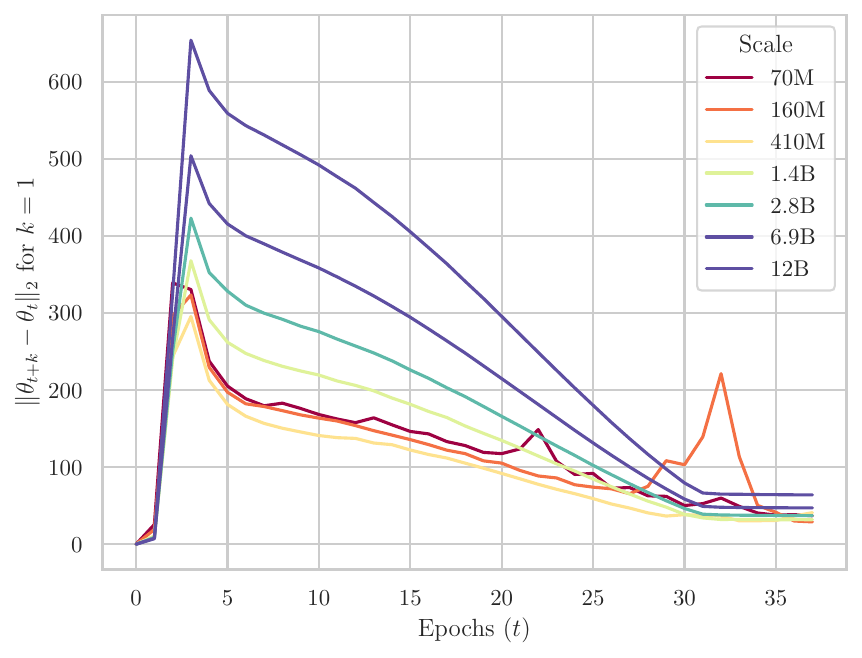}}
	\subfigure[$\|\theta_t-\theta_0\|_2$ ]{\label{fig:}
		\includegraphics[width=0.3\textwidth]{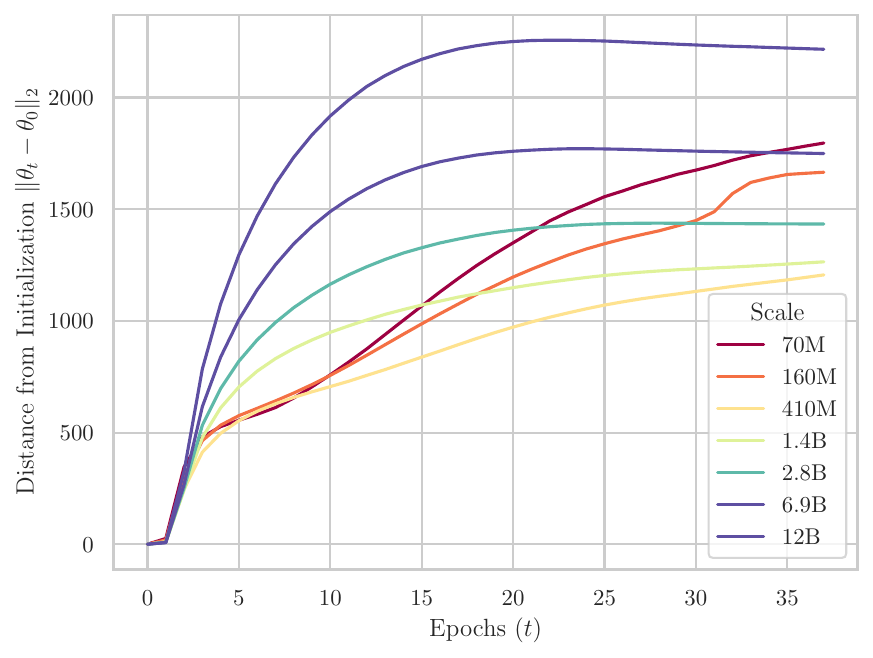}
		\vspace{-2mm}}
	\caption{Norm-based measures of the Trajectory for GPT-NeoX trained on the Pile dataset} 
\end{figure*}

\clearpage
\begin{figure*}[h!]
	\centering
	\subfigure[$\angle(\theta_{t+1}-\theta_t,\theta_t)$]{\label{fig:}
		\includegraphics[width=0.34\textwidth]{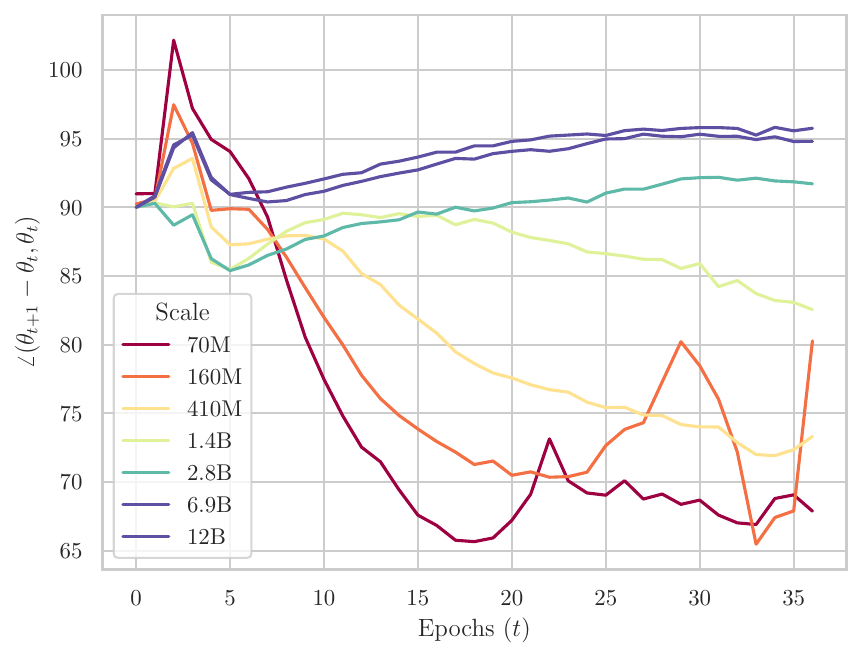}}
	\subfigure[$\angle(\theta_{t+1}-\theta_t,\theta_T-\theta_0)$]{\label{fig:}
		\includegraphics[width=0.34\textwidth]{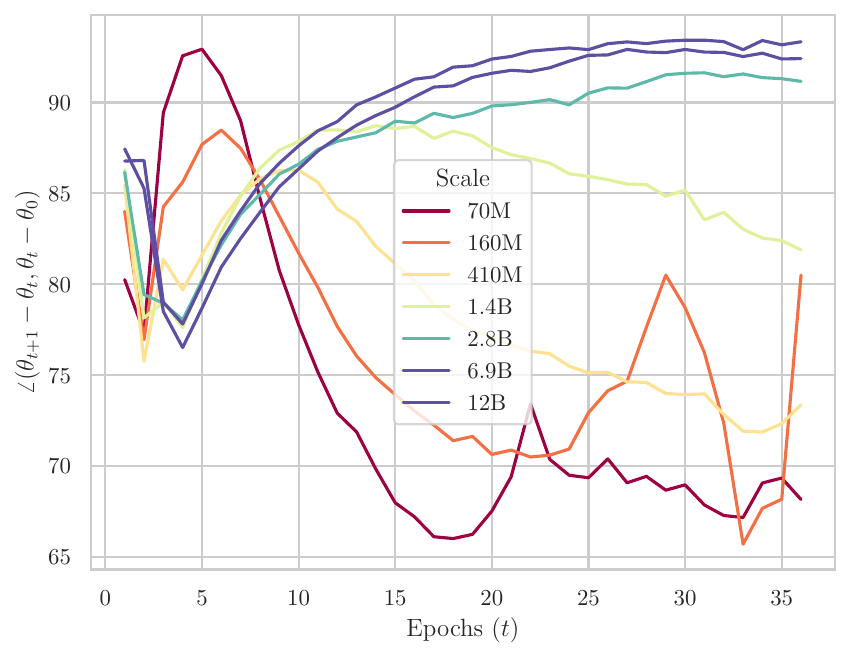}}
	\subfigure[$\angle(\theta_{t+k}-\theta_t,\theta_t-\theta_{t-k})$, for $k=1$ ]{\label{fig:}
		\includegraphics[width=0.34\textwidth]{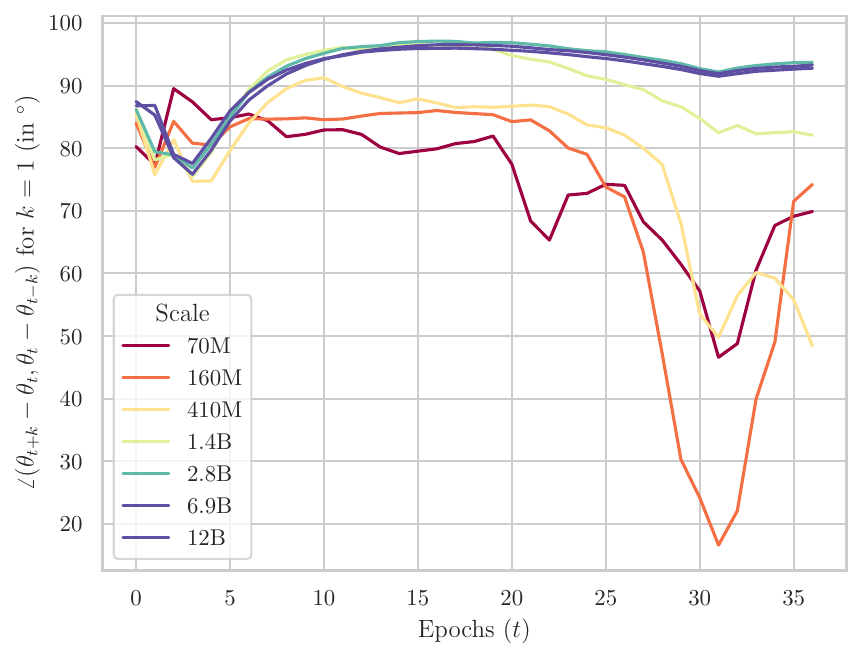}
		\vspace{-2mm}}
	\subfigure[$\angle(\theta_{t}-\theta_0,\theta_T-\theta_0)$]{\label{fig:}
		\includegraphics[width=0.34\textwidth]{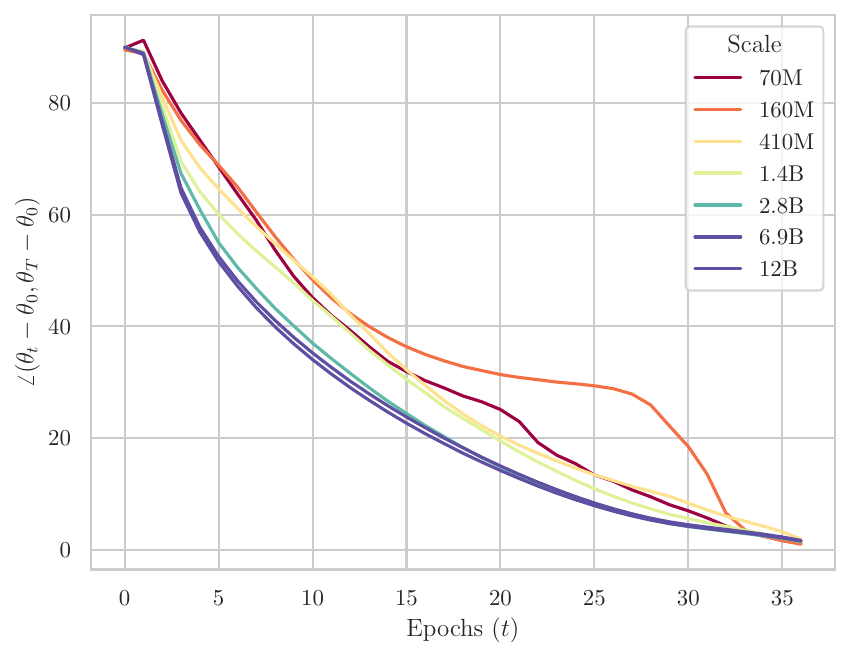}
		\vspace{-2mm}}
	\subfigure[$\angle(\theta_{t+1}-\theta_t, \theta_T-\theta_0)$]{\label{fig:}
		\includegraphics[width=0.34\textwidth]{figures/icml/Scale/ckpt_freq-4_heatmap_from_multi-pythia_dedup_2024-02-01_02-09-47_590002/figures/pdf/angle_theta__t+1_-theta_t,theta_T-theta_0__vs_Epochs__t__across_Scale.pdf}}
	\subfigure[Apex Angle at Initialization $\angle(\theta_t-\theta_0,\theta_1-\theta_0)$ ]{\label{fig:}
		\includegraphics[width=0.34\textwidth]{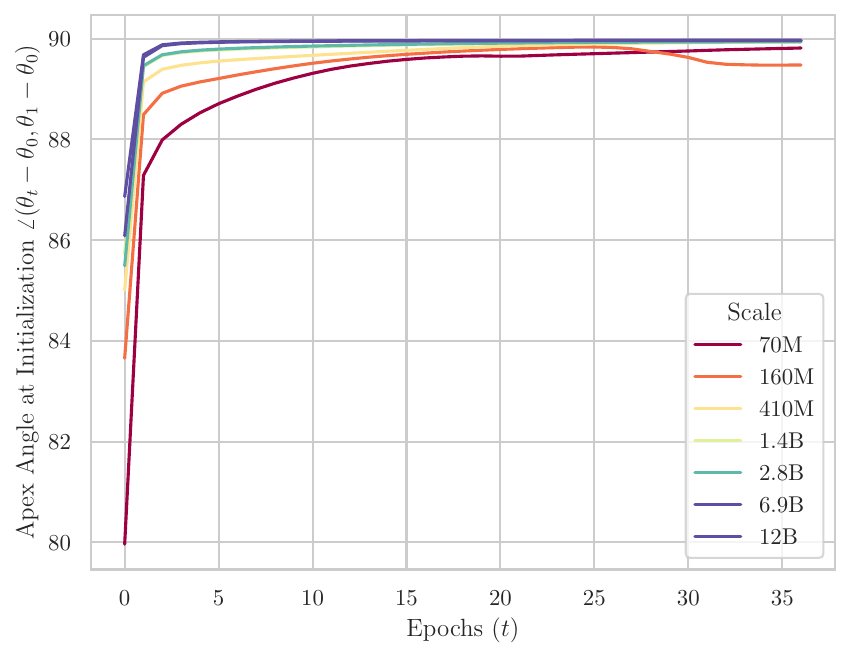}
		\vspace{-2mm}}
	\subfigure[Apex Angle at Origin $\angle(\theta_t,\theta_0)$]{\label{fig:}
		\includegraphics[width=0.34\textwidth]{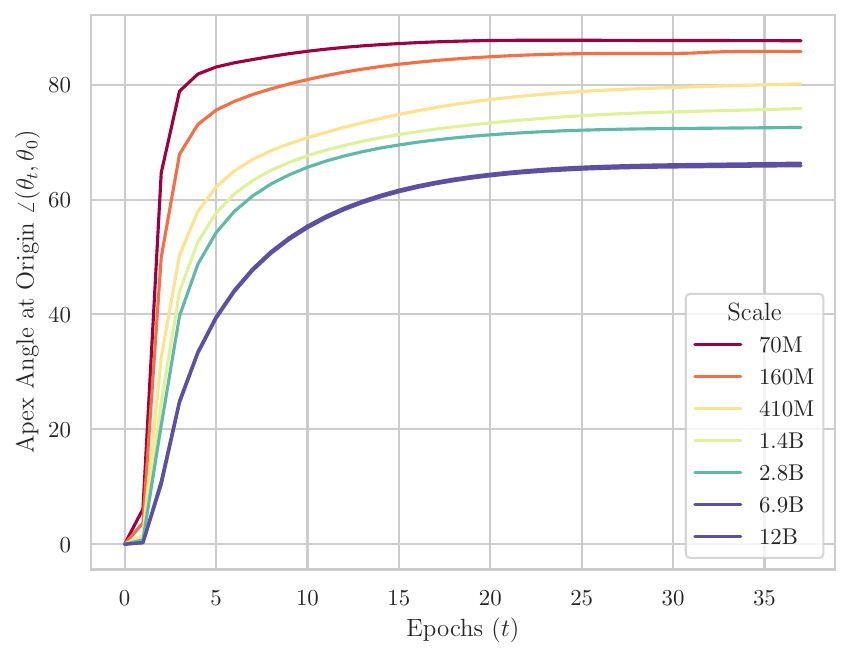}
		\vspace{-2mm}}
	\caption{Angular measures of the Trajectory for GPT-NeoX trained on the Pile dataset} 
\end{figure*}

\begin{figure*}[h!]
	\centering
	\subfigure[Eigenvalues: $\Km$]{\label{fig:}
		\includegraphics[width=0.3\textwidth]{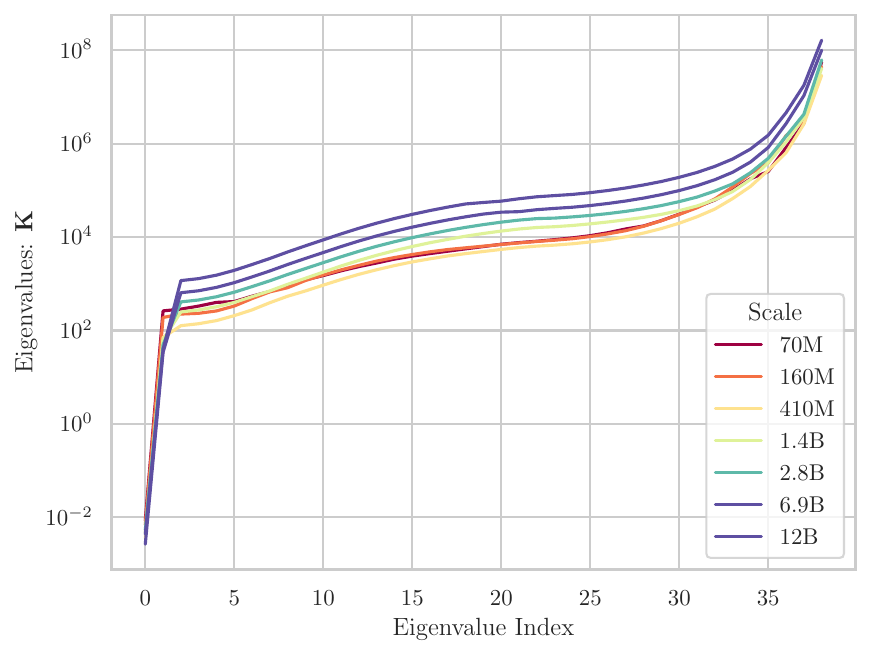}}
	\subfigure[Eigenvalues: $\Km_0$]{\label{fig:}
		\includegraphics[width=0.3\textwidth]{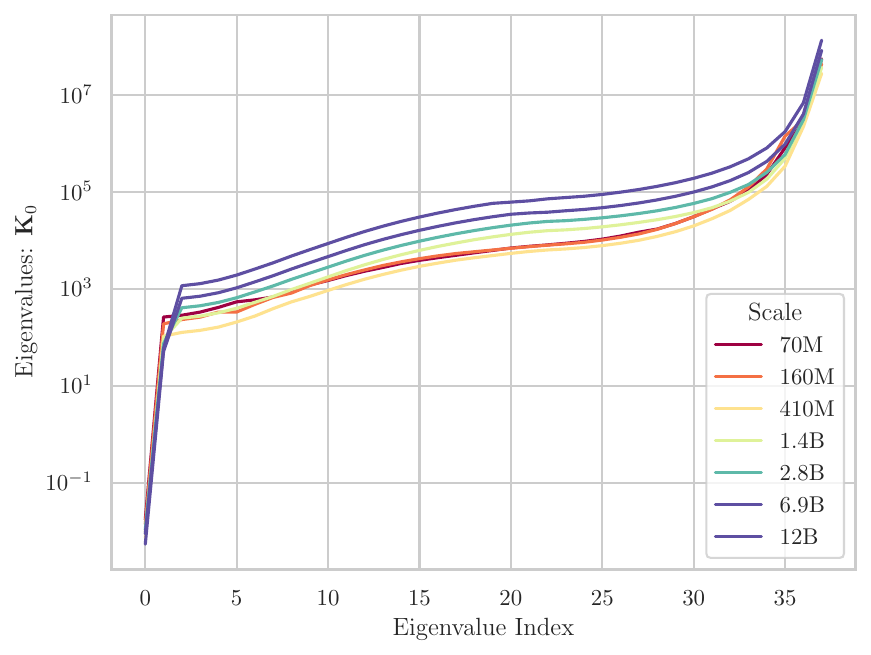}}

	\subfigure[Eigenvalues: $\Cm$ ]{\label{fig:}
		\includegraphics[width=0.3\textwidth]{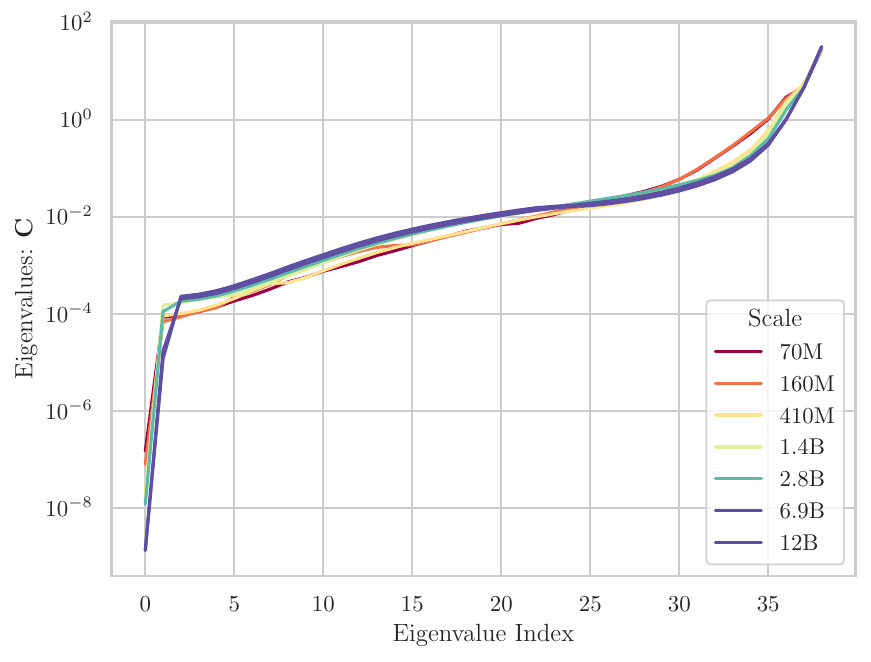}
		\vspace{-2mm}}
	\subfigure[Eigenvalues: $\Cm_0$]{\label{fig:}
		\includegraphics[width=0.3\textwidth]{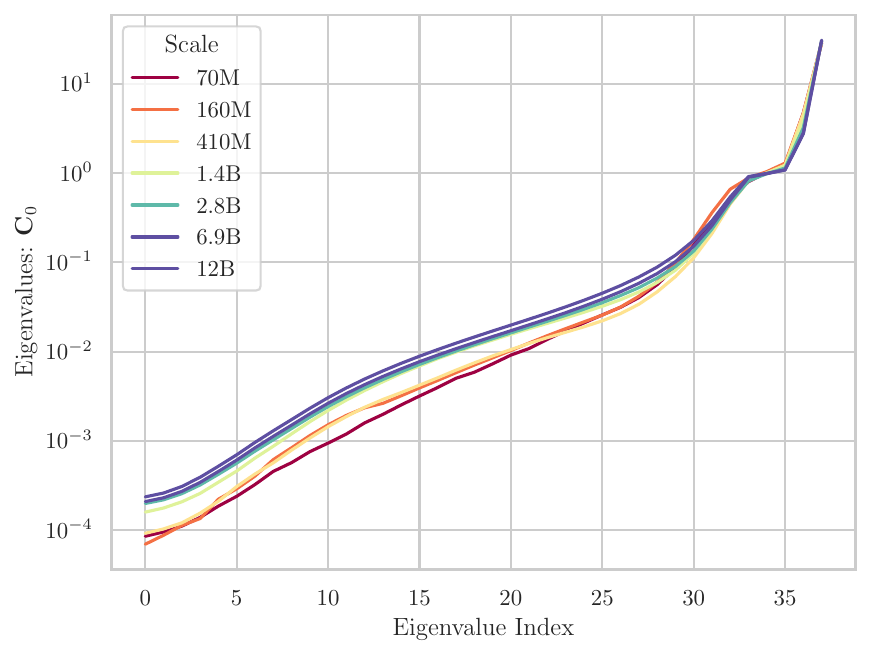}
		\vspace{-2mm}}
	\caption{Spectral measures of the Trajectory for GPT-NeoX trained on the Pile dataset} 
\end{figure*}

\clearpage

\section{Layerwise-Trajectory Maps}\label{fig:f}

	\begin{figure*}[h!]
	\centering
	\subfigure[query-key-value,  bias]{\label{fig:}
		\includegraphics[width=0.8\textwidth]{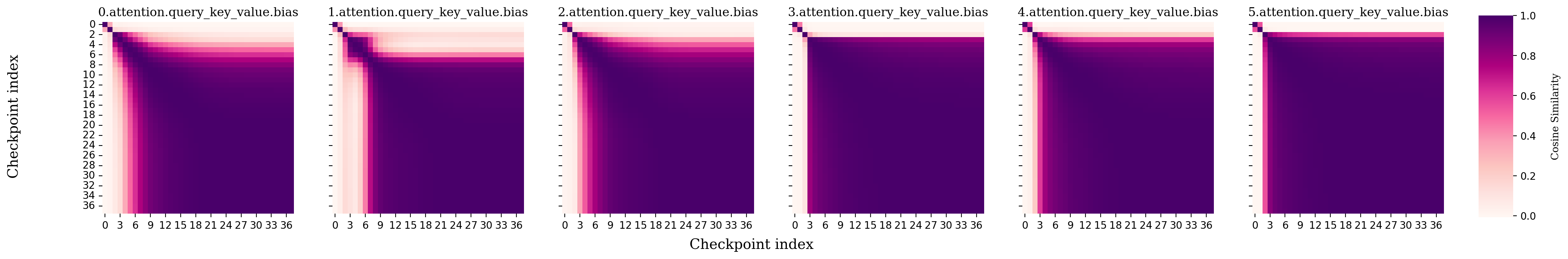}}
	\subfigure[query-key-value, weight]{\label{fig:}
		\includegraphics[width=0.8\textwidth]{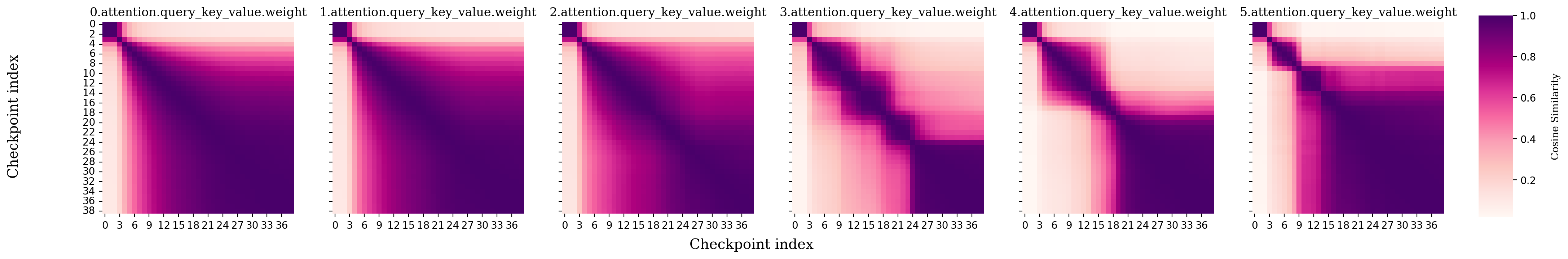}}
	\subfigure[dense, bias]{\label{fig:}
		\includegraphics[width=0.8\textwidth]{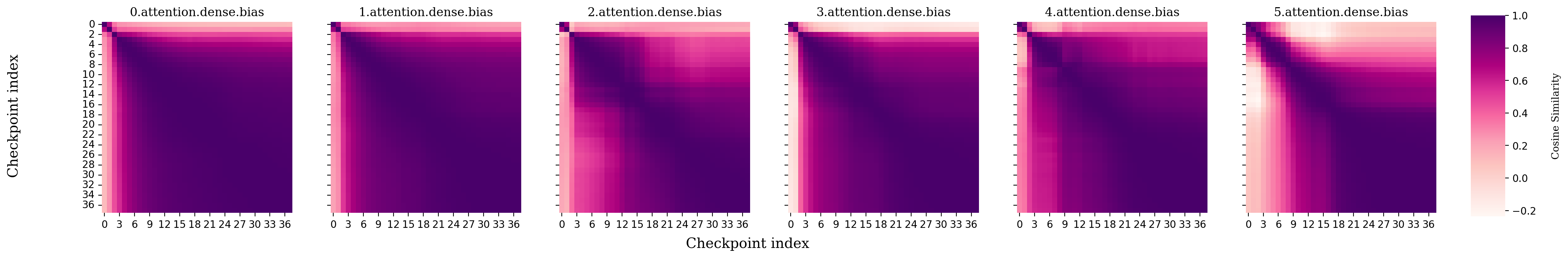}}
	\subfigure[dense, weight]{\label{fig:}
		\includegraphics[width=0.8\textwidth]{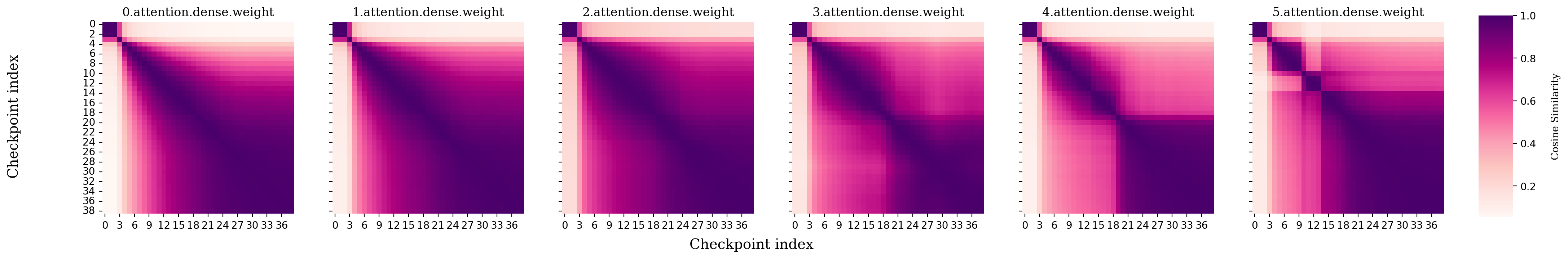}}
	\subfigure[dense-4h-to-h, bias]{\label{fig:}
		\includegraphics[width=0.8\textwidth]{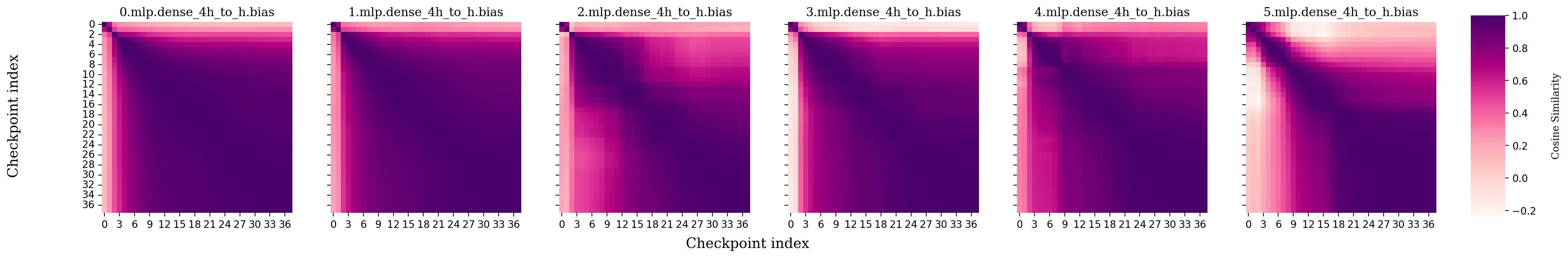}}
	\subfigure[dense-4h-to-h, weight]{\label{fig:}
		\includegraphics[width=0.8\textwidth]{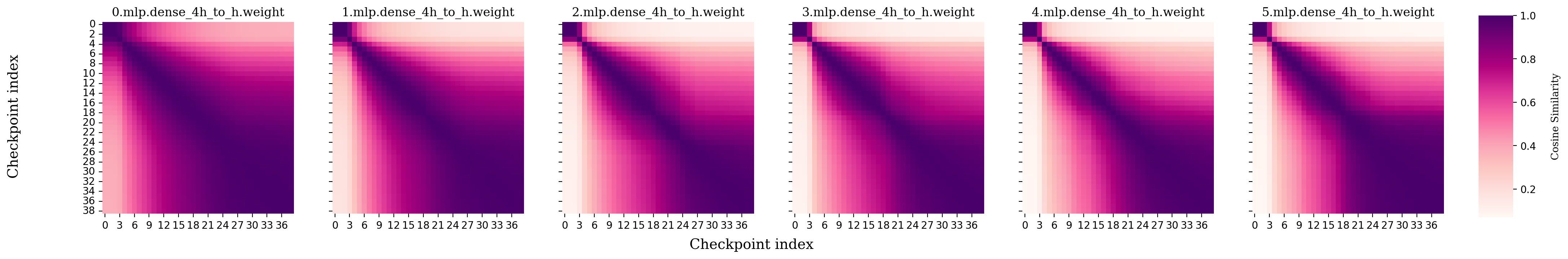}}
	\caption{Layerwise Trajectory Maps, grouped by layer type, for the $14M$ GPT-NeoX model trained on the Pile dataset.} 
\end{figure*}

\begin{figure*}[h!]
	\centering
	\subfigure[dense-h-to-4h, bias]{\label{fig:}
		\includegraphics[width=0.8\textwidth]{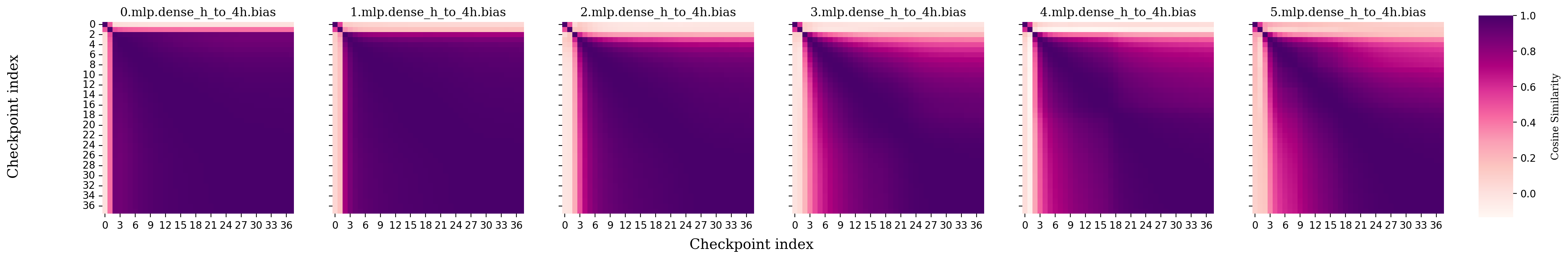}}
	\subfigure[dense-h-to-4h, weight]{\label{fig:}
		\includegraphics[width=0.8\textwidth]{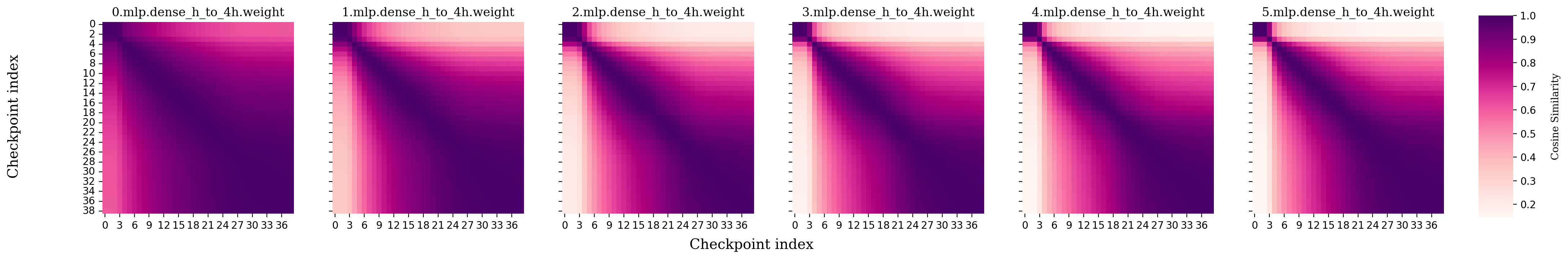}}
	\subfigure[embed-in, weight]{\label{fig:}
		\includegraphics[width=0.2\textwidth]{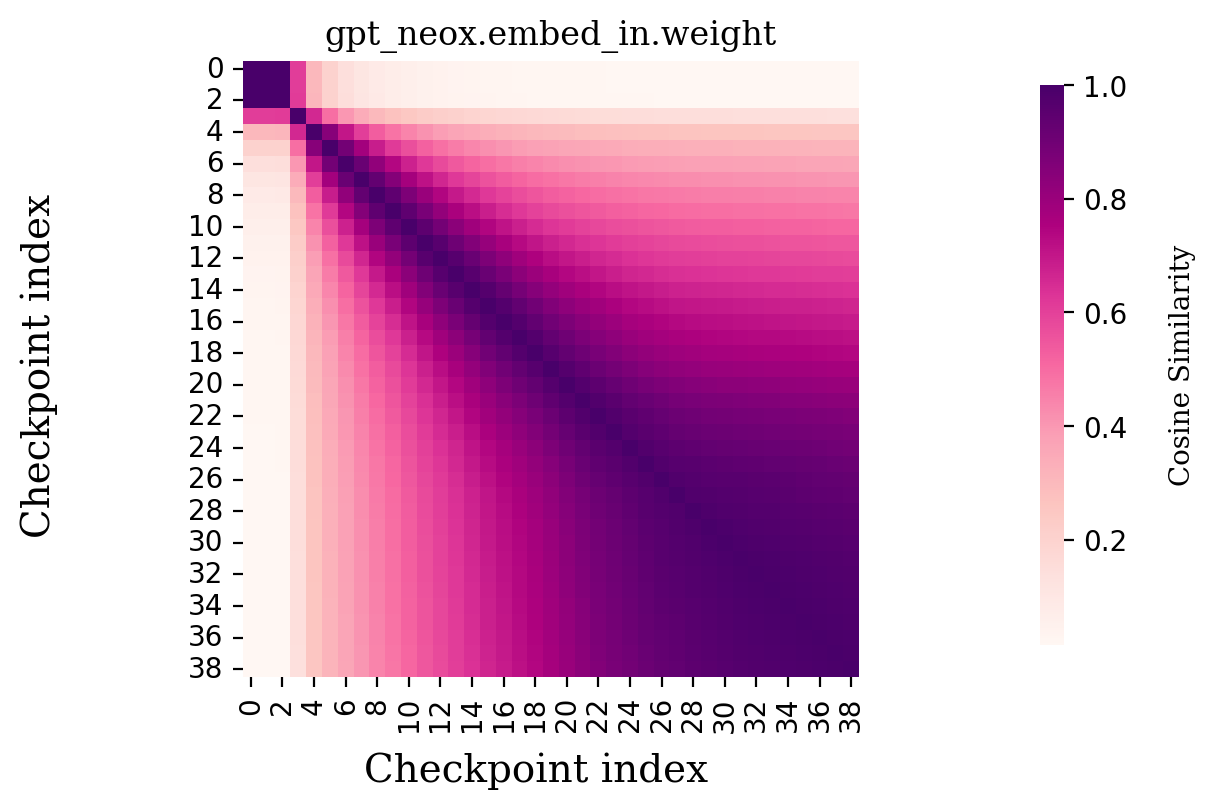}}
	\subfigure[embed-out, weight]{\label{fig:}
		\includegraphics[width=0.2\textwidth]{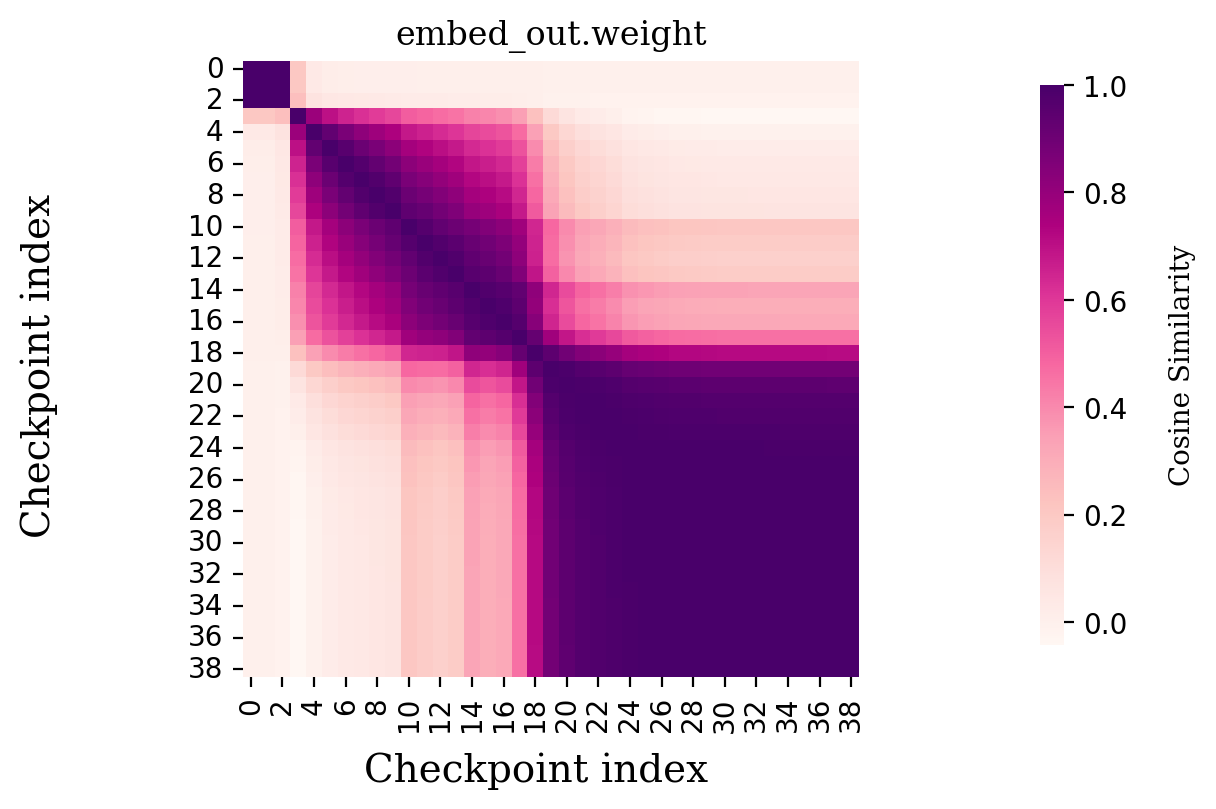}}
	\subfigure[final-layer-norm, bias]{\label{fig:}
		\includegraphics[width=0.2\textwidth]{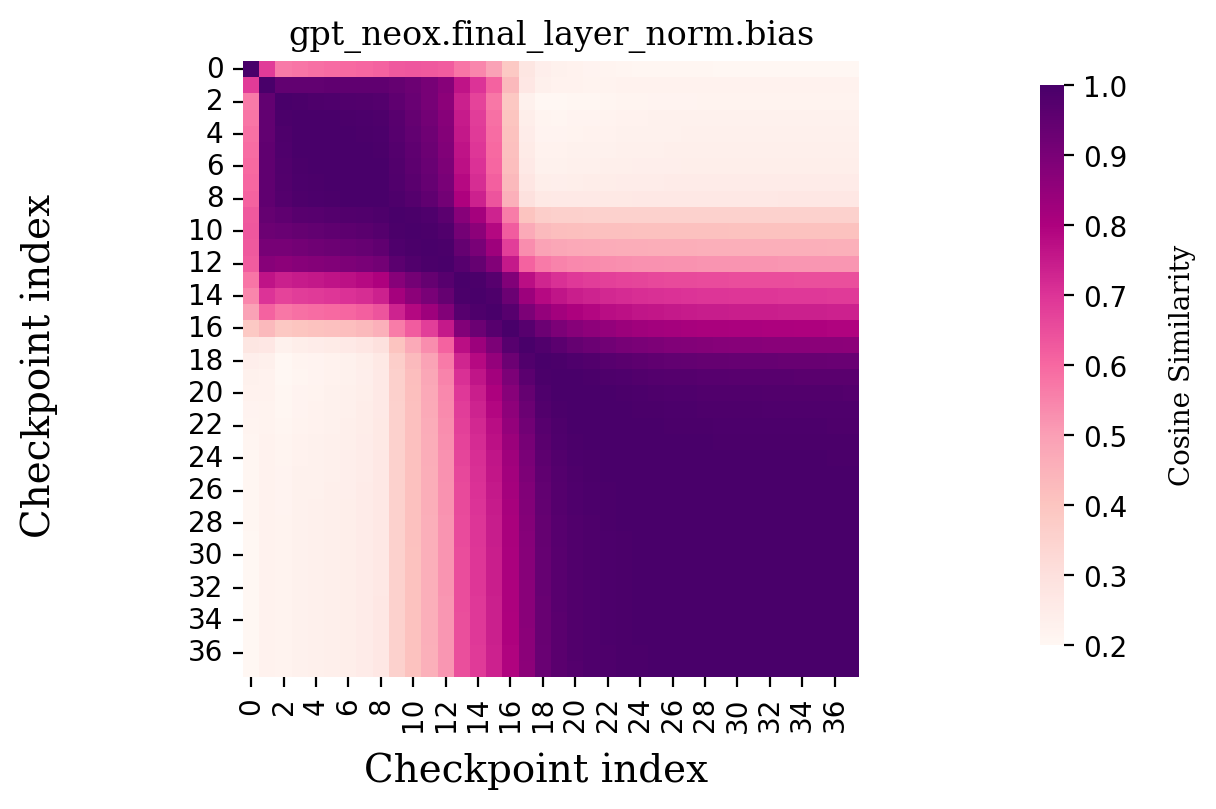}}
	\subfigure[final-layer-norm, weight]{\label{fig:}
		\includegraphics[width=0.2\textwidth]{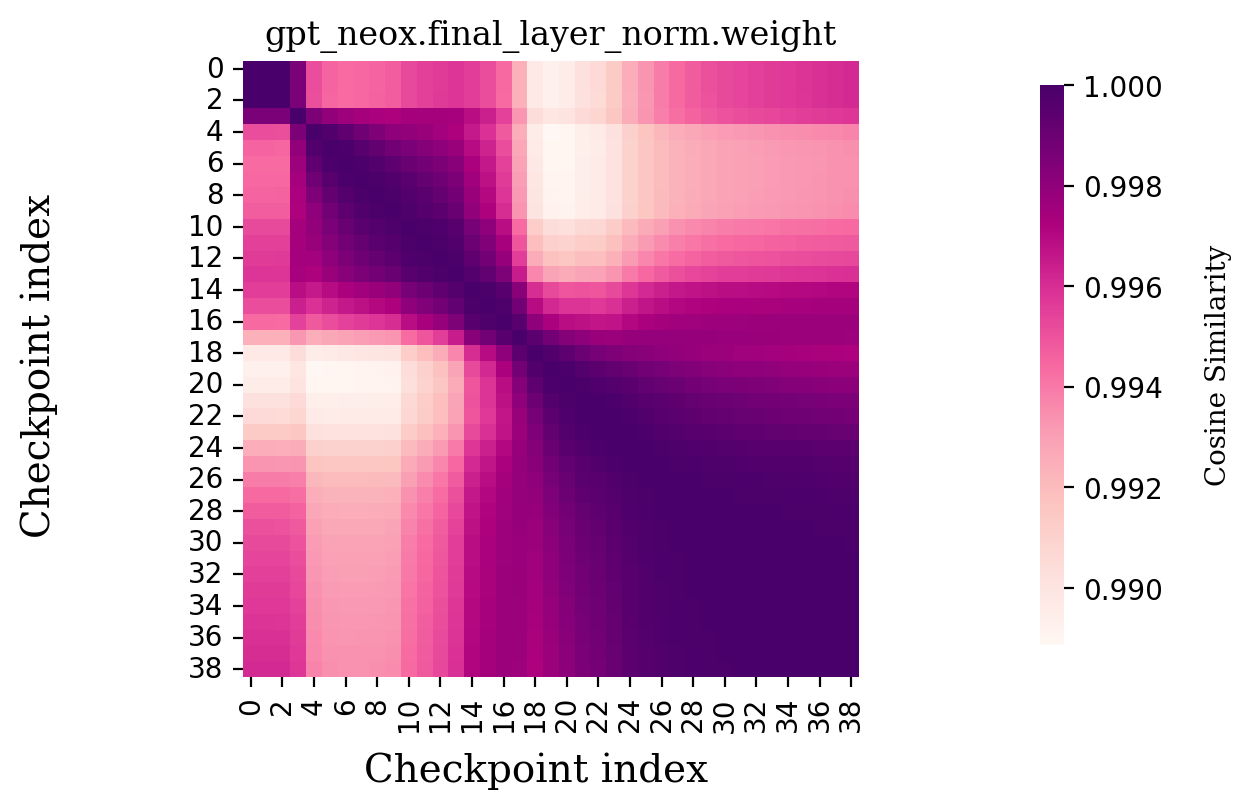}}
	\subfigure[input-layernorm, bias]{\label{fig:}
		\includegraphics[width=0.4\textwidth]{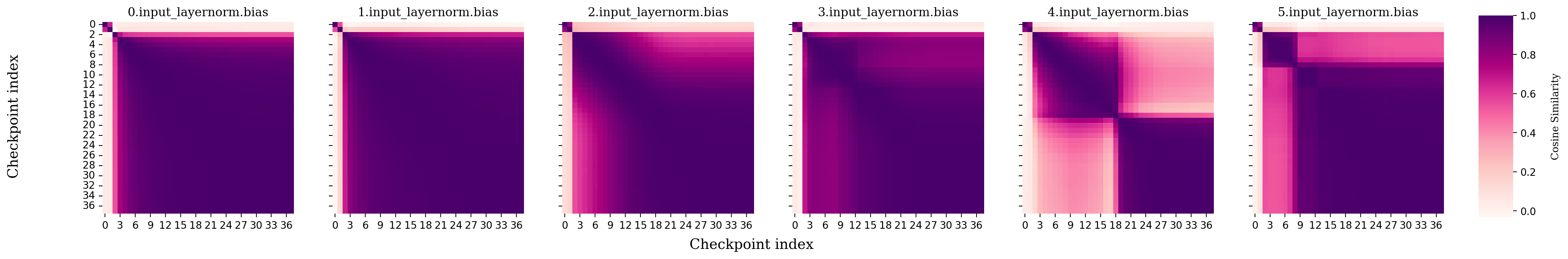}}
	\subfigure[input-layernorm, weight]{\label{fig:}
		\includegraphics[width=0.4\textwidth]{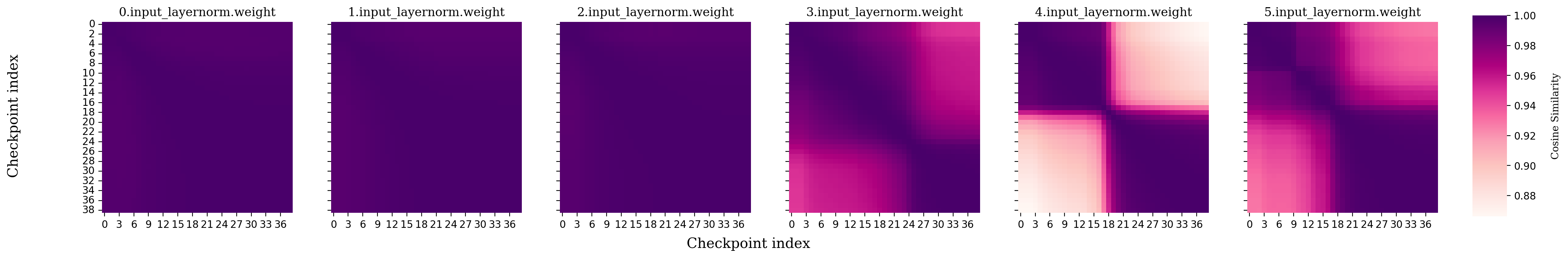}}
	\subfigure[post-attention-layernorm, bias]{\label{fig:}
		\includegraphics[width=0.4\textwidth]{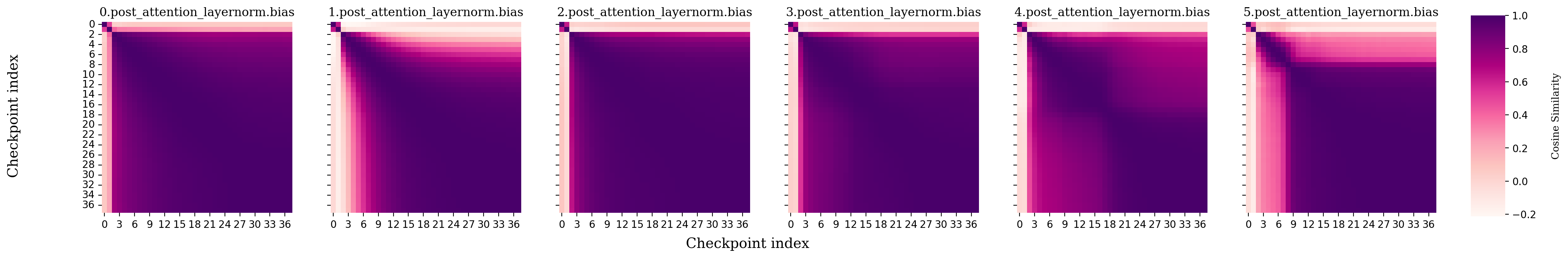}}
	\subfigure[post-attention-layernorm, weight]{\label{fig:}
		\includegraphics[width=0.4\textwidth]{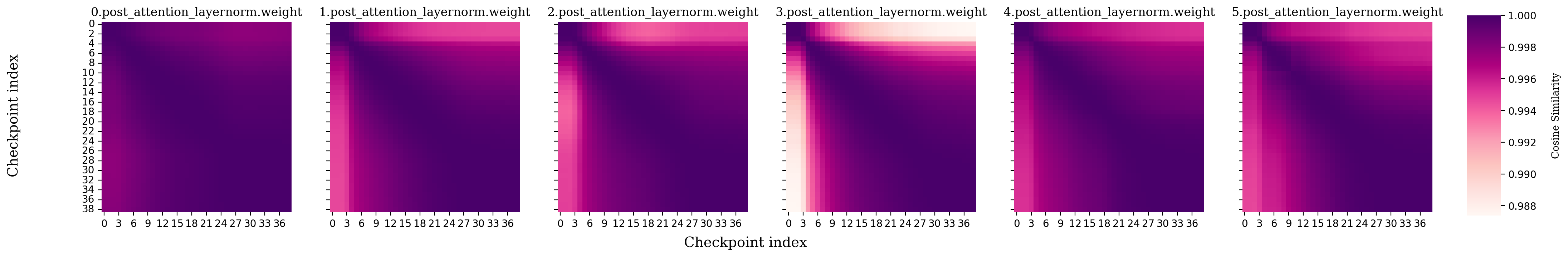}}
	\caption{Layerwise Trajectory Maps, grouped by layer type, for the $14M$ GPT-NeoX model trained on the Pile dataset.} 
\end{figure*}

\clearpage

\section{Trajectory Maps for Grokking}\label{app:grokking}
In grokking~\citep{power2022grokking}, we have that the performance on test samples significantly lags behind the training performance. Below, we look at the trajectory maps in this setting, considering the experimental setup of \url{https://github.com/teddykoker/grokking}.
\begin{figure}[!h]
    \centering
    \subfigure[Upto about 1500 steps]{\includegraphics[width=0.4\textwidth]{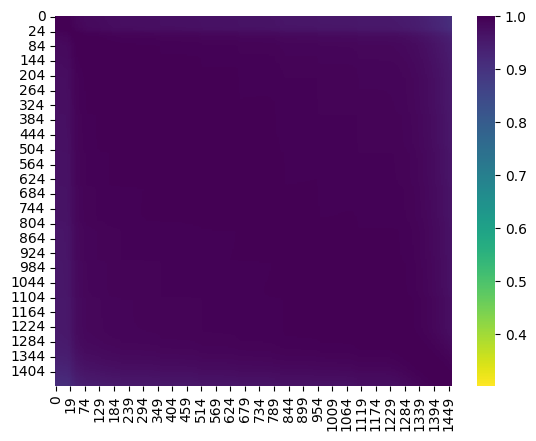}}
    \subfigure[Upto about 2000 steps]{\includegraphics[width=0.4\textwidth]{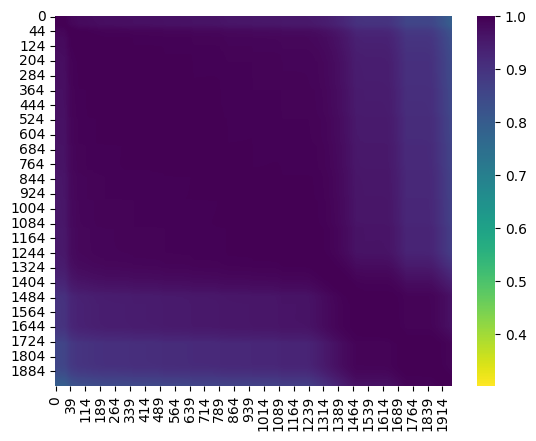}}
    \subfigure[Upto about 4000 epoch]{\includegraphics[width=0.4\textwidth]{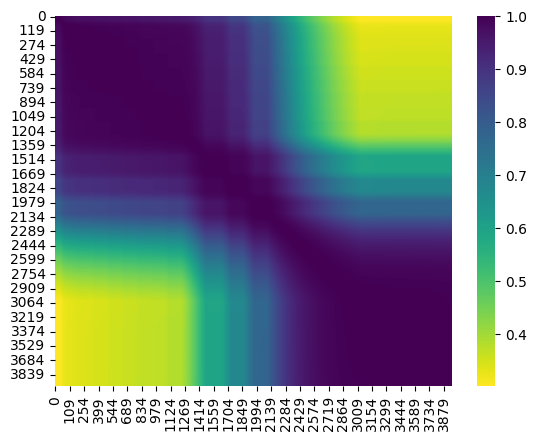}}
    \subfigure[Upto the end of training.
]{\includegraphics[width=0.4\textwidth]{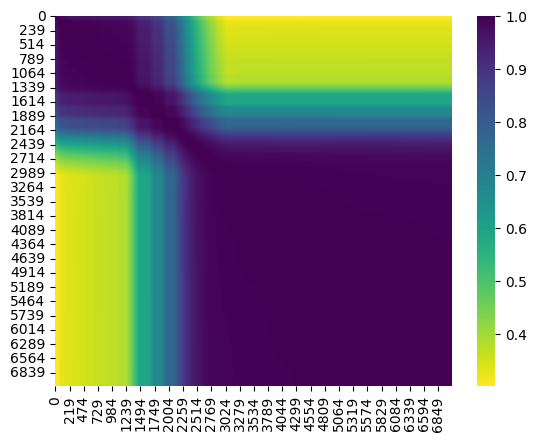}}
    \caption{Trajectory maps during the course of learning. Grokking~\citep{power2022grokking}, or sudden increase in test accuracy while training accuracy is already at a ceiling, occurs where the trajectory map also
shows a transition point.}    \label{fig:grok}
\end{figure}
We can observe in Figure~\ref{fig:grok} that:
\begin{itemize}
    \item Upto about 1500 epochs: Everything is pitch blue. No directional exploration, test accuracy remains, more or less, random.
\item Upto about 2000 epochs: Some directional movement starts to happen, and some initial signs of improvement in test performance.
\item Upto about 4000 epoch: Transition point for directional exploration. Test performance visibly improves.
\end{itemize}

We think that without (appropriate) directional exploration, the training converges to a `lazy’/`shortcut'/`dead-end' like solutions. Moreover, we believe that being `lazy’ in the directional sense is highly intertwined with being `lazy’ in the sense of feature learning~\citep{chizat2020lazy}. Besides, the above experiments show that the resemblance with the lazy regime is more than an analogy.~\citet{kumar2024grokking} have shown that grokking can be seen as the transition from the lazy to the non-lazy (rich) training regime. In particular, we find that the precise part of the training, where the test accuracy first shows a marked growth is also the part where the directional exploration starts to happen. 

\clearpage
\section{Putting Directional Redundancy to Test}
\begin{figure*}[h!]
	\centering
	\includegraphics[width=\textwidth]{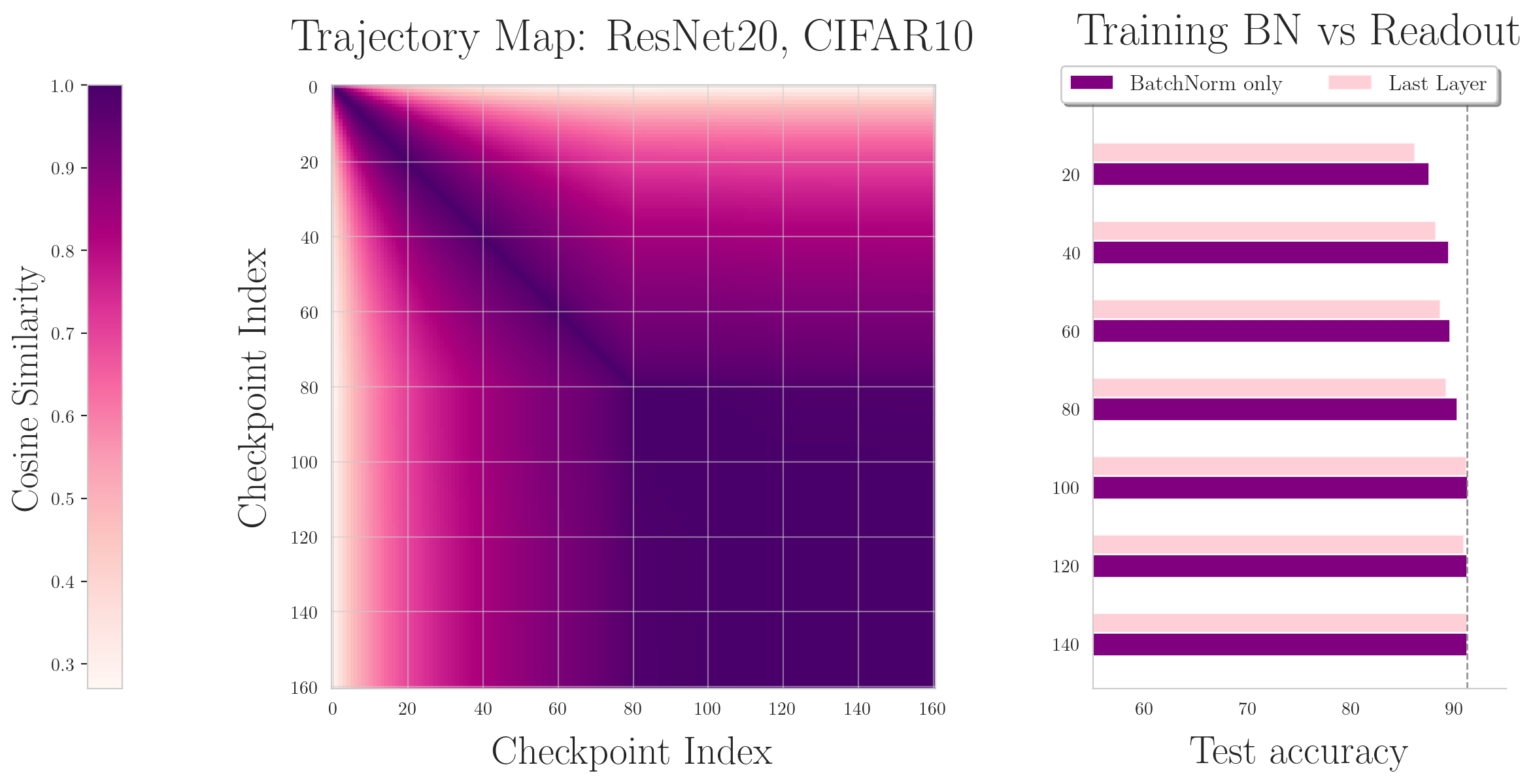}
	\caption{Comparison of training only batch norm parameters with training entire last layer (readout) parameters.\looseness=-1} 
 \label{fig:tm-bn-last}
\end{figure*}

The presented results, and that in the main section, have been averaged over 3 seeds. The standard deviation is never more than $0.05$ (for the batchnorm training, even less), and hence it is difficult to make it out in the plots and has been omitted.

\end{document}

%% file: math_defs.tex
\usepackage{dsfont}

\usepackage{mathabx}

\newcommand{\btheta}{{\pmb \theta}}
\newcommand{\bTheta}{{\pmb \Theta}}

\NewCommandCopy{\olda}{\a}
\renewcommand{\a}{\mathbf a}
\NewCommandCopy{\oldb}{\b}
\renewcommand{\b}{\mathbf b}

\makeatletter
\newtheorem*{rep@theorem}{\rep@title}
\newcommand{\newreptheorem}[2]{%
\newenvironment{rep#1}[1]{%
 \defrep@title{#2 \ref{##1}}%
 \begin{rep@theorem}}%
 {\end{rep@theorem}}}
\makeatother

\newtheorem{theorem}{Theorem}
\usepackage{mdframed}

\newtheorem{lemma}[theorem]{Lemma}

\DeclareMathOperator{\diag}{diag}

\newcommand{\opt}{{\btheta^\star}}

\newcommand{\Reals}[1]{{\mathbb{R}^{#1}}} %

\usepackage[scr]{rsfso}

\def\Cm{{\bf C}}
\def\Dm{{\bf D}}

\def\Im{{\bf I}}
\def\Km{{\bf K}}
\def\Mm{{\bf M}}

\def\Um{{\bf U}}
\def\Zm{{\bf Z}}

\usepackage{xcolor}%
\usepackage{ifmtarg}%
\usepackage{xifthen}%
\usepackage{environ}%
\usepackage{multido}%
\usepackage{lipsum}%
\usepackage{mdframed}
\theoremstyle{plain}

\usepackage{thmtools}
\usepackage{cleveref}

\declaretheoremstyle[%
  spaceabove=-2pt,%
  spacebelow=6pt,%
  headfont=\normalfont\itshape,%
  postheadspace=1em,%
  qed=\qedsymbol%
]{mystyle}



\newtheorem{assump}{Assumption}

\newcommand*{\colorboxed}{}
\def\colorboxed#1#{%
  \colorboxedAux{#1}%
}
\newcommand*{\colorboxedAux}[3]{%
  \begingroup
    \colorlet{cb@saved}{.}%
    \color#1{#2}%
    \boxed{%
      \color{cb@saved}%
      #3%
    }%
  \endgroup
}

\newreptheorem{proposition}{Proposition}
\newreptheorem{lemma}{Lemma}
\newreptheorem{theorem}{Theorem}
\newreptheorem{corollary}{Corollary}
\newreptheorem{assump}{Assumption}

\makeatletter
\newcommand{\neutralize}[1]{\expandafter\let\csname c@#1\endcsname\count@}
\makeatother

%% file: neurips_2024.bbl
\begin{thebibliography}{30}
\providecommand{\natexlab}[1]{#1}
\providecommand{\url}[1]{\texttt{#1}}
\expandafter\ifx\csname urlstyle\endcsname\relax
  \providecommand{\doi}[1]{doi: #1}\else
  \providecommand{\doi}{doi: \begingroup \urlstyle{rm}\Url}\fi

\bibitem[Andriushchenko et~al.(2023)Andriushchenko, D'Angelo, Varre, and
  Flammarion]{andriushchenko2023need}
Maksym Andriushchenko, Francesco D'Angelo, Aditya Varre, and Nicolas
  Flammarion.
\newblock Why do we need weight decay in modern deep learning?, 2023.

\bibitem[Biderman et~al.(2023)Biderman, Schoelkopf, Anthony, Bradley,
  O’Brien, Hallahan, Khan, Purohit, Prashanth, Raff,
  et~al.]{biderman2023pythia}
Stella Biderman, Hailey Schoelkopf, Quentin~Gregory Anthony, Herbie Bradley,
  Kyle O’Brien, Eric Hallahan, Mohammad~Aflah Khan, Shivanshu Purohit,
  USVSN~Sai Prashanth, Edward Raff, et~al.
\newblock Pythia: A suite for analyzing large language models across training
  and scaling.
\newblock In \emph{International Conference on Machine Learning}, pages
  2397--2430. PMLR, 2023.

\bibitem[Black et~al.(2022)Black, Biderman, Hallahan, Anthony, Gao, Golding,
  He, Leahy, McDonell, Phang, et~al.]{black2022gpt}
Sid Black, Stella Biderman, Eric Hallahan, Quentin Anthony, Leo Gao, Laurence
  Golding, Horace He, Connor Leahy, Kyle McDonell, Jason Phang, et~al.
\newblock Gpt-neox-20b: An open-source autoregressive language model.
\newblock \emph{arXiv preprint arXiv:2204.06745}, 2022.

\bibitem[Cao et~al.(2023)Cao, Zou, Li, and Gu]{cao2023implicit}
Yuan Cao, Difan Zou, Yuanzhi Li, and Quanquan Gu.
\newblock The implicit bias of batch normalization in linear models and
  two-layer linear convolutional neural networks.
\newblock In \emph{The Thirty Sixth Annual Conference on Learning Theory},
  pages 5699--5753. PMLR, 2023.

\bibitem[Chen et~al.(2021)Chen, Zhao, Wang, Li, Liu, Li, Yang, and
  Wang]{chen2021spann}
Qi~Chen, Bing Zhao, Haidong Wang, Mingqin Li, Chuanjie Liu, Zengzhong Li, Mao
  Yang, and Jingdong Wang.
\newblock Spann: Highly-efficient billion-scale approximate nearest neighbor
  search, 2021.

\bibitem[Chizat et~al.(2020)Chizat, Oyallon, and Bach]{chizat2020lazy}
Lenaic Chizat, Edouard Oyallon, and Francis Bach.
\newblock On lazy training in differentiable programming, 2020.

\bibitem[Cohen et~al.(2022)Cohen, Kaur, Li, Kolter, and
  Talwalkar]{cohen2022gradient}
Jeremy~M. Cohen, Simran Kaur, Yuanzhi Li, J.~Zico Kolter, and Ameet Talwalkar.
\newblock Gradient descent on neural networks typically occurs at the edge of
  stability, 2022.

\bibitem[Davis et~al.(2014)Davis, Balzer, and Soatto]{davis2014asymmetric}
Damek Davis, Jonathan Balzer, and Stefano Soatto.
\newblock Asymmetric sparse kernel approximations for large-scale visual
  search.
\newblock In \emph{Proceedings of the IEEE Conference on Computer Vision and
  Pattern Recognition}, pages 2107--2114, 2014.

\bibitem[Elhage et~al.(2021)Elhage, Nanda, Olsson, Henighan, Joseph, Mann,
  Askell, Bai, Chen, Conerly, et~al.]{elhage2021mathematical}
Nelson Elhage, Neel Nanda, Catherine Olsson, Tom Henighan, Nicholas Joseph, Ben
  Mann, Amanda Askell, Yuntao Bai, Anna Chen, Tom Conerly, et~al.
\newblock A mathematical framework for transformer circuits.
\newblock \emph{Transformer Circuits Thread}, 1:\penalty0 1, 2021.

\bibitem[Foret et~al.(2021)Foret, Kleiner, Mobahi, and
  Neyshabur]{foret2021sharpnessaware}
Pierre Foret, Ariel Kleiner, Hossein Mobahi, and Behnam Neyshabur.
\newblock Sharpness-aware minimization for efficiently improving
  generalization, 2021.

\bibitem[Frankle et~al.(2021)Frankle, Schwab, and Morcos]{frankle2021training}
Jonathan Frankle, David~J. Schwab, and Ari~S. Morcos.
\newblock Training batchnorm and only batchnorm: On the expressive power of
  random features in cnns, 2021.

\bibitem[Grosse et~al.(2023)Grosse, Bae, Anil, Elhage, Tamkin, Tajdini,
  Steiner, Li, Durmus, Perez, Hubinger, Lukošiūtė, Nguyen, Joseph,
  McCandlish, Kaplan, and Bowman]{grosse2023studying}
Roger Grosse, Juhan Bae, Cem Anil, Nelson Elhage, Alex Tamkin, Amirhossein
  Tajdini, Benoit Steiner, Dustin Li, Esin Durmus, Ethan Perez, Evan Hubinger,
  Kamilė Lukošiūtė, Karina Nguyen, Nicholas Joseph, Sam McCandlish, Jared
  Kaplan, and Samuel~R. Bowman.
\newblock Studying large language model generalization with influence
  functions, 2023.

\bibitem[Gunasekar et~al.(2018)Gunasekar, Lee, Soudry, and
  Srebro]{gunasekar2018characterizing}
Suriya Gunasekar, Jason Lee, Daniel Soudry, and Nathan Srebro.
\newblock Characterizing implicit bias in terms of optimization geometry.
\newblock In \emph{International Conference on Machine Learning}, pages
  1832--1841. PMLR, 2018.

\bibitem[He et~al.(2015)He, Zhang, Ren, and Sun]{he2015delving}
Kaiming He, Xiangyu Zhang, Shaoqing Ren, and Jian Sun.
\newblock Delving deep into rectifiers: Surpassing human-level performance on
  imagenet classification.
\newblock In \emph{Proceedings of the IEEE international conference on computer
  vision}, pages 1026--1034, 2015.

\bibitem[Jacot et~al.(2018)Jacot, Gabriel, and Hongler]{jacot2018neural}
Arthur Jacot, Franck Gabriel, and Cl{\'e}ment Hongler.
\newblock Neural tangent kernel: Convergence and generalization in neural
  networks.
\newblock \emph{Advances in neural information processing systems}, 31, 2018.

\bibitem[Jacot et~al.(2020)Jacot, Gabriel, and Hongler]{jacot2020neural}
Arthur Jacot, Franck Gabriel, and Clément Hongler.
\newblock Neural tangent kernel: Convergence and generalization in neural
  networks, 2020.

\bibitem[Jelassi and Li(2022)]{jelassi2022towards}
Samy Jelassi and Yuanzhi Li.
\newblock Towards understanding how momentum improves generalization in deep
  learning, 2022.
\newblock URL \url{https://openreview.net/forum?id=lf0W6tcWmh-}.

\bibitem[Ji and Telgarsky(2020)]{ji2020directional}
Ziwei Ji and Matus Telgarsky.
\newblock Directional convergence and alignment in deep learning.
\newblock \emph{Advances in Neural Information Processing Systems},
  33:\penalty0 17176--17186, 2020.

\bibitem[Kumar et~al.(2024)Kumar, Bordelon, Gershman, and
  Pehlevan]{kumar2024grokking}
Tanishq Kumar, Blake Bordelon, Samuel~J. Gershman, and Cengiz Pehlevan.
\newblock Grokking as the transition from lazy to rich training dynamics, 2024.

\bibitem[Li et~al.(2019)Li, Ma, and Zhang]{li2019algorithmic}
Yuanzhi Li, Tengyu Ma, and Hongyang Zhang.
\newblock Algorithmic regularization in over-parameterized matrix sensing and
  neural networks with quadratic activations, 2019.

\bibitem[Li et~al.(2020)Li, Wei, and Ma]{li2020explaining}
Yuanzhi Li, Colin Wei, and Tengyu Ma.
\newblock Towards explaining the regularization effect of initial large
  learning rate in training neural networks, 2020.

\bibitem[Liu et~al.(2023)Liu, Huang, and Xu]{liu2023implicit}
Chunrui Liu, Wei Huang, and Richard Yi~Da Xu.
\newblock Implicit bias of deep learning in the large learning rate phase: A
  data separability perspective.
\newblock \emph{Applied Sciences}, 13\penalty0 (6):\penalty0 3961, 2023.

\bibitem[Loshchilov and Hutter(2017)]{loshchilov2017decoupled}
Ilya Loshchilov and Frank Hutter.
\newblock Decoupled weight decay regularization.
\newblock \emph{arXiv preprint arXiv:1711.05101}, 2017.

\bibitem[Merrill et~al.(2020)Merrill, Ramanujan, Goldberg, Schwartz, and
  Smith]{merrill2020effects}
William Merrill, Vivek Ramanujan, Yoav Goldberg, Roy Schwartz, and Noah Smith.
\newblock Effects of parameter norm growth during transformer training:
  Inductive bias from gradient descent.
\newblock \emph{arXiv preprint arXiv:2010.09697}, 2020.

\bibitem[Moroshko et~al.(2020)Moroshko, Gunasekar, Woodworth, Lee, Srebro, and
  Soudry]{moroshko2020implicit}
Edward Moroshko, Suriya Gunasekar, Blake Woodworth, Jason~D. Lee, Nathan
  Srebro, and Daniel Soudry.
\newblock Implicit bias in deep linear classification: Initialization scale vs
  training accuracy, 2020.

\bibitem[Power et~al.(2022)Power, Burda, Edwards, Babuschkin, and
  Misra]{power2022grokking}
Alethea Power, Yuri Burda, Harri Edwards, Igor Babuschkin, and Vedant Misra.
\newblock Grokking: Generalization beyond overfitting on small algorithmic
  datasets, 2022.

\bibitem[Raffel et~al.(2023)Raffel, Shazeer, Roberts, Lee, Narang, Matena,
  Zhou, Li, and Liu]{raffel2023exploring}
Colin Raffel, Noam Shazeer, Adam Roberts, Katherine Lee, Sharan Narang, Michael
  Matena, Yanqi Zhou, Wei Li, and Peter~J. Liu.
\newblock Exploring the limits of transfer learning with a unified text-to-text
  transformer, 2023.

\bibitem[Sagun et~al.(2017)Sagun, Evci, Guney, Dauphin, and
  Bottou]{sagun2017empirical}
Levent Sagun, Utku Evci, V~Ugur Guney, Yann Dauphin, and Leon Bottou.
\newblock Empirical analysis of the hessian of over-parametrized neural
  networks.
\newblock \emph{arXiv preprint arXiv:1706.04454}, 2017.

\bibitem[Singh et~al.(2021)Singh, Bachmann, and Hofmann]{singh2021analytic}
Sidak~Pal Singh, Gregor Bachmann, and Thomas Hofmann.
\newblock Analytic insights into structure and rank of neural network hessian
  maps.
\newblock \emph{Advances in Neural Information Processing Systems},
  34:\penalty0 23914--23927, 2021.

\bibitem[Yang et~al.(2022)Yang, Hu, Babuschkin, Sidor, Liu, Farhi, Ryder,
  Pachocki, Chen, and Gao]{yang2022tensor}
Greg Yang, Edward~J. Hu, Igor Babuschkin, Szymon Sidor, Xiaodong Liu, David
  Farhi, Nick Ryder, Jakub Pachocki, Weizhu Chen, and Jianfeng Gao.
\newblock Tensor programs v: Tuning large neural networks via zero-shot
  hyperparameter transfer, 2022.

\end{thebibliography}
